\documentclass{article}

\PassOptionsToPackage{numbers, compress}{natbib}

\usepackage{PRIMEarxiv}

\usepackage[utf8]{inputenc} 
\usepackage[T1]{fontenc}    
\usepackage{hyperref}       
\usepackage{url}            
\usepackage{booktabs}       
\usepackage{amsfonts}       
\usepackage{nicefrac}       
\usepackage{microtype}      
\usepackage{lipsum}
\usepackage{fancyhdr}       
\usepackage{graphicx}       
\usepackage[utf8]{inputenc}
\usepackage[T1]{fontenc}
\usepackage{hyperref}
\usepackage{url}
\usepackage{booktabs}
\usepackage{amsfonts}
\usepackage{amssymb}
\usepackage{pifont}

\usepackage{nicefrac}
\usepackage{microtype}
\usepackage[table]{xcolor}
\usepackage{graphicx}
\usepackage{multirow}
\usepackage{tabularx}
\usepackage{adjustbox}
\usepackage{siunitx}
\usepackage{float}
\usepackage{caption}
\usepackage{makecell}
\usepackage{tcolorbox}
\usepackage{listings}
\usepackage{rotating}
\usepackage{natbib}
\usepackage{titlesec}

\graphicspath{{media/}}     

\title{SynthForensics: Benchmarking and Evaluating
People-Centric Synthetic Video Deepfakes}

\author{
\vspace{0.2cm}
\begin{tabular}{cccc}
Roberto Leotta$^{*}$ & Salvatore Alfio Sambataro$^{*}$ & Claudio Vittorio Ragaglia & Mirko Casu\\
{\small iCTLab s.r.l.} & {\small University of Catania} & {\small University of Catania} & {\small University of Catania}\\[0.5em]
Yuri Petralia & Francesco Guarnera & Luca Guarnera & Sebastiano Battiato\\
{\small iCTLab s.r.l.} & {\small University of Catania} & {\small University of Catania} & {\small University of Catania}
\end{tabular}\\[0.5cm]
{\tt\small \{roberto.leotta, yuri.petralia\}@ictlab.srl}\\
{\tt\small \{salvatore.sambataro, claudio.ragaglia, mirko.casu\}@phd.unict.it}\\
{\tt\small \{francesco.guarnera, luca.guarnera, sebastiano.battiato\}@unict.it}\\[0.2cm]
{\normalsize $^{*}$These authors contributed equally to this work.}
}

\begin{document}
\maketitle

\addtocontents{toc}{\protect\setcounter{tocdepth}{-1}}

\begin{figure}[!h]
    \centering
     \includegraphics[width=\linewidth]{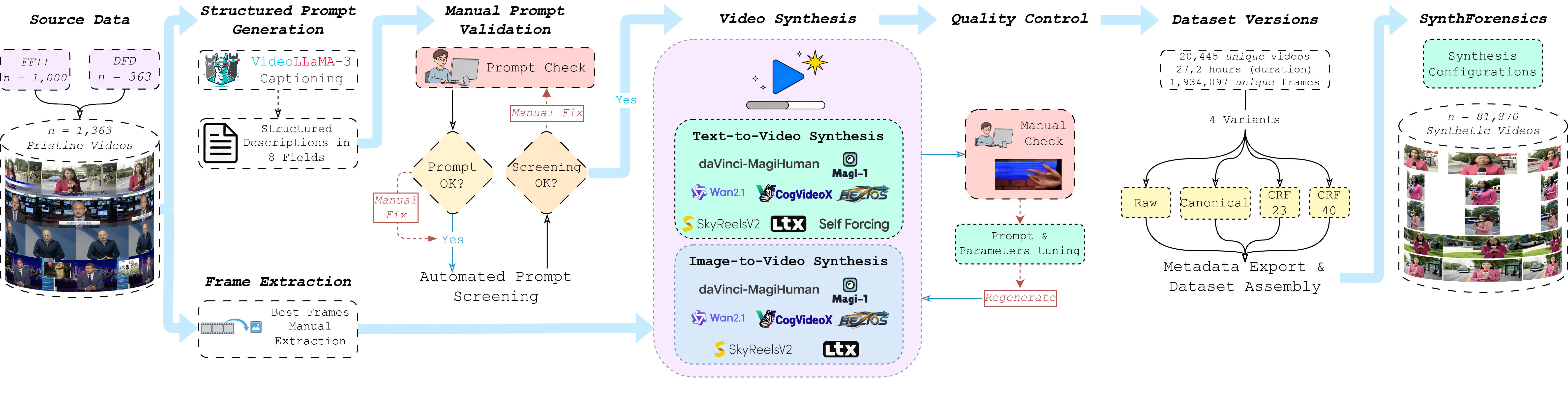}
    \caption{SynthForensics pipeline. From 1{,}363 paired real sources, human-validated captions drive 8 T2V and 7 I2V (with reference frames) generators. Outputs in four compression versions with metadata after second human validation, yielding a people-centric synthetic-video benchmark.}
    \label{fig:pipeline}
\end{figure}

\vspace{3em}

\begin{abstract}
Modern T2V/I2V generators synthesize people increasingly hard to distinguish from authentic footage, while current evaluation suites lag: legacy benchmarks target manipulation-based forgeries, and recent synthetic-video benchmarks prioritize scale over realistic human depiction. We introduce SynthForensics, a people-centric benchmark of $20{,}445$ videos from 8 T2V and 7 I2V open-source generators, paired-source from FF++/DFD reals, two-stage human-validated, in four compression versions with full metadata. In our paired-comparison human study, raters prefer SynthForensics in $71$--$77\%$ of head-to-head comparisons against each of nine existing synthetic-video benchmarks, while facial-quality metrics fall within the FF++/DFD baseline range. Across 15 detectors and three protocols, face-based methods drop $13$--$55$ AUC points (mean $27$) from FF++ to SynthForensics and a further $23$ under aggressive compression; fine-tuning closes the gap at a backward cost on legacy benchmarks; training from scratch shows synthetic and manipulation features largely disjoint for most detectors. We release dataset, pipeline, and code.
\end{abstract}


\newpage

\section{Introduction}
\label{sec:intro}

Modern text-to-video (T2V) and image-to-video (I2V) generative models can synthesize human subjects whose visual fidelity increasingly approaches that of authentic footage, raising forensic and societal concerns. While proprietary systems~\citep{sora2,veo3,kling2,seeddance2} enforce content guardrails and restricted access, the parallel rise of open-source generators~\citep{wan2025wan,yang2024cogvideox,chen2025skyreels,teng2025magi} is democratizing high-fidelity synthesis on consumer-grade hardware~\citep{huang2025selfforcingcode}. Unlike manipulation-based deepfakes, which locally alter existing recordings, purely synthetic generators can fabricate entirely novel individuals from a text prompt or animate a single photograph of a real person through I2V conditioning. This dramatically lowers the barrier for identity fraud, non-consensual content, disinformation, and impersonation; threats that fundamentally require realistic human depiction to succeed.

Despite recent efforts from the forensics community, a critical gap persists. Manipulation-based datasets~\citep{rossler2019faceforensics++,dolhansky2020deepfake} address face-swapping and reenactment artifacts that differ fundamentally from those of modern generative architectures. More recent synthetic video benchmarks~\citep{chen2024demamba,ni2026genvidbench} have expanded scale, yet often mix heterogeneous content (animations, game footage, natural scenes) without ensuring their human subsets reach the quality required for meaningful forensic evaluation~\citep{kundu2025towards,chen2025genworld}.

To address this gap, we present SynthForensics, a rigorously curated, human-centered benchmark for evaluating synthetic video deepfake detection. SynthForensics comprises over 20,000 unique videos from eight T2V and seven I2V architecturally diverse, state-of-the-art open-source generators. Each video is produced through a paired-source protocol where the positive prompt is derived from captioning a real source video in FaceForensics++ (FF++)~\citep{rossler2019faceforensics++} or the Deep Fake Detection (DFD) dataset~\citep{dufour2019google}, ensuring semantic alignment between real and generated counterparts. Every prompt and every generated video underwent two-stage human-in-the-loop validation for semantic fidelity, visual quality, and ethical compliance. Each video is provided in four compression versions with full generation metadata for reproducibility, totaling 81,780 video files.

Our contributions are as follows:
\begin{itemize}
    \item We introduce SynthForensics, a 20,000-video human-centered benchmark from 8 T2V and 7 I2V open-source generators, built via paired-source generation, two-stage human validation, and full metadata release at four compression levels.
    \item We benchmark curation practices, facial realism, and perceptual quality across SynthForensics and existing synthetic-video datasets, characterizing the current people-centric synthetic-video landscape and the forensic threat it poses.
    \item We evaluate state-of-the-art detectors on SynthForensics under zero-shot, fine-tuning, and generalization protocols, quantifying their performance against the latest open-source generators alongside compression robustness and T2V/I2V cross-modality transfer.
\end{itemize}

The paper is organized as follows. Section~\ref{sec:related_work} surveys prior work, Section~\ref{sec:sf_bench} details SynthForensics construction, Section~\ref{sec:comparative} compares it to existing benchmarks, Section~\ref{sec:detection} reports detection results, and Section~\ref{sec:conclusion} summarizes our contributions and findings.

\section{Related Work}
\label{sec:related_work}

Three threads frame the context for SynthForensics: the rapid expansion of generative video architectures, the landscape of benchmarks built around them, and the detectors that probe their artifacts.

\subsection{Evolution of Generative Video Architectures}
\label{subsec:gen_models}

Video synthesis has progressed from early GAN-based generation~\citep{goodfellow2014generative} through diffusion models~\citep{ho2020denoising} to the current Diffusion Transformer (DiT) paradigm~\citep{peebles2023scalable}, which now powers a broad range of generation tasks. Text-to-video models synthesize video from a textual description, while image-to-video models animate a reference image into a video sequence guided by a text prompt. The I2V modality is particularly relevant to the people-centric threat: a single photograph of a real individual can be sufficient to generate a convincing synthetic video.

Foundational open-source contributions to T2V/I2V generation include HunyuanVideo~\citep{kong2024hunyuanvideo}, which pioneered structured prompting via Multi-modal Large Language Models, alongside Wan2.1~\citep{wan2025wan} and CogVideoX~\citep{yang2024cogvideox}. A second wave explores autoregressive long-video generation: MAGI-1~\citep{teng2025magi} adopts purely autoregressive chunk-wise generation, Helios~\citep{yuan2026helios} achieves real-time autoregressive continuation, and SkyReels-V2~\citep{chen2025skyreels} pairs a Diffusion Forcing Transformer with reinforcement-learning post-training. Self-Forcing~\citep{huang2025self} extends this line by training a causal student that bridges the train-test exposure-bias gap and operates with very few inference steps. A third wave extends generation to joint audio-visual transformers: LTX-2.3~\citep{hacohen2026ltx} with an asymmetric dual-stream design, and daVinci-MagiHuman~\citep{chern2026speed} with a single-stream, human-centric variant.

Earlier-generation models such as ModelScope~\citep{wang2023modelscope}, VideoCrafter~\citep{chen2023videocrafter1}, SVD~\citep{blattmann2023stable}, AnimateDiff~\citep{guo2023animatediff}, and DynamiCrafter~\citep{xing2024dynamicrafter} remain present in existing detection benchmarks (Section~\ref{subsec:benchmarks}) but, built on pre-DiT U-Net diffusion architectures, generate at most 25 native frames ($\sim$2 seconds). On the proprietary side, Sora~2~\citep{sora2}, Veo~3.1~\citep{veo3}, Kling~3.0~\citep{kling2}, and SeedDance~2.0~\citep{seeddance2} represent the current closed-source state of the art.

\subsection{Existing Benchmarks for Video Deepfake Detection}
\label{subsec:benchmarks}

The deepfake detection landscape has been shaped by manipulation-focused benchmarks such as FF++~\citep{rossler2019faceforensics++}, DFD~\citep{dufour2019google}, Celeb-DF (CDF)~\citep{li2020celeb,li2025celeb}, DFDC~\citep{dolhansky2020deepfake}, DeeperForensics-1.0~\citep{jiang2020deeperforensics}, WildDeepfake (WildDF)~\citep{zi2020wilddeepfake}, DeepfakeTIMIT (TIMIT)~\citep{korshunov2018deepfakes}, Trusted Media (TM)~\citep{chen2022trusted}, AV-Deepfake1M++~\citep{cai2025av}, and DF40~\citep{yan2024df40}, which target face-swapping and reenactment artifacts fundamentally different from modern T2V/I2V outputs.

Recent benchmarks differ along several axes. In terms of synthetic-video scale, GenVideo~\citep{chen2024demamba} (1.1M), GenVidBench~\citep{ni2026genvidbench} (6.4M, primarily from VidProM~\citep{wang2024vidprom}), AIGVDBench~\citep{ma2026your} (422K, T2V/I2V/V2V), GenVidDet~\citep{ji2024distinguish} (1.2M), and ILLUSION~\citep{thakral2024illusion} (300K) reach the largest sizes. Among smaller efforts, GVF~\citep{ma2025detecting} adopts a paired-source protocol with 964 real-prompt-fake triplets, while GVD~\citep{bai2024ai} evaluates compression robustness on 11,618 videos at multiple H.264 CRF levels. Only DeepAction~\citep{bohacek2025deepaction} and Deepfake-Eval-2024~\citep{chandra2025deepfake} explicitly target people-centric content, with DeepAction focusing on action plausibility rather than facial realism and Deepfake-Eval-2024 scraping social-media videos that include face-swapping, lip-sync, and fully synthetic content; all other benchmarks address different scopes: DVF~\citep{song2024learning}, LOKI~\citep{ye2024loki}, AEGIS~\citep{li2025aegis}, GenWorld~\citep{chen2025genworld}, RobustSora~\citep{wang2025robustsora}, and Ivy-Fake~\citep{jiang2025ivy}.

\subsection{Detection Methodologies}
\label{subsec:detection_methods}

Detection approaches for synthetic and manipulated video can be organized along two axes: the input modality (frame-level versus video-level analysis) and the content scope (face-specific versus general-purpose). Among face-based frame-level detectors, CFM~\citep{luo2023beyond} mines critical forgery features via prior-agnostic augmentation and triplet relation learning, RECCE~\citep{cao2022end} learns reconstruction-classification differences between real and fake faces, ProDet~\citep{cheng2024can} regularizes the real-blendfake-deepfake transition through progressive learning, UCF~\citep{yan2023ucf} disentangles content from common and method-specific forgery components, Effort~\citep{yan2024orthogonal} adapts vision foundation models via orthogonal SVD with frozen principal components, LAA-Net~\citep{nguyen2024laa} localizes vulnerable pixels through heatmap and self-consistency branches, and GenD~\citep{yermakov2026deepfake} fine-tunes only LayerNorm parameters on a hyperspherical metric-learning manifold. Face-based video-level detectors model temporal dynamics: AltFreezing~\citep{wang2023altfreezing} alternately freezes spatial and temporal weights during training, FTCN~\citep{zheng2021exploring} reduces spatial kernels to one to force temporal learning, GenConViT~\citep{deressa2025genconvit} combines a generative branch with ConvNeXt and Swin features, DFD-FCG~\citep{han2025towards} adapts CLIP through a side network with Facial Component Guidance, and FakeSTormer~\citep{nguyen2025vulnerability} adopts a TimeSformer with vulnerability-aware auxiliary branches.

General-purpose video-level detectors include MM-Det~\citep{song2024learning}, which couples LMM reasoning with VQ-VAE~\citep{van2017neural}-amplified diffusion traces, D3~\citep{zheng2025d3}, a training-free method based on second-order temporal features under a Newtonian formulation, NSG-VD~\citep{zhang2025physics}, which detects through a physics-driven Normalized Spatiotemporal Gradient, DeMamba~\citep{chen2024demamba}, a plug-and-play state-space-model module for temporal-spatial inconsistency detection, and UNITE~\citep{kundu2025towards}, which builds on SigLIP for unified manipulation/synthesis detection.

\section{The SynthForensics Benchmark}
\label{sec:sf_bench}

This section details the SynthForensics construction pipeline, visually summarized in Figure~\ref{fig:pipeline}.

\subsection{Source Data and Generative Models}
\label{subsec:source_n_gen}

SynthForensics employs a paired-source protocol using 1,363 pristine videos from FF++ (1,000) and DFD (363). Each pristine video is captioned and the resulting caption is used as the positive prompt to anchor synthetic counterparts in both T2V and I2V modes, controlling high-level semantic variables (scene, subjects, actions) to isolate low-level generative artifacts from semantic confounds.

Among the recent open-source generators surveyed in Section~\ref{subsec:gen_models}, we selected those that produce near-photorealistic human outputs and excluded those that fail this bar or are closed-source. The selected eight T2V generators are Wan2.1 (14B), CogVideoX (5B), SkyReels-V2 (14B), Self-Forcing (1.3B), MAGI-1-Distilled (24B), LTX-2.3 (22B), Helios-Distilled (14B), and daVinci-MagiHuman-Distilled (15B). Seven are also used in I2V mode, with Self-Forcing excluded as it does not support image conditioning. This selection balances scale (1.3B to 24B parameters), efficiency (consumer versus datacenter GPUs), and architectural diversity (standard, autoregressive, diffusion-forcing, distilled, and audio-visual transformers). Pre-DiT U-Net models (Section~\ref{subsec:gen_models}) were excluded as a class in favor of more recent DiT-based architectures. We also exclude proprietary models (Section~\ref{subsec:gen_models}), whose visible watermarks act as trivial detection artifacts, content guardrails prevent unrestricted human-centered generation, and closed-source distribution precludes reproducibility. Full selection criteria and per-generator specifications are reported in Appendix~\ref{app:benchmark-sources}.

\subsection{Structured Conditioning Preparation and Validation}
\label{subsec:conditioning_val}

To produce high-fidelity outputs, each generator requires tailored prompts that match its captioning style. Following HunyuanVideo's multi-dimensional framework~\citep{kong2024hunyuanvideo}, we employed a Vision-Language Model (VLM)~\citep{zhang2025videollama} to generate structured captions for each source video across eight fields: \textit{short description}, \textit{dense description}, \textit{background}, \textit{style}, \textit{shot type}, \textit{camera movement}, \textit{lighting}, and \textit{atmosphere}. From this unified representation we dynamically construct model-specific positive prompts that condition the generators to recreate the source's scene, subjects, and actions by selecting and ordering fields according to each generator's training distribution, and iteratively curate model-specific negative prompts to suppress recurring artifacts (e.g., ``blurry, distorted''). Per-generator templates and negative-prompt details are reported in Appendix~\ref{app:benchmark-prompts}.

Each of the 1,363 structured captions underwent manual field-by-field validation by human annotators to verify consistency with source video content and detect potentially sensitive or harmful content. As a complementary safeguard, human-validated captions were then automatically screened by a Large Language Model (LLM)~\citep{grattafiori2024llama}, which flagged references across seven thematic categories (violence and warfare, weapons, vulnerable individuals, political figures and institutions, geopolitical references, national symbols, and identifiable real persons) potentially missed during human review. Flagged captions were returned to human annotators for rewriting that neutralized sensitive elements while preserving visual and narrative intent. These iterations continued until all captions passed both human and automated screening. Detailed protocols are reported in Appendix~\ref{app:benchmark-validation}.

Additionally, for I2V conditioning, human annotators manually select a reference frame from each source video, prioritizing a large, centered, and sharp face in a neutral and frontal pose that matches the structured caption. The frame extraction protocol is detailed in Appendix~\ref{app:benchmark-i2v-frames}.

\subsection{Video Synthesis and Quality Control}
\label{subsec:vid_synth_qc}

Once prompts and reference frames are validated, we proceed to video generation. Each generator has distinct optimal settings for inference hyperparameters (e.g., classifier-free guidance (CFG) scale, diffusion steps), as well as resolution, frame rate, and duration based on its training data distribution. To ensure they operate at peak performance while emulating modern distribution scenarios on social media and broadcast platforms, we developed a synthesis protocol combining diversity across temporal, spatial, and format parameters with rigorous manual validation for visual quality and ethical compliance. All inference parameters followed each author's guidelines, with per-generator and per-video refinements applied during preliminary experiments and the subsequent manual validation stage (full per-generator parameters in Appendix~\ref{app:benchmark-hyperparams}).

Every synthesized video underwent manual inspection for anatomical inconsistencies, rendering errors, temporal artifacts, semantic coherence with the prompt and reference frame (for I2V), and ethical compliance. Rejected videos triggered iterative refinement of prompts (both positive and negative) and generation parameters until a high-fidelity, coherent, and ethically sound video was accepted. Only videos passing this validation constitute the final 20,445 samples. Detailed inspection criteria are reported in Appendix~\ref{app:benchmark-video-validation}.

\subsection{Dataset Versions, Statistics, and Metadata}
\label{subsec:dataset_statistics}

\begin{table}[t]
\centering
\caption{Overall statistics of the SynthForensics benchmark.}
\label{tab:overall_stats}
\small
\begin{tabularx}{\columnwidth}{Xr}
\toprule
\textbf{Metric} & \textbf{Value} \\
\midrule
Unique videos (T2V) & 10,904 \\
Unique videos (I2V) & 9,541 \\
Total unique synthetic videos & 20,445 \\
Total video files (4 compression versions) & 81,780 \\
Total unique frames & 1,934,097 \\
Total unique duration & $\sim$27.2 hours \\
\midrule
Resolution range & 384$\times$640 to 1088$\times$1920 \\
Frame rate range (FPS) & 8 to 25 \\
Duration range & 4 to 6 s \\
Landscape / Portrait videos & 16,349 / 4,096 \\
\bottomrule
\end{tabularx}
\end{table}

We provide four versions of each video to facilitate robustness analysis. The \textit{Raw} version is the direct, unprocessed generator output. The \textit{Canonical} version re-encodes each video with uniform parameters (H.264 CRF=0, YUV420p, BT.709 color space) while preserving original resolutions, neutralizing potential format-specific confounds. The \textit{CRF23} and \textit{CRF40} versions apply progressive compression to the \textit{Canonical} videos to simulate real-world distribution scenarios. Table~\ref{tab:overall_stats} reports overall statistics. We release comprehensive metadata alongside each video (prompts, negative prompts, random seeds, and all inference hyperparameters). Full statistics, metadata catalog, and qualitative samples are in Appendices~\ref{app:benchmark-stats} and~\ref{app:benchmark-samples}.

\section{Comparative Benchmark Analysis}
\label{sec:comparative}

We characterize SynthForensics relative to nine publicly available synthetic-video benchmarks: GenVideo~\citep{chen2024demamba}, GenVidBench~\citep{ni2026genvidbench}, AIGVDBench~\citep{ma2026your}, DVF~\citep{song2024learning}, GVF~\citep{ma2025detecting}, GVD~\citep{bai2024ai}, LOKI~\citep{ye2024loki}, AEGIS~\citep{li2025aegis}, and DeepAction~\citep{bohacek2025deepaction}. These are selected from the landscape of Section~\ref{subsec:benchmarks} under three inclusion criteria: (i) downloadable data, (ii) labels that isolate fully synthetic videos from partial manipulations such as face-swap or lip-sync, and (iii) content not entirely re-sourced from other benchmarks in our selection; GenVidDet~\citep{ji2024distinguish}, GenWorld~\citep{chen2025genworld}, ILLUSION~\citep{thakral2024illusion}, and RobustSora~\citep{wang2025robustsora} are excluded for (i), Deepfake-Eval-2024~\citep{chandra2025deepfake} for (ii), and Ivy-Fake~\citep{jiang2025ivy} for (iii). FF++~\citep{rossler2019faceforensics++} and DFD~\citep{dufour2019google} pristine videos serve as real-video baselines throughout. To enable a fair comparison, given that SynthForensics is people-centric by construction whereas the selected benchmarks span broader content categories, the per-benchmark aggregates reported in the next subsections are computed on the face-positive subset of each benchmark. Per-benchmark video counts after byte-identical deduplication and the construction of this subset are detailed in Appendix~\ref{app:comparative-preprocessing}.

\subsection{Landmark Stability Analysis}
\label{subsec:landmark_stability}

A realistic face, whether real or synthetic, should admit a clean, anatomically consistent localization of its landmarks. When a generator mangles facial geometry, produces unstable contours across frames, or blends textures without coherent structure, the resulting faces resist precise landmark placement: the model's representation of where each landmark lies becomes diffuse rather than sharply peaked. We quantify this through the 2D Face Alignment Network (2DFAN4)~\citep{bulat2017fan}, treating the peak activation of each per-landmark heatmap as a proxy for landmark localization reliability.

\begin{figure}[t]
\centering
\begin{minipage}[b]{0.48\linewidth}
\centering
\includegraphics[width=\linewidth]{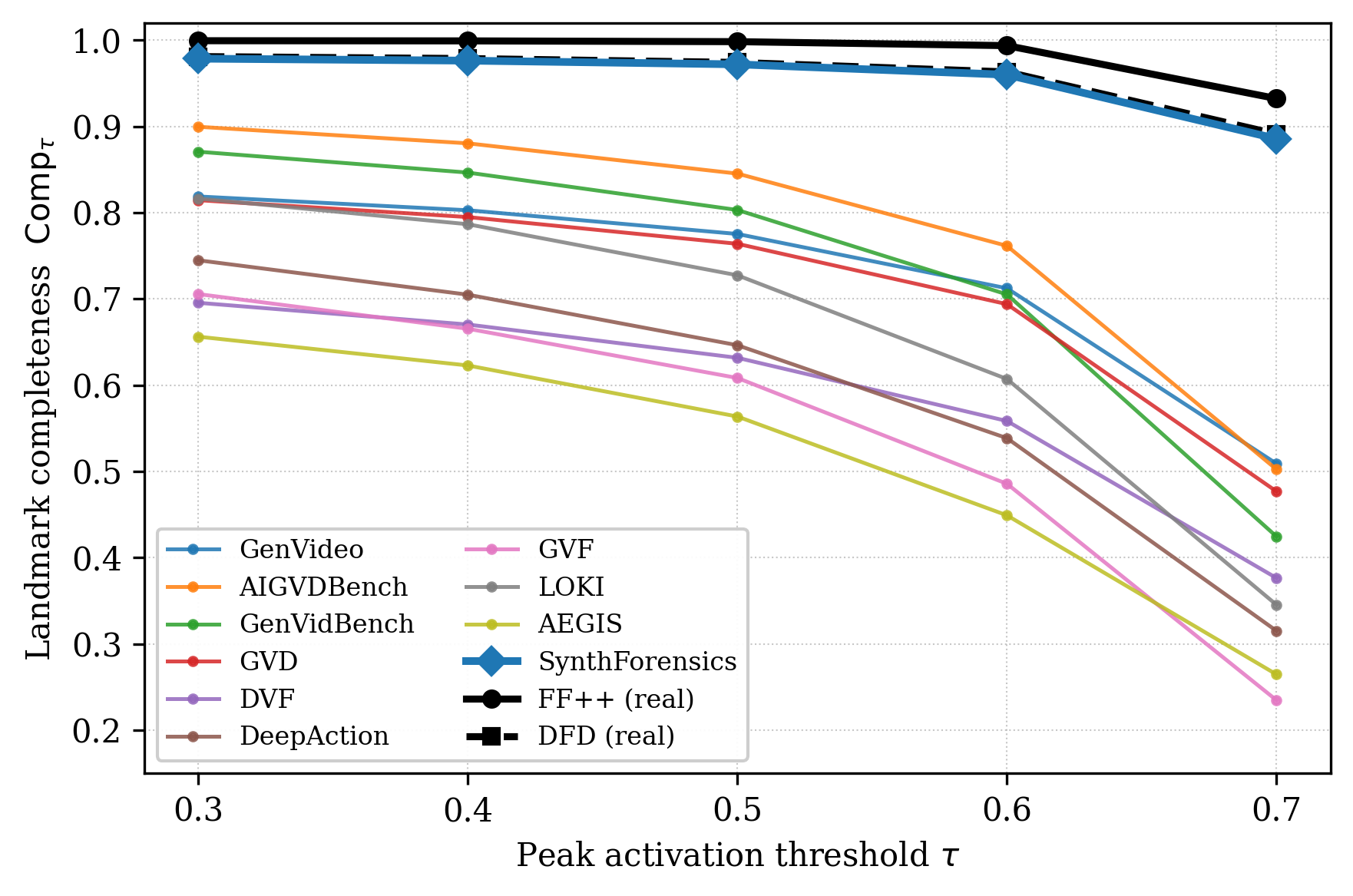}
\captionof{figure}{Mean landmark completeness $\mathrm{Comp}_\tau$ vs threshold $\tau$.}
\label{fig:landmark_curves}
\end{minipage}
\hfill
\begin{minipage}[b]{0.48\linewidth}
\centering
\includegraphics[width=\linewidth]{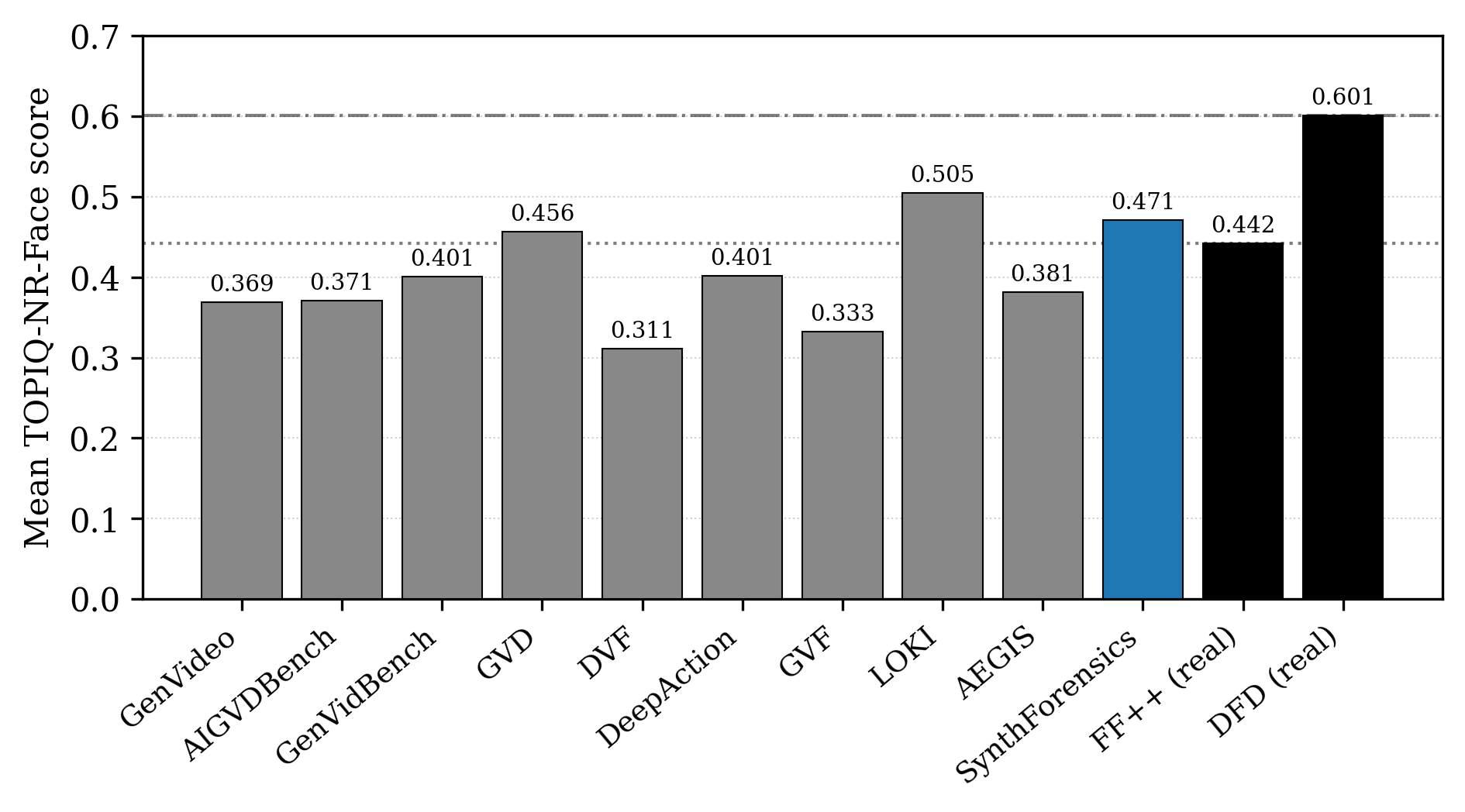}
\captionof{figure}{Per-benchmark mean TOPIQ-NR-Face score $Q$, with FF++ and DFD baselines marked.}
\label{fig:face_quality_bars}
\end{minipage}
\end{figure}

Let $\mathcal{F}_v$ denote the set of 32 frames uniformly sampled from video $v$, and let $\mathcal{F}_v^{\mathrm{det}} \subseteq \mathcal{F}_v$ denote the subset of those frames on which FAN's upstream detector finds a face. For each detected frame $f \in \mathcal{F}_v^{\mathrm{det}}$, let $\{a_{v,f,\ell}\}_{\ell=1}^{68}$, with $a_{v,f,\ell} \in [0,1]$, denote the 68 per-landmark peak activations returned by FAN. We define landmark completeness at threshold $\tau$ as
\begin{equation}
\label{eq:comp_tau}
\mathrm{Comp}_\tau(v) = \frac{1}{|\mathcal{F}_v^{\mathrm{det}}|} \sum_{f \in \mathcal{F}_v^{\mathrm{det}}} \mathbf{1}\!\left[\min_{\ell \in \{1,\dots,68\}} a_{v,f,\ell} \geq \tau\right], \qquad |\mathcal{F}_v^{\mathrm{det}}| > 0,
\end{equation}
i.e.\ the fraction of face-detected frames on which every one of the 68 landmarks reaches peak activation at least $\tau$. Taking the $\min$ over landmarks, rather than a mean, reflects the intuition that a single weakly localized landmark already signals partial loss of face structure. The per-dataset value is the mean of $\mathrm{Comp}_\tau(v)$ over all videos for which $|\mathcal{F}_v^{\mathrm{det}}| > 0$, that is, videos on which FAN detects a face on at least one of the 32 sampled frames (Appendix~\ref{app:comparative-preprocessing}). Figure~\ref{fig:landmark_curves} plots the per-benchmark mean of $\mathrm{Comp}_\tau$ for $\tau \in \{0.3, 0.4, 0.5, 0.6, 0.7\}$.

The three curves corresponding to FF++, DFD, and SynthForensics in Figure~\ref{fig:landmark_curves} remain within five percentage points of one another across the entire range of $\tau$. At $\tau=0.7$ the values are $0.932$ for FF++, $0.891$ for DFD, and $0.886$ for SynthForensics, whereas the best benchmark (GenVideo) reaches only $0.509$; the remaining eight benchmark curves fall further below, down to $0.235$ for GVF. The gap between SynthForensics and the best benchmark therefore exceeds $37$ percentage points at $\tau=0.7$. Experimental details and per-benchmark breakdowns are in Appendix~\ref{app:comparative-landmark}.

\newpage

\subsection{Face Quality Assessment}
\label{subsec:face_quality}

Landmark stability (Section~\ref{subsec:landmark_stability}) captures the geometric coherence of a face but not its perceptual quality. A synthetic face can be anatomically well-placed and still exhibit visible artifacts that leave landmark geometry intact: blurred skin texture, inconsistent lighting, compression banding, or other perceptual distortions. We quantify this second dimension through TOPIQ-NR-Face, the no-reference face variant of the TOPIQ image-quality estimator~\citep{chen2024topiq}, fine-tuned on CGFIQA-40k~\citep{chen2024dslfiqa}, $40{,}000$ real face images annotated with mean opinion scores for perceptual degradations (lighting, compression, blur, defocus, camera motion). The metric produces a scalar score in $[0,1]$ correlated with perceptual quality and is by design independent of authenticity.

Let $\mathcal{F}_v$ denote the set of 32 frames uniformly sampled from video $v$, and let $\mathcal{F}_v^{\mathrm{det}} \subseteq \mathcal{F}_v$ denote the subset of those frames on which TOPIQ-NR-Face's internal detector finds a face. For each detected frame $f \in \mathcal{F}_v^{\mathrm{det}}$, the metric crops the face region and returns a per-frame score $s_{v,f} \in [0,1]$. The per-video quality is the mean over detected frames,
\begin{equation}
\label{eq:face_quality}
Q(v) = \frac{1}{|\mathcal{F}_v^{\mathrm{det}}|} \sum_{f \in \mathcal{F}_v^{\mathrm{det}}} s_{v,f}, \qquad |\mathcal{F}_v^{\mathrm{det}}| > 0.
\end{equation}
The per-dataset value is the mean of $Q(v)$ over all videos for which $|\mathcal{F}_v^{\mathrm{det}}| > 0$, that is, videos on which the detector finds a face on at least one of the 32 sampled frames (Appendix~\ref{app:comparative-preprocessing}).

Figure~\ref{fig:face_quality_bars} reports the per-benchmark means. SynthForensics scores $0.471$, between the real-video baselines (FF++ at $0.442$, DFD at $0.601$). Seven benchmarks fall below both real baselines, ranging from $0.311$ (DVF) to $0.401$ (GenVidBench and DeepAction); two more (GVD at $0.456$, LOKI at $0.505$) sit above FF++ but below DFD. LOKI is the only benchmark whose mean exceeds SynthForensics on point estimate, but its $95\%$ CI overlaps ours at the $n=64$ face-positive subset, and its generator pool is dominated by closed-source models (five of seven) explicitly excluded from SynthForensics (Section~\ref{sec:sf_bench}). Experimental details and per-generator breakdowns are in Appendix~\ref{app:comparative-face-quality}.

\subsection{Human Perceptual Study}
\label{subsec:human_study}

To complement the automatic analyses of Sections~\ref{subsec:landmark_stability} and~\ref{subsec:face_quality}, we conduct a paired-comparison human study following ITU-T P.910~\citep{itu_p910_2023} and T2V-HE~\citep{zhang2024t2vhe}. The stimulus pool consists of 110 video pairs sampled with a fixed seed for reproducibility. Of these, 100 pair SynthForensics with one of the nine benchmarks of Section~\ref{sec:comparative} and 10 with a real video from FF++ or DFD; pairs are evenly distributed across all sources, and all videos in the pool are drawn from the intersection of the face-positive subsets of Sections~\ref{subsec:landmark_stability} and~\ref{subsec:face_quality} so that every stimulus admits a valid face localization under both pipelines. Each session is conducted remotely in an uncontrolled environment.
After an instruction-based training page~\citep{zhang2024t2vhe} and a short demographic questionnaire (age bracket, AI expertise, deepfake familiarity), the participant rates 35 pairs in random order.
Per pair, participants answer three questions: Q1, ``Which video is of better overall quality?''; Q2, ``Which video appears more realistic?''; Q3, ``Which video could be a fake?''. Q1 and Q2 use $\{A, B, \mathrm{Equal}\}$. Q3 uses $\{A, B, \mathrm{Both}, \mathrm{Neither}\}$: prior work on human deepfake detection shows that participants default to ``real'' when uncertain~\citep{groh2022deepfake}, so a 3-option scheme conflates no-detection (neither is fake) with doubt.

\begin{figure}[t]
\centering
\includegraphics[width=\linewidth]{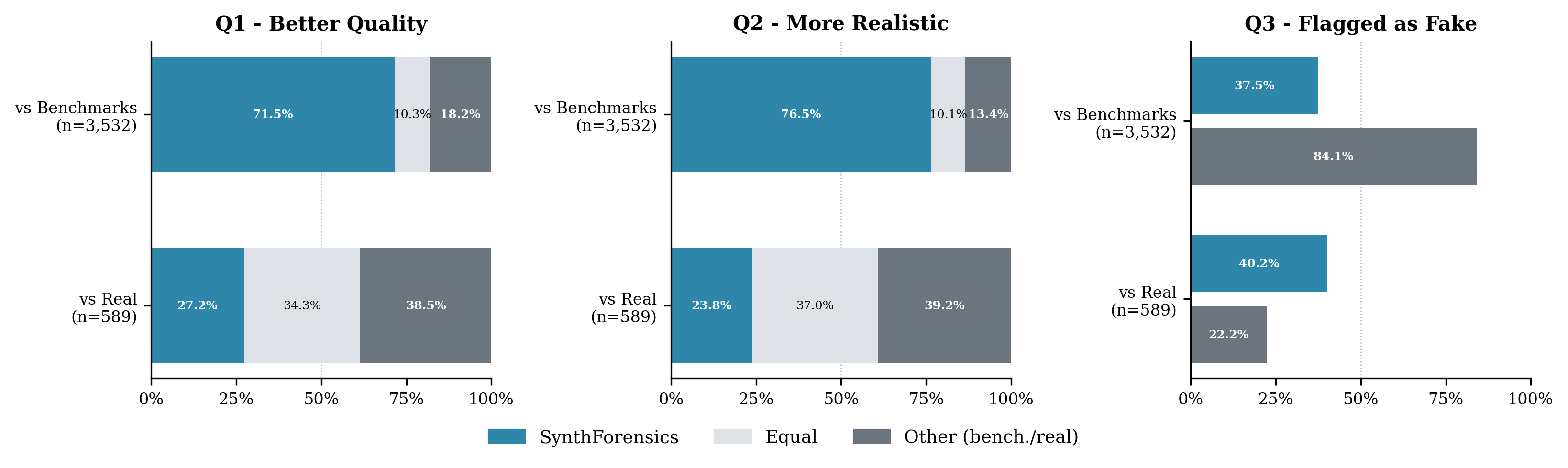}
\caption{Human-study results. Q1 (Better Quality) and Q2 (More Realistic) are stacked $100\%$ bars; Q3 (Flagged as Fake) reports binary fractions. $95\%$ cluster-bootstrap CI in Appendix~\ref{app:comparative-human-study}.}
\label{fig:human_study}
\end{figure}

Of the 159 recruited participants, 118 passed our screening procedure (Appendix~\ref{app:comparative-human-study}), yielding 4{,}121 ratings per question (approximately 37 per pair). For Q1 and Q2 we report \emph{win rates}~\citep{david1988method}, the fraction of pairs in which SynthForensics is preferred over the paired video, while for Q3 we report the binary fraction of videos flagged as fake. Per-benchmark point estimates with 95\% cluster-bootstrap confidence intervals over participants~\citep{efron1979bootstrap} and inter-rater agreement (Krippendorff's $\alpha$~\citep{krippendorff2011computing}) are reported in Appendix~\ref{app:comparative-human-study}; for clarity, Figure~\ref{fig:human_study} aggregates the nine per-benchmark estimates into a single \emph{vs Benchmarks} group and the FF++/DFD pristine pool into \emph{vs Real}. Against existing benchmarks, SynthForensics is preferred by a wide margin both on overall quality (Q1, $71.5\%$) and on realism (Q2, $76.5\%$), and is flagged as fake less than half as often as the competing benchmark videos (Q3, $37.5\%$ vs.\ $84.1\%$; see Appendix~\ref{app:human-study-agreement} for robustness analyses). Against pristine FF++/DFD videos, SynthForensics narrows but does not close the gap with real footage (Q1, $27.2\%$; Q2, $23.8\%$); the corresponding Q3 detection rate ($40.2\%$) sits near the ``cannot decide'' regime~\citep{groh2022deepfake}, indicating that participants struggle to reliably mark SynthForensics videos as synthetic when paired against real footage.

\section{Detection Experiments}
\label{sec:detection}

We evaluate SynthForensics as a benchmark for modern deepfake detectors confronted with purely synthetic human-centered video. Our evaluation leverages SynthForensics alongside three datasets (FF++, DFD, and CDF) providing real counterparts and legacy manipulations used in this section.

\textbf{Training and Validation Sets.} Both pair reals from FF++~\citep{rossler2019faceforensics++} official splits with their SynthForensics counterparts: \textbf{\texttt{SynthForensics-Train}} (train split) drives optimization, and \textbf{\texttt{SynthForensics-Val}} (val split) guides hyperparameter selection and early stopping.

\textbf{Primary Evaluation Sets.} Three testbeds introduce progressively stronger distribution shifts:
\begin{itemize}
    \item \textbf{\texttt{SF-FF++}} (paired baseline): reals from the FF++~\citep{rossler2019faceforensics++} official test split paired with their SynthForensics counterparts; the controlled setting with minimal confounding variables, where both real source and synthetic semantics coincide.
    \item \textbf{\texttt{SF-DFD}} (provenance shift): all 363 DFD~\citep{dufour2019google} reals (lab recordings) paired with their SynthForensics counterparts; isolates whether detectors rely on FF++-specific real-domain artifacts rather than on synthetic structure.
    \item \textbf{\texttt{SF-CDF}} (compound shift): 178 reals from the CDF~\citep{li2020celeb} official test split and 178 randomly sampled SynthForensics videos per generator (89 from SF-FF++, 89 from SF-DFD). This deliberately unpaired testbed assesses robustness when both real content distribution (celebrity footage vs.\ YouTube/lab) and synthetic semantic alignment shift simultaneously.
\end{itemize}
To ensure fair comparison across compression levels, real videos in all Primary Evaluation Sets are re-encoded with the same compression pipeline applied to synthetic videos (Section~\ref{subsec:dataset_statistics}), yielding four matched compression versions (\textit{Raw}, \textit{Canonical}, \textit{CRF23}, \textit{CRF40}).

\textbf{Legacy Benchmark Sets.} To assess backward compatibility and transfer to traditional manipulation-based deepfakes, we evaluate on the reals and manipulations from the official test splits of three benchmarks: \textbf{\texttt{FF++}}, \textbf{\texttt{DFD}}, and \textbf{\texttt{CDF}}.

\newpage

We adopt video-level Area Under the ROC Curve (AUC) as our primary metric, providing threshold-independent assessment; frame-level predictions are aggregated to video-level via mean pooling. For the detectors surveyed in Section~\ref{subsec:detection_methods}, we use the authors' official implementations and weights where available, supplemented by the DeepfakeBench~\citep{yan2023deepfakebench} framework for reproducibility; UNITE~\citep{kundu2025towards} (no public code or weights) and DeMamba~\citep{chen2024demamba} (no weights) are excluded. Details in Appendix~\ref{app:detection-setup}.

\subsection{Zero-Shot Evaluation}
\label{subsec:zero_shot}

\begin{table*}[t]
\centering
\caption{Zero-shot performance (AUC \%) on Legacy Benchmark Sets and Primary Evaluation Sets. \textsuperscript{*}Frame-level detector. \textsuperscript{\textdagger}Our evaluation, \textsuperscript{\textdaggerdbl}third-party results~\citep{yan2024orthogonal,chen2024x2}.}
\label{tab:zeroshot_comprehensive}
\resizebox{\textwidth}{!}{
\begin{tabular}{l l | ccc | ccc | S[table-format=-2.2] S[table-format=-2.2]}
\toprule
& & \multicolumn{3}{c|}{\textbf{Legacy Benchmark Sets}} & \multicolumn{3}{c|}{\textbf{Primary Evaluation Sets}} & \multicolumn{2}{c}{\textbf{Performance Gap}} \\
\cmidrule(lr){3-5} \cmidrule(lr){6-8} \cmidrule(lr){9-10}
\textbf{Detector} & \textbf{Trained on} & \textbf{FF++} & \textbf{DFD} & \textbf{CDF} & \textbf{SF-FF++} & \textbf{SF-DFD} & \textbf{SF-CDF} & {\textbf{vs. FF++}} & {\textbf{Mean}} \\
\midrule
CFM\textsuperscript{*}         & FF++                                                   & 99.56 & 95.21 & 89.65 & 73.54 & 72.41 & 66.03 & -26.02 & -24.15 \\
RECCE\textsuperscript{*}       & \makecell[l]{FF++, DFDC,\\ CDF, WildDF}                & 99.15\textsuperscript{\textdagger} & 89.10\textsuperscript{\textdaggerdbl} & 99.94 & 79.50 & 78.63 & 72.92 & -19.65 & -19.05 \\
ProDet\textsuperscript{*}      & FF++                                                   & 98.77\textsuperscript{\textdagger} & 90.10\textsuperscript{\textdaggerdbl} & 84.48 & 75.35 & 67.74 & 69.42 & -23.41 & -20.28 \\
UCF\textsuperscript{*}         & FF++                                                   & 99.50\textsuperscript{\textdagger} & 94.50 & 82.40\textsuperscript{\textdaggerdbl} & 75.41 & 74.32 & 75.39 & -24.09 & -17.09 \\
Effort\textsuperscript{*}      & FF++                                                   & 98.11\textsuperscript{\textdagger} & 96.50 & 95.60 & 70.31 & 64.53 & 60.21 & -27.79 & -31.72 \\
LAA-Net\textsuperscript{*}     & FF++                                                   & 99.96 & 98.43 & 95.40 & 69.42 & 59.07 & 47.48 & -30.54 & -39.27 \\
GenD\textsuperscript{*}        & FF++                                                   & 98.97\textsuperscript{\textdagger} & 97.00 & 96.00 & 82.35 & 82.27 & 68.83 & -16.62 & -19.51 \\
\addlinespace
AltFreezing                    & FF++                                                   & 98.60 & 98.50 & 89.50 & 56.12 & 71.65 & 50.12 & -42.48 & -36.23 \\
FTCN                           & FF++                                                   & 99.70 & 94.40\textsuperscript{\textdaggerdbl} & 86.90 & 44.76 & 59.82 & 53.69 & -54.94 & -40.91 \\
GenConViT                      & \makecell[l]{FF++, DFDC, TM,\\ CDF, TIMIT}             & 99.60 & 99.95\textsuperscript{\textdagger} & 98.10 & 83.61 & 90.68 & 65.74 & -15.99 & -19.21 \\
DFD-FCG                        & FF++                                                   & 99.57 & 92.94\textsuperscript{\textdagger} & 95.00 & 86.57 & 82.63 & 77.05 & -13.00 & -13.75 \\
FakeSTormer                    & FF++                                                   & 99.90 & 98.90 & 96.50 & 72.59 & 72.83 & 64.30 & -27.31 & -28.53 \\
\midrule
MM-Det                         & DVF                                                    & ---   & ---   & ---   & 49.17 & 55.16 & 41.26 & {---}  & {---}  \\
NSG-VD                         & \makecell[l]{K400~\citep{kay2017kinetics},\\ Pika~\citep{pika_labs}} & ---   & ---   & ---   & 58.95 & 52.13 & 70.58 & {---}  & {---}  \\
D3                             & ---                                                    & ---   & ---   & ---   & 48.85 & 49.40 & 61.80 & {---}  & {---}  \\
\bottomrule
\end{tabular}
}
\end{table*}

Most current deepfake detectors were built for manipulation-based content, but the emerging threat is purely synthetic video. This protocol quantifies how much of their original capability survives the shift. To this end, we evaluate the 12 face-based detectors of Section~\ref{subsec:detection_methods} zero-shot on both Primary Evaluation Sets and Legacy Benchmark Sets. We additionally test 3 detectors purpose-built for synthetic video detection on the Primary Evaluation Sets (Legacy sets fall outside the design scope of synthetic-video generic methods) to ensure observed gap is not specific to face-forensics paradigm.

Table~\ref{tab:zeroshot_comprehensive} shows that face-based detectors lose on average 27 AUC points moving from FF++ Legacy to SF-FF++, with FTCN and AltFreezing collapsing by over 40 points; gaps widen further on SF-DFD and SF-CDF, with LAA-Net falling to 47\% on the compound-shift set. Purpose-built synthetic-video methods (MM-Det, NSG-VD, D3) perform poorly as well, with AUC between 41\% and 71\% across all Primary Sets and frequently at or below random chance. A compression ablation across the standardized \textit{Canonical}, \textit{CRF23}, and \textit{CRF40} versions of every Primary Evaluation Set shows the average AUC across the 15 detectors on SF-FF++ collapsing from 68\% (\textit{Canonical}) to 45\% (\textit{CRF40}). Full details and ablation studies are in Appendix~\ref{app:detection-zeroshot}.

\subsection{Fine-Tuning}
\label{subsec:fine_tuning}

\begin{table}[t]
\centering
\begin{minipage}[t]{0.49\linewidth}
\centering
\caption{Fine-tuning efficacy: zero-shot vs.\ fine-tuned AUC on SF-FF++ (Gain $=$ FT $-$ ZS) and backward AUC on FF++. \textsuperscript{*}Frame-level detector.}
\label{tab:finetuning_results}
\footnotesize
\begin{tabular*}{\linewidth}{@{\extracolsep{\fill}} l c c c c}
\toprule
\textbf{Detector} & \makecell{\textbf{Zero-}\\\textbf{Shot}} & \makecell{\textbf{Fine-}\\\textbf{Tuned}} & \textbf{Gain} & \textbf{Backward} \\
\midrule
RECCE\textsuperscript{*}       & 79.50 & 98.04 & +18.54 & 88.83 \\
ProDet\textsuperscript{*}      & 75.35 & 96.14 & +20.79 & 84.15 \\
UCF\textsuperscript{*}         & 75.41 & 96.77 & +21.36 & 92.55 \\
Effort\textsuperscript{*}      & 70.31 & 96.40 & +26.09 & 92.36 \\
GenD\textsuperscript{*}        & 82.35 & 80.60 & $-1.75$ & 98.23 \\
AltFreezing                    & 56.12 & 99.12 & +43.00 & 65.52 \\
FTCN                           & 44.76 & 98.36 & +53.60 & 57.07 \\
GenConViT                      & 83.61 & 97.56 & +13.95 & 86.39 \\
DFD-FCG                        & 86.57 & 92.81 & $+6.24$ & 99.37 \\
\bottomrule
\end{tabular*}
\end{minipage}%
\hfill
\begin{minipage}[t]{0.49\linewidth}
\centering
\caption{Training from scratch efficacySS: in-domain, out-of-domain, and legacy transfer AUC \%. \textsuperscript{*}Frame-level detector.}
\label{tab:retraining_results}
\footnotesize
\begin{tabular*}{\linewidth}{@{\extracolsep{\fill}} l c c c}
\toprule
\textbf{Detector} & \makecell{\textbf{In-}\\\textbf{Domain}} & \makecell{\textbf{Out-of-}\\\textbf{Domain}} & \textbf{Legacy} \\
\midrule
RECCE\textsuperscript{*}       & 97.21 & 94.65 & 65.25 \\
ProDet\textsuperscript{*}      & 78.91 & 80.92 & 54.57 \\
UCF\textsuperscript{*}         & 96.84 & 94.33 & 60.82 \\
Effort\textsuperscript{*}      & 97.83 & 94.64 & 62.97 \\
GenD\textsuperscript{*}        & 56.81 & 55.47 & 50.11 \\
AltFreezing                    & 99.24 & 98.78 & 63.08 \\
FTCN                           & 98.14 & 97.74 & 53.75 \\
GenConViT                      & 94.36 & 87.58 & 60.19 \\
DFD-FCG                        & 95.52 & 90.19 & 83.41 \\
\bottomrule
\end{tabular*}
\end{minipage}
\end{table}

Fine-tuning asks how much of the zero-shot gap of Section~\ref{subsec:zero_shot} can be closed when detectors are given direct access to SynthForensics, and at what cost to their original manipulation-detection capability. We fine-tune with a plain protocol, without retention-preserving techniques such as student-teacher distillation, so that the backward drop on Legacy Benchmark Sets is measured directly rather than masked by the training scheme.

We fine-tune nine face-based detectors on SynthForensics-Train: RECCE, ProDet, UCF, Effort, and GenD at frame-level; AltFreezing, FTCN, GenConViT, and DFD-FCG at video-level. The other six are excluded by construction (Appendix~\ref{app:detection-trained-pool}): CFM (no training code), LAA-Net and FakeSTormer (blending-mask supervision), D3 (training-free), NSG-VD (non-parametric MMD), MM-Det (diffusion-only design). Fine-tuned detectors are evaluated at \textit{Raw} compression on SF-FF++ to assess learnability and on FF++ to measure backward compatibility (Table~\ref{tab:finetuning_results}).

Fine-tuning closes the zero-shot gap on SF-FF++ for nearly every detector, lifting AUC to $96$--$99\%$ (GenD is the sole outlier, with LayerNorm-only adaptation losing $1.75$pp). Backward compatibility on FF++ separates the pool: video-level AltFreezing and FTCN, with the largest gains ($+43$ and $+54$pp), pay the steepest cost ($65.5\%$ and $57.1\%$); frame-level methods retain $84$--$99\%$ of FF++ capability. Full fine-tuning details and results are in Appendix~\ref{app:detection-finetuning}.

\subsection{Training from Scratch}
\label{subsec:re_training}

To push the analysis further and simulate the rapid emergence of new generative architectures, we train detectors from scratch on a partition of the SynthForensics-Train generator pool, holding out a subset as unseen at training time. The training partition, the \emph{In-Domain} pool, consists of CogVideoX, Wan2.1, LTX-2.3, and MAGI-1 in both T2V and I2V modalities, covering the dominant DiT and autoregressive paradigms. The held-out \emph{Out-of-Domain} pool, never seen during training, comprises SkyReels-V2, Self-Forcing, Helios, and daVinci-MagiHuman, covering more recent paradigms (diffusion forcing, distilled causal, autoregressive continuation, audio-visual).

We train from scratch the same nine face-based detectors as in Section~\ref{subsec:fine_tuning}, with identical inclusion and exclusion criteria (Appendix~\ref{app:detection-trained-pool}). Evaluation is performed along the three axes of Table~\ref{tab:retraining_results} on \textit{Raw}-compression test samples never seen during training. \emph{In-Domain} reports the mean AUC on the SF-FF++ test split restricted to the generators used for training, measuring performance on familiar generators but novel samples. \emph{Out-of-Domain} reports the mean AUC on the SF-FF++ test split restricted to the held-out generators, measuring performance on generators never seen during training. \emph{Legacy} reports the mean AUC on the Legacy Benchmark Sets (FF++, DFD, CDF), measuring backward compatibility with traditional benchmarks.

Training from scratch closes the In-Domain and Out-of-Domain gap for seven detectors, all reaching $87$--$99\%$ AUC, with video-level methods (AltFreezing, FTCN) holding within $0.5$pp across the two pools. ProDet ($78.9/80.9$) and GenD ($56.8/55.5$) lag behind, both losing substantial ground relative to fine-tuning. Legacy transfer collapses uniformly to $50$--$65\%$, with DFD-FCG the only exception ($83.4\%$): the discriminative features learned on fully-synthetic videos do not generalize to traditional manipulation benchmarks. Full training-from-scratch details and results are in Appendix~\ref{app:detection-retraining}.

\section{Conclusion}
\label{sec:conclusion}

We presented SynthForensics, a people-centric benchmark of $20{,}445$ videos from 8 T2V and 7 I2V open-source generators, paired-source from $1{,}363$ FF++/DFD reals, human-validated, in four compression versions with full reproducibility metadata. Our comparative analysis ranks SynthForensics above each of the nine existing synthetic-video benchmarks on landmark stability and human paired comparisons (win rates above $71\%$ on quality and realism, fake-flag rate $\sim 38\%$ vs $\sim 84\%$); face quality in the FF++/DFD baseline range. Face-based detectors drop $13$--$55$pp (mean $27$) zero-shot on SF-FF++ and $23$pp more under \textit{CRF40}; fine-tuning recovers the gap at a backward cost on legacy benchmarks; training from scratch reveals that synthetic-detection and manipulation-detection feature spaces barely overlap for most detectors. Dataset, pipeline, and code are publicly available.

\newpage

{
\bibliographystyle{plainnat}
\bibliography{references}

@String(CVPR= {IEEE Conf. Comput. Vis. Pattern Recog.})

@String(NIPS= {Adv. Neural Inform. Process. Syst.})

@String(ICME = {Int. Conf. Multimedia and Expo})

@String(ICLR = {Int. Conf. Learn. Represent.})

@String(AAAI = {AAAI})

@String(CVPR  = {CVPR})

@String(NIPS  = {NeurIPS})

@String(ICME  =	{ICME})

@String(ICLR  = {ICLR})

@misc{huang2025selfforcingcode,
  author       = {Huang, Xun and Li, Zhengqi and He, Guande and Zhou, Mingyuan and Shechtman, Eli},
  title        = {{Self-Forcing: Official Implementation}},
  year         = {2025},
  howpublished = {\url{https://github.com/guandeh17/Self-Forcing}},
  note         = {Accessed: 2025-11-09}
}

@misc{dufour2019google,
  author       = {Dufour, Nick and Gully, Andrew},
  title        = {{Contributing Data to Deepfake Detection Research}},
  year         = {2019},
  howpublished = {\url{https://ai.googleblog.com/2019/09/contributing-data-to-deepfake-detection.html}},
  note         = {Accessed: 2025-11-05}
}

@misc{kling2,
  author       = {{Kuaishou Technology}},
  title        = {{Kling}},
  year         = {2024},
  howpublished = {\url{https://klingai.com/global/}},
  note         = {Accessed: 2025-11-05}
}

@misc{sora2,
  author       = {{OpenAI}},
  title        = {{Sora 2}},
  year         = {2024},
  howpublished = {\url{https://openai.com/index/sora-2/}},
  note         = {Accessed: 2025-11-05}
}

@misc{veo3,
  author       = {{Google}},
  title        = {{New Veo Updates and More AI Progress in Video and Audio}},
  year         = {2024},
  howpublished = {\url{https://blog.google/technology/ai/veo-updates-flow/}},
  note         = {Accessed: 2025-11-05}
}

@article{wan2025wan,
  title={Wan: Open and advanced large-scale video generative models},
  author={Wan, Team and Wang, Ang and Ai, Baole and Wen, Bin and Mao, Chaojie and Xie, Chen-Wei and Chen, Di and Yu, Feiwu and Zhao, Haiming and Yang, Jianxiao and others},
  journal={arXiv preprint arXiv:2503.20314},
  year={2025}
}

@article{chen2025skyreels,
  title={Skyreels-v2: Infinite-length film generative model},
  author={Chen, Guibin and Lin, Dixuan and Yang, Jiangping and Lin, Chunze and Zhu, Junchen and Fan, Mingyuan and Zhang, Hao and Chen, Sheng and Chen, Zheng and Ma, Chengcheng and others},
  journal={arXiv preprint arXiv:2504.13074},
  year={2025}
}

@article{yang2024cogvideox,
  title={Cogvideox: Text-to-video diffusion models with an expert transformer},
  author={Yang, Zhuoyi and Teng, Jiayan and Zheng, Wendi and Ding, Ming and Huang, Shiyu and Xu, Jiazheng and Yang, Yuanming and Hong, Wenyi and Zhang, Xiaohan and Feng, Guanyu and others},
  journal={arXiv preprint arXiv:2408.06072},
  year={2024}
}

@article{teng2025magi,
  title={Magi-1: Autoregressive video generation at scale},
  author={Teng, Hansi and Jia, Hongyu and Sun, Lei and Li, Lingzhi and Li, Maolin and Tang, Mingqiu and Han, Shuai and Zhang, Tianning and Zhang, WQ and Luo, Weifeng and others},
  journal={arXiv preprint arXiv:2505.13211},
  year={2025}
}

@inproceedings{yin2025causvid,
    title={From Slow Bidirectional to Fast Autoregressive Video Diffusion Models},
    author={Yin, Tianwei and Zhang, Qiang and Zhang, Richard and Freeman, William T and Durand, Fredo and Shechtman, Eli and Huang, Xun},
    booktitle={CVPR},
    year={2025}
}

@article{guo2023animatediff,
  title={Animatediff: Animate your personalized text-to-image diffusion models without specific tuning},
  author={Guo, Yuwei and Yang, Ceyuan and Rao, Anyi and Liang, Zhengyang and Wang, Yaohui and Qiao, Yu and Agrawala, Maneesh and Lin, Dahua and Dai, Bo},
  journal={arXiv preprint arXiv:2307.04725},
  year={2023}
}

@article{ma2025step,
  title={{Step-Video-T2V Technical Report: The Practice, Challenges, and Future of Video Foundation Model}},
  author={Ma, Guoqing and Huang, Haoyang and Yan, Kun and Chen, Liangyu and Duan, Nan and Yin, Shengming and others},
  journal={arXiv preprint arXiv:2502.10248},
  year={2025}
}

@article{kong2024hunyuanvideo,
  title={Hunyuanvideo: A systematic framework for large video generative models},
  author={Kong, Weijie and Tian, Qi and Zhang, Zijian and Min, Rox and Dai, Zuozhuo and Zhou, Jin and Xiong, Jiangfeng and Li, Xin and Wu, Bo and Zhang, Jianwei and others},
  journal={arXiv preprint arXiv:2412.03603},
  year={2024}
}

@article{goodfellow2014generative,
  title={Generative adversarial nets},
  author={Goodfellow, Ian J and Pouget-Abadie, Jean and Mirza, Mehdi and Xu, Bing and Warde-Farley, David and Ozair, Sherjil and Courville, Aaron and Bengio, Yoshua},
  journal={Advances in neural information processing systems},
  volume={27},
  year={2014}
}

@article{ho2020denoising,
  title={Denoising diffusion probabilistic models},
  author={Ho, Jonathan and Jain, Ajay and Abbeel, Pieter},
  journal={Advances in neural information processing systems},
  volume={33},
  pages={6840--6851},
  year={2020}
}

@article{deressa2025genconvit,
  title={GenConViT: Deepfake video detection using generative convolutional vision transformer},
  author={Deressa, Deressa Wodajo and Mareen, Hannes and Lambert, Peter and Atnafu, Solomon and Akhtar, Zahid and Van Wallendael, Glenn},
  journal={Applied Sciences},
  volume={15},
  number={12},
  pages={6622},
  year={2025},
  publisher={MDPI}
}

@inproceedings{han2025towards,
  title={Towards more general video-based deepfake detection through facial component guided adaptation for foundation model},
  author={Han, Yue-Hua and Huang, Tai-Ming and Hua, Kai-Lung and Chen, Jun-Cheng},
  booktitle={Proceedings of the IEEE/CVF conference on computer vision and pattern recognition},
  pages={22995--23005},
  year={2025}
}

@inproceedings{ni2026genvidbench,
  title={GenVidBench: A 6-Million Benchmark for AI-Generated Video Detection},
  author={Ni, Zhenliang and Yan, Qiangyu and Huang, Mouxiao and Yuan, Tianning and Tang, Yehui and Hu, Hailin and Chen, Xinghao and Wang, Yunhe},
  booktitle={Proceedings of the AAAI Conference on Artificial Intelligence},
  volume={40},
  number={18},
  pages={15582--15590},
  year={2026}
}

@inproceedings{cai2025av,
  title={Av-deepfake1m++: A large-scale audio-visual deepfake benchmark with real-world perturbations},
  author={Cai, Zhixi and Kuckreja, Kartik and Ghosh, Shreya and Chuchra, Akanksha and Khan, Muhammad Haris and Tariq, Usman and Gedeon, Tom and Dhall, Abhinav},
  booktitle={Proceedings of the 33rd ACM International Conference on Multimedia},
  pages={13686--13691},
  year={2025}
}

@inproceedings{rossler2019faceforensics++,
  title={{FaceForensics++: Learning to Detect Manipulated Facial Images}},
  author={Rössler, Andreas and Cozzolino, Davide and Verdoliva, Luisa and Riess, Christian and Thies, Justus and Niessner, Matthias},
  booktitle={Proceedings of the IEEE/CVF International Conference on Computer Vision},
  pages={1--11},
  year={2019}
}

@article{dolhansky2020deepfake,
  title={The deepfake detection challenge (dfdc) dataset},
  author={Dolhansky, Brian and Bitton, Joanna and Pflaum, Ben and Lu, Jikuo and Howes, Russ and Wang, Menglin and Ferrer, Cristian Canton},
  journal={arXiv preprint arXiv:2006.07397},
  year={2020}
}

@inproceedings{li2020celeb,
  title={{Celeb-DF: A Large-Scale Challenging Dataset for DeepFake Forensics}},
  author={Li, Yuezun and Yang, Xin and Sun, Pu and Qi, Honggang and Lyu, Siwei},
  booktitle={Proceedings of the IEEE/CVF Conference on Computer Vision and Pattern Recognition},
  pages={3204-3213},
  year={2020}
}

@article{li2025celeb,
  title={Celeb-df++: A large-scale challenging video deepfake benchmark for generalizable forensics},
  author={Li, Yuezun and Zhu, Delong and Cui, Xinjie and Lyu, Siwei},
  journal={arXiv preprint arXiv:2507.18015},
  year={2025}
}

@inproceedings{jiang2020deeperforensics,
  title={Deeperforensics-1.0: A large-scale dataset for real-world face forgery detection},
  author={Jiang, Liming and Li, Ren and Wu, Wayne and Qian, Chen and Loy, Chen Change},
  booktitle={Proceedings of the IEEE/CVF conference on computer vision and pattern recognition},
  pages={2889--2898},
  year={2020}
}

@inproceedings{zi2020wilddeepfake,
  title={Wilddeepfake: A challenging real-world dataset for deepfake detection},
  author={Zi, Bojia and Chang, Minghao and Chen, Jingjing and Ma, Xingjun and Jiang, Yu-Gang},
  booktitle={Proceedings of the 28th ACM international conference on multimedia},
  pages={2382--2390},
  year={2020}
}

@article{zheng2025open,
  title={{Open-Sora 2.0: Training a Commercial-Level Video Generation Model in \$200k}},
  author={Zheng, Zangwei and Peng, Xiangyu and Lou, Yuxuan and Shen, Chenhui and Young, Tom and Guo, Xinying and Wang, Binluo and Xu, Hang and Liu, Hongxin and Jiang, Mingyan and others},
  journal={arXiv preprint arXiv:2503.09642},
  year={2025}
}

@article{luo2023beyond,
  title={Beyond the prior forgery knowledge: Mining critical clues for general face forgery detection},
  author={Luo, Anwei and Kong, Chenqi and Huang, Jiwu and Hu, Yongjian and Kang, Xiangui and Kot, Alex C},
  journal={IEEE Transactions on Information Forensics and Security},
  volume={19},
  pages={1168--1182},
  year={2023},
  publisher={IEEE}
}

@inproceedings{cao2022end,
  title={End-to-end reconstruction-classification learning for face forgery detection},
  author={Cao, Junyi and Ma, Chao and Yao, Taiping and Chen, Shen and Ding, Shouhong and Yang, Xiaokang},
  booktitle={Proceedings of the IEEE/CVF conference on computer vision and pattern recognition},
  pages={4113--4122},
  year={2022}
}

@article{cheng2024can,
  title={Can we leave deepfake data behind in training deepfake detector?},
  author={Cheng, Jikang and Yan, Zhiyuan and Zhang, Ying and Luo, Yuhao and Wang, Zhongyuan and Li, Chen},
  journal={Advances in Neural Information Processing Systems},
  volume={37},
  pages={21979--21998},
  year={2024}
}

@inproceedings{yan2023ucf,
  title={Ucf: Uncovering common features for generalizable deepfake detection},
  author={Yan, Zhiyuan and Zhang, Yong and Fan, Yanbo and Wu, Baoyuan},
  booktitle={Proceedings of the IEEE/CVF international conference on computer vision},
  pages={22412--22423},
  year={2023}
}

@inproceedings{yan2024orthogonal,
  title     = {{Orthogonal Subspace Decomposition for Generalizable AI-Generated Image Detection}},
  author    = {Zhiyuan Yan and Jiangming Wang and Peng Jin and Ke-Yue Zhang and Chengchun Liu and Shen Chen and Taiping Yao and Shouhong Ding and Baoyuan Wu and Li Yuan},
  booktitle = {Proceedings of the 42nd International Conference on Machine Learning},
  year      = {2025},
  series    = {Proceedings of Machine Learning Research},
  publisher = {PMLR}
}

@article{zhang2025videollama,
  title={{VideoLLaMA 3: Frontier Multimodal Foundation Models for Image and Video Understanding}},
  author={Boqiang Zhang and Kehan Li and Zesen Cheng and Zhiqiang Hu and Yuqian Yuan and Guanzheng Chen and Sicong Leng and Yuming Jiang and Hang Zhang and Xin Li and Peng Jin and Wenqi Zhang and Fan Wang and Lidong Bing and Deli Zhao},
  journal={arXiv preprint arXiv:2501.13106},
  year={2025}
}

@inproceedings{huang2024vbench,
  title={Vbench: Comprehensive benchmark suite for video generative models},
  author={Huang, Ziqi and He, Yinan and Yu, Jiashuo and Zhang, Fan and Si, Chenyang and Jiang, Yuming and Zhang, Yuanhan and Wu, Tianxing and Jin, Qingyang and Chanpaisit, Nattapol and others},
  booktitle={Proceedings of the IEEE/CVF Conference on Computer Vision and Pattern Recognition},
  pages={21807--21818},
  year={2024}
}

@article{huang2024vbenchpp,
  title={{VBench++}: Comprehensive and Versatile Benchmark Suite for Video Generative Models},
  author={Huang, Ziqi and Zhang, Fan and Xu, Xiaojie and He, Yinan and Yu, Jiashuo and Dong, Ziyue and Ma, Qianli and Chanpaisit, Nattapol and Si, Chenyang and Jiang, Yuming and Wang, Yaohui and Chen, Xinyuan and Chen, Ying-Cong and Wang, Limin and Lin, Dahua and Qiao, Yu and Liu, Ziwei},
  journal={arXiv preprint arXiv:2411.13503},
  year={2024}
}

@article{zheng2025vbench2,
  title={{VBench-2.0}: Advancing Video Generation Benchmark Suite for Intrinsic Faithfulness},
  author={Zheng, Dian and Huang, Ziqi and Liu, Hongbo and Zou, Kai and He, Yinan and Zhang, Fan and Gu, Lulu and Zhang, Yuanhan and He, Jingwen and Zheng, Wei-Shi and Qiao, Yu and Liu, Ziwei},
  journal={arXiv preprint arXiv:2503.21755},
  year={2025}
}

@article{huang2025self,
  title={Self forcing: Bridging the train-test gap in autoregressive video diffusion},
  author={Huang, Xun and Li, Zhengqi and He, Guande and Zhou, Mingyuan and Shechtman, Eli},
  journal={arXiv preprint arXiv:2506.08009},
  year={2025}
}

@inproceedings{peebles2023scalable,
  title={Scalable diffusion models with transformers},
  author={Peebles, William and Xie, Saining},
  booktitle={Proceedings of the IEEE/CVF international conference on computer vision},
  pages={4195--4205},
  year={2023}
}

@inproceedings{kundu2025towards,
  title={{Towards a Universal Synthetic Video Detector: From Face or Background Manipulations to Fully AI-Generated Content}},
  author={Kundu, Rohit and Xiong, Hao and Mohanty, Vishal and Balachandran, Athula and Roy-Chowdhury, Amit K.},
  booktitle={Proceedings of the Computer Vision and Pattern Recognition Conference},
  pages={28050--28060},
  year={2025}
}

@inproceedings{wang2023altfreezing,
  title={Altfreezing for more general video face forgery detection},
  author={Wang, Zhendong and Bao, Jianmin and Zhou, Wengang and Wang, Weilun and Li, Houqiang},
  booktitle={Proceedings of the IEEE/CVF conference on computer vision and pattern recognition},
  pages={4129--4138},
  year={2023}
}

@inproceedings{zheng2021exploring,
  title={Exploring temporal coherence for more general video face forgery detection},
  author={Zheng, Yinglin and Bao, Jianmin and Chen, Dong and Zeng, Ming and Wen, Fang},
  booktitle={Proceedings of the IEEE/CVF international conference on computer vision},
  pages={15044--15054},
  year={2021}
}

@article{grattafiori2024llama,
  title={{The Llama 3 Herd of Models}},
  author={Grattafiori, Aaron and Dubey, Abhimanyu and Jauhri, Abhinav and Pandey, Abhinav and Kadian, Abhishek and Al-Dahle, Ahmad and Letman, Aiesha and Mathur, Akhil and Schelten, Alan and Vaughan, Alex and others},
  journal={arXiv preprint arXiv:2407.21783},
  year={2024}
}

@article{yan2023deepfakebench,
  title={{DeepfakeBench: A Comprehensive Benchmark of Deepfake Detection}},
  author={Yan, Zhiyuan and Zhang, Yong and Yuan, Xinhang and Lyu, Siwei and Wu, Baoyuan},
  journal={Advances in Neural Information Processing Systems},
  volume={36},
  pages={4534--4565},
  year={2023}
}

@article{korshunov2018deepfakes,
  title={Deepfakes: a new threat to face recognition? assessment and detection},
  author={Korshunov, Pavel and Marcel, S{\'e}bastien},
  journal={arXiv preprint arXiv:1812.08685},
  year={2018}
}

@inproceedings{chen2022trusted,
  title={Trusted media challenge dataset and user study},
  author={Chen, Weiling and Chua, Sheng Lun Benjamin and Winkler, Stefan and Ng, See-Kiong},
  booktitle={Proceedings of the 31st ACM International Conference on Information \& Knowledge Management},
  pages={3873--3877},
  year={2022}
}

@inproceedings{chen2024x2,
  title={{X2-DFD: A framework for explainable and extendable Deepfake Detection}},
  author={Chen, Yize and Yan, Zhiyuan and Cheng, Guangliang and Zhao, Kangran and Lyu, Siwei and Wu, Baoyuan},
  booktitle = {Advances in Neural Information Processing Systems},
  pages = {83792--83839},
  volume = {38},
  year={2025}
}

@article{lipman2022flow,
  title={Flow matching for generative modeling},
  author={Lipman, Yaron and Chen, Ricky TQ and Ben-Hamu, Heli and Nickel, Maximilian and Le, Matt},
  journal={arXiv preprint arXiv:2210.02747},
  year={2022}
}

@inproceedings{zhao2023unipc,
  title={{UniPC}: A Unified Predictor-Corrector Framework for Fast Sampling of Diffusion Models},
  author={Zhao, Wenliang and Bai, Lujia and Rao, Yongming and Zhou, Jie and Lu, Jiwen},
  booktitle=NIPS,
  year={2023}
}

@article{lu2022dpmpp,
  title={{DPM-Solver++}: Fast Solver for Guided Sampling of Diffusion Probabilistic Models},
  author={Lu, Cheng and Zhou, Yuhao and Bao, Fan and Chen, Jianfei and Li, Chongxuan and Zhu, Jun},
  journal={arXiv preprint arXiv:2211.01095},
  year={2022}
}

@inproceedings{kingma2013auto,
  title={Auto-encoding variational bayes},
  author={Kingma, Diederik P and Welling, Max},
  booktitle=ICLR,
  year={2014}
}

@article{van2017neural,
  title={Neural discrete representation learning},
  author={Van Den Oord, Aaron and Vinyals, Oriol and others},
  journal={Advances in neural information processing systems},
  volume={30},
  year={2017}
}

@article{gebru2021datasheets,
  title={Datasheets for datasets},
  author={Gebru, Timnit and Morgenstern, Jamie and Vecchione, Briana and Vaughan, Jennifer Wortman and Wallach, Hanna and Iii, Hal Daum{\'e} and Crawford, Kate},
  journal={Communications of the ACM},
  volume={64},
  number={12},
  pages={86--92},
  year={2021},
  publisher={ACM New York, NY, USA}
}

@article{chen2024demamba,
  title={Demamba: Ai-generated video detection on million-scale genvideo benchmark},
  author={Chen, Haoxing and Hong, Yan and Huang, Zizheng and Xu, Zhuoer and Gu, Zhangxuan and Li, Yaohui and Lan, Jun and Zhu, Huijia and Zhang, Jianfu and Wang, Weiqiang and others},
  journal={arXiv preprint arXiv:2405.19707},
  year={2024}
}

@article{chen2025genworld,
  title={Genworld: Towards detecting ai-generated real-world simulation videos},
  author={Chen, Weiliang and Zheng, Wenzhao and Zheng, Yu and Chen, Lei and Zhou, Jie and Lu, Jiwen and Duan, Yueqi},
  journal={arXiv preprint arXiv:2506.10975},
  year={2025}
}

@article{ma2026your,
  title={Your One-Stop Solution for AI-Generated Video Detection},
  author={Ma, Long and Xue, Zihao and Wang, Yan and Yan, Zhiyuan and Xu, Jin and Jiang, Xiaorui and Yu, Haiyang and Liao, Yong and Bi, Zhen},
  journal={arXiv preprint arXiv:2601.11035},
  year={2026}
}

@article{song2024learning,
  title={On learning multi-modal forgery representation for diffusion generated video detection},
  author={Song, Xiufeng and Guo, Xiao and Zhang, Jiache and Li, Qirui and Bai, Lei and Liu, Xiaoming and Zhai, Guangtao and Liu, Xiaohong},
  journal={Advances in Neural Information Processing Systems},
  volume={37},
  pages={122054--122077},
  year={2024}
}

@misc{seeddance2,
  author       = {{ByteDance}},
  title        = {{SeedDance 2.0}},
  year         = {2025},
  howpublished = {\url{https://seedance2-ai.io/}},
  note         = {Accessed: 2026-04-15}
}

@inproceedings{ma2025detecting,
  title={Detecting ai-generated video via frame consistency},
  author={Ma, Long and Yan, Zhiyuan and Guo, Qinglang and Liao, Yong and Yu, Haiyang and Zhou, Pengyuan},
  booktitle={2025 IEEE International Conference on Multimedia and Expo (ICME)},
  pages={1--6},
  year={2025},
  organization={IEEE}
}

@inproceedings{bai2024ai,
  title={Ai-generated video detection via spatial-temporal anomaly learning},
  author={Bai, Jianfa and Lin, Man and Cao, Gang and Lou, Zijie},
  booktitle={Chinese Conference on Pattern Recognition and Computer Vision (PRCV)},
  pages={460--470},
  year={2024},
  organization={Springer}
}

@article{ye2024loki,
  title={Loki: A comprehensive synthetic data detection benchmark using large multimodal models},
  author={Ye, Junyan and Zhou, Baichuan and Huang, Zilong and Zhang, Junan and Bai, Tianyi and Kang, Hengrui and He, Jun and Lin, Honglin and Wang, Zihao and Wu, Tong and others},
  journal={arXiv preprint arXiv:2410.09732},
  year={2024}
}

@inproceedings{li2025aegis,
  title={AEGIS: Authenticity Evaluation Benchmark for AI-Generated Video Sequences},
  author={Li, Jieyu and Zhang, Xin and Zhou, Joey Tianyi},
  booktitle={Proceedings of the 33rd ACM International Conference on Multimedia},
  pages={13346--13353},
  year={2025}
}

@article{bohacek2025deepaction,
  title={{Human Action CLIPs: Detecting AI-Generated Human Motion}},
  author={Bohacek, Matyas and Farid, Hany},
  journal={arXiv preprint arXiv:2412.00526},
  year={2025}
}

@article{yan2024df40,
  title={Df40: Toward next-generation deepfake detection},
  author={Yan, Zhiyuan and Yao, Taiping and Chen, Shen and Zhao, Yandan and Fu, Xinghe and Zhu, Junwei and Luo, Donghao and Wang, Chengjie and Ding, Shouhong and Wu, Yunsheng and others},
  journal={Advances in Neural Information Processing Systems},
  volume={37},
  pages={29387--29434},
  year={2024}
}

@article{ji2024distinguish,
  title={Distinguish any fake videos: Unleashing the power of large-scale data and motion features},
  author={Ji, Lichuan and Lin, Yingqi and Huang, Zhenhua and Han, Yan and Xu, Xiaogang and Wu, Jiafei and Wang, Chong and Liu, Zhe},
  journal={arXiv preprint arXiv:2405.15343},
  year={2024}
}

@inproceedings{thakral2024illusion,
  title={ILLUSION: Unveiling truth with a comprehensive multi-modal, multi-lingual deepfake dataset},
  author={Thakral, Kartik and Ranjan, Rishabh and Singh, Akanksha and Jain, Akshat and Vatsa, Mayank and Singh, Richa},
  booktitle={The Thirteenth International Conference on Learning Representations},
  year={2024}
}

@article{wang2025robustsora,
  title={RobustSora: De-Watermarked Benchmark for Robust AI-Generated Video Detection},
  author={Wang, Zhuo and Liu, Xiliang and Sun, Ligang},
  journal={arXiv preprint arXiv:2512.10248},
  year={2025}
}

@article{jiang2025ivy,
  title={Ivy-fake: A unified explainable framework and benchmark for image and video aigc detection},
  author={Jiang, Changjiang and Dong, Wenhui and Zhang, Zhonghao and Si, Chenyang and Yu, Fengchang and Peng, Wei and Yuan, Xinbin and Bi, Yifei and Zhao, Ming and Zhou, Zian and others},
  journal={arXiv preprint arXiv:2506.00979},
  year={2025}
}

@inproceedings{wang2025tipi2v,
  title={{TIP-I2V: A Million-Scale Real Text and Image Prompt Dataset for Image-to-Video Generation}},
  author={Wang, Wenhao and Yang, Yi},
  booktitle={Proceedings of the IEEE/CVF International Conference on Computer Vision},
  year={2025}
}

@article{wang2024vidprom,
  title={Vidprom: A million-scale real prompt-gallery dataset for text-to-video diffusion models},
  author={Wang, Wenhao and Yang, Yi},
  journal={Advances in Neural Information Processing Systems},
  volume={37},
  pages={65618--65642},
  year={2024}
}

@article{chandra2025deepfake,
  title={Deepfake-eval-2024: A multi-modal in-the-wild benchmark of deepfakes circulated in 2024},
  author={Chandra, Nuria Alina and Murtfeldt, Ryan and Qiu, Lin and Karmakar, Arnab and Lee, Hannah and Tanumihardja, Emmanuel and Farhat, Kevin and Caffee, Ben and Paik, Sejin and Lee, Changyeon and others},
  journal={arXiv preprint arXiv:2503.02857},
  year={2025}
}

@article{yuan2026helios,
  title={{Helios: Real Real-Time Long Video Generation Model}},
  author={Yuan, Shenghai and Yin, Yuanyang and Li, Zongjian and Huang, Xinwei and Yang, Xiao and Yuan, Li},
  journal={arXiv preprint arXiv:2603.04379},
  year={2026}
}

@article{chern2026speed,
  title={Speed by Simplicity: A Single-Stream Architecture for Fast Audio-Video Generative Foundation Model},
  author={Chern, Ethan and Teng, Hansi and Sun, Hanwen and Wang, Hao and Pan, Hong and Jia, Hongyu and Su, Jiadi and Li, Jin and Yu, Junjie and Liu, Lijie and others},
  journal={arXiv preprint arXiv:2603.21986},
  year={2026}
}

@article{wang2023modelscope,
  title={Modelscope text-to-video technical report},
  author={Wang, Jiuniu and Yuan, Hangjie and Chen, Dayou and Zhang, Yingya and Wang, Xiang and Zhang, Shiwei},
  journal={arXiv preprint arXiv:2308.06571},
  year={2023}
}

@article{chen2023videocrafter1,
  title={Videocrafter1: Open diffusion models for high-quality video generation},
  author={Chen, Haoxin and Xia, Menghan and He, Yingqing and Zhang, Yong and Cun, Xiaodong and Yang, Shaoshu and Xing, Jinbo and Liu, Yaofang and Chen, Qifeng and Wang, Xintao and others},
  journal={arXiv preprint arXiv:2310.19512},
  year={2023}
}

@article{blattmann2023stable,
  title={Stable video diffusion: Scaling latent video diffusion models to large datasets},
  author={Blattmann, Andreas and Dockhorn, Tim and Kulal, Sumith and Mendelevitch, Daniel and Kilian, Maciej and Lorenz, Dominik and Levi, Yam and English, Zion and Voleti, Vikram and Letts, Adam and others},
  journal={arXiv preprint arXiv:2311.15127},
  year={2023}
}

@article{ma2024latte,
  title={{Latte: Latent Diffusion Transformer for Video Generation}},
  author={Ma, Xin and Wang, Yaohui and Jia, Gengyun and Chen, Xinyuan and Liu, Ziwei and Li, Yuan-Fang and Chen, Cunjian and Qiao, Yu},
  journal={arXiv preprint arXiv:2401.03048},
  year={2024}
}

@inproceedings{xing2024dynamicrafter,
  title={Dynamicrafter: Animating open-domain images with video diffusion priors},
  author={Xing, Jinbo and Xia, Menghan and Zhang, Yong and Chen, Haoxin and Yu, Wangbo and Liu, Hanyuan and Liu, Gongye and Wang, Xintao and Shan, Ying and Wong, Tien-Tsin},
  booktitle={European Conference on Computer Vision},
  pages={399--417},
  year={2024},
  organization={Springer}
}

@article{hacohen2026ltx,
  title={LTX-2: Efficient Joint Audio-Visual Foundation Model},
  author={HaCohen, Yoav and Brazowski, Benny and Chiprut, Nisan and Bitterman, Yaki and Kvochko, Andrew and Berkowitz, Avishai and Shalem, Daniel and Lifschitz, Daphna and Moshe, Dudu and Porat, Eitan and others},
  journal={arXiv preprint arXiv:2601.03233},
  year={2026}
}

@inproceedings{nguyen2024laa,
  title={Laa-net: Localized artifact attention network for quality-agnostic and generalizable deepfake detection},
  author={Nguyen, Dat and Mejri, Nesryne and Singh, Inder Pal and Kuleshova, Polina and Astrid, Marcella and Kacem, Anis and Ghorbel, Enjie and Aouada, Djamila},
  booktitle={Proceedings of the IEEE/CVF Conference on Computer Vision and Pattern Recognition},
  pages={17395--17405},
  year={2024}
}

@inproceedings{yermakov2026deepfake,
  title={Deepfake detection that generalizes across benchmarks},
  author={Yermakov, Andrii and Cech, Jan and Matas, Jiri and Fritz, Mario},
  booktitle={Proceedings of the IEEE/CVF Winter Conference on Applications of Computer Vision},
  pages={773--783},
  year={2026}
}

@inproceedings{nguyen2025vulnerability,
  title={Vulnerability-Aware Spatio-Temporal Learning for Generalizable Deepfake Video Detection},
  author={Nguyen, Dat and Astrid, Marcella and Kacem, Anis and Ghorbel, Enjie and Aouada, Djamila},
  booktitle={Proceedings of the IEEE/CVF International Conference on Computer Vision},
  pages={10786--10796},
  year={2025}
}

@inproceedings{zheng2025d3,
  title={D3: Training-free ai-generated video detection using second-order features},
  author={Zheng, Chende and Suo, Ruiqi and Lin, Chenhao and Zhao, Zhengyu and Yang, Le and Liu, Shuai and Yang, Minghui and Wang, Cong and Shen, Chao},
  booktitle={Proceedings of the IEEE/CVF International Conference on Computer Vision},
  pages={12852--12862},
  year={2025}
}

@article{zhang2025physics,
  title={Physics-driven spatiotemporal modeling for ai-generated video detection},
  author={Zhang, Shuhai and Lian, ZiHao and Yang, Jiahao and Li, Daiyuan and Pang, Guoxuan and Liu, Feng and Han, Bo and Li, Shutao and Tan, Mingkui},
  journal={arXiv preprint arXiv:2510.08073},
  year={2025}
}

@inproceedings{bulat2017fan,
  title={{How Far are We from Solving the 2D \& 3D Face Alignment Problem? (and a Dataset of 230,000 3D Facial Landmarks)}},
  author={Bulat, Adrian and Tzimiropoulos, Georgios},
  booktitle={Proceedings of the IEEE International Conference on Computer Vision},
  pages={1021--1030},
  year={2017}
}

@article{chen2024topiq,
  title={Topiq: A top-down approach from semantics to distortions for image quality assessment},
  author={Chen, Chaofeng and Mo, Jiadi and Hou, Jingwen and Wu, Haoning and Liao, Liang and Sun, Wenxiu and Yan, Qiong and Lin, Weisi},
  journal={IEEE Transactions on Image Processing},
  volume={33},
  pages={2404--2418},
  year={2024},
  publisher={IEEE}
}

@inproceedings{chen2024dslfiqa,
  title={{DSL-FIQA: Assessing Facial Image Quality via Dual-Set Degradation Learning and Landmark-Guided Transformer}},
  author={Chen, Wei-Ting and Krishnan, Gurunandan and Gao, Qiang and Kuo, Sy-Yen and Ma, Sizhuo and Wang, Jian},
  booktitle={Proceedings of the IEEE/CVF Conference on Computer Vision and Pattern Recognition},
  pages={2931-2941},
  year={2024}
}

@article{rao1967ties,
  title={{Ties in Paired-Comparison Experiments: A Generalization of the Bradley-Terry Model}},
  author={Rao, P. V. and Kupper, L. L.},
  journal={Journal of the American Statistical Association},
  volume={62},
  number={317},
  pages={194--204},
  year={1967}
}

@article{bradley1952rank,
  title={{Rank Analysis of Incomplete Block Designs: I. The Method of Paired Comparisons}},
  author={Bradley, Ralph Allan and Terry, Milton E.},
  journal={Biometrika},
  volume={39},
  number={3/4},
  pages={324--345},
  year={1952}
}

@techreport{itu_bt500_2023,
  title={{Methodology for the subjective assessment of the quality of television pictures}},
  author={{ITU-R}},
  institution={International Telecommunication Union},
  number={Recommendation ITU-R BT.500-15},
  year={2023}
}

@inproceedings{zhang2024t2vhe,
  title={{Rethinking human evaluation protocol for text-to-video models: enhancing reliability, reproducibility, and practicality}},
  author={Zhang, Tianle and Ma, Langtian and Yan, Yuchen and Zhang, Yuchen and Wang, Kai and Yang, Yue and Guo, Ziyao and Shao, Wenqi and You, Yang and Qiao, Yu and Luo, Ping and Zhang, Kaipeng},
  booktitle={Proceedings of the 38th International Conference on Neural Information Processing Systems},
  series = {NIPS '24},
  year={2024}
}

@article{groh2022deepfake,
  title={{Deepfake detection by human crowds, machines, and machine-informed crowds}},
  author={Groh, Matthew and Epstein, Ziv and Firestone, Chaz and Picard, Rosalind},
  journal={Proceedings of the National Academy of Sciences},
  volume={119},
  number={1},
  pages={e2110013119},
  year={2022}
}

@techreport{itu_p910_2023,
  title={{Subjective video quality assessment methods for multimedia applications}},
  author={{ITU-T}},
  institution={International Telecommunication Union},
  number={Recommendation ITU-T P.910},
  url={https://www.itu.int/rec/t-rec-p.910/en},
  year={2023}
}

@article{kay2017kinetics,
  title={{The Kinetics Human Action Video Dataset}},
  author={Will Kay and Joao Carreira and Karen Simonyan and Brian Zhang and Chloe Hillier and Sudheendra Vijayanarasimhan and Fabio Viola and Tim Green and Trevor Back and Paul Natsev and Mustafa Suleyman and Andrew Zisserman},
  journal={arXiv preprint arXiv:1705.06950},
  year={2017}
}

@misc{pika_labs,
  title={{Pika}},
  author={{Pika Labs}},
  year={2023},
  howpublished={\url{https://pika.art}}
}

@article{fleiss1971kappa,
  title={{Measuring nominal scale agreement among many raters}},
  author={Fleiss, Joseph L.},
  journal={Psychological Bulletin},
  volume={76},
  number={5},
  pages={378--382},
  year={1971}
}

@article{wiegand2003overview,
  title={{Overview of the H.264/AVC video coding standard}},
  author={Wiegand, Thomas and Sullivan, Gary J. and Bjontegaard, Gisle and Luthra, Ajay},
  journal={IEEE Transactions on Circuits and Systems for Video Technology},
  volume={13},
  number={7},
  pages={560--576},
  year={2003}
}

@book{david1988method,
  title={{The Method of Paired Comparisons}},
  author={David, Herbert Aron},
  year={1988},
  publisher={Oxford University Press},
  edition={2nd}
}

@Inbook{efron1979bootstrap,
  title={{Bootstrap Methods: Another Look at the Jackknife}},
  author={Efron, Bradley},
  bookTitle="Breakthroughs in Statistics: Methodology and Distribution",
  publisher="Springer New York",
  address="New York, NY",
  pages="569--593",
  year={1992}
}

@techreport{krippendorff2011computing,
  title={{Computing Krippendorff's Alpha-Reliability}},
  author={Krippendorff, Klaus},
  institution={Annenberg School for Communication, University of Pennsylvania},
  year={2011}
}

@article{peyre2019computational,
  title={{Computational Optimal Transport: With Applications to Data Science}},
  author={Peyr{\'e}, Gabriel and Cuturi, Marco},
  journal={Foundations and Trends in Machine Learning},
  volume={11},
  number={5--6},
  pages={355--607},
  year={2019},
  publisher={Now Publishers, Inc.}
}
}

\clearpage
\appendix
\addtocontents{toc}{\protect\setcounter{tocdepth}{2}}

\renewcommand{\contentsname}{Appendices}
\tableofcontents
\clearpage

\section{Datasheet for SynthForensics}
\label{app:datasheet}

This appendix follows the datasheets-for-datasets framework of~\citet{gebru2021datasheets}, summarizing the construction, intended use, distribution, and maintenance of SynthForensics. Where applicable, we point to the appendix sections that document each aspect in full detail rather than duplicating their content.

\subsection{Motivation}
\label{app:datasheet-motivation}

SynthForensics is a people-centric benchmark for evaluating deepfake-detection methods against the latest generation of state-of-the-art open-source synthetic-video generators. The motivation, threat model, and target research community are described in Section~\ref{sec:intro}; the generator-pool selection criteria and source-data sourcing are detailed in Appendix~\ref{app:benchmark-sources}.

\subsection{Composition}
\label{app:datasheet-composition}

SynthForensics consists of $20{,}445$ unique synthetic videos generated by $8$ T2V and $7$ I2V open-source models, paired-source from $1{,}363$ FF++ and DFD pristine sources, and released in four compression versions per video for a total of $81{,}780$ video files. Each video is paired with a JSON sidecar containing the full generation pipeline. Per-generator counts (modality, orientation), summary statistics, and the metadata schema are reported in Section~\ref{subsec:dataset_statistics} and Appendix~\ref{app:benchmark-stats}.

\subsection{Collection Process}
\label{app:datasheet-collection}

SynthForensics is built on a paired-source protocol: each generated video is anchored to one of $1{,}363$ pristine source videos drawn from FaceForensics++~\citep{rossler2019faceforensics++} ($1{,}000$ raw real videos) and the Deep Fake Detection dataset~\citep{dufour2019google} ($363$ actor-recorded pristine videos) (Appendix~\ref{app:benchmark-sources}). For each source video, a Vision-Language Model~\citep{zhang2025videollama} extracts a structured caption that is then manually validated by human annotators and screened by an LLM-based~\citep{grattafiori2024llama} safety filter across seven sensitive thematic categories (Appendices~\ref{app:benchmark-prompts} and~\ref{app:benchmark-validation}); for I2V, a reference frame is additionally selected by human annotators from the first half of the source video (Appendix~\ref{app:benchmark-i2v-frames}). The validated structured caption (and reference frame, for I2V) is then fed to each of the $8$ T2V and $7$ I2V open-source generators (Appendix~\ref{app:benchmark-hyperparams}) to produce one synthetic video per (source, generator, modality) combination. Each generated video is then validated by five human annotators against an anatomical, temporal, rendering, semantic, and ethical-compliance rubric, with rejected outputs triggering iterative refinement until acceptance (Appendix~\ref{app:benchmark-video-validation}). Generation was distributed across two on-site servers and an AWS instance, equipped in aggregate with 2 NVIDIA RTX PRO 6000 Blackwell Server Edition (96 GB), 3 NVIDIA A100 80GB PCIe, and 1 NVIDIA H100 80GB GPUs (full hardware specifications in Appendix~\ref{app:detection-hardware}).

\subsection{Preprocessing, Cleaning, and Labeling}
\label{app:datasheet-preprocessing}

Per-generator inference hyperparameters, resolution and duration settings, and pipeline-specific details (e.g., LoRA strengths, scheduler choices, distillation modes) are documented in Appendix~\ref{app:benchmark-hyperparams}. Each generated video is inspected by five human annotators against a rubric covering anatomical, temporal, rendering, semantic, and ethical compliance criteria; rejected outputs trigger iterative refinement of prompts and inference parameters until acceptance (Appendix~\ref{app:benchmark-video-validation}). The released videos are produced in four compression versions (\textit{Raw}, \textit{Canonical} at H.264 CRF=$0$ with standardized BT.709 encoding, \textit{CRF23}, \textit{CRF40}), described in Section~\ref{subsec:dataset_statistics}.

\subsection{Uses}
\label{app:datasheet-uses}

SynthForensics is intended as an evaluation benchmark for deepfake-detection methods on people-centric synthetic content. The compatible task scenarios include zero-shot evaluation of pre-trained detectors, fine-tuning evaluation of detector adaptability to modern synthetic content, and training-from-scratch evaluation of generator-distribution generalization, all demonstrated in Section~\ref{sec:detection}. Prohibited uses, the threat model surrounding the release, broader societal impacts, and risks of misuse are discussed in Appendices~\ref{app:limitations-limitations} and~\ref{app:limitations-impact}.

\subsection{Distribution}
\label{app:datasheet-distribution}

The dataset is distributed through a gated Hugging Face repository under the Creative Commons Attribution-NonCommercial 4.0 International (CC BY-NC 4.0) license, consistent with the upstream non-commercial research terms of use of FF++ and DFD. Access requires explicit acceptance of the dataset terms of use; the gating mechanism is automatic with no manual approval required. Full license terms, the data usage agreement, and access-control mechanisms are documented in Appendix~\ref{app:limitations-mitigation}.

\subsection{Maintenance}
\label{app:datasheet-maintenance}

SynthForensics is released as a static snapshot of the open-source synthetic-video state of the art at submission time. We plan rolling releases that incorporate future state-of-the-art open-source generators after they pass the same selection and validation pipeline (Appendix~\ref{app:limitations-limitations}). Errors, misuse reports, and access-revocation requests are handled through the maintainer contact channel released alongside the dataset; the underlying monitoring and revocation policies are described in Appendix~\ref{app:limitations-mitigation}.

\clearpage
\section{Benchmark Construction: Extended Details}
\label{app:benchmark}

This appendix extends Section~\ref{sec:sf_bench} with the technical specifications and protocols governing the construction of SynthForensics. Appendix~\ref{app:benchmark-sources} documents the source data and the per-generator selection process. Appendix~\ref{app:benchmark-prompts} details the structured prompt generation and adaptation pipeline. Appendix~\ref{app:benchmark-validation} describes the manual prompt validation protocol with human annotators and LLM-based safety screening. Appendix~\ref{app:benchmark-i2v-frames} specifies the reference frame selection procedure for I2V. Appendix~\ref{app:benchmark-hyperparams} reports per-generator specifications and inference hyperparameters. Appendix~\ref{app:benchmark-video-validation} outlines the manual video validation protocol. Finally, Appendix~\ref{app:benchmark-stats} reports per-generator statistics, the released metadata catalog, and quality assessment, while Appendix~\ref{app:benchmark-samples} presents qualitative visual samples.

\subsection{Source Data and Generator Pool}
\label{app:benchmark-sources}

The 1,363 pristine source videos draw from two established deepfake-detection corpora: FaceForensics++ (FF++)~\citep{rossler2019faceforensics++}, from which we use the 1,000 raw real videos, and the Deep Fake Detection (DFD) dataset~\citep{dufour2019google}, from which we use all 363 actor-recorded pristine videos. Both corpora are people-centric by construction, with every video featuring one or more identifiable subjects performing actions, and they are publicly distributed for non-commercial research and widely adopted as real-video baselines in the forensic community, ensuring continuity with prior detection benchmarks.

\begin{figure}[!p]
    \centering
    \setlength{\tabcolsep}{1pt}
    \renewcommand{\arraystretch}{0.5}
    \begin{tabular}{c ccccc}
        & $t=0$ & $t=1$ & $t=2$ & $t=3$ & $t=4$ \\
        \rotatebox{90}{\scriptsize \textbf{\textsc{OpenSora 2.0}}} &
        \includegraphics[width=0.16\textwidth]{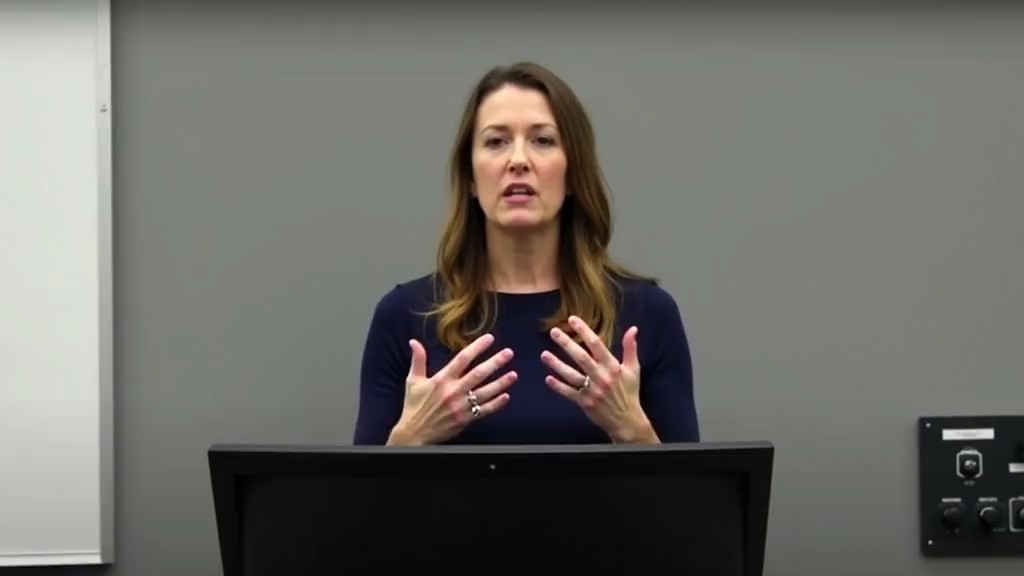} &
        \includegraphics[width=0.16\textwidth]{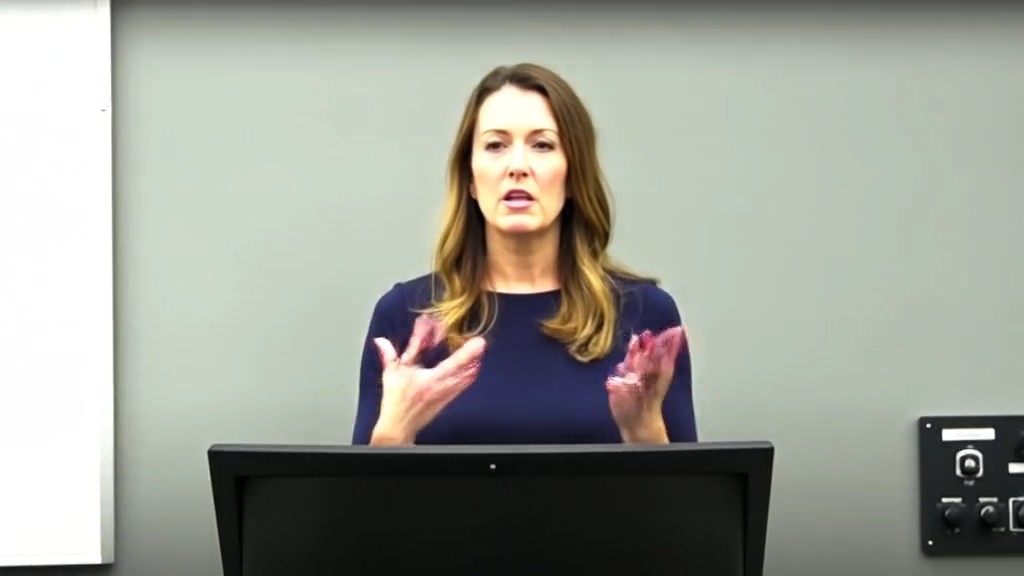} &
        \includegraphics[width=0.16\textwidth]{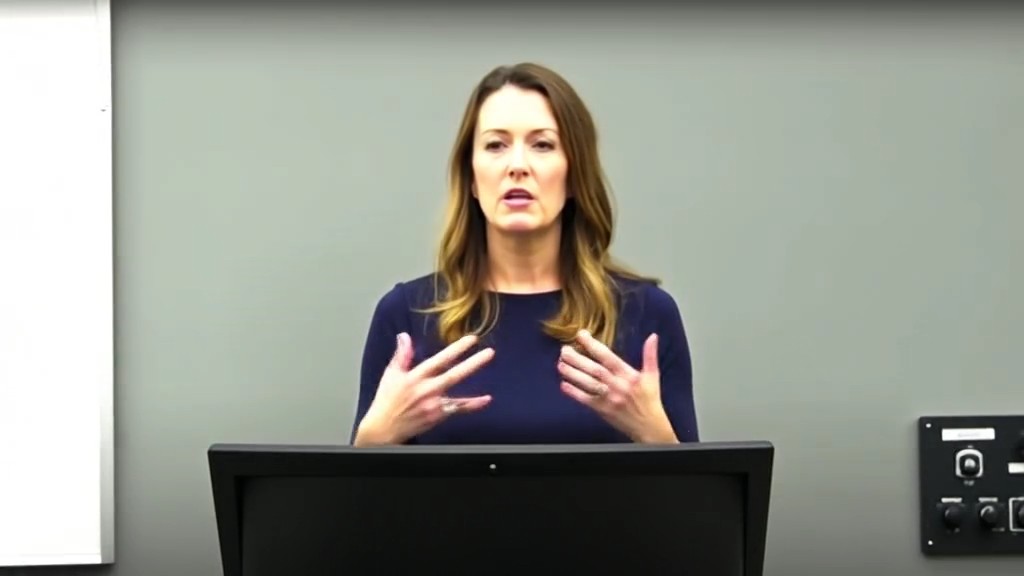} &
        \includegraphics[width=0.16\textwidth]{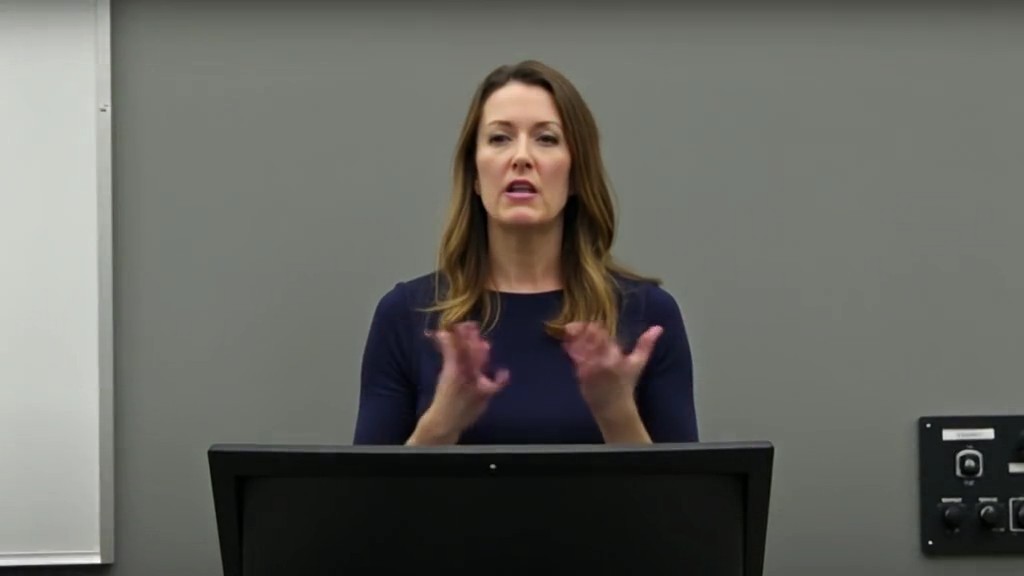} &
        \includegraphics[width=0.16\textwidth]{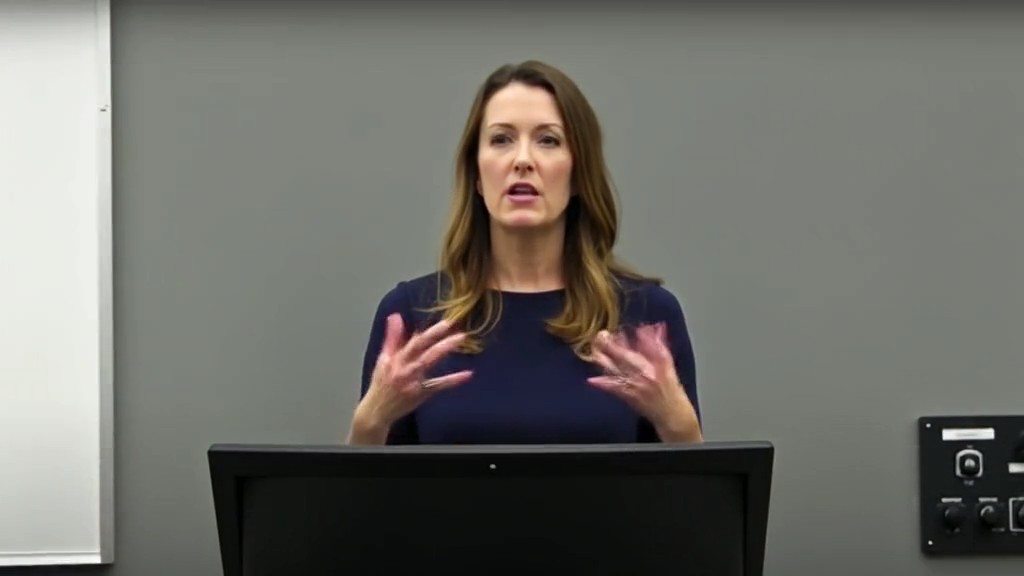} \\
        \rotatebox{90}{\scriptsize \textbf{\textsc{CausVid}}} &
        \includegraphics[width=0.16\textwidth]{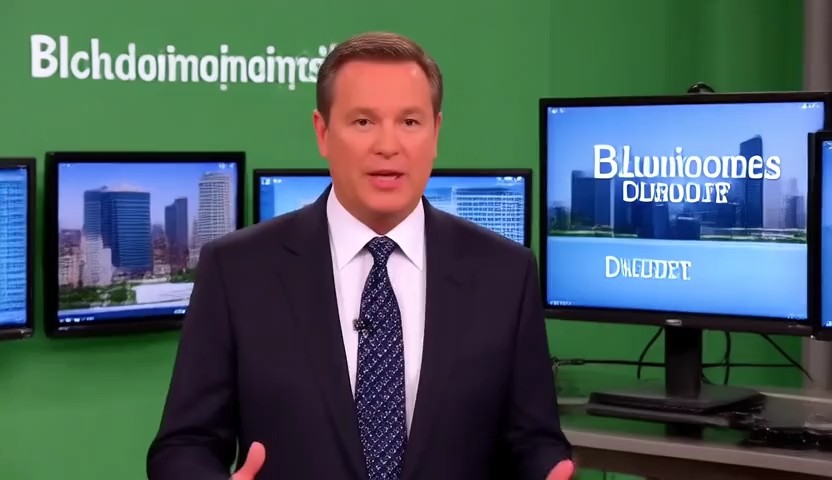} &
        \includegraphics[width=0.16\textwidth]{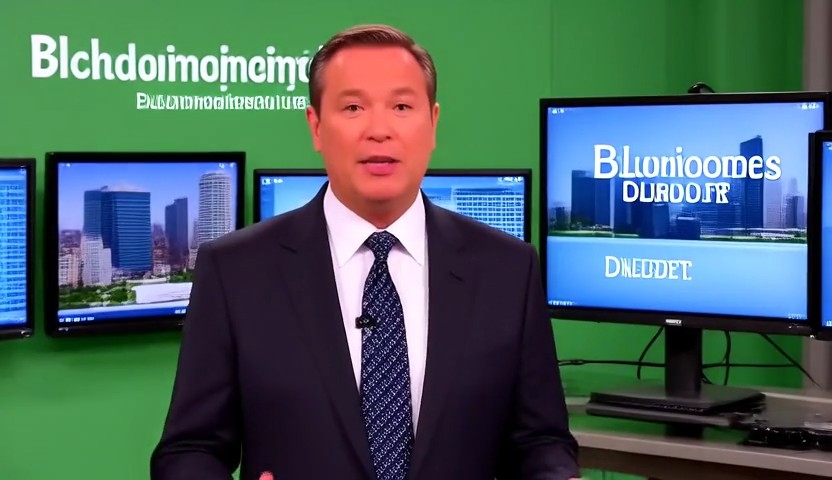} &
        \includegraphics[width=0.16\textwidth]{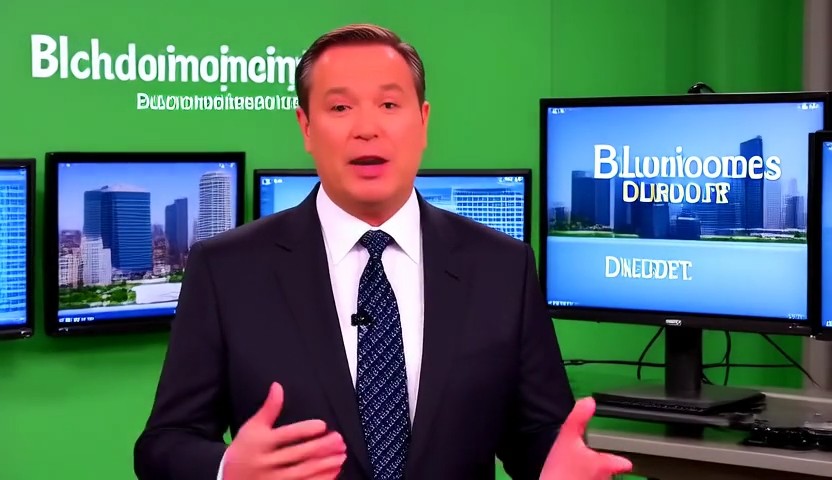} &
        \includegraphics[width=0.16\textwidth]{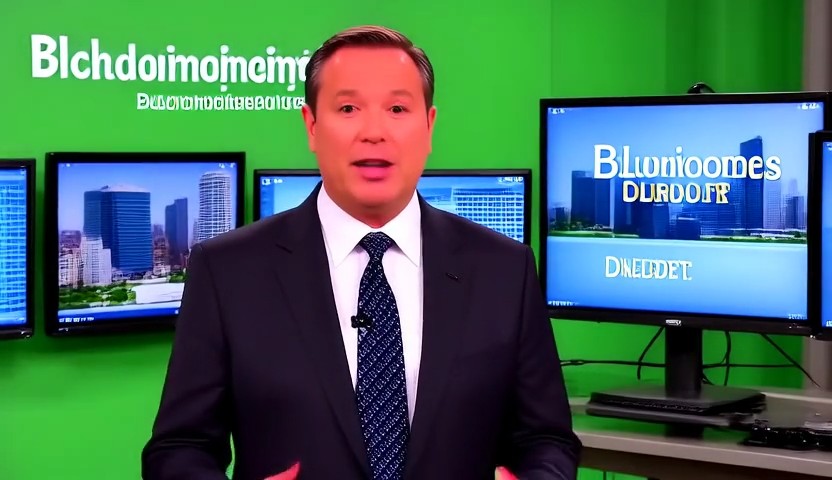} &
        \includegraphics[width=0.16\textwidth]{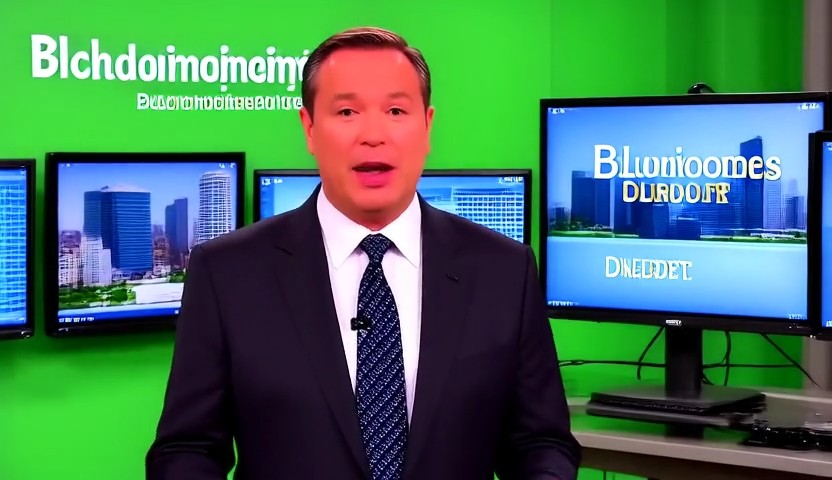} \\
        \rotatebox{90}{\scriptsize \textbf{\textsc{Step-Video-T2V}}} &
        \includegraphics[width=0.16\textwidth]{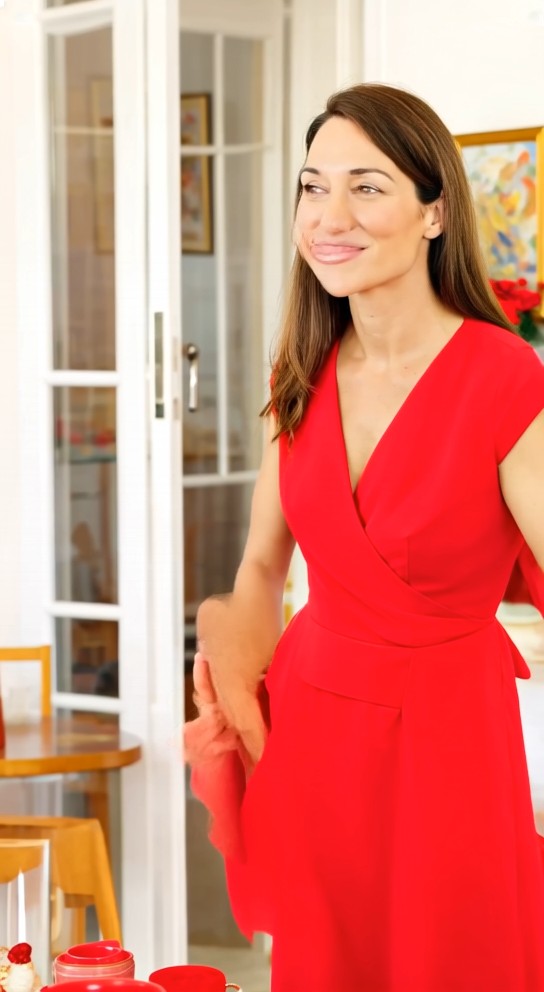} &
        \includegraphics[width=0.16\textwidth]{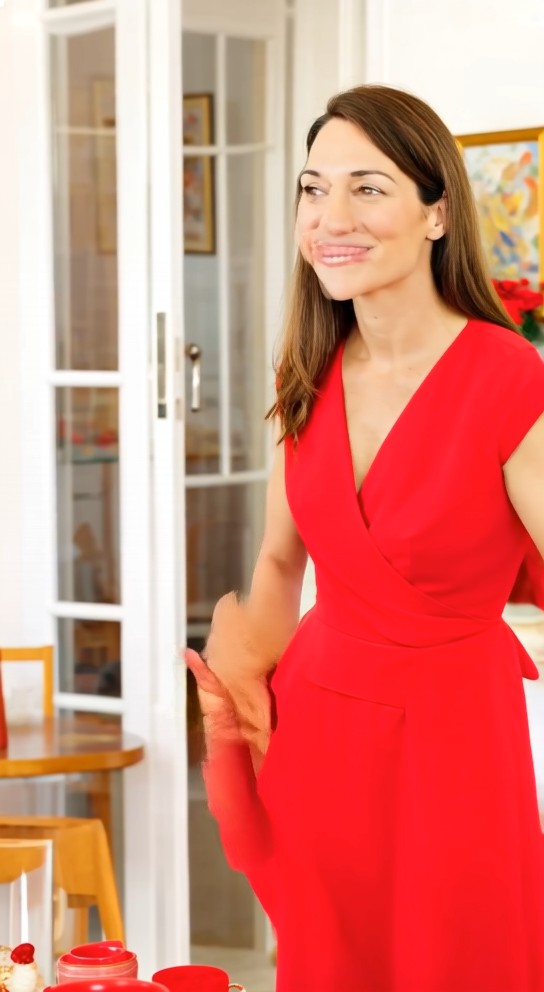} &
        \includegraphics[width=0.16\textwidth]{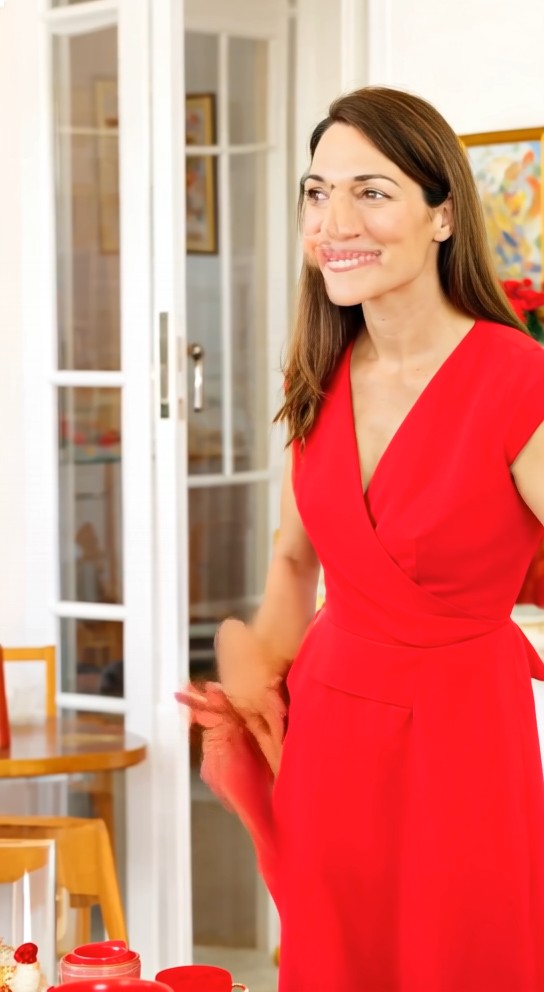} &
        \includegraphics[width=0.16\textwidth]{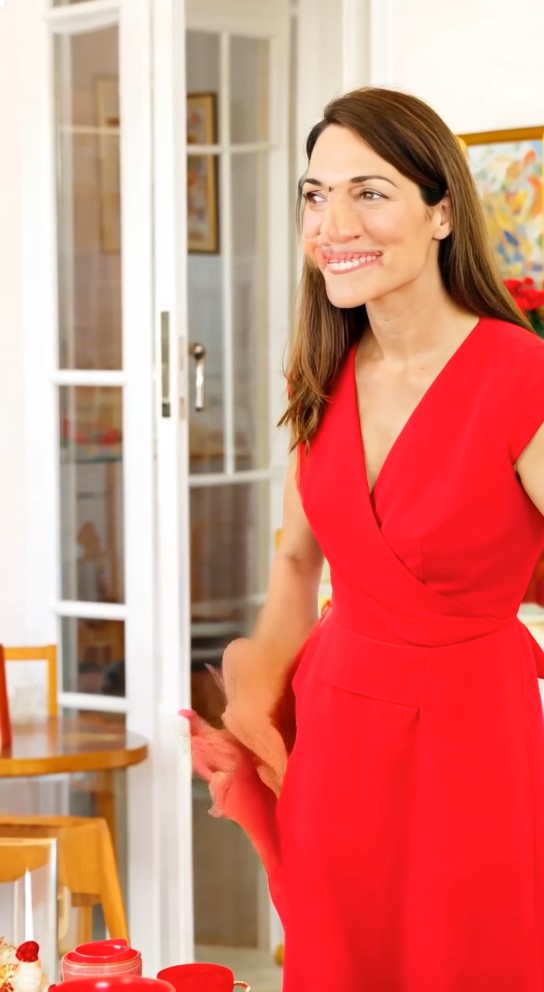} &
        \includegraphics[width=0.16\textwidth]{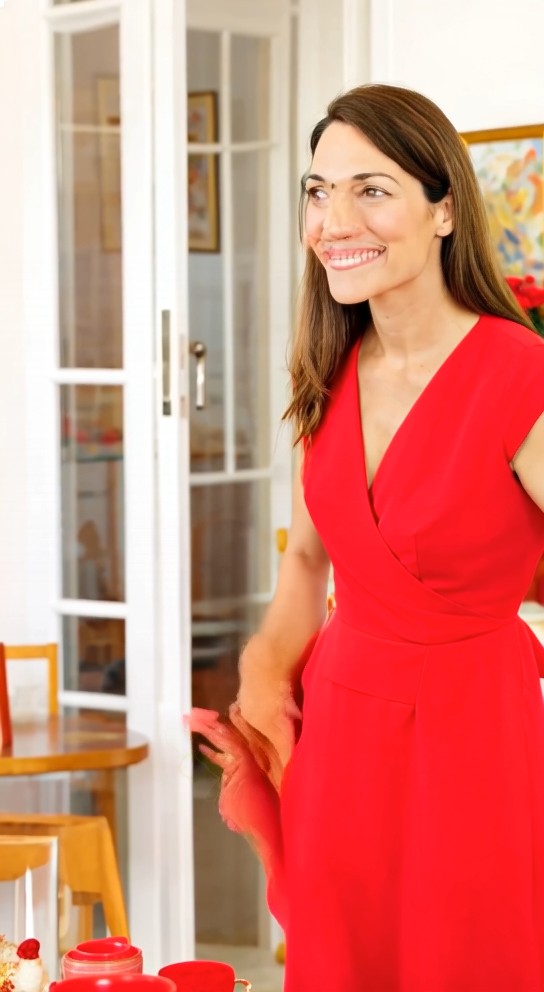} \\
    \end{tabular}
    \caption{Qualitative failure analysis of excluded models. Representative artifacts that led to the exclusion of OpenSora 2.0, CausVid, and Step-Video-T2V. OpenSora 2.0 exhibits temporal flickering, texture boiling, and frequent static generation with frozen subjects. CausVid suffers from color saturation artifacts and inconsistent background rendering. Step-Video-T2V displays severe anatomical deformation (``melting faces'') on human subjects.}
    \label{fig:rejected_examples}
\end{figure}

To ensure the benchmark represents the realistic threat posed by near-photorealistic synthetic-video systems that current detection methods must address, rather than easily identifiable low-quality outputs, generator selection from the open-source landscape surveyed in Section~\ref{subsec:gen_models} combined architectural eligibility with empirical quality assessment. We restricted the candidate pool to architectures capable of multi-second human-centric synthesis at modern fidelity, which excluded pre-DiT U-Net models such as ModelScope~\citep{wang2023modelscope}, VideoCrafter~\citep{chen2023videocrafter1}, SVD~\citep{blattmann2023stable}, AnimateDiff~\citep{guo2023animatediff}, and DynamiCrafter~\citep{xing2024dynamicrafter} and early DiT models with limited native length such as Latte~\citep{ma2024latte}, all of which generate native sequences of at most 25 frames per inference pass ($\sim$2 seconds at typical playback rates). HunyuanVideo~\citep{kong2024hunyuanvideo} was further excluded due to licensing restrictions that prohibit research use and dataset distribution. The remaining eleven candidates were tested for quality: Wan2.1~\citep{wan2025wan}, CogVideoX~\citep{yang2024cogvideox}, SkyReels-V2~\citep{chen2025skyreels}, Self-Forcing~\citep{huang2025self}, MAGI-1~\citep{teng2025magi}, LTX-2.3~\citep{hacohen2026ltx}, Helios-Distilled~\citep{yuan2026helios}, daVinci-MagiHuman-Distilled~\citep{chern2026speed}, OpenSora 2.0~\citep{zheng2025open}, Step-Video-T2V~\citep{ma2025step}, and CausVid~\citep{yin2025causvid}. For each candidate, we generated 1,363 T2V videos using the validated structured prompts produced by the protocol detailed in Appendices~\ref{app:benchmark-prompts} and~\ref{app:benchmark-validation}, then manually inspected the outputs for four artifact categories: malformed anatomy (distorted facial features, impossible limb configurations), temporal flickering (frame-to-frame instability), physically implausible motion (unnatural trajectories, violation of kinematic constraints), and static generation (lack of meaningful temporal dynamics or frozen subjects). Candidates whose rejection rate exceeded $25\%$ were excluded, where the rejection rate is defined as the proportion of rejected videos over the 1,363 generated per candidate. OpenSora 2.0, Step-Video-T2V, and CausVid surpassed this threshold, while the eight retained models remained below it. Per-generator rejection statistics are reported in Table~\ref{tab:model_rejection_stats_v2}, and Figure~\ref{fig:rejected_examples} shows representative artifacts from the excluded models.

The $25\%$ rejection-rate threshold operationalizes our near-photorealistic threat model: a generator that requires manual rejection on more than one in four outputs does not credibly support the people-centric attack scenarios (impersonation, fraud, non-consensual content) where outputs failing human inspection are self-defeating. The threshold was set independently of the measured per-generator rejection rates; the empirical gap separating retained candidates ($\leq 19\%$) from excluded ones ($\geq 31\%$) in Table~\ref{tab:model_rejection_stats_v2} is reported as a post-hoc robustness check, indicating that the choice does not depend on the precise value within this band. The dominant failure modes of the three excluded models in Figure~\ref{fig:rejected_examples} are intrinsic generator limitations rather than prompt-driven: OpenSora~2.0 produces frequent static generation and texture boiling, CausVid exhibits persistent color saturation artifacts, and Step-Video-T2V deforms human anatomy on the majority of outputs. All eleven candidates received the same per-generator prompt-structure tuning (Appendix~\ref{app:benchmark-prompts}) and negative-prompt baseline (Appendix~\ref{app:benchmark-prompts-negative}) during the selection phase, so a captioning-style mismatch between protocol and generator cannot account for the order-of-magnitude difference in rejection rate between the retained and excluded clusters.

\clearpage

\begin{table}[!p]
    \centering
    \small
    \caption{Per-generator quality assessment results for the eleven tested candidates, sorted by rejection rate. Models above the double line were retained for SynthForensics; those below were excluded for exceeding the $25\%$ rejection threshold.}
    \label{tab:model_rejection_stats_v2}
    \begin{tabular}{lrc}
        \toprule
        \textbf{Model} & \textbf{Rejected Videos} & \textbf{Rejection Rate} \\
        \midrule
        Wan2.1                            & 22  & 1.6\%  \\
        LTX-2.3                             & 27  & 2.0\%  \\
        MAGI-1-Distilled                  & 47  & 3.4\%  \\
        SkyReels-V2                       & 53  & 3.9\%  \\
        Helios-Distilled                  & 61  & 4.5\%  \\
        Self-Forcing                      & 123 & 9.0\%  \\
        CogVideoX                         & 177 & 13.0\% \\
        daVinci-MagiHuman-Distilled       & 255 & 18.7\% \\
        \midrule\midrule
        CausVid                           & 430 & 31.5\% \\
        OpenSora 2.0                      & 551 & 40.4\% \\
        Step-Video-T2V                    & 779 & 57.1\% \\
        \bottomrule
    \end{tabular}
\end{table}

\clearpage

\subsection{Structured Prompt Generation and Adaptation Details}
\label{app:benchmark-prompts}

Our prompting strategy addresses a critical challenge in multi-generator benchmarking: architecture-specific prompt alignment. The eight generative models selected for SynthForensics, Wan2.1~\citep{wan2025wan}, SkyReels-V2~\citep{chen2025skyreels}, Self-Forcing~\citep{huang2025self}, CogVideoX~\citep{yang2024cogvideox}, MAGI-1-Distilled~\citep{teng2025magi}, LTX-2.3~\citep{hacohen2026ltx}, Helios-Distilled~\citep{yuan2026helios}, and daVinci-MagiHuman-Distilled~\citep{chern2026speed}, are trained on distinct captioning distributions; for example, SkyReels-V2 favors cinematic descriptors at the start, while Wan2.1 leads with stylistic and framing descriptors. A static, monolithic caption would fail to elicit optimal performance across all models. To address these diverse requirements while ensuring semantic consistency and high visual fidelity, we developed a comprehensive protocol covering both positive and negative conditioning.

\subsubsection{Positive Prompt Construction}

To ensure semantic consistency, we focused primarily on establishing a robust pipeline for positive prompt construction. We adopted a semantic decomposition strategy inspired by HunyuanVideo~\citep{kong2024hunyuanvideo}: instead of relying on a single unstructured caption, we employed the Vision-Language Model (VLM) VideoLLaMA 3~\citep{zhang2025videollama} to extract a unified set of eight semantic fields from each source video:

\begin{itemize}
    \item \textit{Short Description}: A concise synopsis summarizing the primary subject and spatial context in 1--2 sentences, used for high-level semantic anchoring.
    \item \textit{Dense Description}: The comprehensive narrative layer of the prompt, capturing fine-grained details of the subject's appearance, specific actions, and the dynamic evolution of the scene over time.
    \item \textit{Background}: A dedicated description of the environment in which the subject is situated. This field isolates fixed and mobile environmental elements to maintain depth consistency and prevent subject-background bleeding.
    \item \textit{Style}: Specifies the visual aesthetic or media genre (e.g., ``Documentary'', ``Cinematic'', ``News Broadcast''), enforcing stylistic coherence with the source distribution.
    \item \textit{Shot Type}: Identifies the framing size of the shot (e.g., ``Aerial shot'', ``Close-up'', ``Medium shot'').
    \item \textit{Camera Movement}: Captures the camera dynamics descriptors (e.g., ``Pan'', ``Zoom'', ``Static shot'').
    \item \textit{Lighting}: Details the illumination properties of the scene, capturing specific attributes such as light source direction, hardness (e.g., ``Chiaroscuro''), and color temperature.
    \item \textit{Atmosphere}: Encapsulates the emotional resonance or mood of the video (e.g., ``Tense'', ``Serene'', ``Mysterious''), guiding the generator's tone beyond mere physical description.
\end{itemize}

\begin{figure}[!h]
\centering
\begin{tcolorbox}[
    colback=gray!5,
    colframe=gray!40,
    boxrule=0.5pt,
    arc=2mm,
    left=6pt, right=6pt, top=5pt, bottom=5pt,
    fontupper=\footnotesize\ttfamily
]
System Instruction: You are a specialized assistant for advanced video analysis. Your task is to provide structured captions in JSON format for real videos. Follow the guidelines below carefully to produce accurate, detailed, and multidimensional descriptions.

\vspace{0.4em}
Structured Caption Format: \\
For each analyzed video, you must generate a structured caption in JSON format that includes the following components:

\vspace{0.4em}
\{
    \vspace{0.2em}
    \hspace*{1em} "short\_description": "A brief description that captures the main essence of the scene in 1-2 sentences",

    \vspace{0.3em}
    \hspace*{1em} "dense\_description": "A comprehensive narrative of the scene. Describe the subject's appearance, their specific actions, and interactions",

    \vspace{0.3em}
    \hspace*{1em} "background": "Detailed description of the environment in which the subject is situated, including fixed and mobile elements",

    \vspace{0.3em}
    \hspace*{1em} "style": "Characterization of the video style (e.g., documentary, cinematic, realistic, sci-fi, etc.)",

    \vspace{0.3em}
    \hspace*{1em} "shot\_type": "Framing size of the shot (e.g., close-up, medium shot, aerial shot)",

    \vspace{0.3em}
    \hspace*{1em} "camera\_movement": "Camera dynamics (e.g., static shot, pan, tilt, zoom, tracking)",

    \vspace{0.3em}
    \hspace*{1em} "lighting": "Description of the lighting conditions in the video (e.g., natural light, artificial, chiaroscuro, backlight, etc.)",

    \vspace{0.3em}
    \hspace*{1em} "atmosphere": "Description of the video atmosphere (e.g., cozy, tense, mysterious, serene, etc.)"
    \vspace{0.2em}
\}
\end{tcolorbox}
\caption{System prompt provided to VideoLLaMA 3~\citep{zhang2025videollama} to extract the structured positive prompts that guide synthetic video generation. The schema enforces an 8-field decomposition that separates aesthetic, semantic, and technical video attributes.}
\label{fig:system_prompt_v2}
\end{figure}

The extraction process is governed by the system instruction shown in Figure~\ref{fig:system_prompt_v2}, which enforces a strict JSON schema for high semantic and visual precision. The decomposition into discrete fields lets us programmatically select and reorder subsets to match each target generator's optimal conditioning distribution, as detailed in Table~\ref{tab:prompt_structures_v2}, and reduces VLM hallucinations by forcing the model to categorize visual information rather than emit a single unstructured caption. Figure~\ref{fig:json_example_v2} illustrates a representative raw structured caption produced by VideoLLaMA 3 for one of our source videos.

\begin{table}[!p]
    \centering
    \caption{Model-specific prompt structures. The fields, derived from the unified caption, were ordered and combined to align with each generator's optimal input format.}
    \label{tab:prompt_structures_v2}
    \footnotesize
    \begin{tabularx}{\columnwidth}{@{}l >{\raggedright\arraybackslash}X@{}}
    \toprule
    \textbf{Model} & \textbf{Field Composition and Order} \\
    \midrule
    Wan2.1, Self-Forcing, Helios-Distilled & \texttt{style, shot\_type, short\_description, dense\_description, background, lighting, atmosphere, camera\_movement} \\
    \addlinespace
    SkyReels-V2 & \texttt{dense\_description, style, shot\_type, camera\_movement, background, atmosphere, lighting} \\
    \addlinespace
    MAGI-1-Distilled & \texttt{short\_description, dense\_description, background, style, shot\_type, lighting, atmosphere} \\
    \addlinespace
    CogVideoX & \texttt{dense\_description, style, shot\_type, background} \\
    \addlinespace
    LTX-2.3 & \texttt{short\_description, dense\_description, background, shot\_type, camera\_movement, lighting, atmosphere} \\
    \addlinespace
    daVinci-MagiHuman-Distilled & \texttt{dense\_description, background, shot\_type, lighting, atmosphere, style} \\
    \bottomrule
    \end{tabularx}
\end{table}

\begin{figure}[!p]
\centering
\begin{lstlisting}[basicstyle=\ttfamily\footnotesize, frame=single, breaklines=true]
{
  "short_description": "A news anchor is sitting in a studio, ready to deliver the news.",
  "dense_description": "The video starts with a news anchor seated in a chair. She is wearing a pink blazer over a black top and has shoulder-length dark hair. The background features a blue and white backdrop with a window showing a blue sky. There are text overlays on the screen indicating weather information and other news updates.",
  "background": "The background includes a blue and white curtain with a window that shows a blue sky. There is also a graphic overlay with text in Chinese characters.",
  "style": "Realistic, News Broadcast",
  "shot_type": "Medium shot",
  "camera_movement": "Static camera",
  "lighting": "Artificial, bright studio lighting",
  "atmosphere": "Professional"
}
\end{lstlisting}
\caption{Representative example of the raw structured caption extracted from a source video, following the schema enforced by the system prompt in Figure~\ref{fig:system_prompt_v2}.}
\label{fig:json_example_v2}
\end{figure}

\clearpage

\subsubsection{Negative Prompt Optimization}
\label{app:benchmark-prompts-negative}

While positive prompts guide the semantic content, negative prompts steer generation away from undesired visual patterns and reinforce perceptual fidelity. For each generator, we initialized the negative prompt from the author-recommended strings, preserved in their native language to remain on-distribution with the generator's text encoder, and augmented them with global English-language keywords targeting recurring artifacts observed in early outputs; the resulting baselines are listed in Table~\ref{tab:baseline_negatives_v2}, normalized to English for readability uniformity. The strings actually fed to Wan2.1 and Self-Forcing, and recorded in the released JSON sidecars, are the original Chinese defaults of the upstream Wan2.1 release; SkyReels-V2, CogVideoX, and LTX-2.3 use English-language defaults verbatim. The distilled versions of MAGI-1, Helios, and daVinci-MagiHuman do not require negative prompts: in distilled mode the guidance signal is baked directly into the model weights ($\textrm{cfg}=1$), and negative prompts are effectively ignored. These baselines were further refined per-video and per-generator during the human-in-the-loop video validation stage (Appendix~\ref{app:benchmark-video-validation}), where annotators added tailored keywords to each rejected output to suppress its specific artifacts during regeneration.

\begin{table}[!h]
    \centering
    \footnotesize
    \caption{Negative-prompt baselines used for each generator. Each baseline combines the author-recommended strings with global English-language keywords added to target recurring artifacts. The Wan2.1 and Self-Forcing entries are shown in English translation for uniformity; the strings actually fed to the generators, and recorded in the released JSON sidecars, are the original Chinese defaults of the upstream Wan2.1 release.}
    \label{tab:baseline_negatives_v2}
    \begin{tabularx}{\columnwidth}{lX}
        \toprule
        \textbf{Model} & \textbf{Baseline Negative Prompt} \\
        \midrule
        \makecell[l]{Wan2.1 \\ \& Self-Forcing} & \texttt{Vibrant colors, overexposed, static, blurry details, artwork, painting, image, still, overall grayish, worst quality, low quality, JPEG compression residue, ugly, incomplete, extra fingers, poorly drawn hands, poorly drawn faces, deformed, disfigured, distorted limbs, fingers fused together, static image, cluttered background, three legs, walking backwards} \\
        \addlinespace
        SkyReels-V2 & \texttt{Bright tones, overexposed, static, blurred details, paintings, images, overall gray, worst quality, low quality, JPEG compression residue, ugly, incomplete, extra fingers, poorly drawn hands, poorly drawn faces, deformed, disfigured, misshapen limbs, fused fingers, still picture, messy background, three legs, walking backwards} \\
        \addlinespace
        CogVideoX & \texttt{Overexposed, static, blurred, worst quality, low quality, JPEG compression, ugly, incomplete, extra fingers, poorly drawn, deformed, disfigured, misshapen limbs, fused fingers, messy background, three legs, blurred eyes} \\
        \addlinespace
        LTX-2.3 & \texttt{blurry, out of focus, overexposed, underexposed, low contrast, washed out colors, excessive noise, grainy texture, poor lighting, flickering, motion blur, distorted proportions, unnatural skin tones, deformed facial features, asymmetrical face, missing facial features, extra limbs, disfigured hands, wrong hand count, artifacts around text, inconsistent perspective, camera shake, incorrect depth of field, background too sharp, background clutter, distracting reflections, harsh shadows, inconsistent lighting direction, color banding, cartoonish rendering, 3D CGI look, unrealistic materials, uncanny valley effect, incorrect ethnicity, wrong gender, exaggerated expressions, wrong gaze direction, mismatched lip sync, silent or muted audio, distorted voice, robotic voice, echo, background noise, off-sync audio, incorrect dialogue, added dialogue, repetitive speech, jittery movement, awkward pauses, incorrect timing, unnatural transitions, inconsistent framing, tilted camera, flat lighting, inconsistent tone, cinematic oversaturation, stylized filters, or AI artifacts} \\
        \addlinespace
        MAGI-1-Distilled & - \\
        \addlinespace
        Helios-Distilled & - \\
        \addlinespace
        \makecell[l]{daVinci-MagiHuman \\ -Distilled} & - \\
        \bottomrule
    \end{tabularx}
\end{table}

\newpage

\subsection{Manual Prompt Validation Details}
\label{app:benchmark-validation}

To ensure the benchmark's semantic fidelity and ethical compliance, every structured caption produced by the protocol of Section~\ref{app:benchmark-prompts} underwent a rigorous review process prior to video generation. We employed a team of three independent annotators to validate the 1,363 captions through field-by-field evaluation against two primary criteria: semantic consistency with the source video (correcting VLM hallucinations) and the absence of overt sensitive or harmful content. If a caption was flagged by at least one annotator, the rewriting task was assigned to the flagging annotator (or randomly assigned in cases of multiple flags) to neutralize the issue while preserving visual and narrative intent. This iterative loop continued until all captions passed without flags from any annotator. Of the initial 1,363 captions, 70 (5.1\%) were flagged during manual review and subsequently revised.

Following this human review, we deployed a Large Language Model (LLM)~\citep{grattafiori2024llama} as an additional automated guardrail to detect subtle sensitive content that might have been missed during manual inspection. The LLM applied the system instruction in Figure~\ref{fig:safety_prompt_v2}, flagging content across seven thematic categories. An additional 38 captions (2.8\%) were flagged during automated screening and were returned to human annotators for rewriting that neutralized sensitive elements while preserving visual and narrative intent. These iterations continued until all 1,363 captions passed both human and automated screening.

\begin{figure}[!h]
\centering
\begin{tcolorbox}[
    colback=gray!5,
    colframe=gray!40,
    boxrule=0.5pt,
    arc=2mm,
    left=6pt, right=6pt, top=5pt, bottom=5pt,
    fontupper=\footnotesize\ttfamily
]
System Instruction: You are given JSON files containing multiple textual metadata fields (short\_description, dense\_description, background, style, shot\_type, camera\_movement, lighting, atmosphere) that describe visual content in videos, generated from a captioning model. Your task is to review each textual field and identify any occurrence of terms or references related to sensitive, NSFW, or politically/ethically controversial content. The keywords and expressions listed below are provided solely as examples: you must also identify any term that conveys similar meaning or carries comparable sensitive implications, including those expressed through slight variations in spelling, morphology, or wording.

\vspace{0.5em}
Sensitive content includes, but is not limited to, references to:

\begin{itemize}
    \item Violence and warfare, such as: war, aggression, conflict, battle, attack, bomb, explosion, missile, drone, airstrike, invasion, homicide, occupation, military, soldier, combat, casualty, terrorism, terrorist.
    \item Weapons, such as: gun, rifle, pistol, weapon, firearm, shooting, sniper, assault, grenade, ammunition.
    \item Vulnerable individuals and human rights concerns, such as: child, kid, minor, baby, orphan, refugee, displaced, abuse, trafficking, exploitation.
    \item Political figures and political content, including names like Trump, Putin, or roles and institutions such as president, prime minister, government.
    \item Geopolitical references, such as Ukraine, Russia, Iran, China, Europe.
    \item Symbolic references to national identity, including the term ``flag''. This should only be considered sensitive if it explicitly refers to a national or geopolitical entity.
    \item References to real existing persons, including but not limited to public figures, celebrities, politicians, or any known individual. This includes any explicit naming or implicit mention that can identify such a person.
\end{itemize}
\end{tcolorbox}
\caption{System instruction used for the automated safety screening performed with the LLM~\citep{grattafiori2024llama}. The model acts as an adversarial filter, flagging content across seven specific risk categories.}
\label{fig:safety_prompt_v2}
\end{figure}

\newpage

\subsection{Reference Frame Selection for I2V}
\label{app:benchmark-i2v-frames}

\begin{figure}[!t]
\centering
\definecolor{frameGreen}{RGB}{34,139,34}
\definecolor{frameRed}{RGB}{200,50,50}
\setlength{\fboxsep}{0pt}
\setlength{\fboxrule}{2.5pt}

\begin{minipage}{\linewidth}
\centering
{\scriptsize \textbf{\textsc{FF++/000}}}\\[2pt]
\fcolorbox{frameGreen}{white}{\includegraphics[width=0.22\linewidth]{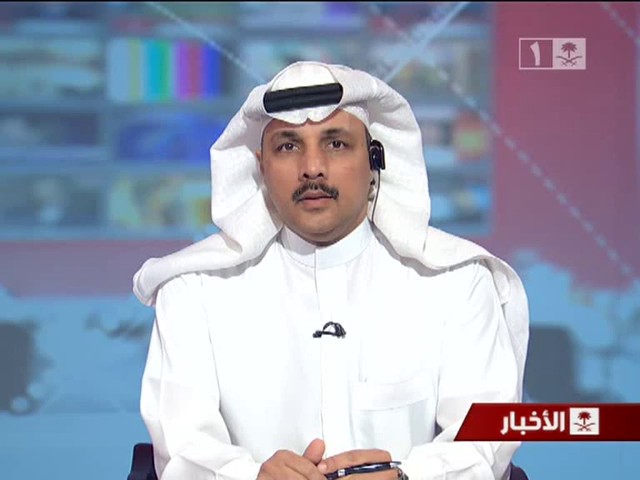}}\hspace{2pt}
\fcolorbox{frameRed}{white}{\includegraphics[width=0.22\linewidth]{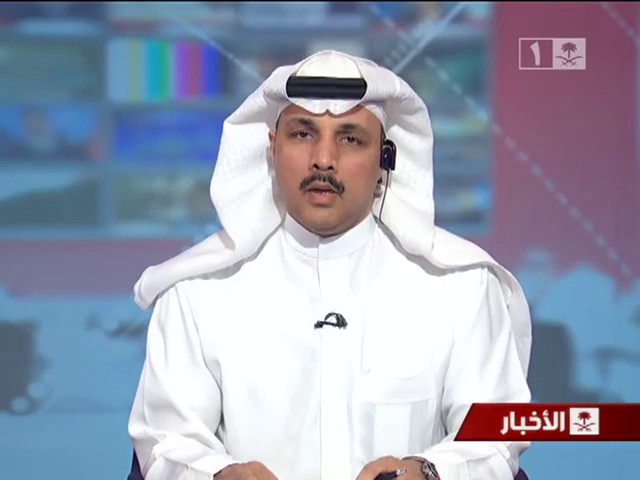}}
\end{minipage}

\vspace{0.6em}

\begin{minipage}{\linewidth}
\centering
{\scriptsize \textbf{\textsc{FF++/240}}}\\[2pt]
\fcolorbox{frameRed}{white}{\includegraphics[width=0.22\linewidth]{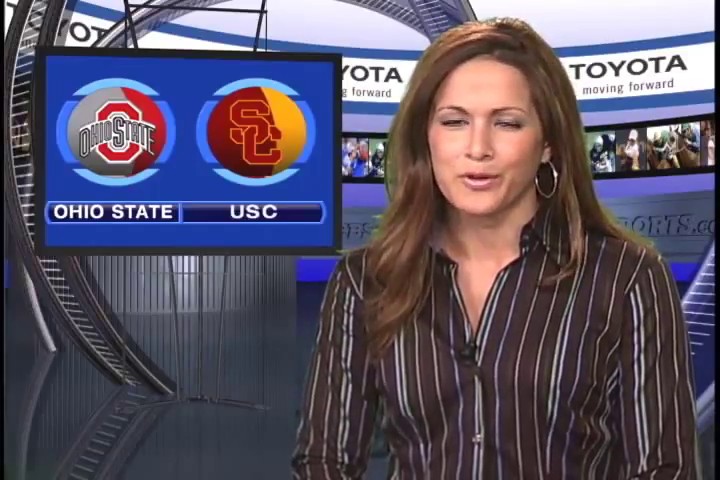}}\hspace{2pt}
\fcolorbox{frameRed}{white}{\includegraphics[width=0.22\linewidth]{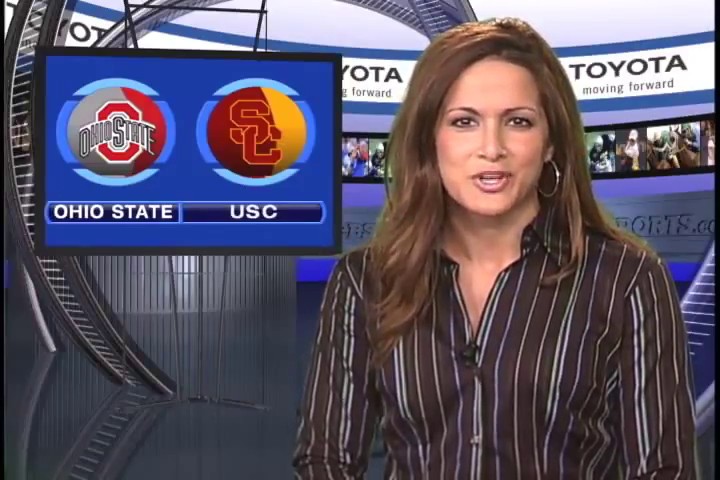}}\hspace{2pt}
\fcolorbox{frameGreen}{white}{\includegraphics[width=0.22\linewidth]{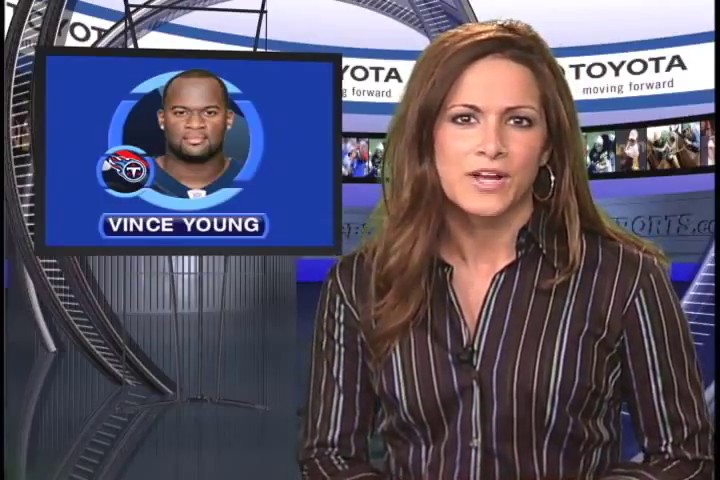}}\hspace{2pt}
\fcolorbox{frameRed}{white}{\includegraphics[width=0.22\linewidth]{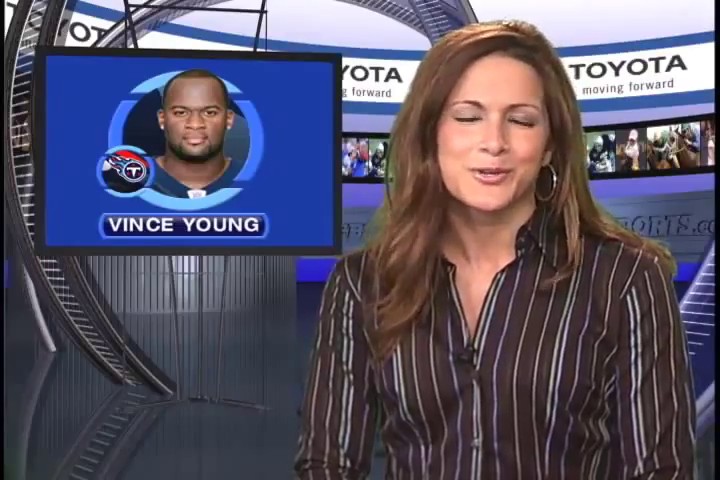}}
\end{minipage}

\vspace{0.6em}

\begin{minipage}{\linewidth}
\centering
{\scriptsize \textbf{\textsc{DFD/01\_\_kitchen\_still}}}\\[2pt]
\fcolorbox{frameGreen}{white}{\includegraphics[width=0.22\linewidth]{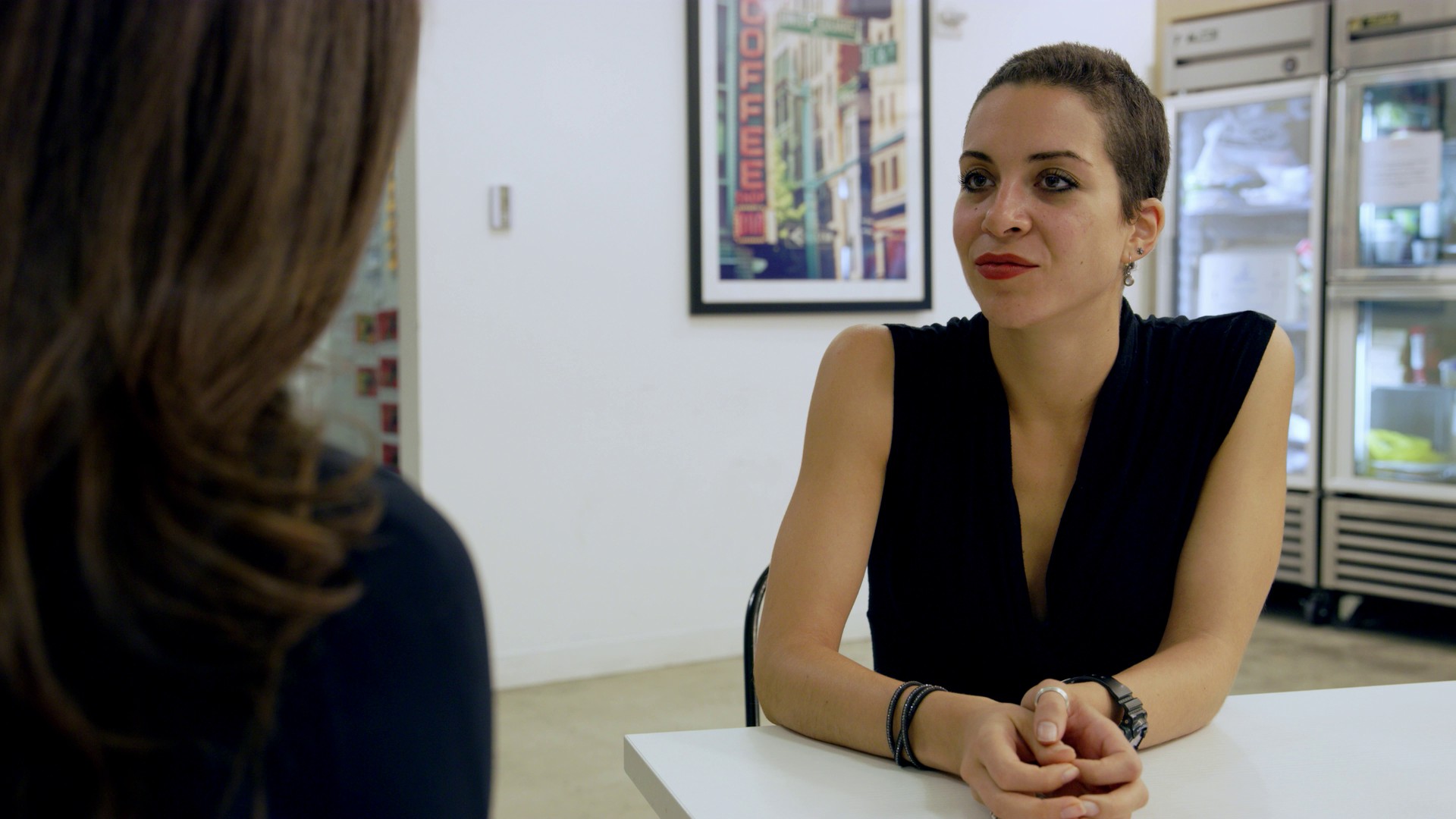}}\hspace{2pt}
\fcolorbox{frameRed}{white}{\includegraphics[width=0.22\linewidth]{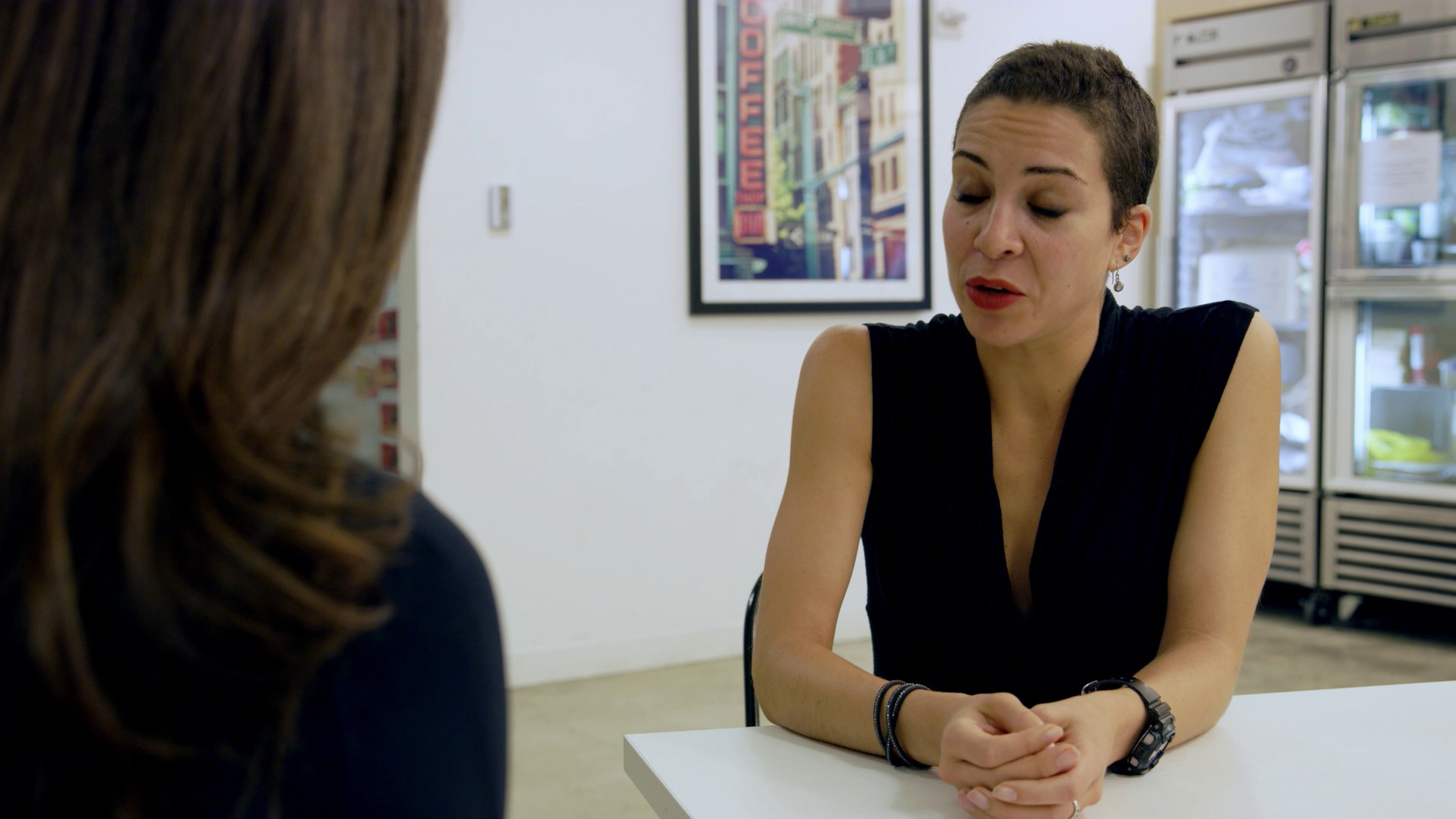}}\hspace{2pt}
\fcolorbox{frameRed}{white}{\includegraphics[width=0.22\linewidth]{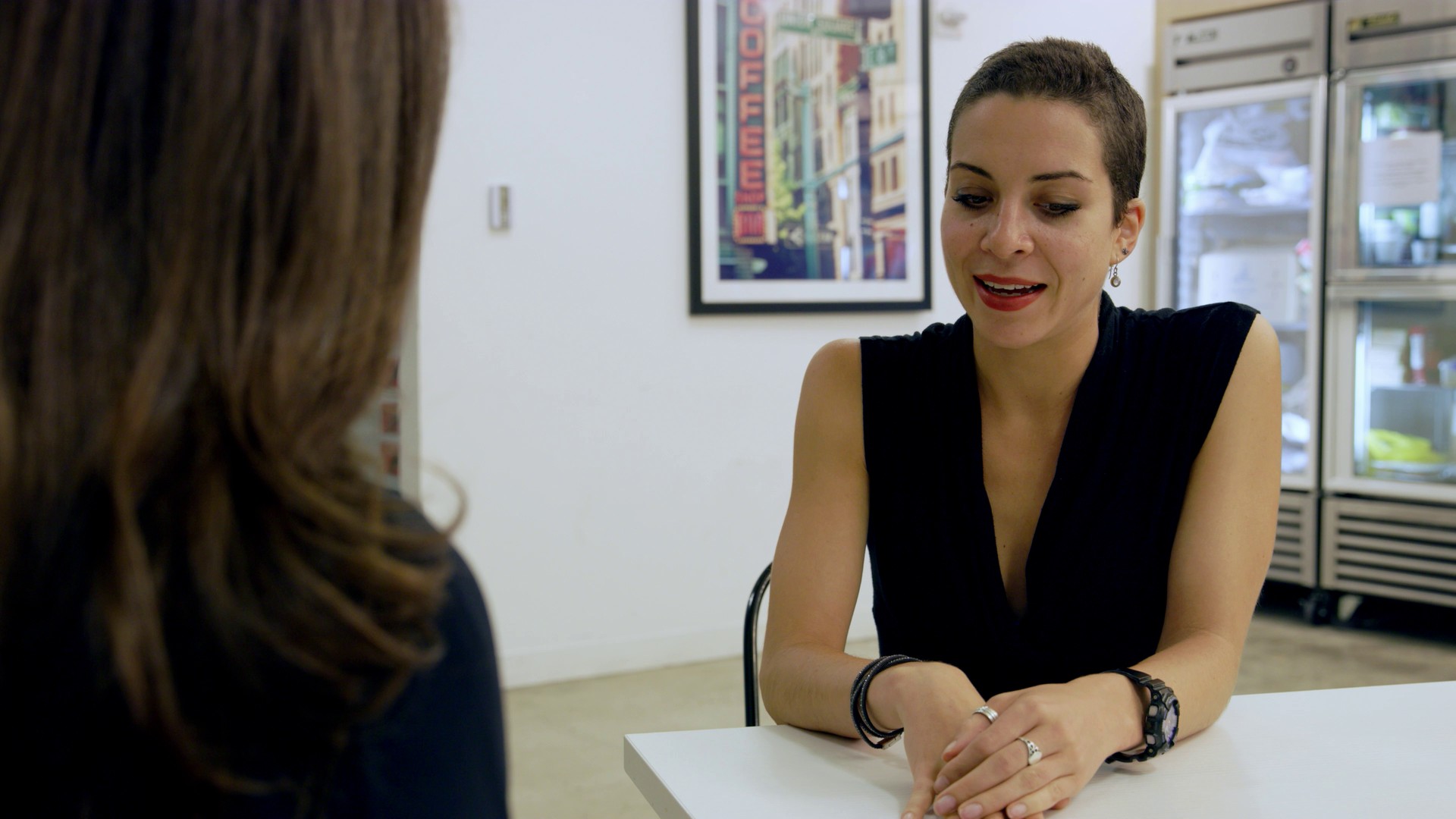}}\hspace{2pt}
\fcolorbox{frameRed}{white}{\includegraphics[width=0.22\linewidth]{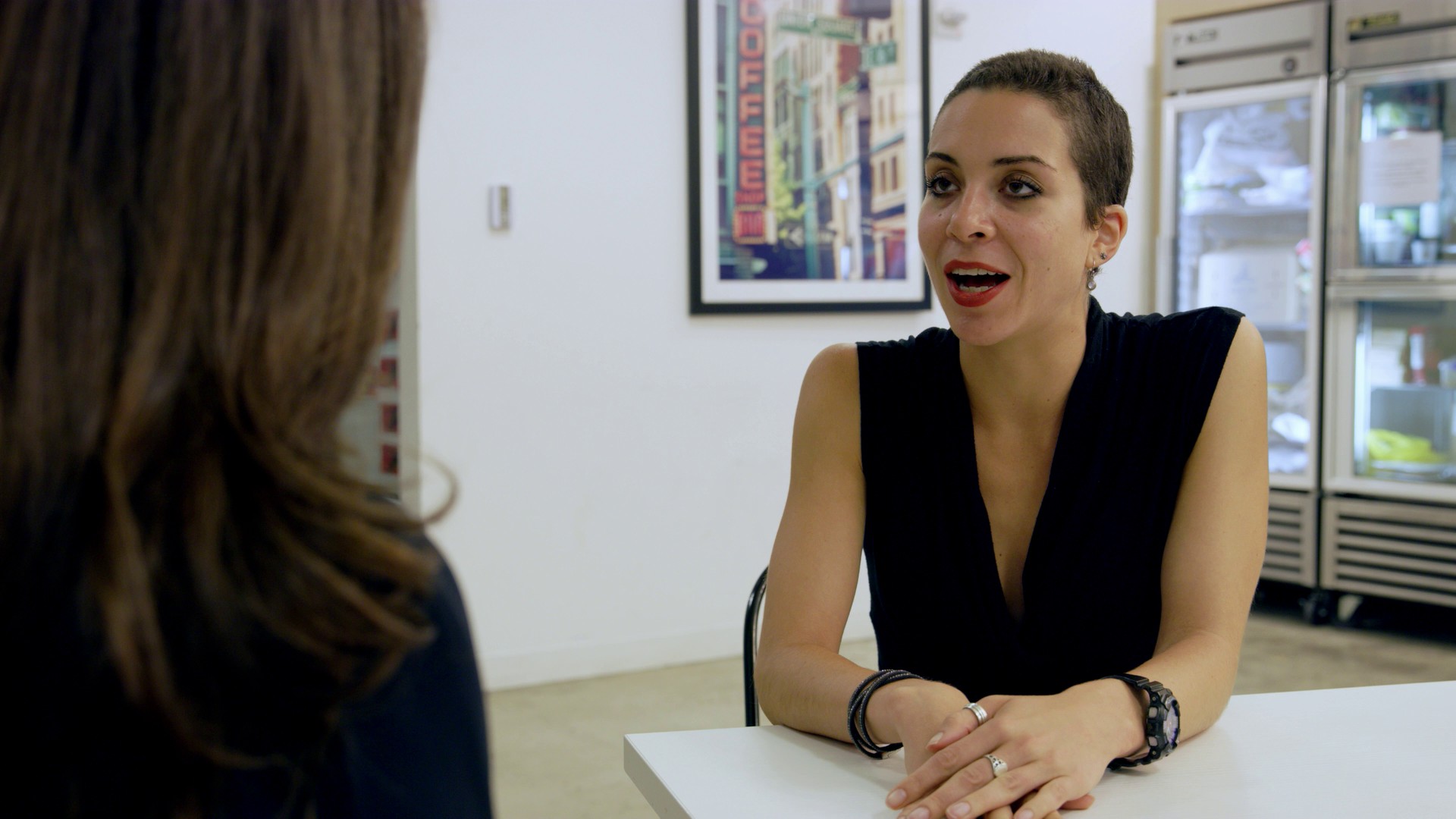}}
\end{minipage}

\vspace{0.6em}

\begin{minipage}{\linewidth}
\centering
{\scriptsize \textbf{\textsc{DFD/26\_\_exit\_phone\_room}}}\\[2pt]
\fcolorbox{frameRed}{white}{\includegraphics[width=0.22\linewidth]{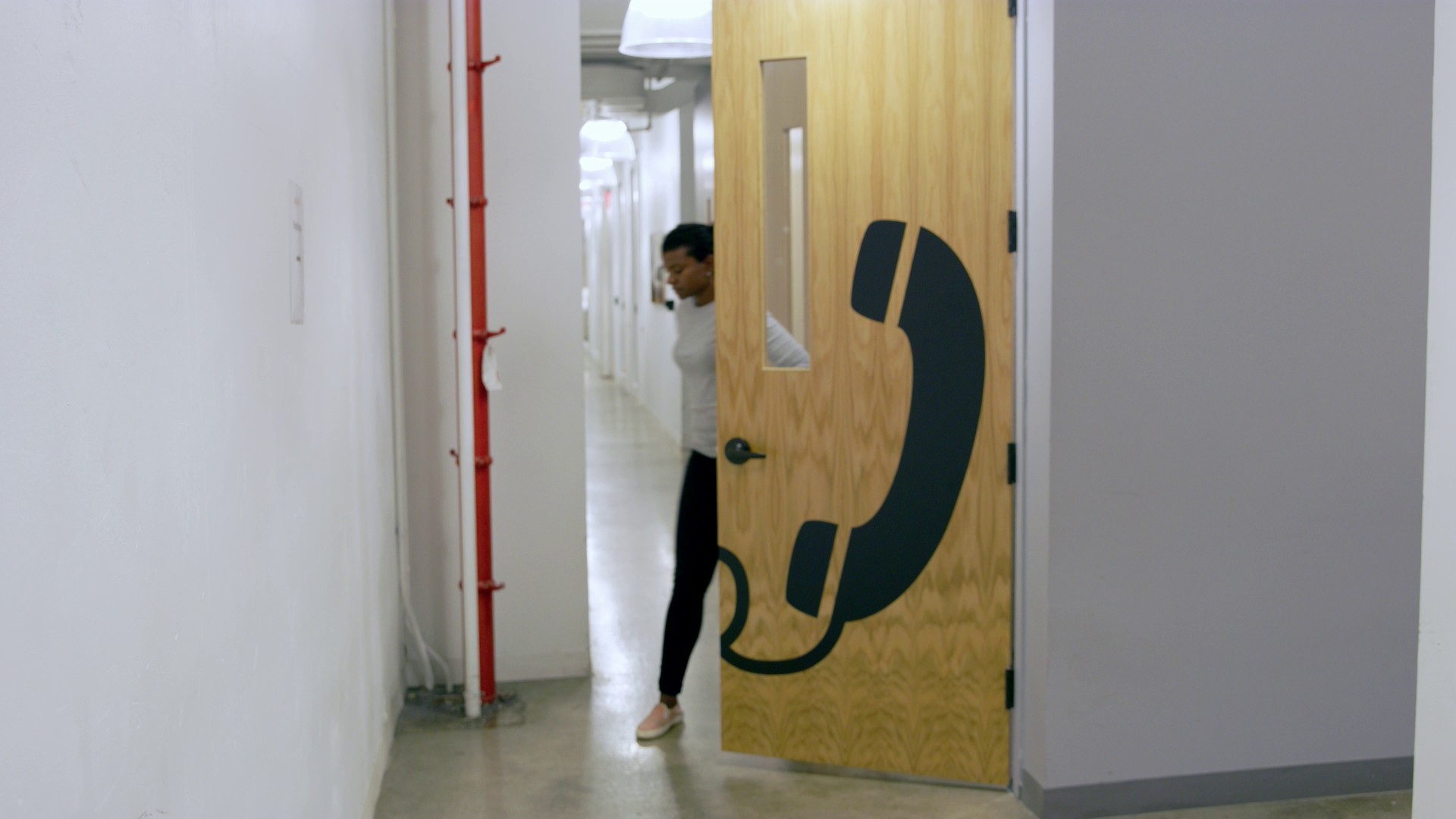}}\hspace{2pt}
\fcolorbox{frameGreen}{white}{\includegraphics[width=0.22\linewidth]{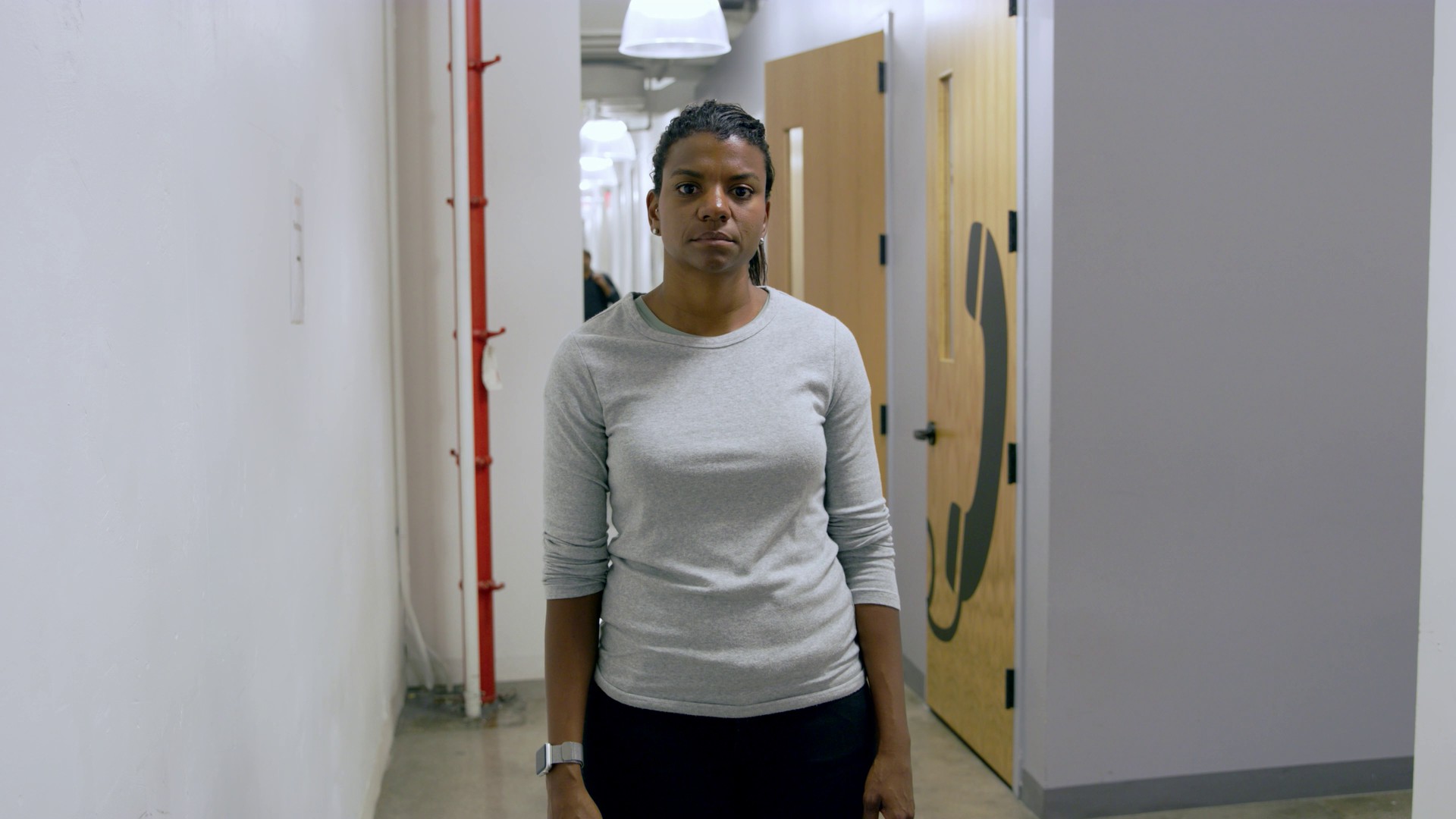}}
\end{minipage}

\caption{Reference frame selection for I2V on four representative source videos. All I-frames extracted from each source are shown in temporal order; the frame chosen by the human annotator as the I2V conditioning reference is bordered in green, while rejected candidates are bordered in red.}
\label{fig:i2v_frame_selection}
\end{figure}

For I2V generation, the reference frame fixes the subject's identity, anatomy, composition, and lighting from the first frame onward, so a poorly chosen frame propagates artifacts throughout the entire synthetic clip. To minimize this risk, human annotators select the reference frame manually. For each source video, candidate frames are restricted to the I-frames~\citep{wiegand2003overview} of the first half of the duration, since I-frames are encoded independently in the video bitstream and therefore provide artifact-free reconstruction, unlike P- and B-frames which carry inter-frame compression artifacts that would propagate into the I2V output. Restricting selection to the first half preserves the remainder as an independent ground-truth segment for evaluation. Within this range, annotators inspect the source video alongside the structured caption and select the frame that best matches the caption's description of subject, action, and scene, prioritizing a large, centered, and sharp face in a neutral and frontal pose. Figure~\ref{fig:i2v_frame_selection} illustrates the selection on four representative source videos.

\subsection{Per-Generator Specifications and Hyperparameters}
\label{app:benchmark-hyperparams}


This appendix expands Section~\ref{subsec:vid_synth_qc} by detailing the per-generator setup, fully reported in Tables~\ref{tab:gen_specs_inference_v2} and~\ref{tab:gen_specs_output_v2}.

The eight selected generators span four architectural variants of the DiT paradigm~\citep{peebles2023scalable}. Wan2.1~\citep{wan2025wan} and CogVideoX~\citep{yang2024cogvideox} follow the standard bidirectional DiT design with a 3D causal VAE~\citep{kingma2013auto} compressing space-time tokens; Wan2.1 uses cross-attention text conditioning, while CogVideoX adopts an expert-AdaLN variant with separate modulation parameters for text and vision tokens within shared transformer weights. SkyReels-V2~\citep{chen2025skyreels}, MAGI-1~\citep{teng2025magi}, and Helios~\citep{yuan2026helios} are autoregressive long-video models: SkyReels-V2 post-trains a Wan2.1 backbone with a Diffusion Forcing scheme that uses a non-decreasing per-frame noise schedule to extend video beyond the training-time horizon, MAGI-1 generates fixed-length latent chunks one at a time conditioning each chunk on previously decoded ones, and Helios formulates long-video generation as video continuation, denoising a fresh segment conditioned on prior frames. Self-Forcing~\citep{huang2025self} couples diffusion with a causal student trained via self-rollout to eliminate the train/test exposure-bias gap, yielding a distilled model that operates with very few denoising steps. LTX-2.3~\citep{hacohen2026ltx} and daVinci-MagiHuman~\citep{chern2026speed} extend the bidirectional DiT paradigm with joint audio-visual generation but realize it differently: LTX-2.3 uses an asymmetric dual-stream architecture that jointly denoises audio and video latents, while daVinci-MagiHuman adopts a single-stream Transformer specialized for human-centric content, processing text, video, and audio tokens within a unified self-attention sequence under a sandwich architecture and a timestep-free denoising scheme. The distilled checkpoints of MAGI-1, Helios, and daVinci-MagiHuman bake the classifier-free guidance signal directly into the model weights, allowing inference at $\textrm{cfg}=1$ without negative prompts.

The eight generators differ substantially across all inference hyperparameters reported in Table~\ref{tab:gen_specs_inference_v2}. Diffusion-step counts span from 8 + 5 super-resolution (daVinci-MagiHuman) and 16 (MAGI-1) up to 100 (CogVideoX), with Wan2.1 at 50 (T2V) or 40 (I2V), SkyReels-V2 and Self-Forcing at 50, and LTX-2.3 at 30 (with distilled LoRA at half strength); the distilled Helios checkpoint runs a three-stage pyramid scheduler with [2, 2, 3] effective steps. Classifier-free guidance ranges from $\textrm{cfg}=1$ (distilled MAGI-1, Helios, and daVinci-MagiHuman) to $\textrm{cfg}=6.0$ (CogVideoX, SkyReels-V2), with intermediate values of 5.0 for Wan2.1 and 3.0 for LTX-2.3 and Self-Forcing. Temporal shift values range from 3.0 (Wan2.1 I2V at 480p) to 8.0 (SkyReels-V2) where applicable. Schedulers also differ: Wan2.1 uses UniPC~\citep{zhao2023unipc}; CogVideoX uses CogVideoXDPMScheduler, a DPM++~\citep{lu2022dpmpp} variant with trailing timestep spacing; SkyReels-V2, LTX-2.3, and daVinci-MagiHuman use Flow-Matching~\citep{lipman2022flow} variants; MAGI-1 uses flow-matching with custom logit-normal step weighting~\citep{teng2025magi}; Helios uses a UniPC-based HeliosScheduler~\citep{yuan2026helios}; and Self-Forcing employs a specialized few-step rollout scheme~\citep{huang2025self}.

Per-generator output dimensions are summarized in Table~\ref{tab:gen_specs_output_v2}. The temporal sampling spans 8 FPS (CogVideoX) to 25 FPS (daVinci-MagiHuman); durations were set to 5 seconds where the generator's native temporal stride permits, and to the closest feasible value (4 or 6 seconds) otherwise. Aspect ratio and spatial resolution are also generator-dependent: CogVideoX and Helios are trained in landscape only and output at $720 \times 480$ and $640 \times 384$ respectively, while the remaining six generators support both orientations natively at resolutions ranging from $832 \times 480 / 480 \times 832$ (Wan2.1) up to $1920 \times 1088 / 1088 \times 1920$ (daVinci-MagiHuman with super-resolution). For T2V we sample orientation uniformly at random within each generator's supported aspect ratios; for I2V the orientation follows the aspect ratio of the human-selected reference frame (Appendix~\ref{app:benchmark-i2v-frames}).

For Image-to-Video synthesis, the human-selected reference frame (Appendix~\ref{app:benchmark-i2v-frames}) is supplied as the first-frame conditioning signal to each I2V generator. Conditioning mechanisms vary across architectures: a clean VAE-encoded prefix (LTX-2.3, Helios), VAE-encoded first-frame latents with model-specific supplementary channels (Wan2.1, SkyReels-V2), or model-internal interfaces (CogVideoX, MAGI-1, daVinci-MagiHuman). Reference-frame preprocessing also differs. CogVideoX, SkyReels-V2, Helios, and MAGI-1 receive a center crop to match the target dimensions, performed externally before invocation; Wan2.1, LTX-2.3, and daVinci-MagiHuman handle the resize and crop internally. Self-Forcing is excluded from the I2V protocol because its training treats the entire latent sequence as part of the autoregressive rollout, exposing no clean-prefix interface for an external image.

Generation was distributed across three servers, equipped in aggregate with 2 NVIDIA RTX PRO 6000 Blackwell Server Edition (96 GB), 3 NVIDIA A100 80GB PCIe, and 1 NVIDIA H100 80GB GPUs (full hardware specifications in Appendix~\ref{app:detection-hardware}). Average inference time per video varies considerably across architectures (Table~\ref{tab:gen_specs_output_v2}): approximately 10 seconds for Self-Forcing, 30 seconds for Helios, 2.5 minutes for daVinci-MagiHuman, 3.5 minutes for LTX-2.3, 5 minutes for CogVideoX, 14 minutes for Wan2.1, 17 minutes for MAGI-1, and 24 minutes for SkyReels-V2.

\clearpage
\begin{table}[!t]
\centering
\footnotesize
\caption{Architectural family and inference hyperparameters for the eight generators in SynthForensics. T2V and I2V values are reported separately when they differ; otherwise, the value applies to both modes.}
\label{tab:gen_specs_inference_v2}
\setlength{\tabcolsep}{4pt}
\renewcommand{\arraystretch}{1.6}
\begin{tabular*}{\textwidth}{@{\extracolsep{\fill}}l|ccccccc@{}}
\toprule
\textbf{Generator} & \textbf{Release} & \textbf{Architecture} & \textbf{Model version} & \makecell[c]{\textbf{Steps}\\\textbf{(T2V/I2V)}} & \textbf{CFG} & \makecell[c]{\textbf{Shift}\\\textbf{(T2V/I2V)}} & \textbf{Scheduler} \\
\midrule
Wan2.1 & Mar 2025 & \makecell[c]{Cross-attn\\DiT} & \makecell[c]{14B (T2V) /\\14B-480P (I2V)} & 50 / 40 & 5.0 & 5.0 / 3.0 & UniPC \\
\midrule
CogVideoX & Aug 2024 & \makecell[c]{Expert-AdaLN\\DiT} & 5B & 100 & 6.0 & --- & DPM++ \\
\midrule
SkyReels-V2 & Apr 2025 & DFoT & 14B & 50 & 6.0 & 8.0 & \makecell[c]{Flow-\\Matching} \\
\midrule
MAGI-1 & May 2025 & \makecell[c]{AR\\chunk-wise} & 24B-distill & 16 & 1.0 & --- & \makecell[c]{Flow +\\logit-normal} \\
\midrule
LTX-2.3 & Mar 2026 & \makecell[c]{Asymmetric\\dual-stream DiT} & \makecell[c]{22B +\\distill-LoRA} & 30\textsuperscript{$\ddagger$} & 3.0 & --- & \makecell[c]{Flow-\\Matching} \\
\midrule
Helios & Mar 2026 & \makecell[c]{AR\\continuation} & 14B-distill & {[2,2,3]}\textsuperscript{$\S$} & 1.0 & --- & \makecell[c]{Helios-\\Scheduler} \\
\midrule
daVinci-MagiHuman & Mar 2026 & \makecell[c]{Single-stream\\sandwich} & 15B-distill & \makecell[c]{8 + 5\\SR\textsuperscript{$\P$}} & \makecell[c]{1.0 /\\2.0 SR} & --- & \makecell[c]{Flow-\\Matching} \\
\midrule
Self-Forcing & Jun 2025 & \makecell[c]{Few-step\\distilled} & 1.3B & 50 & 3.0 & 5.0 & \makecell[c]{Few-step\\rollout} \\
\bottomrule
\end{tabular*}

\vspace{0.6em}
\scriptsize
\textsuperscript{$\ddagger$} LTX-2.3 uses a two-stage inference pipeline: a base DiT denoises a low-resolution latent (30 steps), followed by a dedicated spatial upscaler. \\
\textsuperscript{$\S$} Helios distilled uses a three-stage pyramid scheduler with [2, 2, 3] effective steps per stage. \\
\textsuperscript{$\P$} daVinci-MagiHuman uses a two-stage inference pipeline: base generator (8 steps, cfg=1) followed by a super-resolution upscaler (5 steps, cfg=2).
\end{table}

\begin{table}[!b]
\centering
\footnotesize
\caption{Video output dimensions, average per-video inference time (Avg Inf. Time), and hardware used for the eight generators in SynthForensics. Resolution is reported in landscape (L) and portrait (P) variants.}
\label{tab:gen_specs_output_v2}
\setlength{\tabcolsep}{4pt}
\renewcommand{\arraystretch}{1.6}
\begin{tabular*}{\textwidth}{@{\extracolsep{\fill}}l|cccccc@{}}
\toprule
\textbf{Generator} & \textbf{Resolution (L/P)} & \textbf{FPS} & \textbf{Duration} & \textbf{Frames} & \textbf{Avg Inf. Time} & \textbf{Hardware} \\
\midrule
Wan2.1 & \makecell[c]{832$\times$480 /\\480$\times$832} & 16 & 5\,s & 81 & 14\,min & A100 \\
\midrule
CogVideoX & \makecell[c]{720$\times$480\\(L only)} & 8 & 6\,s & 49 & 5\,min & \makecell[c]{A100,\\Blackwell} \\
\midrule
SkyReels-V2 & \makecell[c]{960$\times$544 /\\544$\times$960} & 24 & 4\,s & 97 & 24\,min & \makecell[c]{A100,\\Blackwell} \\
\midrule
MAGI-1 & \makecell[c]{1280$\times$720 /\\720$\times$1280} & 24 & 4\,s & 96 & 17\,min & \makecell[c]{H100,\\Blackwell} \\
\midrule
LTX-2.3 & \makecell[c]{1536$\times$1024 /\\1024$\times$1536} & 24 & 5\,s & 121 & 3.5\,min & \makecell[c]{Blackwell,\\A100} \\
\midrule
Helios & \makecell[c]{640$\times$384\\(L only)} & 24 & 4\,s & 99 & 30\,s & A100 \\
\midrule
daVinci-MagiHuman & \makecell[c]{1920$\times$1088 /\\1088$\times$1920} & 25 & 5\,s & 125 & 2.5\,min & Blackwell \\
\midrule
Self-Forcing & \makecell[c]{832$\times$480 /\\480$\times$832} & 16 & 5\,s & 81 & 10\,s & A100 \\
\bottomrule
\end{tabular*}
\end{table}
\clearpage

\subsection{Manual Video Validation}
\label{app:benchmark-video-validation}

Despite the high baseline quality of the selected generators (Appendix~\ref{app:benchmark-sources}), diffusion sampling remains stochastic and produces occasional failures. Every synthesized video was therefore independently evaluated by five human annotators, blinded to the source generator to minimize bias, against a fixed rejection rubric. Videos flagged by at least one annotator triggered regeneration through a targeted refinement protocol; only those passing the rubric entered the final benchmark.

\subsubsection{Inspection Criteria}

Annotators flagged and rejected any video exhibiting artifacts in the following categories:
\begin{itemize}
    \item \textit{Anatomical inconsistencies}: malformed limbs (e.g., polydactyly), facial asymmetry, unnatural eye rendering, or ``melting'' skin textures during movement.
    \item \textit{Temporal artifacts}: high-frequency flickering (texture boiling), inconsistent background warping, objects disappearing or reappearing without physical cause, and static generation characterized by frozen subjects or lack of meaningful temporal dynamics.
    \item \textit{Rendering errors}: glitches, black frames, or severe compression-like artifacts inherent to the VAE decoding step.
    \item \textit{Semantic coherence}: failure to execute the action described in the positive prompt (e.g., subject emotion mismatch), camera movements contradicting the metadata, or, for I2V, output diverging from the reference frame's identity, pose, or scene.
    \item \textit{Ethical compliance}: emergence of unsafe or sensitive content in the generated video (e.g., nudity, sexual or violent imagery, recognizable real-person likenesses); such content can arise through stochastic diffusion sampling even when the conditioning prompt and reference frame had passed the screening of Appendix~\ref{app:benchmark-validation}.
    \item \textit{Audio fidelity (joint audio-visual generators)}: for LTX-2.3 and daVinci-MagiHuman, the synchronized audio track was inspected for natural quality and absence of unsafe or sensitive content.
\end{itemize}

\subsubsection{Iterative Refinement Protocol}

We adopted a conservative validation strategy: any video flagged by at least one annotator triggered rejection and regeneration. Rejected videos underwent targeted correction based on the failure mode, applied through a hierarchical strategy:
\begin{enumerate}
    \item \textit{Prompt rewriting}: for failures in semantic coherence or ethical compliance, the positive prompt was manually rewritten to reduce ambiguity or enforce stricter safety constraints.
    \item \textit{Negative prompt augmentation}: for anatomical or rendering artifacts, the model-specific negative prompts were augmented with targeted keywords (e.g., ``deformed iris'', ``glitch'') to suppress the defect in the next generation pass; for distilled checkpoints (MAGI-1, Helios, daVinci-MagiHuman) operating at $\textrm{cfg}=1$ this step was skipped, since negative prompts are ignored.
    \item \textit{Parameter adjustment}: for temporal artifacts or weak adherence to instructions, the per-generator inference hyperparameters reported in Table~\ref{tab:gen_specs_inference_v2} were tuned: classifier-free guidance was raised by $+0.5$ to $+1.0$ to strengthen semantic adherence, the diffusion-step count was increased to resolve under-generated details, and temporal shift values were adjusted to stabilize motion trajectories where supported.
\end{enumerate}

This iterative loop of generation, inspection, and refinement was repeated until a high-fidelity, coherent, and ethically sound video was successfully generated for every sample. Only the videos successfully passing this validation constitute the final 20,445 samples of the benchmark.

\subsection{Dataset Statistics, Metadata, and Quality Assessment}
\label{app:benchmark-stats}

This appendix details the per-generator composition of the SynthForensics dataset, the metadata released alongside each video, and the quantitative quality assessment of the generated content.

\newpage

\subsubsection{Per-Generator Statistics}

Table~\ref{tab:per_gen_stats_v2} reports the per-generator video counts in SynthForensics, broken down by modality (T2V, I2V) and orientation (landscape, portrait); per-generator output specifications (resolution, frame rate, duration) are reported separately in Table~\ref{tab:gen_specs_output_v2}. Each generator produces 1,363 unique videos per supported modality, matching the 1,363 source pristine videos. For T2V, orientation is sampled uniformly at random within each generator's supported aspect ratios (Section~\ref{subsec:vid_synth_qc}), yielding an approximately balanced landscape/portrait split for the six generators that support both orientations, and entirely landscape outputs for CogVideoX and Helios (which are restricted to a single landscape resolution). For I2V, orientation follows the aspect ratio of the human-selected reference frame; since reference frames almost always depict close-up faces in landscape source footage, I2V outputs are dominated by landscape orientation across all seven I2V generators. Self-Forcing is excluded from the I2V protocol (Appendix~\ref{app:benchmark-hyperparams}).

\begin{table}[!ht]
\centering
\footnotesize
\caption{Per-generator video counts in SynthForensics, broken down by modality and orientation. L = landscape, P = portrait.}
\label{tab:per_gen_stats_v2}
\setlength{\tabcolsep}{6pt}
\renewcommand{\arraystretch}{1.2}
\begin{tabular*}{\textwidth}{@{\extracolsep{\fill}}l|cc|cc|r@{}}
\toprule
& \multicolumn{2}{c|}{\textbf{T2V}} & \multicolumn{2}{c|}{\textbf{I2V}} & \\
\cmidrule(lr){2-3} \cmidrule(lr){4-5}
\textbf{Generator} & \textbf{L} & \textbf{P} & \textbf{L} & \textbf{P} & \textbf{Total} \\
\midrule
Wan2.1 & 689 & 674 & 1,361 & 2 & 2,726 \\
CogVideoX & 1,363 & --- & 1,363 & --- & 2,726 \\
SkyReels-V2 & 702 & 661 & 1,361 & 2 & 2,726 \\
MAGI-1 & 665 & 698 & 1,363 & --- & 2,726 \\
LTX-2.3 & 703 & 660 & 1,361 & 2 & 2,726 \\
Helios & 1,363 & --- & 1,363 & --- & 2,726 \\
daVinci-MagiHuman & 667 & 696 & 1,361 & 2 & 2,726 \\
Self-Forcing & 664 & 699 & --- & --- & 1,363 \\
\midrule
\textbf{Total} & \textbf{6,816} & \textbf{4,088} & \textbf{9,533} & \textbf{8} & \textbf{20,445} \\
\bottomrule
\end{tabular*}
\end{table}

\subsubsection{Released Metadata Catalog}

For each generated video, in both T2V and I2V modalities, we release a JSON sidecar that captures the complete generation pipeline; for I2V videos, the human-selected reference frame (Appendix~\ref{app:benchmark-i2v-frames}) is also released alongside the video and JSON.

The sidecar groups fields into five categories:
\begin{itemize}
    \item \textit{Generator identification}: model name, model version, HuggingFace commit hash, and SHA-256 checksums of weight shards.
    \item \textit{Source-data references}: the structured-caption JSON used to construct the prompt (which encodes the link to the originating FF++ or DFD pristine video) and, for I2V, the reference-frame filename.
    \item \textit{Conditioning signals}: the positive and negative prompts as actually fed to the model, and the random seed.
    \item \textit{Inference configuration}: the full model, runtime, and engine settings (diffusion steps, CFG scale, temporal shift, scheduler, output dimensions, and any model-specific parameters).
    \item \textit{Environment metadata}: Python, PyTorch, CUDA, and Flash-Attention versions, GPU model, and the generation script's repository commit.
\end{itemize}
Per-video generation time is also recorded. Together these fields enable rerunning of each video on compatible hardware (up to driver and library nondeterminism) and full traceability of every artifact in the benchmark. Figure~\ref{fig:json_example_metadata_v2} shows a representative example.

\clearpage
\begin{figure}[!p]
\centering
\begin{lstlisting}[basicstyle=\ttfamily\scriptsize, frame=single, breaklines=true]
{
  "generator": "LTX-2.3",
  "model_version": "ltx-2.3-22b-dev",
  "pipeline": "ti2vid_two_stages",
  "mode": "i2v",
  "checkpoint": "ltx-2.3-22b-dev.safetensors",
  "distilled_lora": "ltx-2.3-22b-distilled-lora-384.safetensors",
  "distilled_lora_strength": 0.5,
  "spatial_upsampler": "ltx-2.3-spatial-upscaler-x2-1.0.safetensors",
  "text_encoder": "gemma-3-12b",
  "prompt": "A man in traditional white attire sitting and speaking, likely in a news studio setting. [...]",
  "negative_prompt": "blurry, out of focus, overexposed, underexposed, low contrast, washed out colors, [...]",
  "seed": 1364115978,
  "width": 1536, "height": 1024, "num_frames": 121, "frame_rate": 24,
  "num_inference_steps": 30,
  "video_cfg_scale": 3.0, "video_stg_scale": 1.0, "video_stg_blocks": [28],
  "video_rescale_scale": 0.7, "video_skip_step": 0, "a2v_guidance_scale": 3.0,
  "audio_cfg_scale": 7.0, "audio_stg_scale": 1.0, "audio_stg_blocks": [28],
  "audio_rescale_scale": 0.7, "audio_skip_step": 0, "v2a_guidance_scale": 3.0,
  "quantization": null, "dtype": "bfloat16",
  "source_caption": "000.json", "source_image": "000.png",
  "image_conditioning_frame_idx": 0, "image_conditioning_strength": 1.0, "image_conditioning_crf": 0,
  "orientation": "landscape",
  "environment": {
    "repo_url": "https://github.com/Lightricks/LTX-2.git",
    "repo_commit": "9e8a28e17ac4dd9e49695223d50753a1ebda36fe",
    "model_id": "ltx-2.3-22b-dev",
    "model_url": "https://huggingface.co/Lightricks/LTX-2.3",
    "model_hf_commit": "5a9c1c680bc66c159f708143bf274739961ecd08",
    "model_sha256": {
      "ltx-2.3-22b-dev.safetensors": "7ab7225325bc403448ea84b6db2269811a880e5118cd2ee2b6282a93d585016f",
      [...]
    },
    "text_encoder_id": "gemma-3-12b",
    "text_encoder_url": "https://huggingface.co/google/gemma-3-12b-it-qat-q4_0-unquantized",
    "text_encoder_hf_commit": "68f7ee4fbd59087436ada77ed2d62f373fdd4482",
    "text_encoder_sha256": {
      "model-00001-of-00005.safetensors": "e6fb899db428481aafb45a20130457df6e247e7cb03b7d9f01ee4bc2a9a08138",
      [...]
    },
    "python_version": "3.12.12", "torch_version": "2.7.1+cu128",
    "flash_attn_version": "2.8.3", "cuda_version": "12.8",
    "gpu_name": "NVIDIA A100 80GB PCIe",
    "key_deps": {"ltx-core": "1.0.0", "ltx-pipelines": "1.0.0", "xformers": "0.0.31.post1"}
  },
  "generation_time_seconds": 264.0
}
\end{lstlisting}
\caption{Example JSON sidecar for an I2V video generated by LTX-2.3 from FaceForensics++ source video \#000. The positive and negative prompts and the weight-shard checksums are truncated for readability; the released JSONs contain the complete fields.}
\label{fig:json_example_metadata_v2}
\end{figure}
\clearpage

\subsubsection{VBench Quality Assessment}

Quantitative perceptual evaluation of synthetic video generators has become standard practice in the field, and the VBench benchmark suite has emerged as the established reference. We adopt VBench across its three releases, VBench~1.0~\citep{huang2024vbench}, VBench++~\citep{huang2024vbenchpp}, and VBench-2.0~\citep{zheng2025vbench2}, to obtain an objective, reproducible measure of the perceptual fidelity and intrinsic faithfulness of the videos in SynthForensics.

VBench~1.0 organizes video quality into 16 hierarchical dimensions partitioned into a \emph{Video Quality} group (per-frame and per-clip perceptual properties) and a \emph{Video--Condition Consistency} group (alignment between the generated video and the textual prompt). We adopt all seven Video Quality dimensions and the temporal-style dimension from the Condition Consistency group, for a total of eight metrics:
\begin{itemize}
    \item \emph{Subject consistency}: temporal consistency of the main subject's appearance.
    \item \emph{Background consistency}: temporal consistency of the background.
    \item \emph{Temporal flickering}: absence of high-frequency frame-to-frame instability.
    \item \emph{Motion smoothness}: smoothness of object and subject movement across frames.
    \item \emph{Dynamic degree}: amount of meaningful motion present in the video.
    \item \emph{Aesthetic quality}: artistic and aesthetic appeal of the rendered frames.
    \item \emph{Imaging quality}: per-frame technical quality (sharpness, color, noise).
    \item \emph{Temporal style}: temporal coherence of the depicted style across frames.
\end{itemize}
The remaining eight Condition Consistency dimensions (object class, multiple objects, human action, color, spatial relationship, scene, appearance style, and overall consistency) are excluded because they evaluate textual-prompt alignment rather than perceptual fidelity, which is the primary concern of a deepfake-detection benchmark; moreover, our paired-source protocol uses model-specific prompt construction (Appendix~\ref{app:benchmark-prompts}), so condition-consistency comparisons across generators would be confounded by prompt-template differences.

VBench++ extends VBench~1.0 to image-to-video synthesis with three additional I2V-specific dimensions: I2V subject consistency, I2V background consistency, and camera motion (whether the generated video adheres to camera-control instructions in the prompt). We adopt the first two:
\begin{itemize}
    \item \emph{I2V subject consistency}: similarity of the main subject in the generated video to its appearance in the reference frame.
    \item \emph{I2V background consistency}: similarity of the background in the generated video to that of the reference frame.
\end{itemize}
Both dimensions are computed only on the I2V outputs, where the reference-frame conditioning is well defined. The camera-motion dimension is excluded for the same reason as the Condition Consistency dimensions of VBench~1.0: it measures adherence to a textual instruction rather than perceptual fidelity, and prompt-template differences across generators would confound the comparison.

VBench-2.0 shifts the focus from superficial faithfulness to \emph{intrinsic faithfulness}, evaluating compliance with physical laws, anatomical correctness, controllability, creativity, and commonsense reasoning through generalist vision--language models combined with specialist evaluators. We restrict our evaluation to the three sub-dimensions of the \emph{Human Fidelity} group, since SynthForensics is a people-centric benchmark:
\begin{itemize}
    \item \emph{Human anatomy}: anatomical and biological correctness of the depicted subjects (limbs, joints, facial features).
    \item \emph{Human identity}: preservation of subject identity across frames.
    \item \emph{Human clothes}: realism and consistency of clothing across frames.
\end{itemize}
The remaining VBench-2.0 categories (controllability, creativity, physics, commonsense) target generation properties orthogonal to deepfake-realism assessment and are therefore excluded.

Each metric is computed across the four compression versions (\textit{Raw}, \textit{Canonical}, \textit{CRF23}, \textit{CRF40}) to quantify how compression affects both perceptual and intrinsic quality. Per-generator results are reported in Tables~\ref{tab:vbench1_master_v2}, \ref{tab:vbenchpp_master_v2}, and~\ref{tab:vbench2_master_v2}.

Across the three VBench releases, the SynthForensics generators achieve consistently high perceptual quality. VBench~1.0 subject consistency, background consistency, temporal flickering, and motion smoothness all exceed 0.95 on \textit{Raw} outputs across generators, while dynamic degree spreads widely (0.10 for daVinci-MagiHuman to 0.58 for Helios), reflecting different motion biases. VBench++ I2V-specific scores are uniformly high (0.97--0.99), confirming that I2V generators faithfully preserve the reference frame. VBench-2.0 human-identity and human-clothes scores are near-perfect across generators, but human-anatomy scores are lower and more variable (0.86--0.96 on \textit{Raw} T2V), with daVinci-MagiHuman the lowest despite its human-centric specialization, suggesting that even tailored generators occasionally produce anatomical inconsistencies. Compression robustness varies by metric: in line with expectations, aesthetic quality, imaging quality, and dynamic degree show a slight decline at \textit{CRF40} yet remain within high-quality bounds, while subject/background consistency and motion smoothness remain nearly unchanged.

\clearpage
\begin{table}[!p]
\centering
\scriptsize
\caption{VBench~1.0 scores for all eight generators across the four compression versions and both T2V and I2V modalities. Higher is better.}
\label{tab:vbench1_master_v2}
\setlength{\tabcolsep}{3pt}
\renewcommand{\arraystretch}{1.05}
\begin{tabular}{l c l|cccccccc}
\toprule
\textbf{Generator} & \textbf{Mode} & \textbf{Compression} & \textbf{\makecell{Subj.\\cons.}} & \textbf{\makecell{Bg.\\cons.}} & \textbf{\makecell{Temp.\\flicker.}} & \textbf{\makecell{Motion\\smooth.}} & \textbf{\makecell{Dynamic\\degree}} & \textbf{\makecell{Aesth.\\qual.}} & \textbf{\makecell{Imag.\\qual.}} & \textbf{\makecell{Temp.\\style}} \\
\midrule
\multirow{4}{*}{Wan2.1} & \multirow{4}{*}{T2V} & Raw & 0.978 & 0.968 & 0.989 & 0.993 & 0.326 & 0.548 & 0.726 & 0.157 \\
 & & Canonical & 0.978 & 0.966 & 0.989 & 0.993 & 0.322 & 0.541 & 0.725 & 0.158 \\
 & & CRF23 & 0.978 & 0.955 & 0.990 & 0.993 & 0.315 & 0.519 & 0.726 & 0.158 \\
 & & CRF40 & 0.975 & 0.961 & 0.991 & 0.994 & 0.244 & 0.497 & 0.681 & 0.160 \\
\midrule
\multirow{4}{*}{CogVideoX} & \multirow{4}{*}{T2V} & Raw & 0.963 & 0.951 & 0.975 & 0.987 & 0.485 & 0.489 & 0.719 & 0.217 \\
 & & Canonical & 0.963 & 0.949 & 0.975 & 0.987 & 0.486 & 0.478 & 0.718 & 0.217 \\
 & & CRF23 & 0.963 & 0.948 & 0.976 & 0.987 & 0.484 & 0.476 & 0.717 & 0.217 \\
 & & CRF40 & 0.958 & 0.953 & 0.978 & 0.987 & 0.440 & 0.463 & 0.670 & 0.219 \\
\midrule
\multirow{4}{*}{SkyReels-V2} & \multirow{4}{*}{T2V} & Raw & 0.982 & 0.973 & 0.992 & 0.995 & 0.329 & 0.528 & 0.720 & 0.213 \\
 & & Canonical & 0.982 & 0.971 & 0.992 & 0.995 & 0.328 & 0.521 & 0.719 & 0.214 \\
 & & CRF23 & 0.982 & 0.969 & 0.993 & 0.995 & 0.315 & 0.508 & 0.717 & 0.214 \\
 & & CRF40 & 0.979 & 0.970 & 0.994 & 0.996 & 0.241 & 0.488 & 0.655 & 0.218 \\
\midrule
\multirow{4}{*}{MAGI-1} & \multirow{4}{*}{T2V} & Raw & 0.980 & 0.956 & 0.989 & 0.994 & 0.307 & 0.522 & 0.721 & 0.225 \\
 & & Canonical & 0.980 & 0.956 & 0.989 & 0.994 & 0.308 & 0.507 & 0.721 & 0.226 \\
 & & CRF23 & 0.980 & 0.954 & 0.989 & 0.994 & 0.309 & 0.513 & 0.720 & 0.226 \\
 & & CRF40 & 0.977 & 0.961 & 0.991 & 0.994 & 0.255 & 0.507 & 0.690 & 0.229 \\
\midrule
\multirow{4}{*}{LTX-2.3} & \multirow{4}{*}{T2V} & Raw & 0.968 & 0.959 & 0.992 & 0.995 & 0.274 & 0.507 & 0.723 & 0.227 \\
 & & Canonical & 0.968 & 0.959 & 0.992 & 0.995 & 0.274 & 0.504 & 0.726 & 0.227 \\
 & & CRF23 & 0.968 & 0.960 & 0.992 & 0.995 & 0.273 & 0.504 & 0.725 & 0.227 \\
 & & CRF40 & 0.966 & 0.960 & 0.993 & 0.995 & 0.233 & 0.499 & 0.700 & 0.230 \\
\midrule
\multirow{4}{*}{Helios} & \multirow{4}{*}{T2V} & Raw & 0.976 & 0.962 & 0.987 & 0.994 & 0.575 & 0.521 & 0.694 & 0.158 \\
 & & Canonical & 0.976 & 0.963 & 0.987 & 0.994 & 0.575 & 0.526 & 0.695 & 0.158 \\
 & & CRF23 & 0.976 & 0.961 & 0.987 & 0.994 & 0.573 & 0.516 & 0.693 & 0.158 \\
 & & CRF40 & 0.971 & 0.961 & 0.989 & 0.994 & 0.491 & 0.475 & 0.619 & 0.159 \\
\midrule
\multirow{4}{*}{daVinci-MagiHuman} & \multirow{4}{*}{T2V} & Raw & 0.989 & 0.982 & 0.997 & 0.997 & 0.097 & 0.547 & 0.731 & 0.196 \\
 & & Canonical & 0.989 & 0.982 & 0.997 & 0.997 & 0.097 & 0.542 & 0.733 & 0.195 \\
 & & CRF23 & 0.989 & 0.979 & 0.997 & 0.996 & 0.095 & 0.540 & 0.732 & 0.196 \\
 & & CRF40 & 0.987 & 0.975 & 0.998 & 0.997 & 0.051 & 0.532 & 0.713 & 0.200 \\
\midrule
\multirow{4}{*}{Self-Forcing} & \multirow{4}{*}{T2V} & Raw & 0.977 & 0.953 & 0.990 & 0.994 & 0.150 & 0.544 & 0.729 & 0.154 \\
 & & Canonical & 0.976 & 0.951 & 0.990 & 0.994 & 0.149 & 0.529 & 0.728 & 0.154 \\
 & & CRF23 & 0.977 & 0.946 & 0.990 & 0.994 & 0.146 & 0.527 & 0.727 & 0.154 \\
 & & CRF40 & 0.973 & 0.956 & 0.992 & 0.995 & 0.121 & 0.514 & 0.663 & 0.154 \\
\midrule
\multirow{4}{*}{Wan2.1} & \multirow{4}{*}{I2V} & Raw & 0.966 & 0.958 & 0.988 & 0.992 & 0.289 & 0.468 & 0.716 & 0.160 \\
 & & Canonical & 0.966 & 0.959 & 0.988 & 0.992 & 0.289 & 0.467 & 0.718 & 0.160 \\
 & & CRF23 & 0.967 & 0.958 & 0.989 & 0.993 & 0.284 & 0.463 & 0.717 & 0.160 \\
 & & CRF40 & 0.963 & 0.960 & 0.991 & 0.993 & 0.243 & 0.442 & 0.658 & 0.159 \\
\midrule
\multirow{4}{*}{CogVideoX} & \multirow{4}{*}{I2V} & Raw & 0.969 & 0.955 & 0.983 & 0.989 & 0.232 & 0.461 & 0.696 & 0.199 \\
 & & Canonical & 0.969 & 0.955 & 0.983 & 0.989 & 0.233 & 0.466 & 0.700 & 0.199 \\
 & & CRF23 & 0.969 & 0.953 & 0.983 & 0.989 & 0.234 & 0.458 & 0.698 & 0.199 \\
 & & CRF40 & 0.967 & 0.959 & 0.985 & 0.990 & 0.187 & 0.446 & 0.647 & 0.198 \\
\midrule
\multirow{4}{*}{SkyReels-V2} & \multirow{4}{*}{I2V} & Raw & 0.967 & 0.962 & 0.991 & 0.995 & 0.302 & 0.482 & 0.718 & 0.199 \\
 & & Canonical & 0.967 & 0.963 & 0.991 & 0.995 & 0.302 & 0.481 & 0.720 & 0.199 \\
 & & CRF23 & 0.969 & 0.964 & 0.992 & 0.995 & 0.296 & 0.475 & 0.719 & 0.199 \\
 & & CRF40 & 0.966 & 0.964 & 0.993 & 0.995 & 0.246 & 0.453 & 0.673 & 0.201 \\
\midrule
\multirow{4}{*}{MAGI-1} & \multirow{4}{*}{I2V} & Raw & 0.964 & 0.949 & 0.988 & 0.994 & 0.311 & 0.479 & 0.705 & 0.076 \\
 & & Canonical & 0.964 & 0.950 & 0.988 & 0.994 & 0.311 & 0.478 & 0.708 & 0.075 \\
 & & CRF23 & 0.964 & 0.947 & 0.989 & 0.994 & 0.309 & 0.471 & 0.706 & 0.222 \\
 & & CRF40 & 0.961 & 0.953 & 0.990 & 0.994 & 0.269 & 0.462 & 0.670 & 0.224 \\
\midrule
\multirow{4}{*}{LTX-2.3} & \multirow{4}{*}{I2V} & Raw & 0.961 & 0.953 & 0.992 & 0.995 & 0.343 & 0.474 & 0.710 & 0.224 \\
 & & Canonical & 0.961 & 0.954 & 0.992 & 0.995 & 0.343 & 0.471 & 0.713 & 0.224 \\
 & & CRF23 & 0.961 & 0.954 & 0.992 & 0.995 & 0.349 & 0.469 & 0.712 & 0.224 \\
 & & CRF40 & 0.959 & 0.956 & 0.993 & 0.995 & 0.298 & 0.457 & 0.687 & 0.227 \\
\midrule
\multirow{4}{*}{Helios} & \multirow{4}{*}{I2V} & Raw & 0.970 & 0.962 & 0.990 & 0.994 & 0.553 & 0.468 & 0.691 & 0.159 \\
 & & Canonical & 0.970 & 0.963 & 0.990 & 0.994 & 0.552 & 0.472 & 0.694 & 0.159 \\
 & & CRF23 & 0.971 & 0.961 & 0.990 & 0.994 & 0.554 & 0.464 & 0.690 & 0.158 \\
 & & CRF40 & 0.966 & 0.961 & 0.992 & 0.995 & 0.449 & 0.431 & 0.593 & 0.157 \\
\midrule
\multirow{4}{*}{daVinci-MagiHuman} & \multirow{4}{*}{I2V} & Raw & 0.981 & 0.970 & 0.995 & 0.996 & 0.170 & 0.475 & 0.705 & 0.196 \\
 & & Canonical & 0.981 & 0.970 & 0.995 & 0.996 & 0.169 & 0.475 & 0.708 & 0.196 \\
 & & CRF23 & 0.981 & 0.971 & 0.996 & 0.996 & 0.160 & 0.473 & 0.707 & 0.196 \\
 & & CRF40 & 0.979 & 0.969 & 0.996 & 0.996 & 0.088 & 0.459 & 0.689 & 0.200 \\
\bottomrule
\end{tabular}
\end{table}

\clearpage
\begin{table}[!p]
\centering
\footnotesize
\caption{VBench++ I2V-specific scores for the seven I2V generators across the four compression versions. Higher is better.}
\label{tab:vbenchpp_master_v2}
\setlength{\tabcolsep}{6pt}
\renewcommand{\arraystretch}{1.15}
\begin{tabular}{l l|cc}
\toprule
\textbf{Generator} & \textbf{Compression} & \textbf{\makecell{I2V subject\\consistency}} & \textbf{\makecell{I2V background\\consistency}} \\
\midrule
\multirow{4}{*}{Wan2.1} & Raw & 0.986 & 0.987 \\
 & Canonical & 0.985 & 0.987 \\
 & CRF23 & 0.984 & 0.985 \\
 & CRF40 & 0.957 & 0.969 \\
\midrule
\multirow{4}{*}{CogVideoX} & Raw & 0.974 & 0.968 \\
 & Canonical & 0.973 & 0.968 \\
 & CRF23 & 0.973 & 0.967 \\
 & CRF40 & 0.955 & 0.958 \\
\midrule
\multirow{4}{*}{SkyReels-V2} & Raw & 0.979 & 0.982 \\
 & Canonical & 0.978 & 0.981 \\
 & CRF23 & 0.977 & 0.980 \\
 & CRF40 & 0.952 & 0.967 \\
\midrule
\multirow{4}{*}{MAGI-1} & Raw & 0.978 & 0.981 \\
 & Canonical & 0.977 & 0.981 \\
 & CRF23 & 0.977 & 0.980 \\
 & CRF40 & 0.958 & 0.970 \\
\midrule
\multirow{4}{*}{LTX-2.3} & Raw & 0.981 & 0.975 \\
 & Canonical & 0.981 & 0.975 \\
 & CRF23 & 0.981 & 0.975 \\
 & CRF40 & 0.966 & 0.968 \\
\midrule
\multirow{4}{*}{Helios} & Raw & 0.972 & 0.972 \\
 & Canonical & 0.972 & 0.972 \\
 & CRF23 & 0.971 & 0.971 \\
 & CRF40 & 0.942 & 0.956 \\
\midrule
\multirow{4}{*}{daVinci-MagiHuman} & Raw & 0.983 & 0.984 \\
 & Canonical & 0.982 & 0.983 \\
 & CRF23 & 0.982 & 0.983 \\
 & CRF40 & 0.968 & 0.975 \\
\bottomrule
\end{tabular}
\end{table}

\clearpage
\begin{table}[!p]
\centering
\scriptsize
\caption{VBench-2.0 human-centric scores for all eight generators across the four compression versions and both T2V and I2V modalities. Higher is better.}
\label{tab:vbench2_master_v2}
\setlength{\tabcolsep}{4pt}
\renewcommand{\arraystretch}{1.05}
\begin{tabular}{l c l|ccc}
\toprule
\textbf{Generator} & \textbf{Mode} & \textbf{Compression} & \textbf{Hum. anat.} & \textbf{Hum. ident.} & \textbf{Hum. clothes} \\
\midrule
\multirow{4}{*}{Wan2.1} & \multirow{4}{*}{T2V} & Raw & 0.928 & 0.973 & 0.999 \\
 & & Canonical & 0.924 & 0.973 & 0.999 \\
 & & CRF23 & 0.954 & 0.974 & 0.999 \\
 & & CRF40 & 0.981 & 0.969 & 0.999 \\
\midrule
\multirow{4}{*}{CogVideoX} & \multirow{4}{*}{T2V} & Raw & 0.944 & 0.966 & 0.988 \\
 & & Canonical & 0.942 & 0.968 & 0.989 \\
 & & CRF23 & 0.949 & 0.962 & 0.989 \\
 & & CRF40 & 0.984 & 0.942 & 0.988 \\
\midrule
\multirow{4}{*}{SkyReels-V2} & \multirow{4}{*}{T2V} & Raw & 0.946 & 0.985 & 0.997 \\
 & & Canonical & 0.943 & 0.985 & 0.997 \\
 & & CRF23 & 0.962 & 0.982 & 0.998 \\
 & & CRF40 & 0.987 & 0.974 & 0.999 \\
\midrule
\multirow{4}{*}{MAGI-1} & \multirow{4}{*}{T2V} & Raw & 0.935 & 0.983 & 1.000 \\
 & & Canonical & 0.930 & 0.982 & 1.000 \\
 & & CRF23 & 0.939 & 0.984 & 0.999 \\
 & & CRF40 & 0.974 & 0.968 & 0.999 \\
\midrule
\multirow{4}{*}{LTX-2.3} & \multirow{4}{*}{T2V} & Raw & 0.950 & 0.952 & 0.995 \\
 & & Canonical & 0.946 & 0.951 & 0.995 \\
 & & CRF23 & 0.952 & 0.949 & 0.994 \\
 & & CRF40 & 0.981 & 0.927 & 0.996 \\
\midrule
\multirow{4}{*}{Helios} & \multirow{4}{*}{T2V} & Raw & 0.963 & 0.968 & 0.997 \\
 & & Canonical & 0.960 & 0.971 & 0.997 \\
 & & CRF23 & 0.962 & 0.968 & 0.997 \\
 & & CRF40 & 0.980 & 0.905 & 0.996 \\
\midrule
\multirow{4}{*}{daVinci-MagiHuman} & \multirow{4}{*}{T2V} & Raw & 0.861 & 0.992 & 0.999 \\
 & & Canonical & 0.854 & 0.993 & 0.999 \\
 & & CRF23 & 0.907 & 0.990 & 0.999 \\
 & & CRF40 & 0.945 & 0.983 & 1.000 \\
\midrule
\multirow{4}{*}{Self-Forcing} & \multirow{4}{*}{T2V} & Raw & 0.947 & 0.955 & 1.000 \\
 & & Canonical & 0.943 & 0.956 & 1.000 \\
 & & CRF23 & 0.961 & 0.953 & 1.000 \\
 & & CRF40 & 0.988 & 0.966 & 1.000 \\
\midrule
\multirow{4}{*}{Wan2.1} & \multirow{4}{*}{I2V} & Raw & 0.909 & 0.979 & 1.000 \\
 & & Canonical & 0.907 & 0.979 & 1.000 \\
 & & CRF23 & 0.961 & 0.978 & 1.000 \\
 & & CRF40 & 0.988 & 0.942 & 1.000 \\
\midrule
\multirow{4}{*}{CogVideoX} & \multirow{4}{*}{I2V} & Raw & 0.955 & 0.983 & 0.978 \\
 & & Canonical & 0.953 & 0.982 & 0.977 \\
 & & CRF23 & 0.954 & 0.983 & 0.978 \\
 & & CRF40 & 0.984 & 0.969 & 0.979 \\
\midrule
\multirow{4}{*}{SkyReels-V2} & \multirow{4}{*}{I2V} & Raw & 0.949 & 0.980 & 0.993 \\
 & & Canonical & 0.946 & 0.980 & 0.994 \\
 & & CRF23 & 0.971 & 0.978 & 0.994 \\
 & & CRF40 & 0.992 & 0.959 & 0.994 \\
\midrule
\multirow{4}{*}{MAGI-1} & \multirow{4}{*}{I2V} & Raw & 0.958 & 0.984 & 0.999 \\
 & & Canonical & 0.955 & 0.985 & 1.000 \\
 & & CRF23 & 0.958 & 0.984 & 0.999 \\
 & & CRF40 & 0.974 & 0.973 & 1.000 \\
\midrule
\multirow{4}{*}{LTX-2.3} & \multirow{4}{*}{I2V} & Raw & 0.957 & 0.957 & 0.985 \\
 & & Canonical & 0.955 & 0.956 & 0.984 \\
 & & CRF23 & 0.960 & 0.955 & 0.982 \\
 & & CRF40 & 0.984 & 0.947 & 0.985 \\
\midrule
\multirow{4}{*}{Helios} & \multirow{4}{*}{I2V} & Raw & 0.979 & 0.964 & 0.996 \\
 & & Canonical & 0.978 & 0.964 & 0.996 \\
 & & CRF23 & 0.979 & 0.962 & 0.997 \\
 & & CRF40 & 0.988 & 0.887 & 0.997 \\
\midrule
\multirow{4}{*}{daVinci-MagiHuman} & \multirow{4}{*}{I2V} & Raw & 0.924 & 0.995 & 0.998 \\
 & & Canonical & 0.921 & 0.995 & 0.997 \\
 & & CRF23 & 0.950 & 0.995 & 0.997 \\
 & & CRF40 & 0.977 & 0.989 & 0.998 \\
\bottomrule
\end{tabular}
\end{table}
\clearpage

\subsection{Qualitative Visual Samples}
\label{app:benchmark-samples}

\begin{figure}[!p]
    \centering
    \setlength{\tabcolsep}{1pt}
    \renewcommand{\arraystretch}{0.5}
    \begin{tabular}{c ccccc}
        & $t=0$ & $t=1$ & $t=2$ & $t=3$ & $t=4$ \\
        \rotatebox{90}{\tiny \textbf{\textsc{Pristine}}} &
        \includegraphics[width=0.17\textwidth]{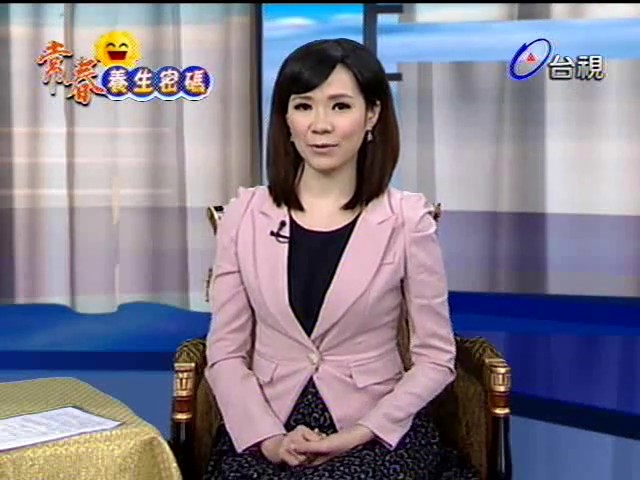} &
        \includegraphics[width=0.17\textwidth]{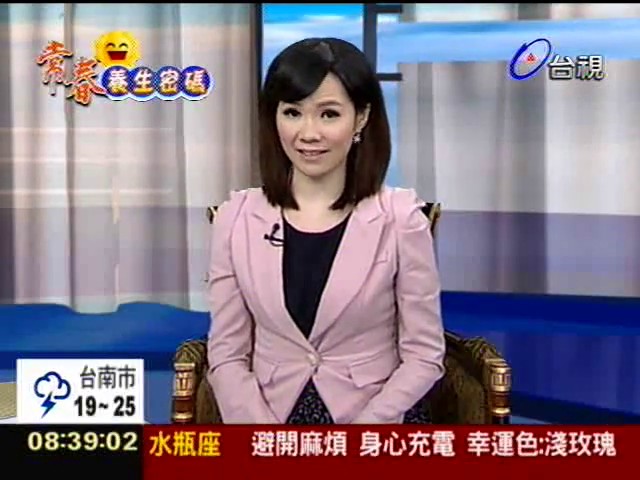} &
        \includegraphics[width=0.17\textwidth]{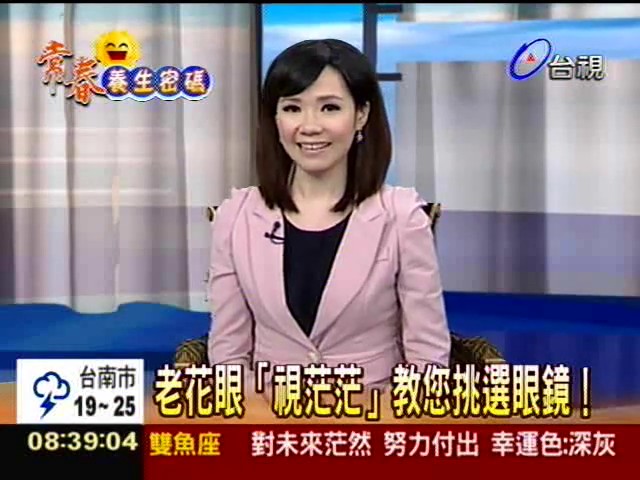} &
        \includegraphics[width=0.17\textwidth]{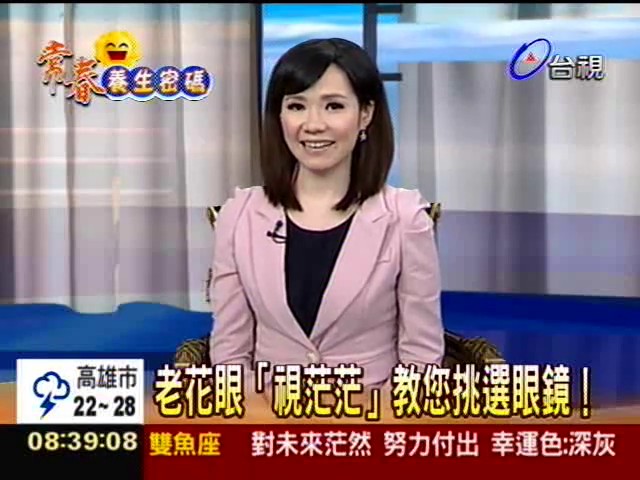} &
        \includegraphics[width=0.17\textwidth]{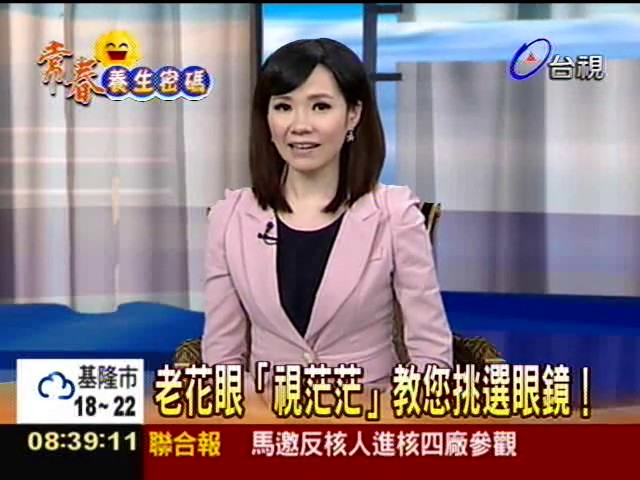} \\
        
        \rotatebox{90}{\tiny \textbf{\textsc{Wan2.1}}} &
        \includegraphics[width=0.17\textwidth]{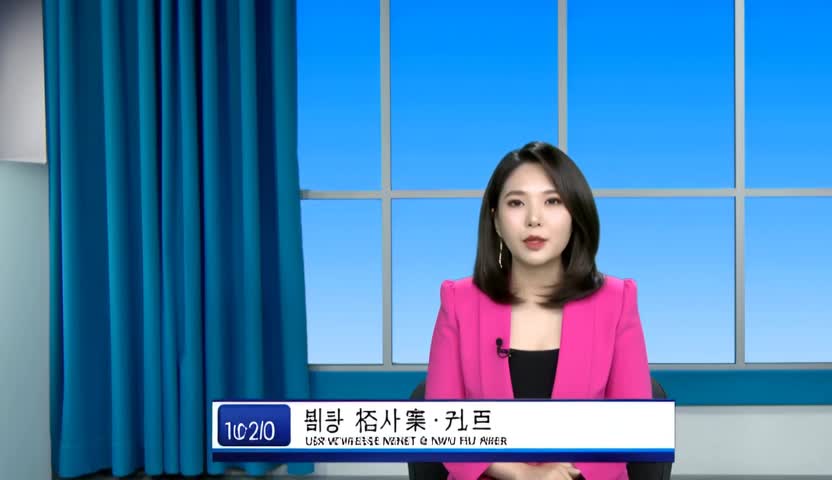} &
        \includegraphics[width=0.17\textwidth]{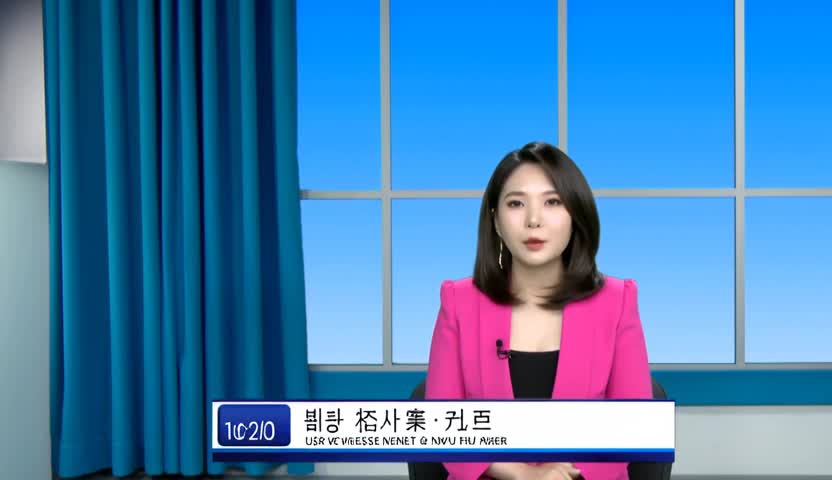} &
        \includegraphics[width=0.17\textwidth]{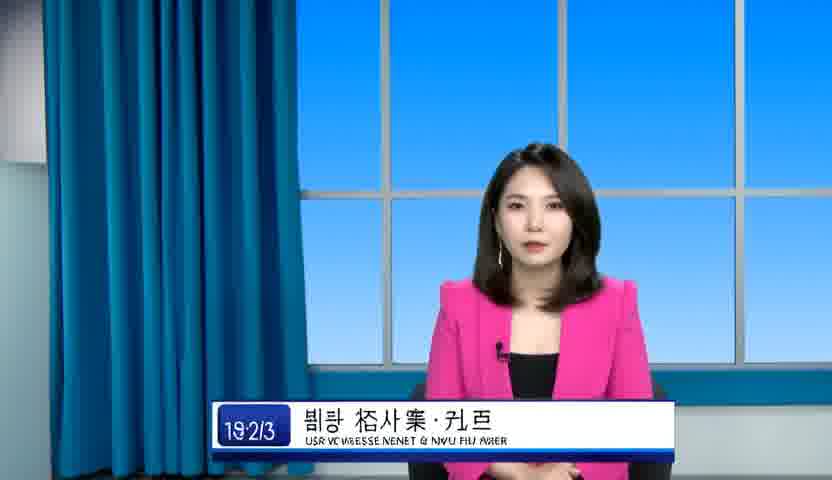} &
        \includegraphics[width=0.17\textwidth]{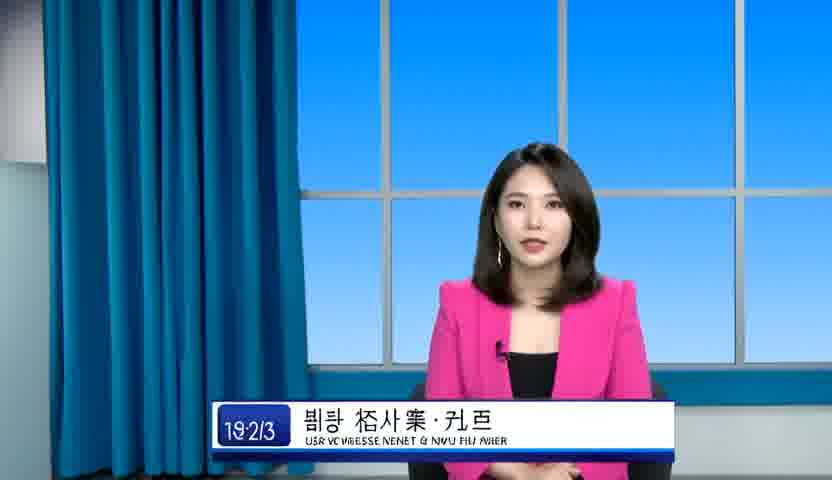} &
        \includegraphics[width=0.17\textwidth]{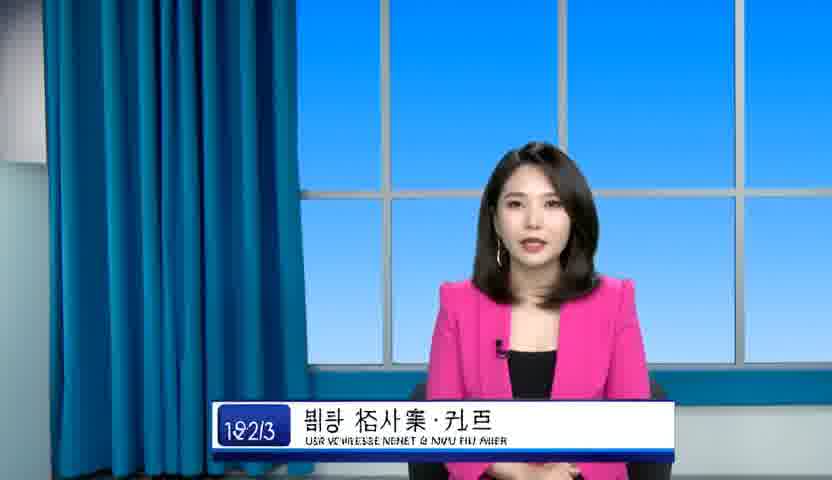} \\
        
        \rotatebox{90}{\tiny \textbf{\textsc{CogVideoX}}} &
        \includegraphics[width=0.17\textwidth]{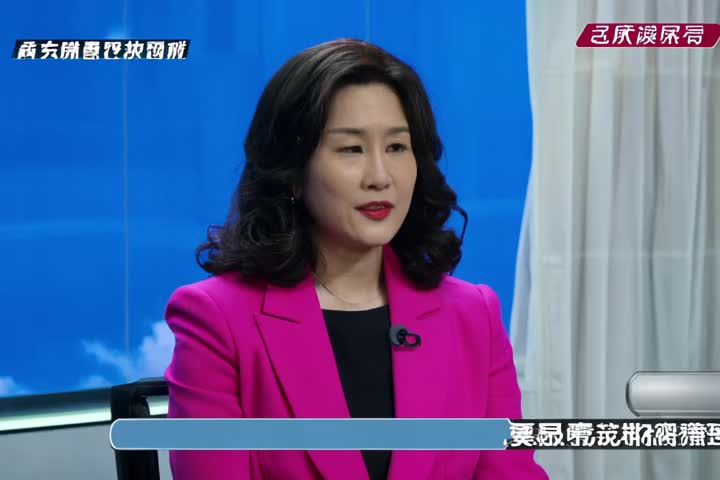} &
        \includegraphics[width=0.17\textwidth]{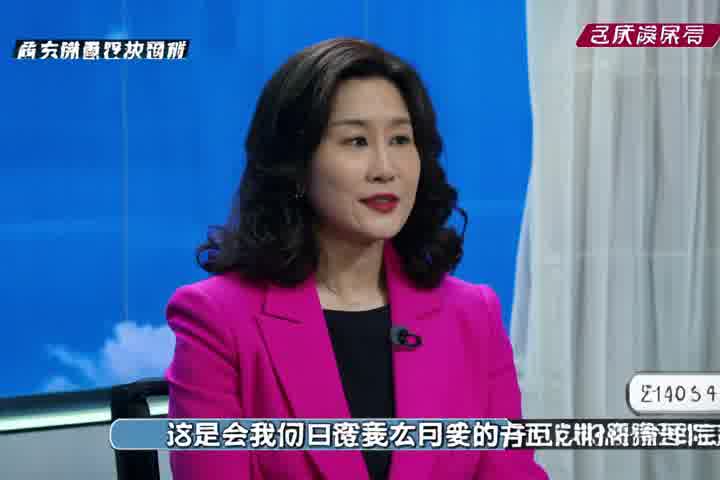} &
        \includegraphics[width=0.17\textwidth]{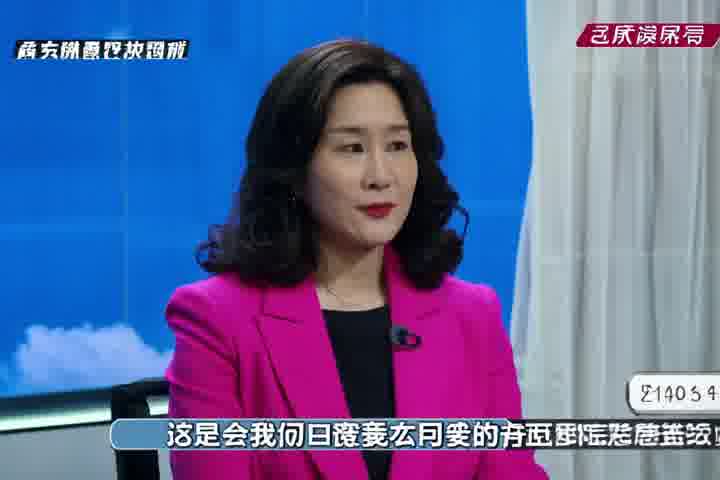} &
        \includegraphics[width=0.17\textwidth]{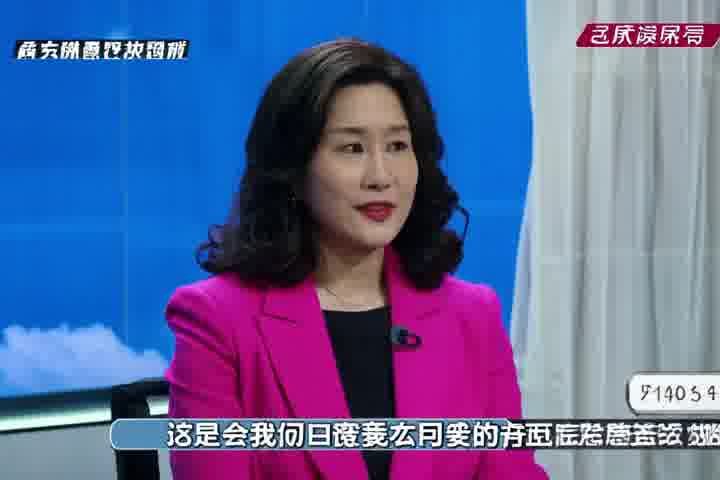} &
        \includegraphics[width=0.17\textwidth]{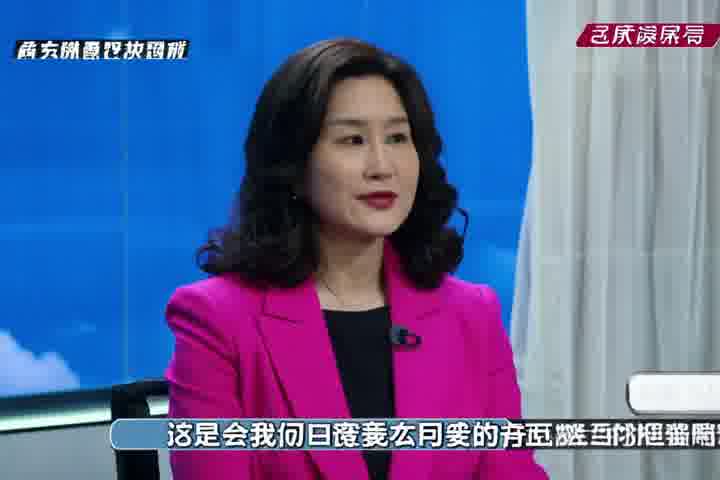} \\
        
        \rotatebox{90}{\tiny \textbf{\textsc{SkyReels-V2}}} &
        \includegraphics[width=0.17\textwidth]{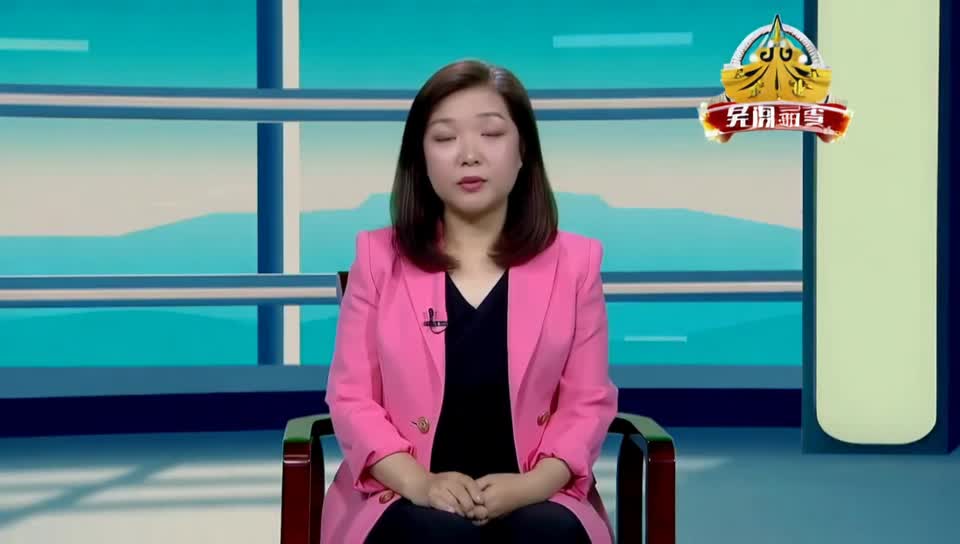} &
        \includegraphics[width=0.17\textwidth]{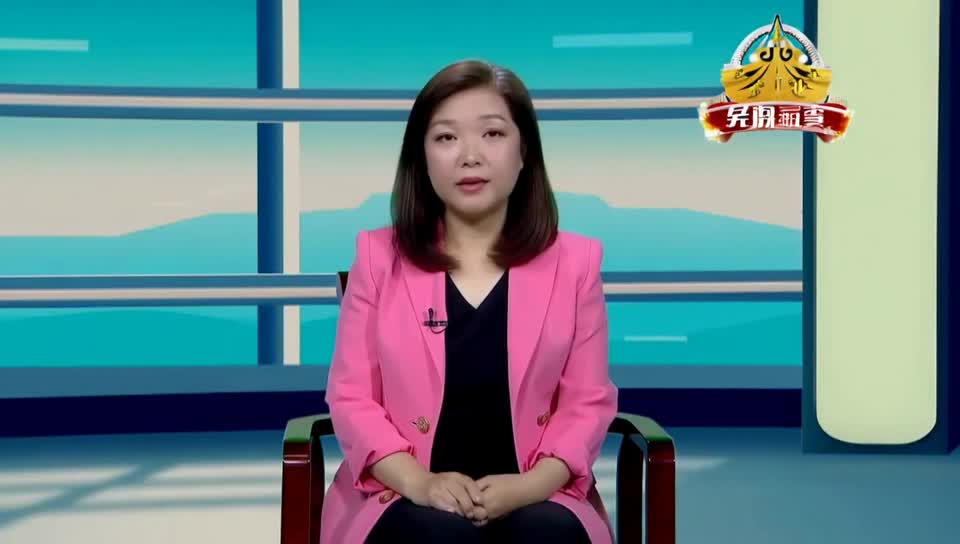} &
        \includegraphics[width=0.17\textwidth]{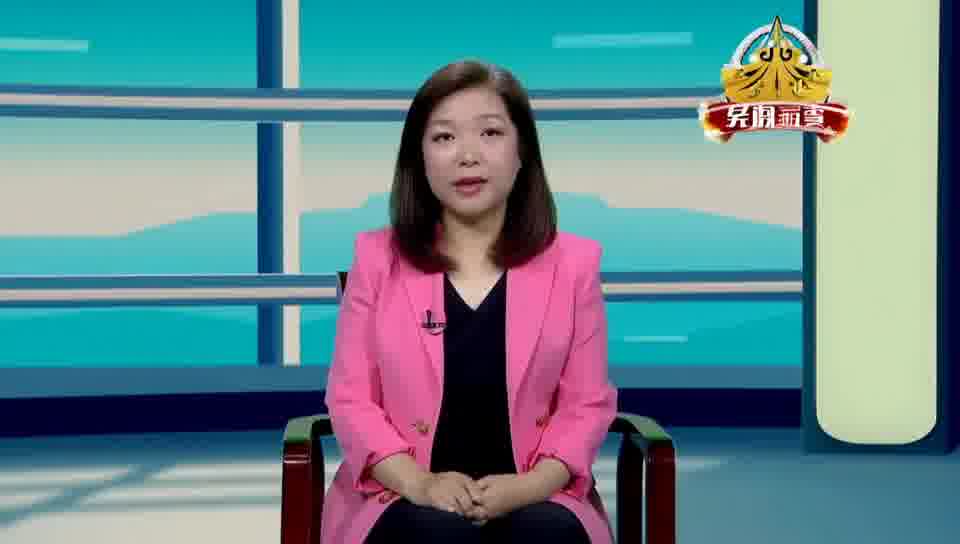} &
        \includegraphics[width=0.17\textwidth]{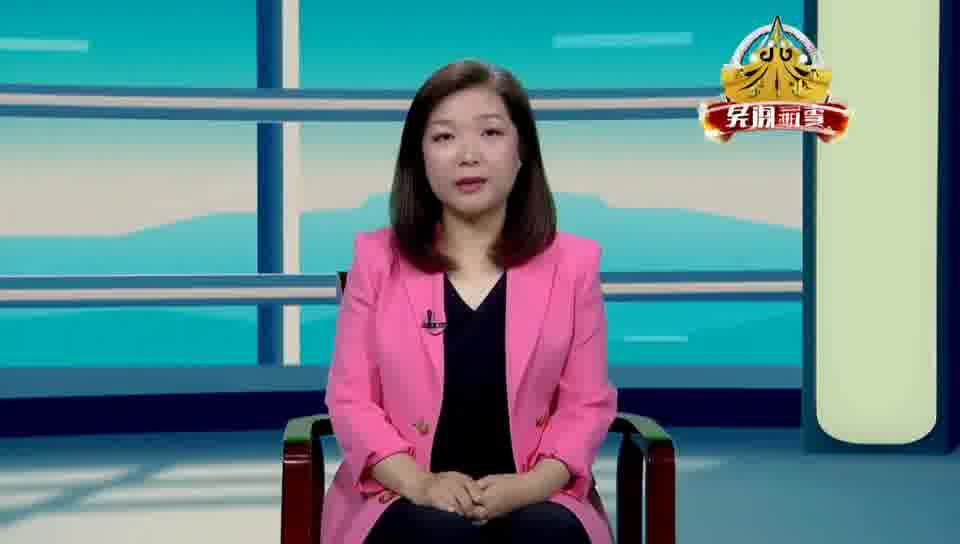} &
        \includegraphics[width=0.17\textwidth]{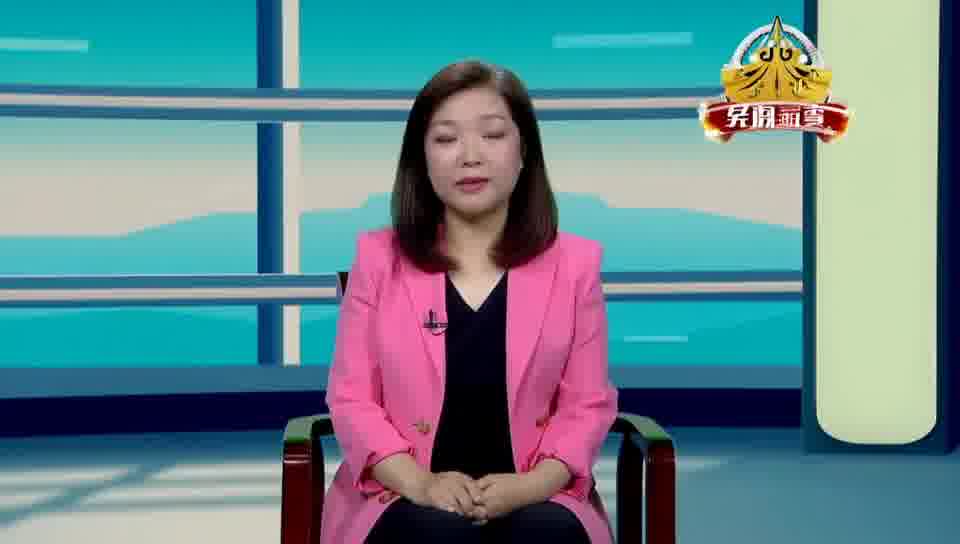} \\
        
        \rotatebox{90}{\tiny \textbf{\textsc{MAGI-1}}} &
        \includegraphics[width=0.17\textwidth]{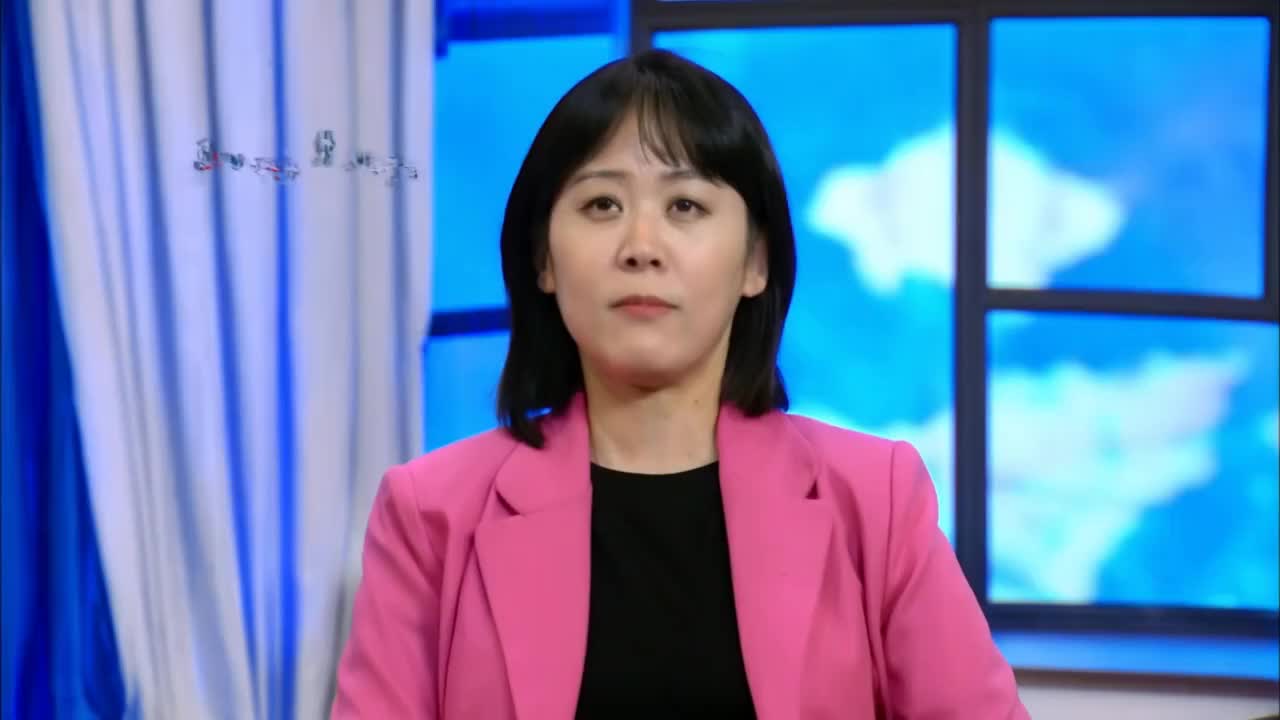} &
        \includegraphics[width=0.17\textwidth]{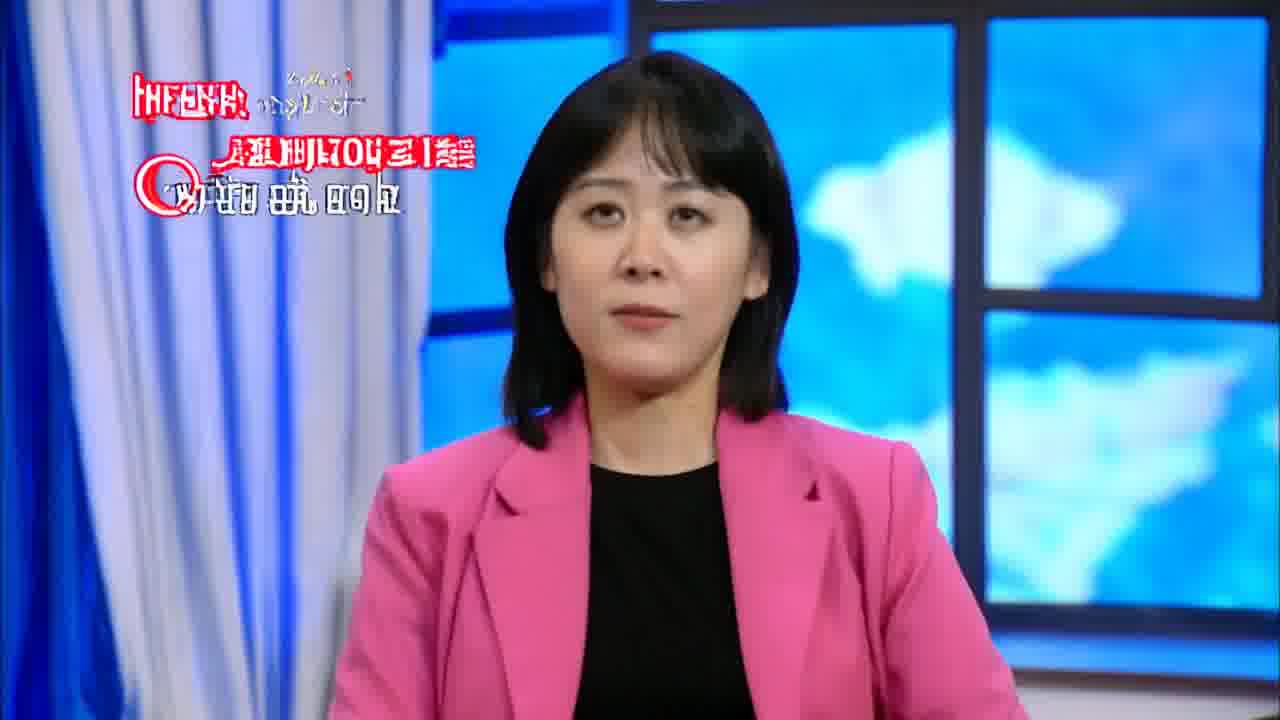} &
        \includegraphics[width=0.17\textwidth]{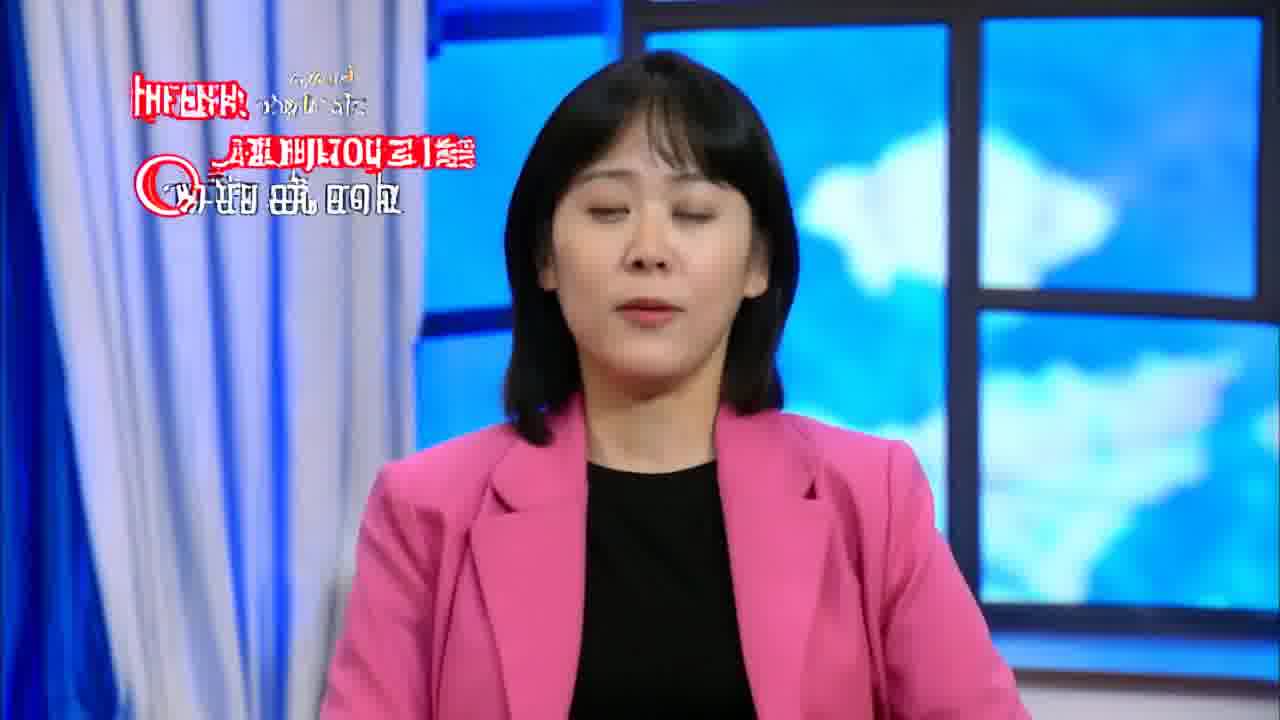} &
        \includegraphics[width=0.17\textwidth]{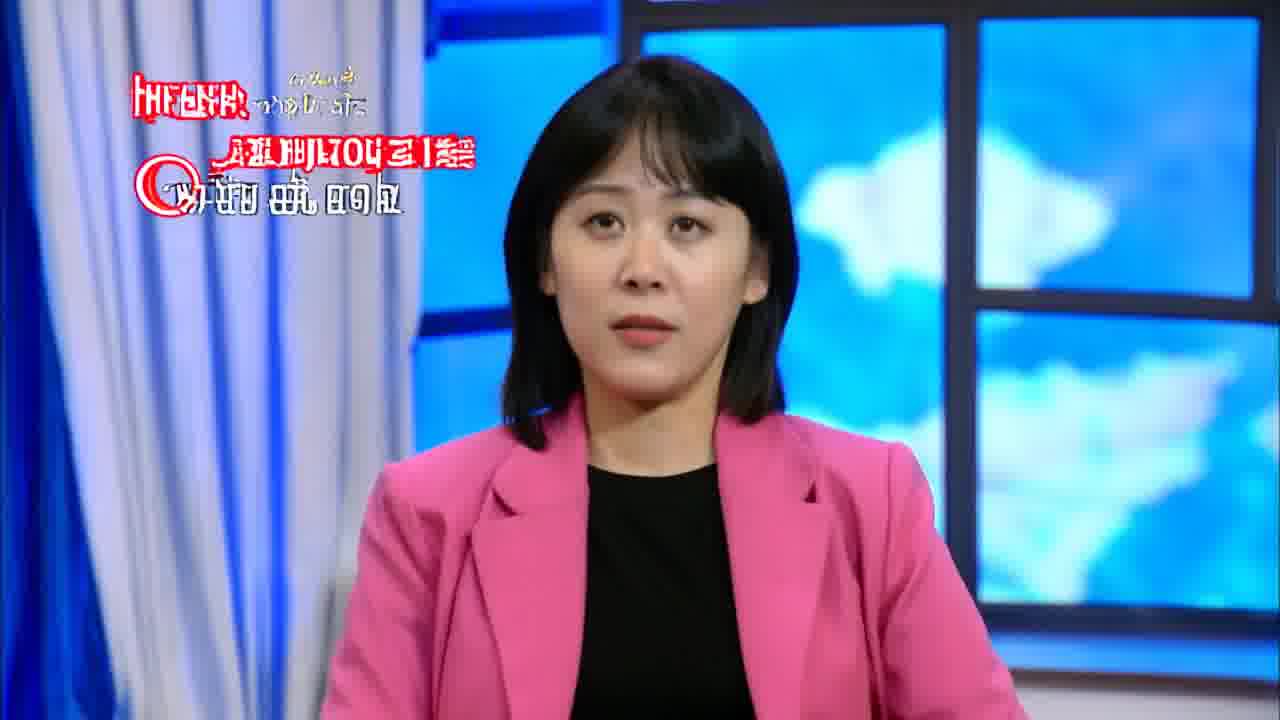} &
        \includegraphics[width=0.17\textwidth]{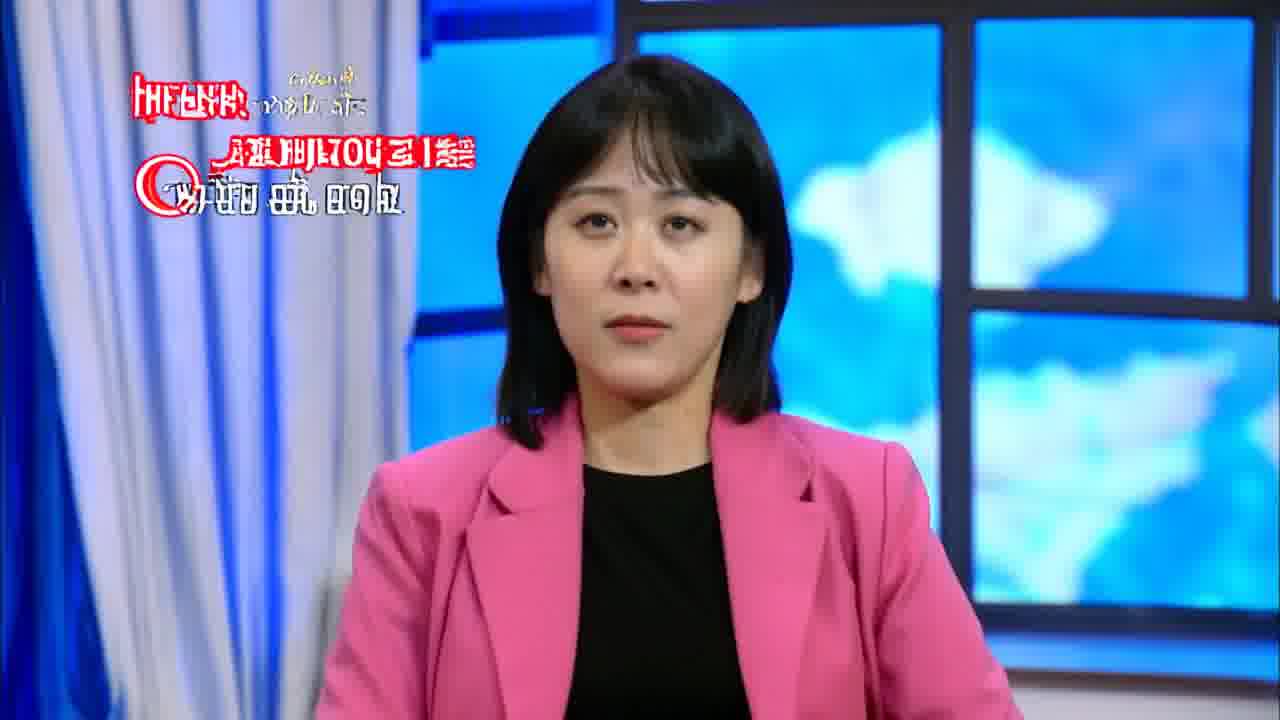} \\
        
        \rotatebox{90}{\tiny \textbf{\textsc{LTX-2.3}}} &
        \includegraphics[width=0.17\textwidth]{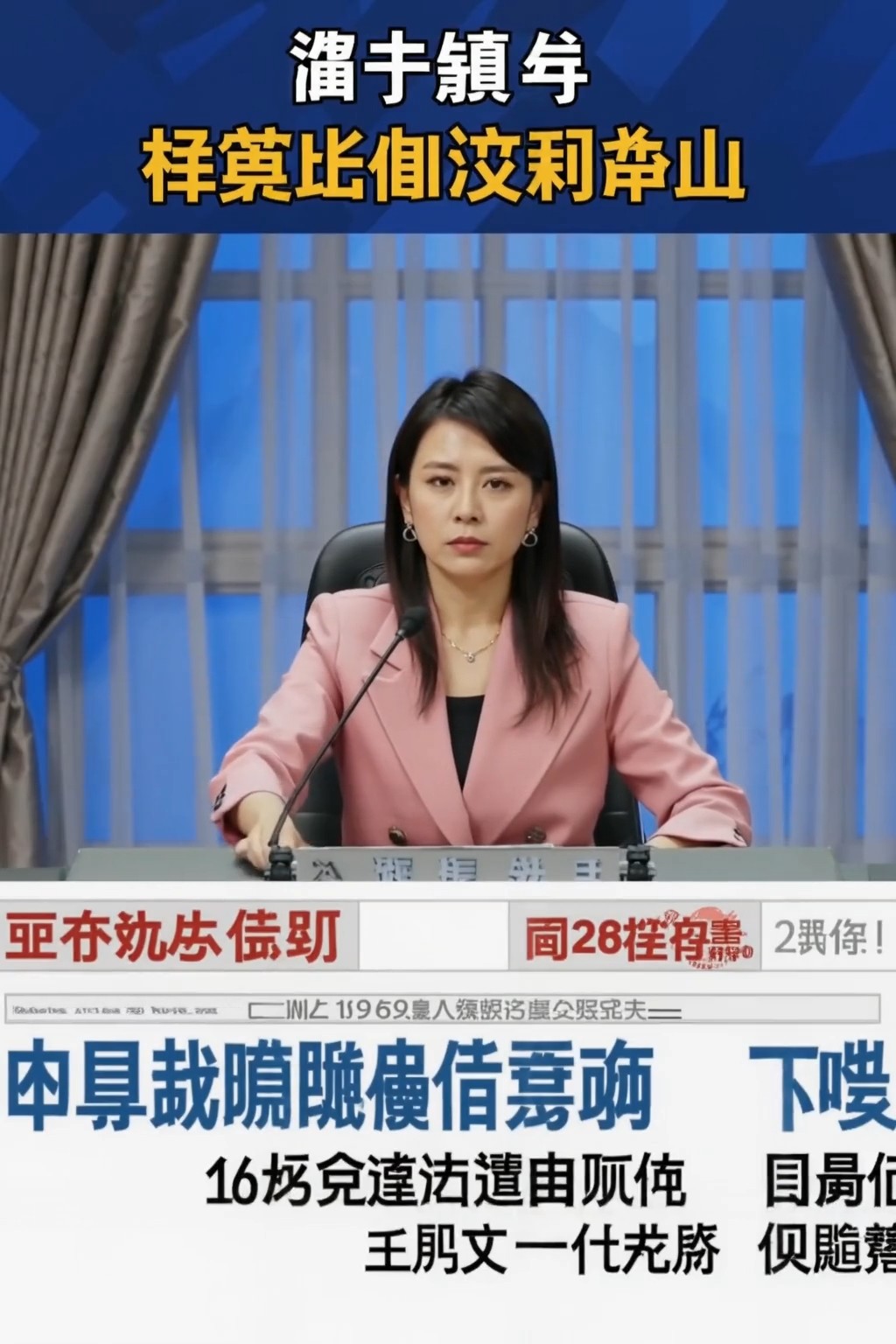} &
        \includegraphics[width=0.17\textwidth]{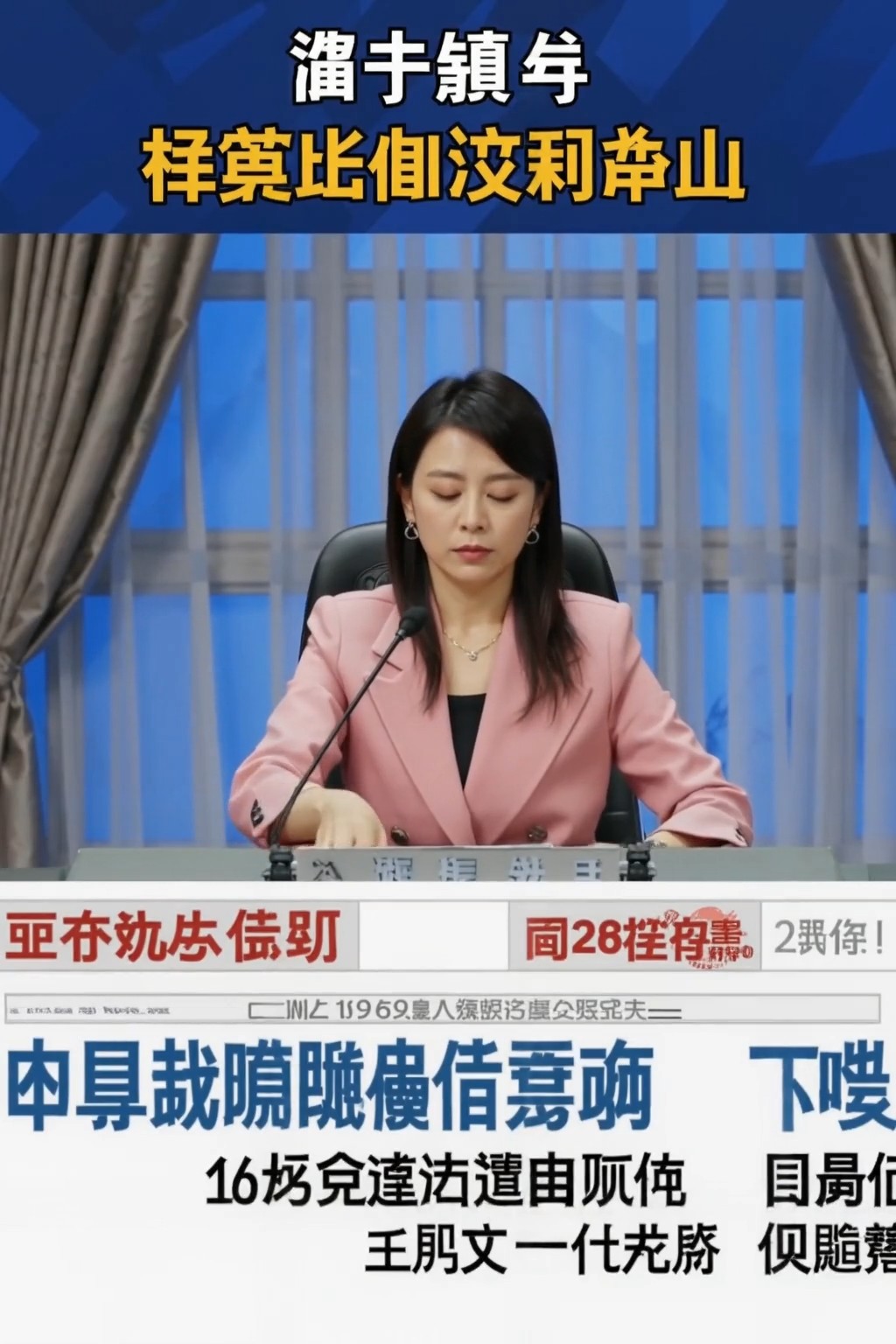} &
        \includegraphics[width=0.17\textwidth]{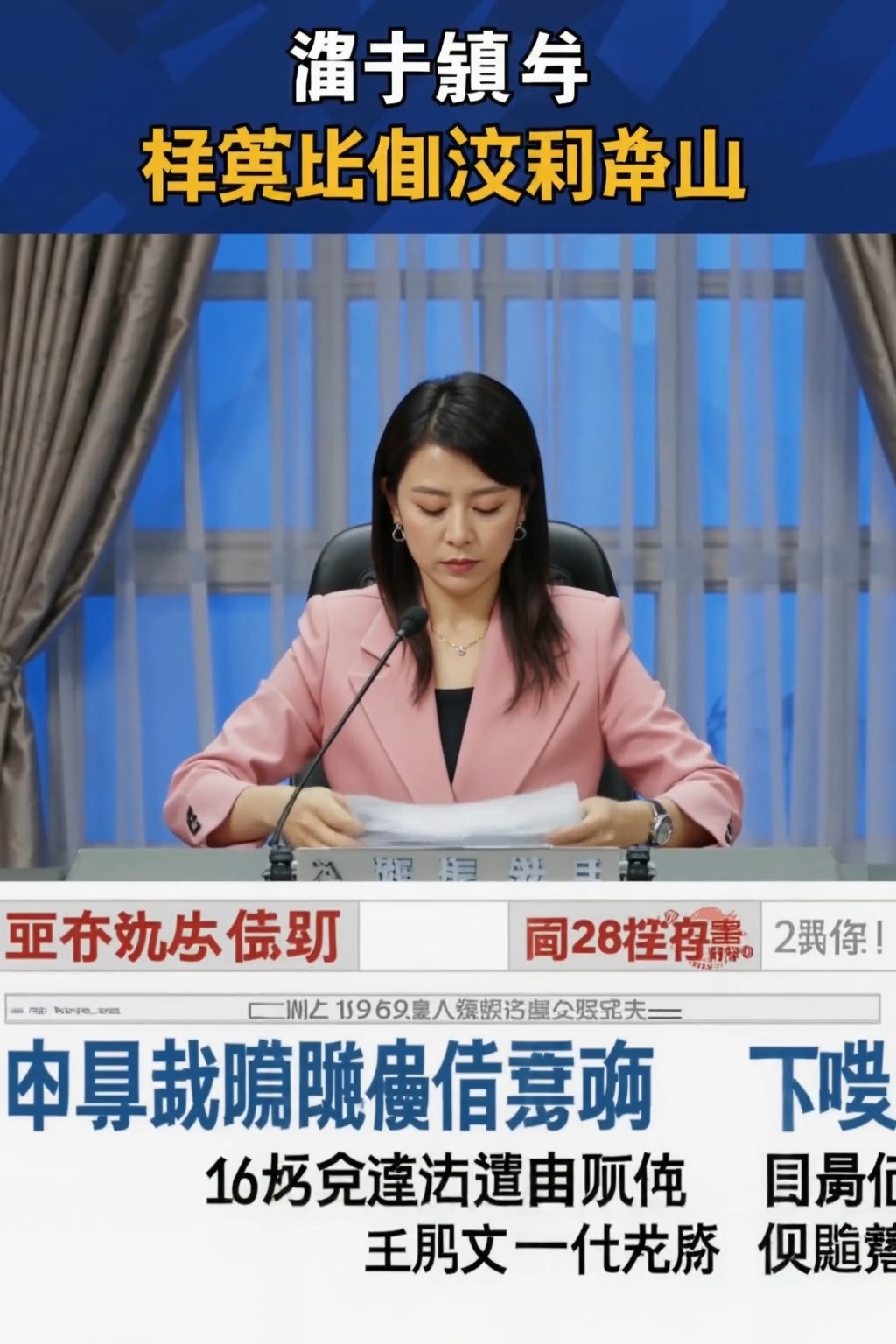} &
        \includegraphics[width=0.17\textwidth]{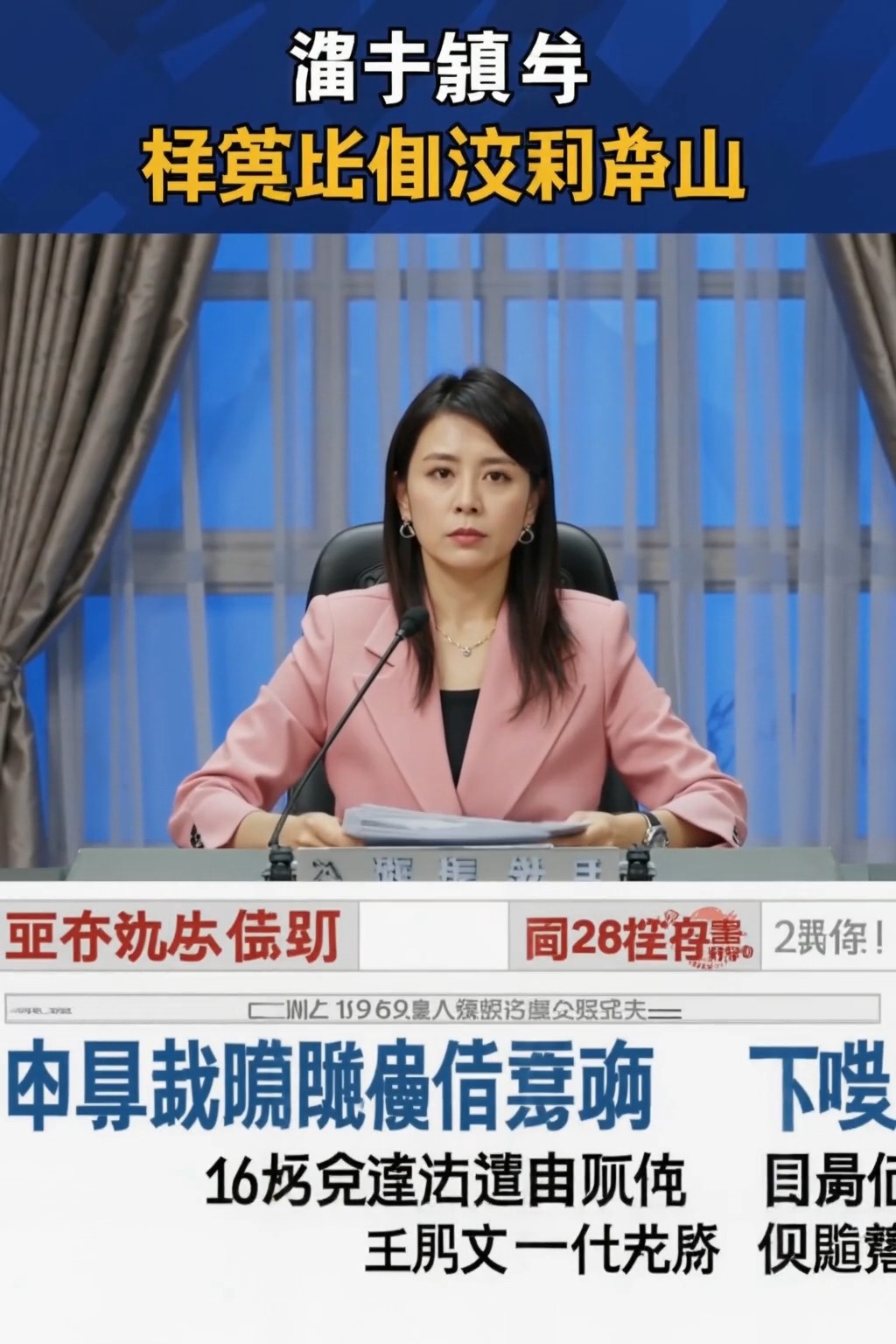} &
        \includegraphics[width=0.17\textwidth]{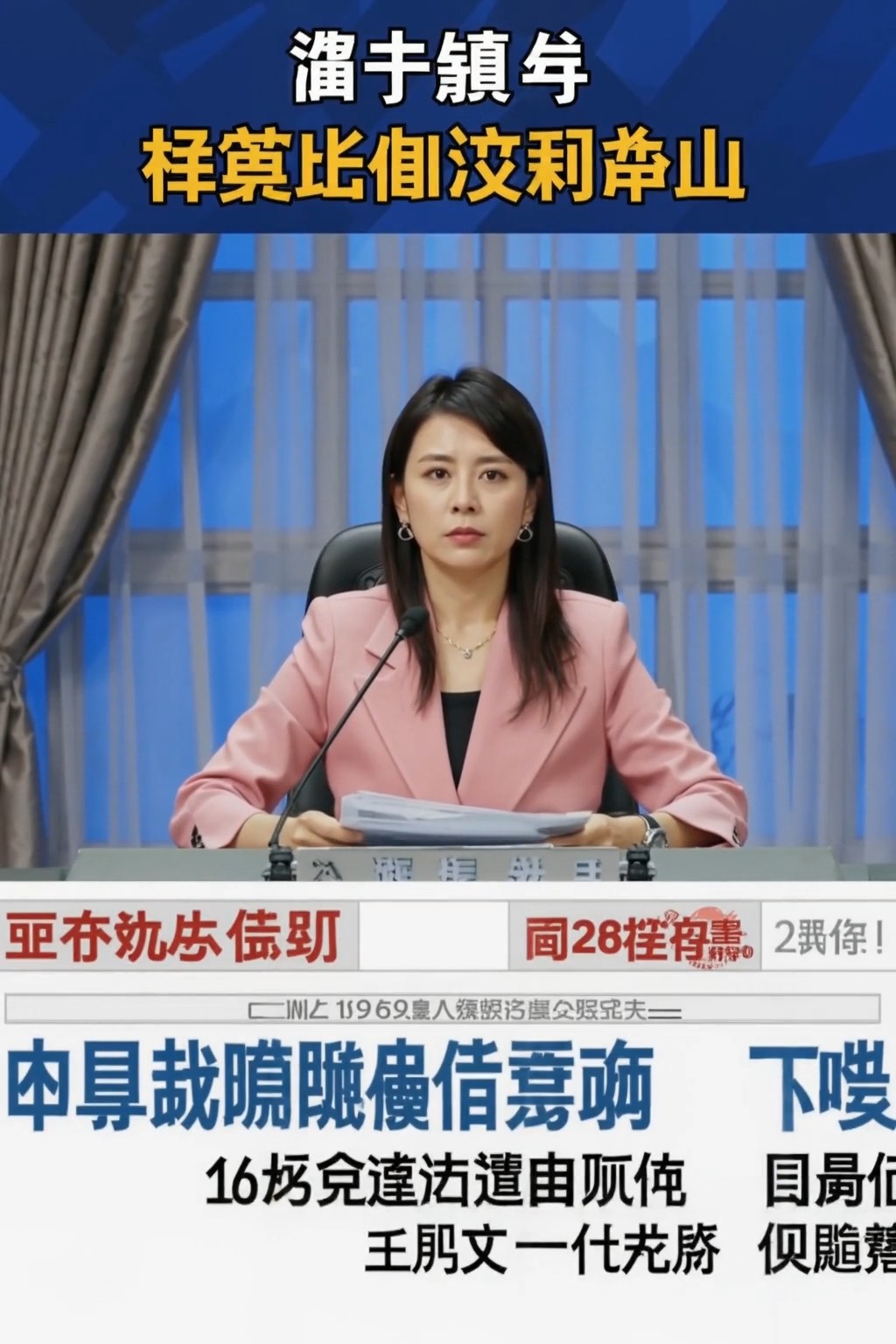} \\
        
        \rotatebox{90}{\tiny \textbf{\textsc{Helios}}} &
        \includegraphics[width=0.17\textwidth]{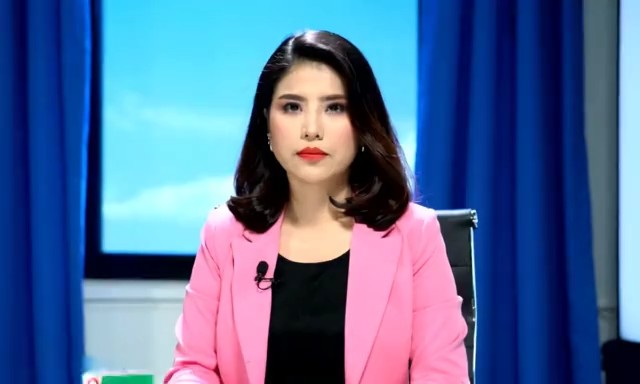} &
        \includegraphics[width=0.17\textwidth]{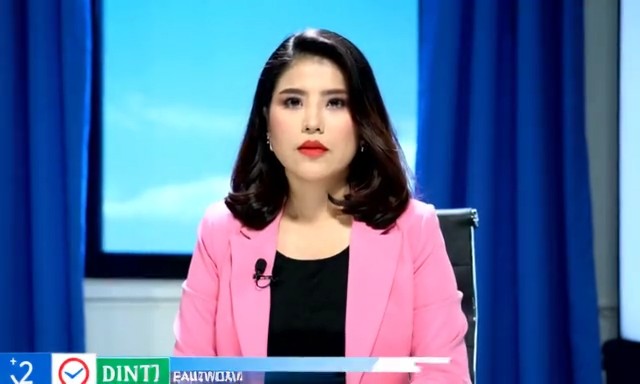} &
        \includegraphics[width=0.17\textwidth]{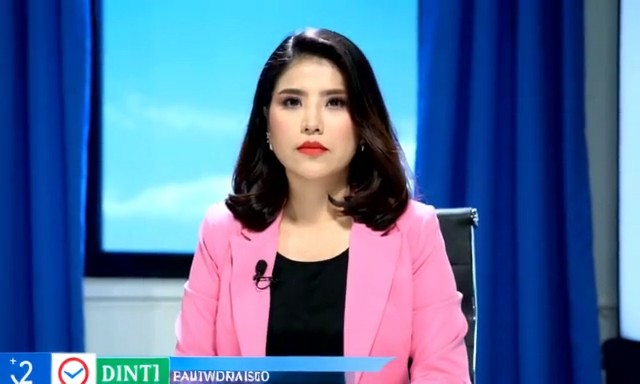} &
        \includegraphics[width=0.17\textwidth]{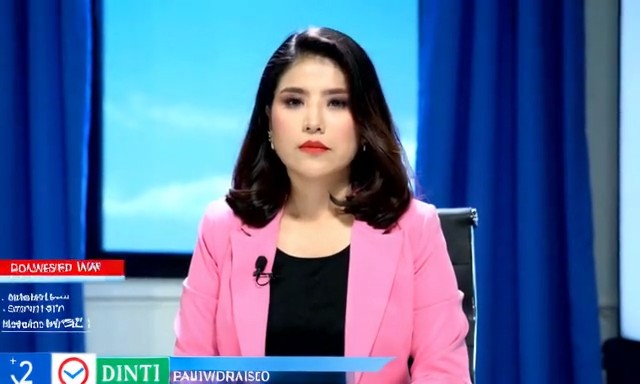} &
        \includegraphics[width=0.17\textwidth]{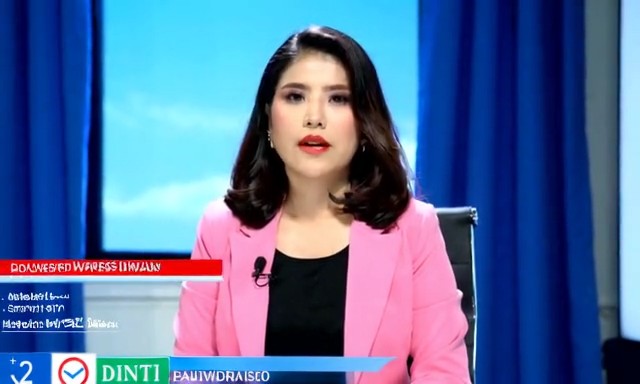} \\
        
        \rotatebox{90}{\tiny \textbf{\textsc{daVinci}}} &
        \includegraphics[width=0.17\textwidth]{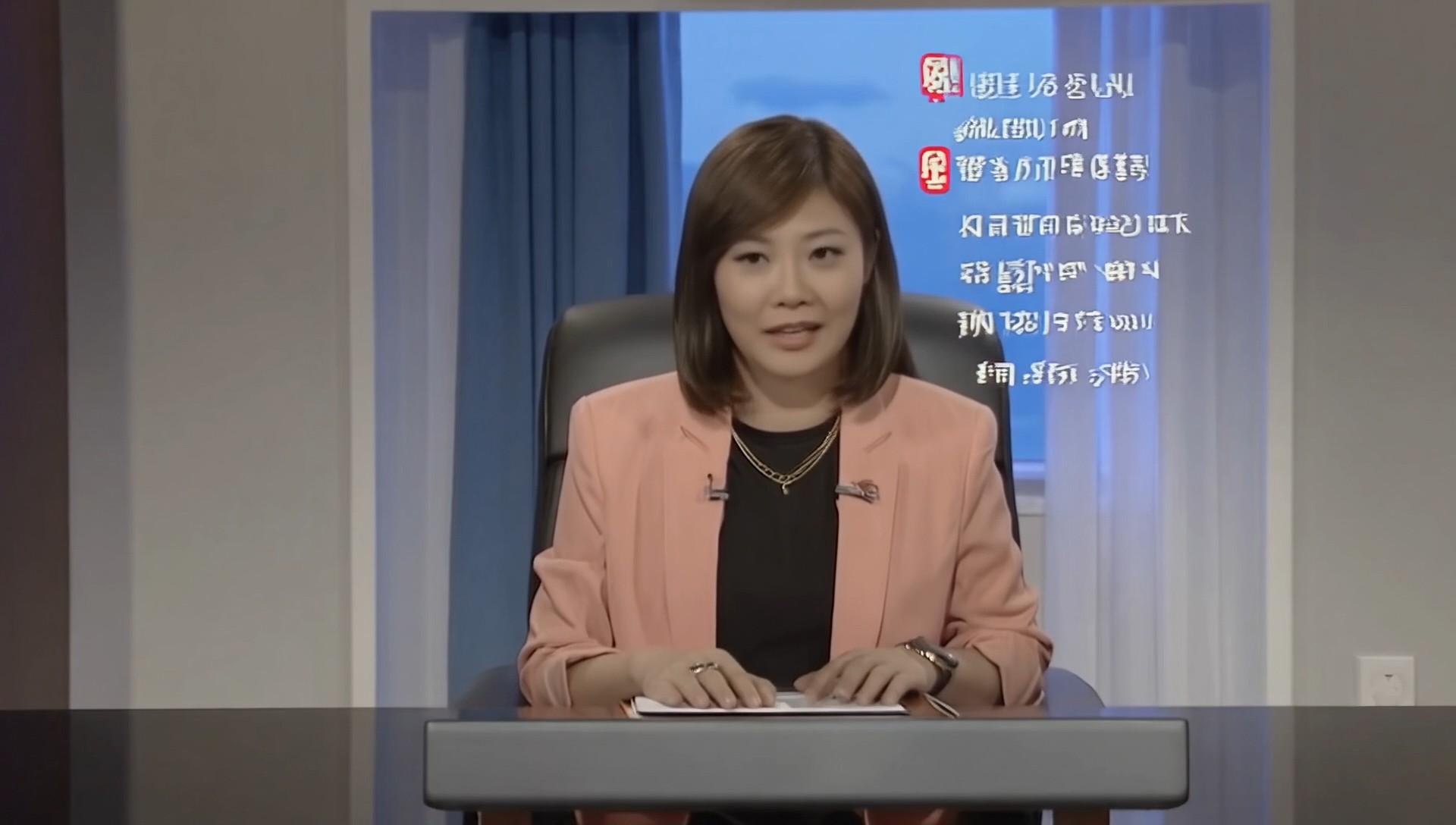} &
        \includegraphics[width=0.17\textwidth]{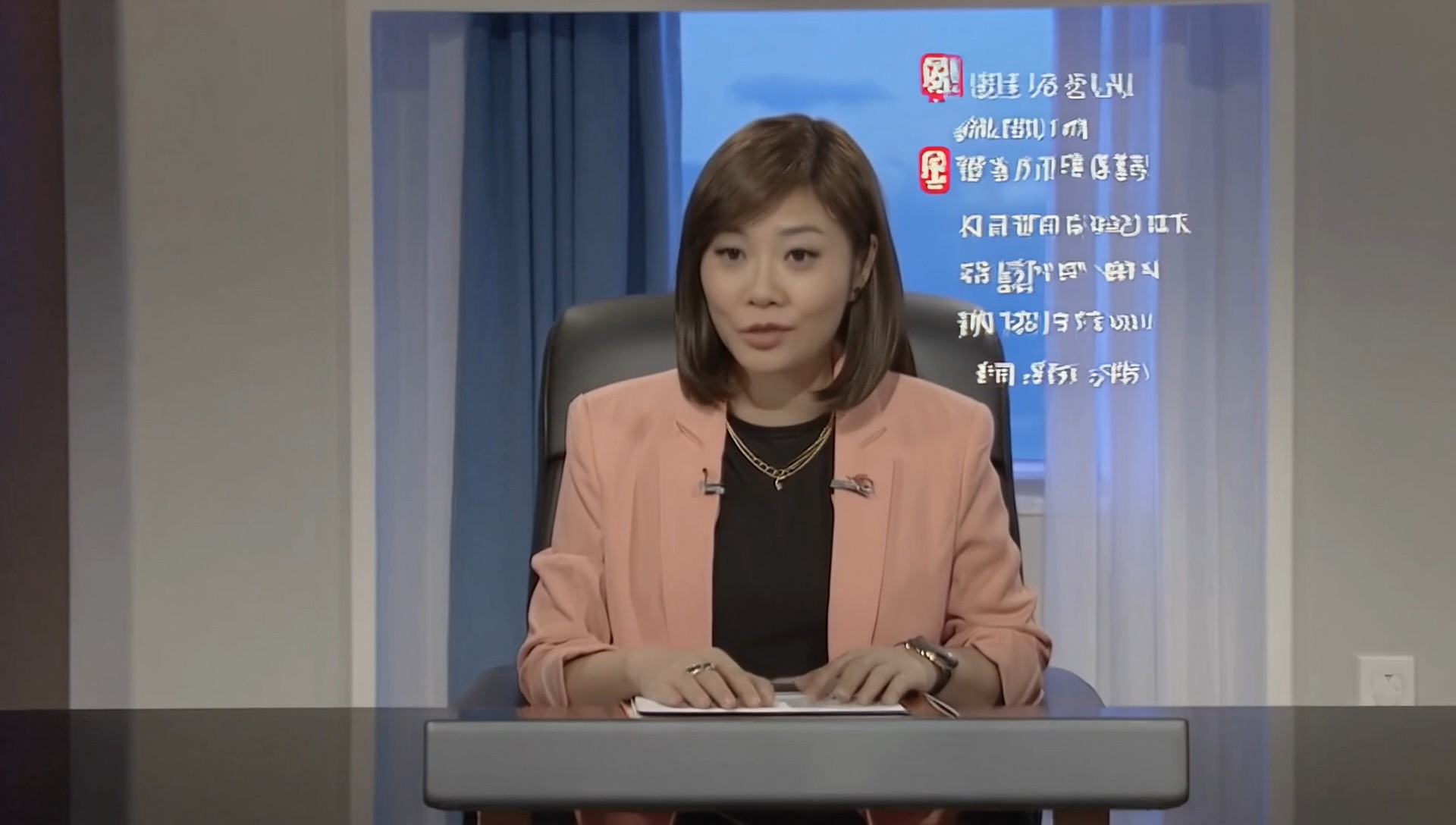} &
        \includegraphics[width=0.17\textwidth]{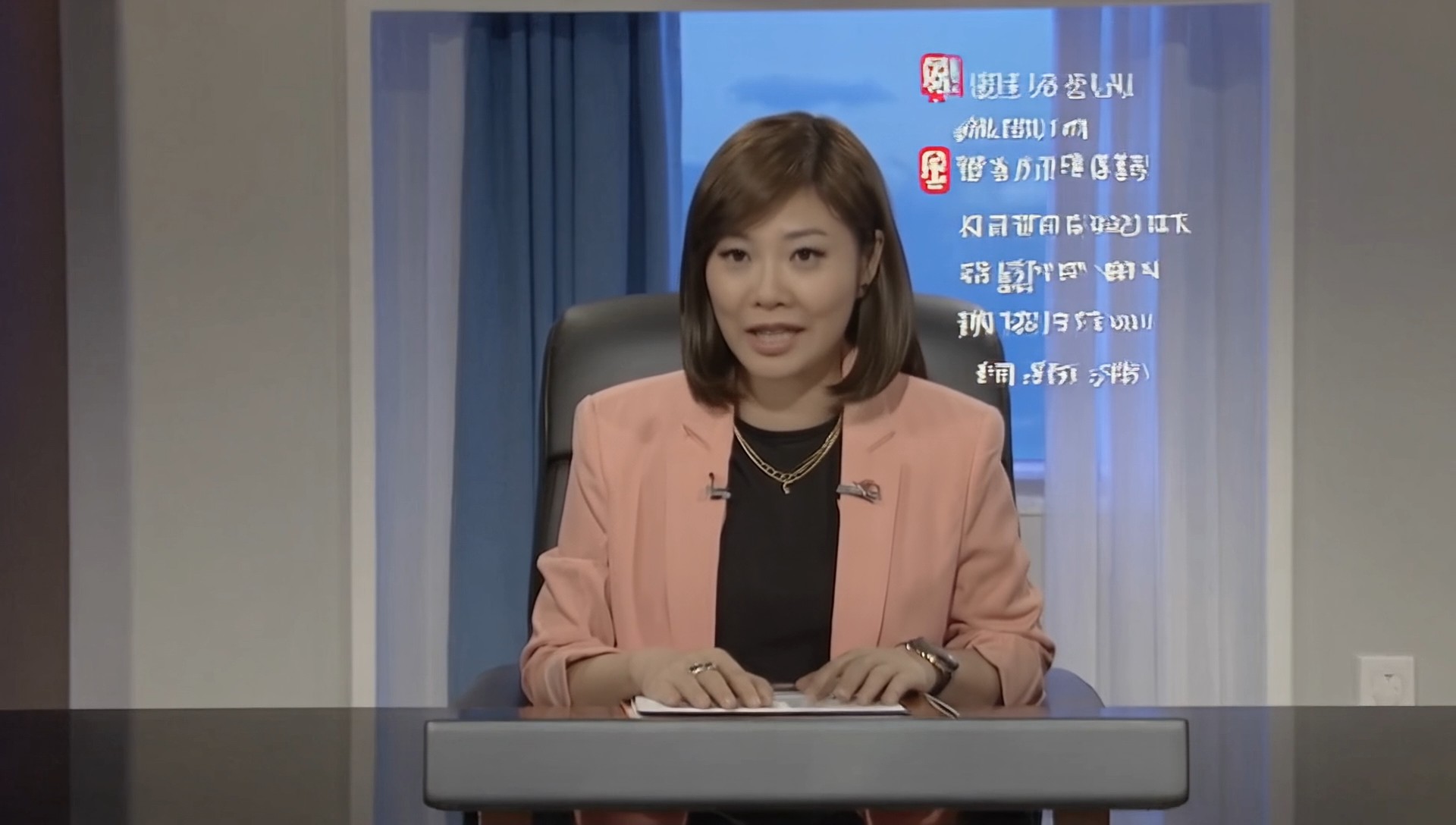} &
        \includegraphics[width=0.17\textwidth]{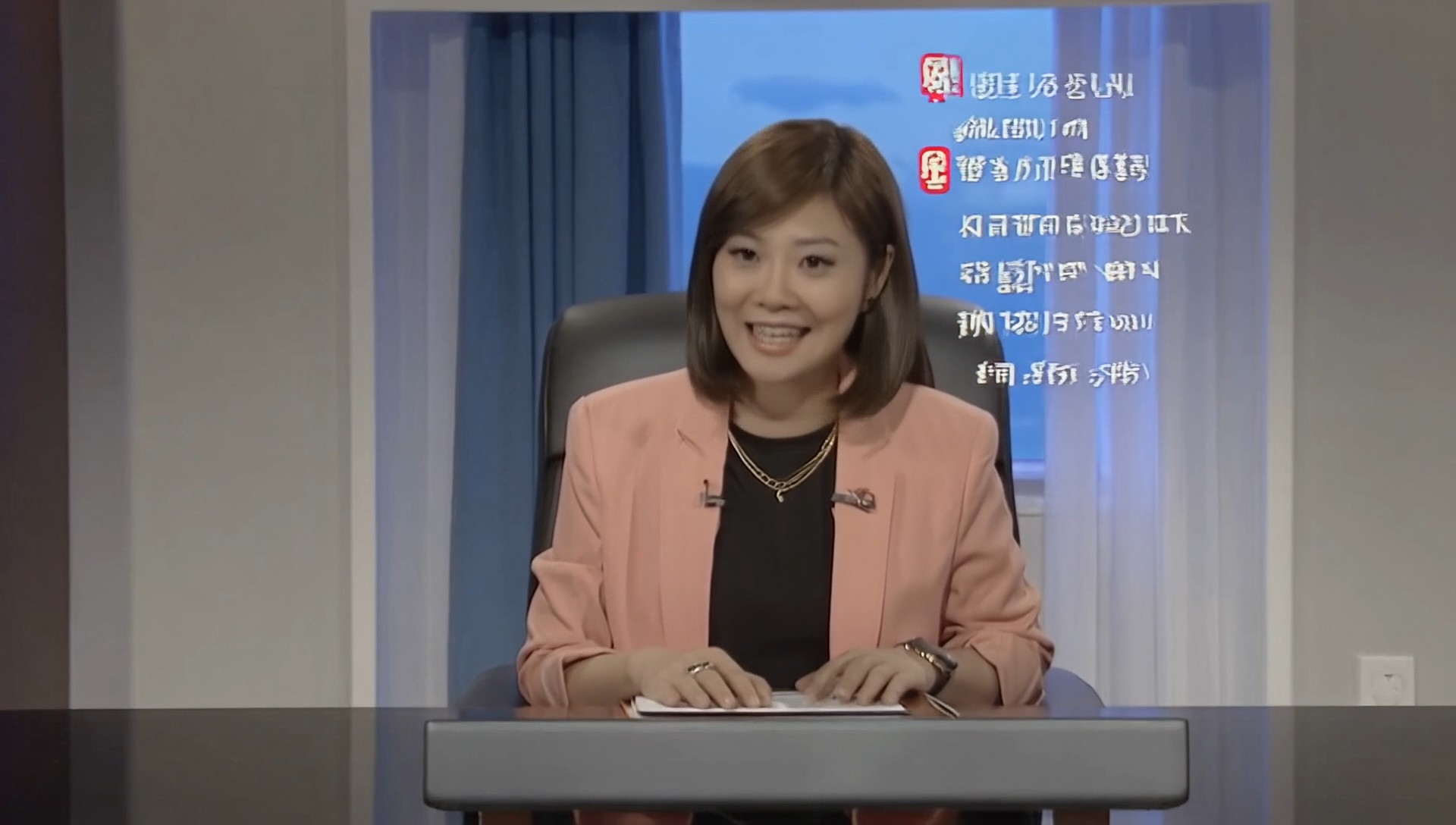} &
        \includegraphics[width=0.17\textwidth]{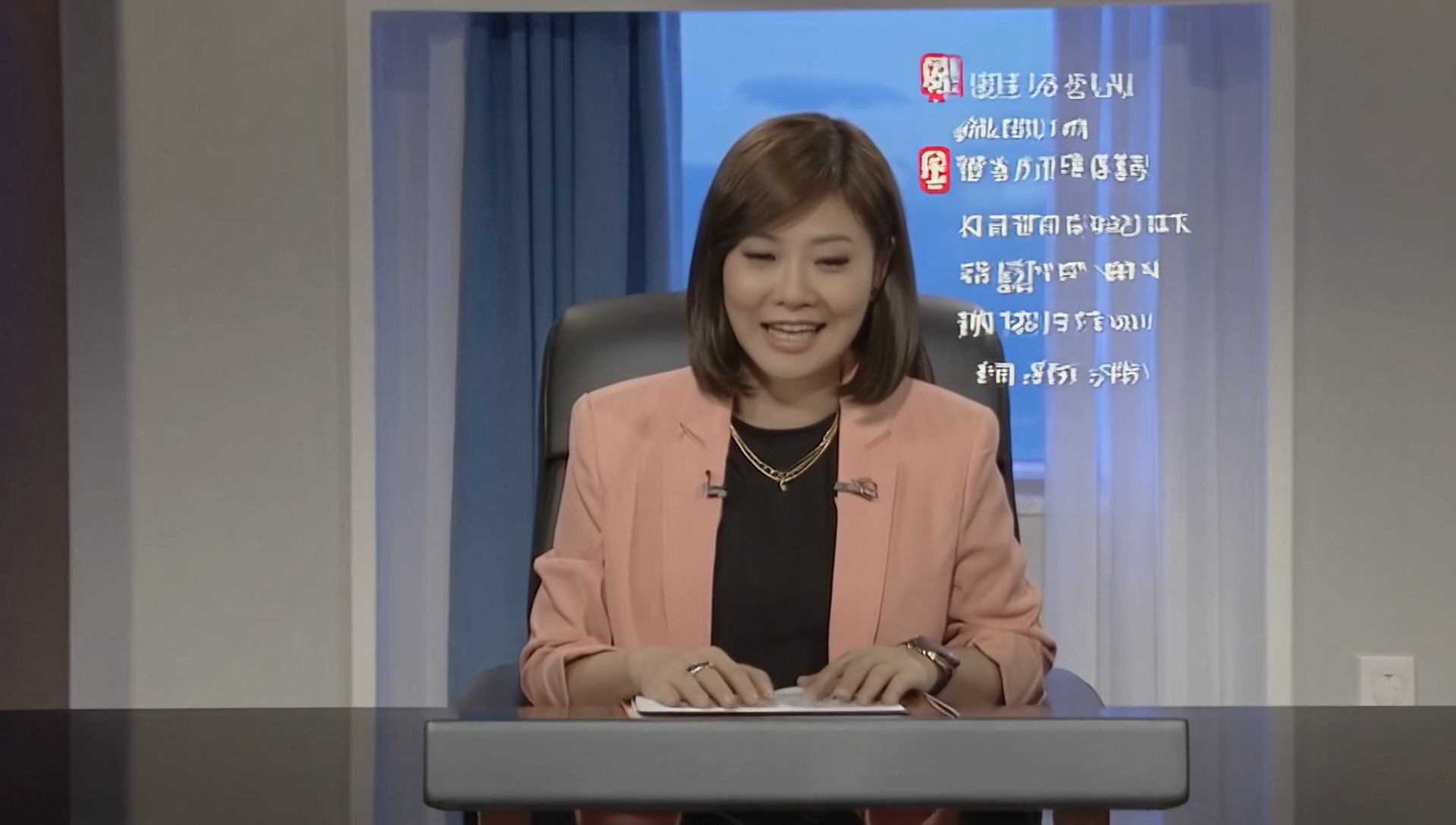} \\
        
        \rotatebox{90}{\tiny \textbf{\textsc{Self-Forcing}}} &
        \includegraphics[width=0.17\textwidth]{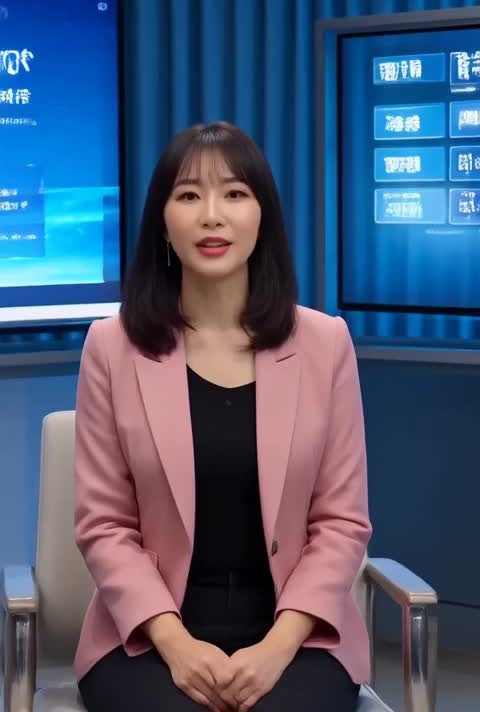} &
        \includegraphics[width=0.17\textwidth]{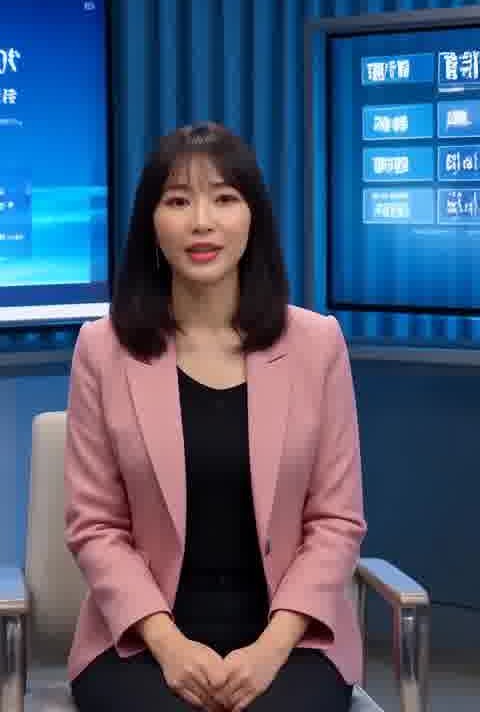} &
        \includegraphics[width=0.17\textwidth]{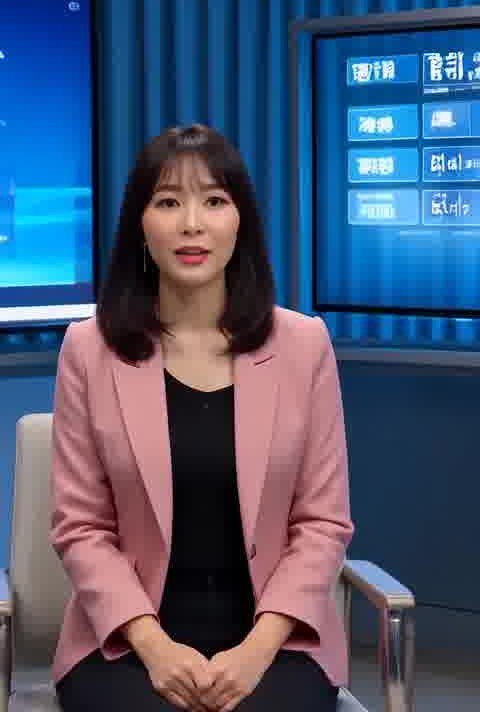} &
        \includegraphics[width=0.17\textwidth]{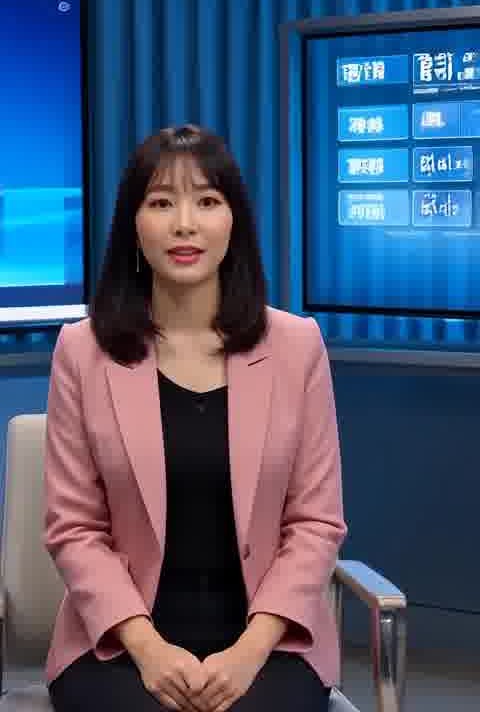} &
        \includegraphics[width=0.17\textwidth]{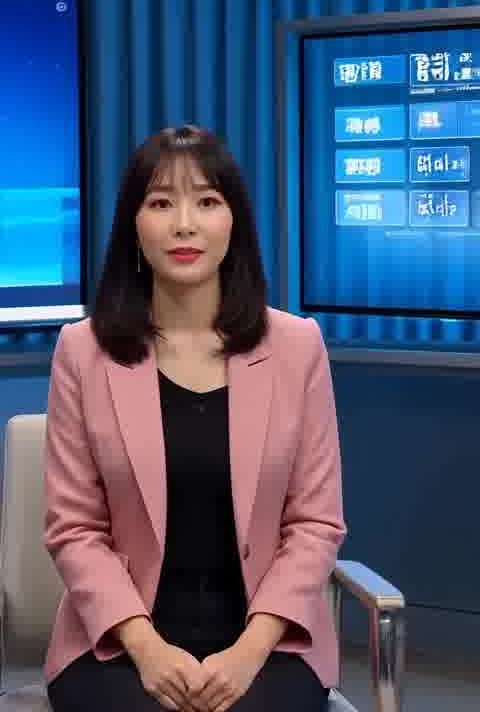} \\
    \end{tabular}
    \caption{Qualitative T2V comparison on FF++ pristine video 271 (news anchor). Frame sequences ($t=0$ to $t=4$) comparing the source video (top row) with synthetic outputs from the eight T2V generators. Despite differing architectural priors, all models correctly interpret the prompt (e.g., pink blazer, newsroom setting) while maintaining high identity consistency and stable background rendering over time.}
    \label{fig:qualitative_271_t2v_v2}
\end{figure}

\begin{figure}[!p]
    \centering
    \setlength{\tabcolsep}{1pt}
    \renewcommand{\arraystretch}{0.5}
    \begin{tabular}{c ccccc}
        & $t=0$ & $t=1$ & $t=2$ & $t=3$ & $t=4$ \\
        \rotatebox{90}{\tiny \textbf{\textsc{Pristine}}} &
        \includegraphics[width=0.14\textwidth]{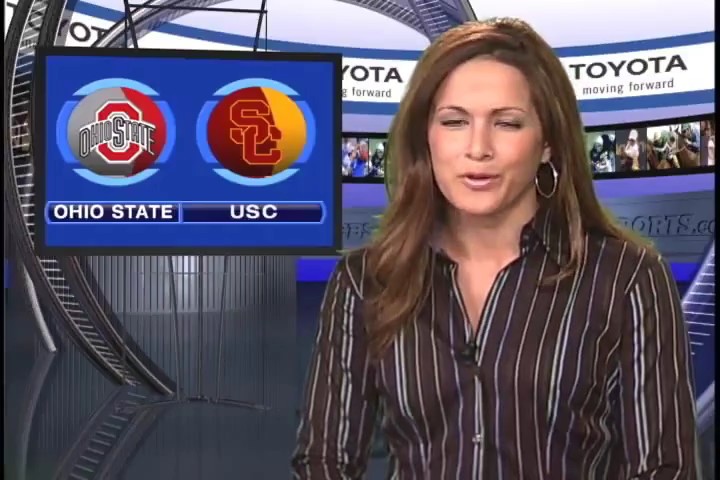} &
        \includegraphics[width=0.14\textwidth]{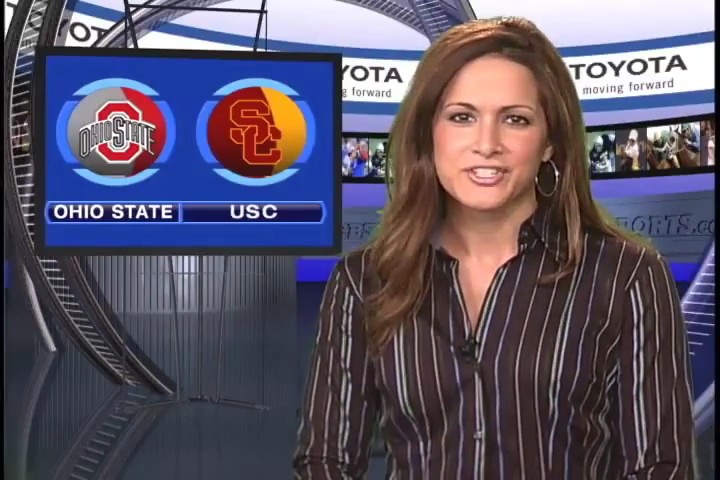} &
        \includegraphics[width=0.14\textwidth]{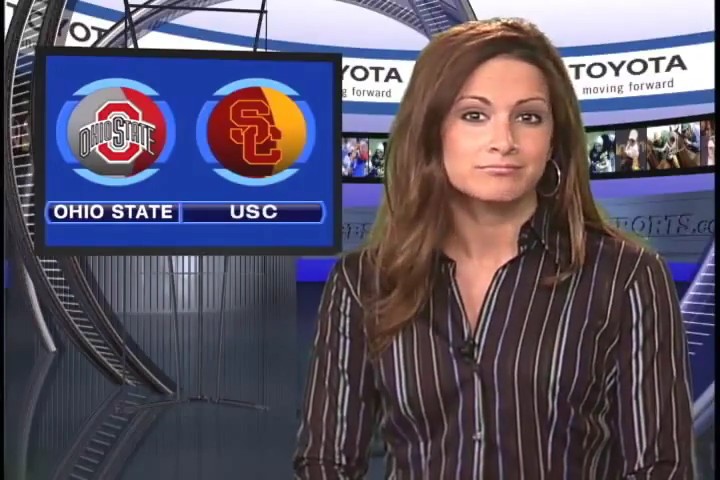} &
        \includegraphics[width=0.14\textwidth]{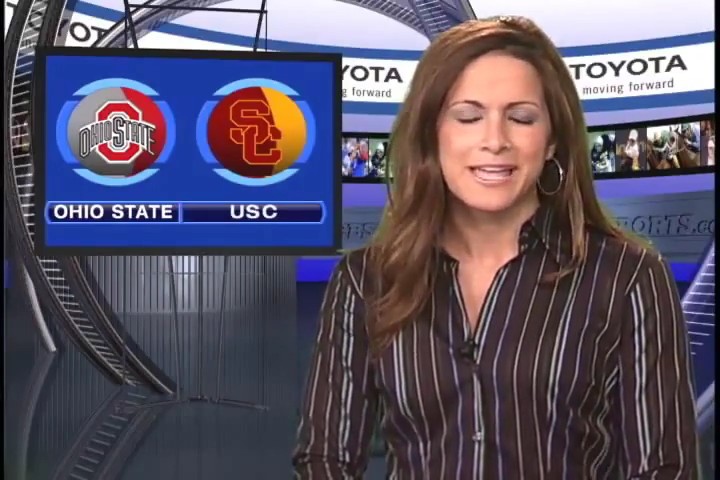} &
        \includegraphics[width=0.14\textwidth]{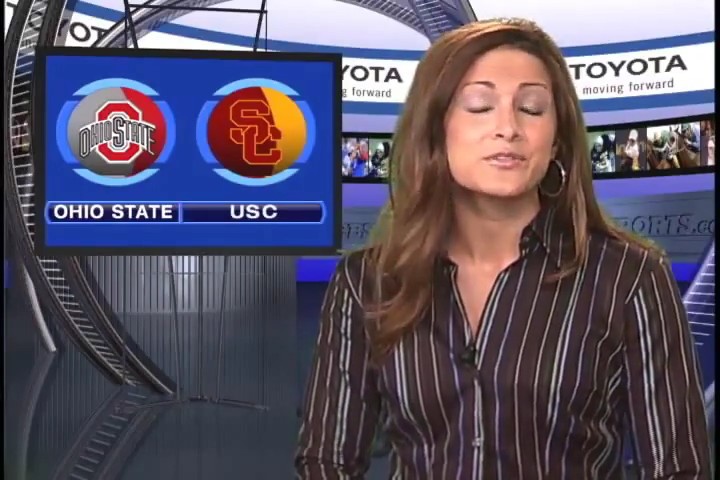} \\
        
        \rotatebox{90}{\tiny \textbf{\textsc{Wan2.1}}} &
        \includegraphics[width=0.14\textwidth]{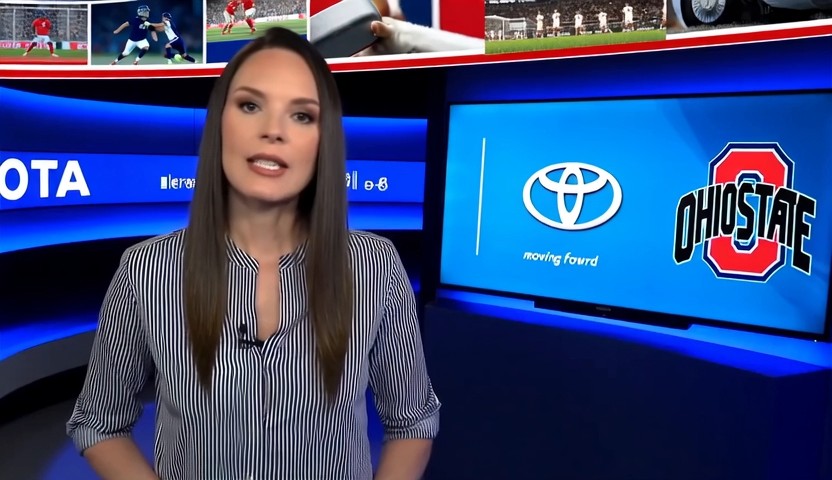} &
        \includegraphics[width=0.14\textwidth]{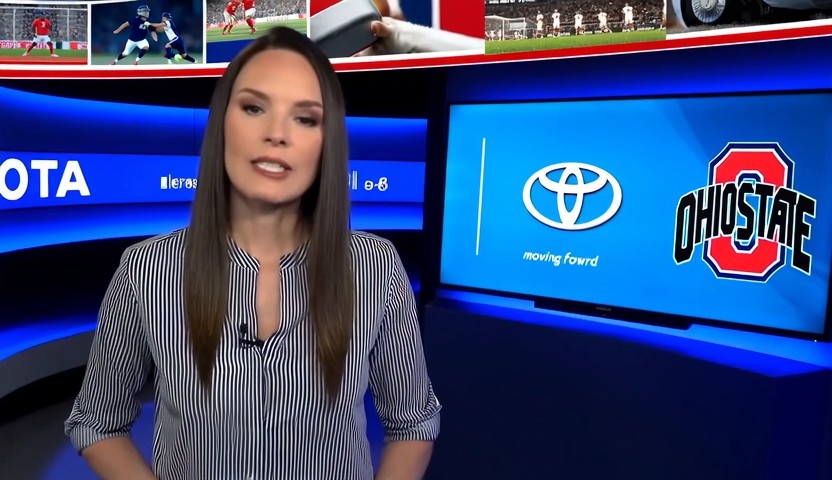} &
        \includegraphics[width=0.14\textwidth]{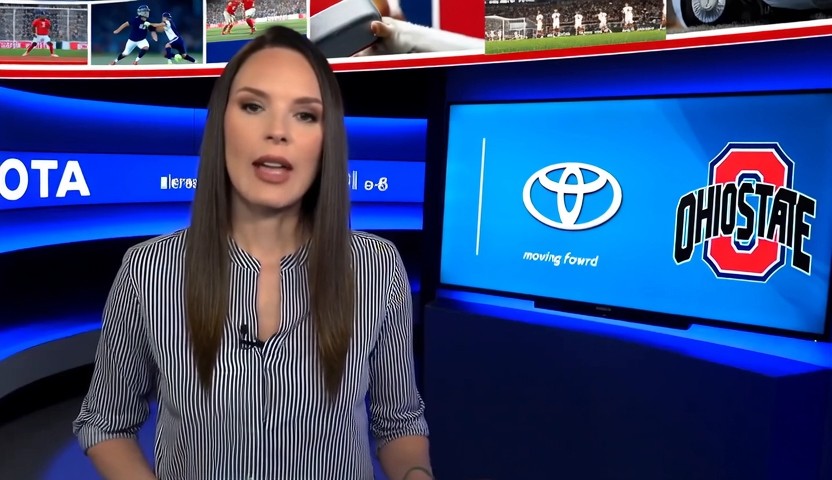} &
        \includegraphics[width=0.14\textwidth]{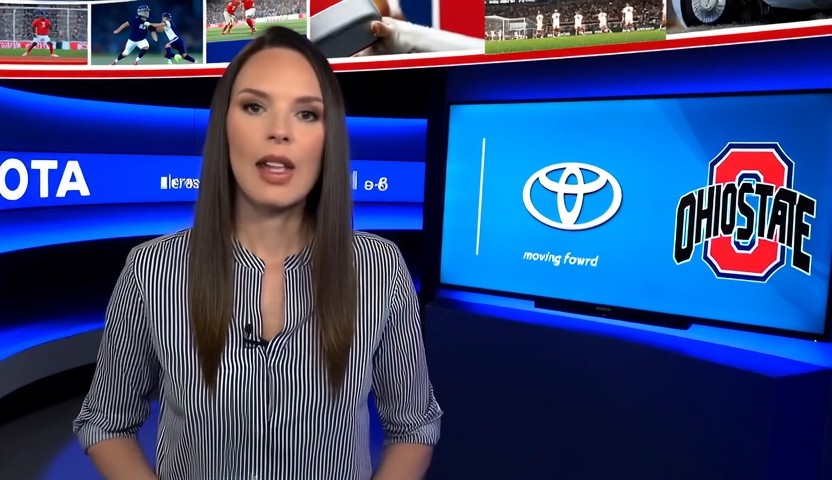} &
        \includegraphics[width=0.14\textwidth]{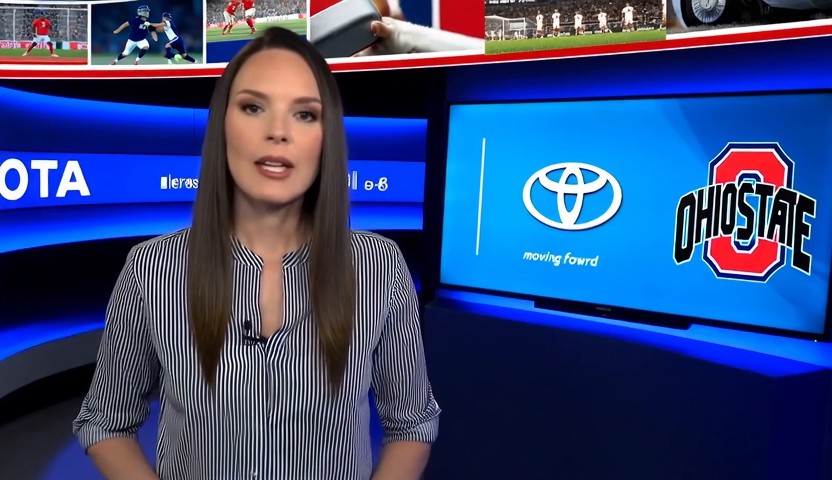} \\
        
        \rotatebox{90}{\tiny \textbf{\textsc{CogVideoX}}} &
        \includegraphics[width=0.14\textwidth]{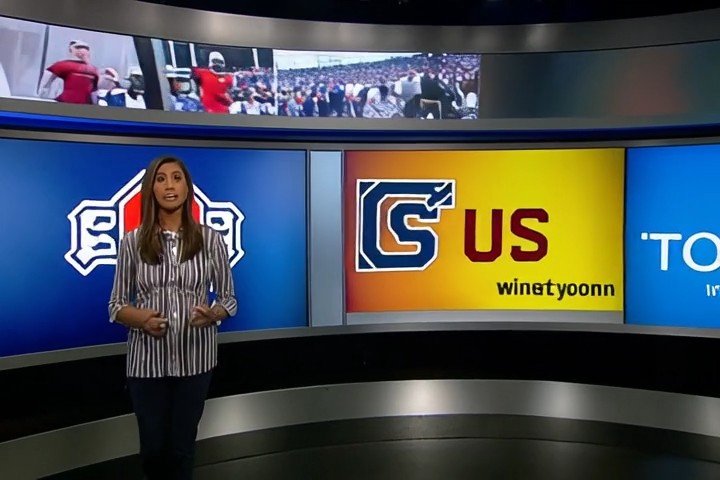} &
        \includegraphics[width=0.14\textwidth]{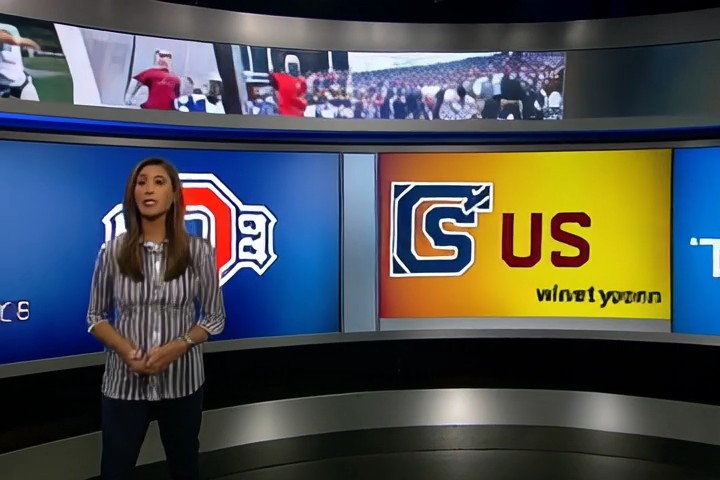} &
        \includegraphics[width=0.14\textwidth]{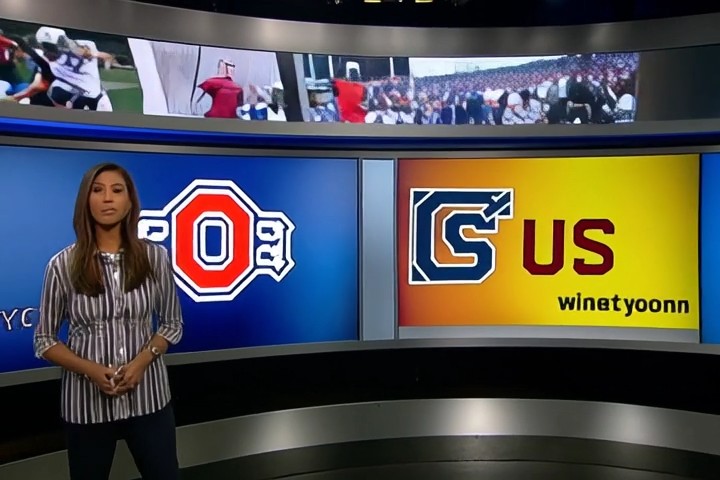} &
        \includegraphics[width=0.14\textwidth]{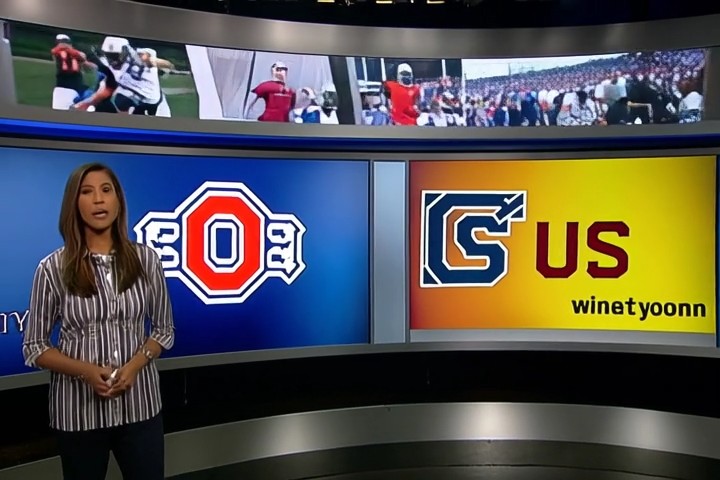} &
        \includegraphics[width=0.14\textwidth]{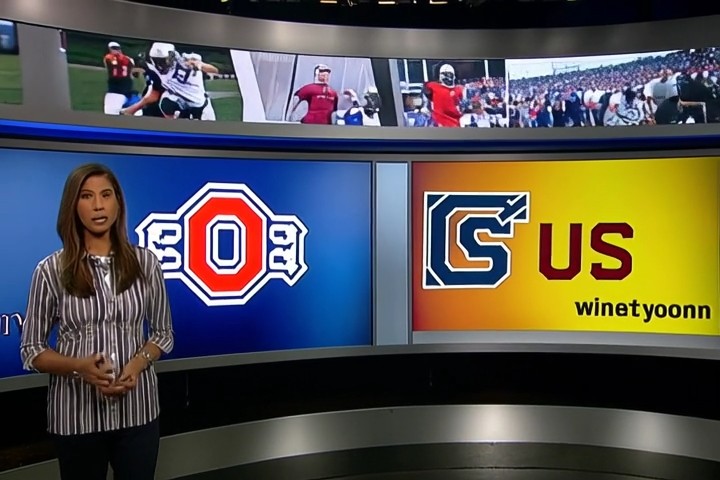} \\
        
        \rotatebox{90}{\tiny \textbf{\textsc{SkyReels-V2}}} &
        \includegraphics[width=0.14\textwidth]{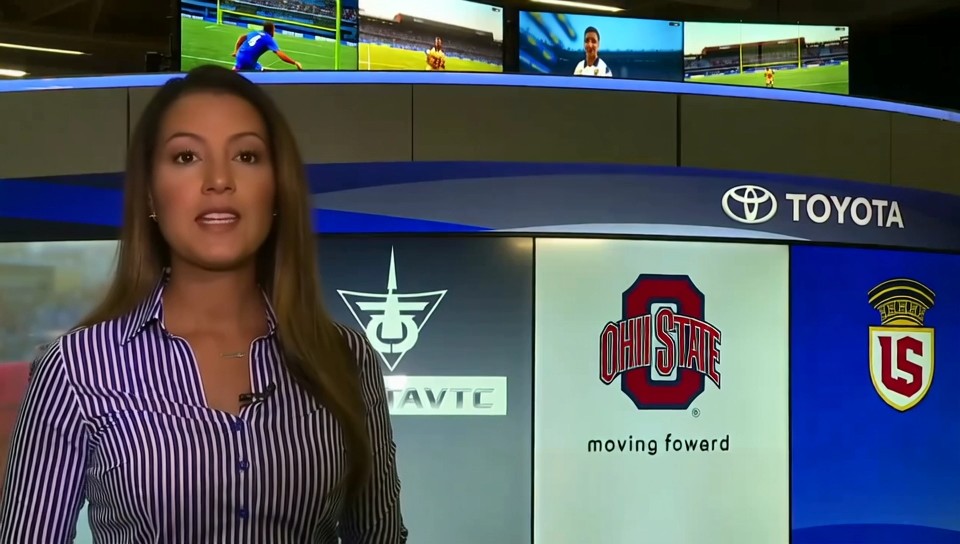} &
        \includegraphics[width=0.14\textwidth]{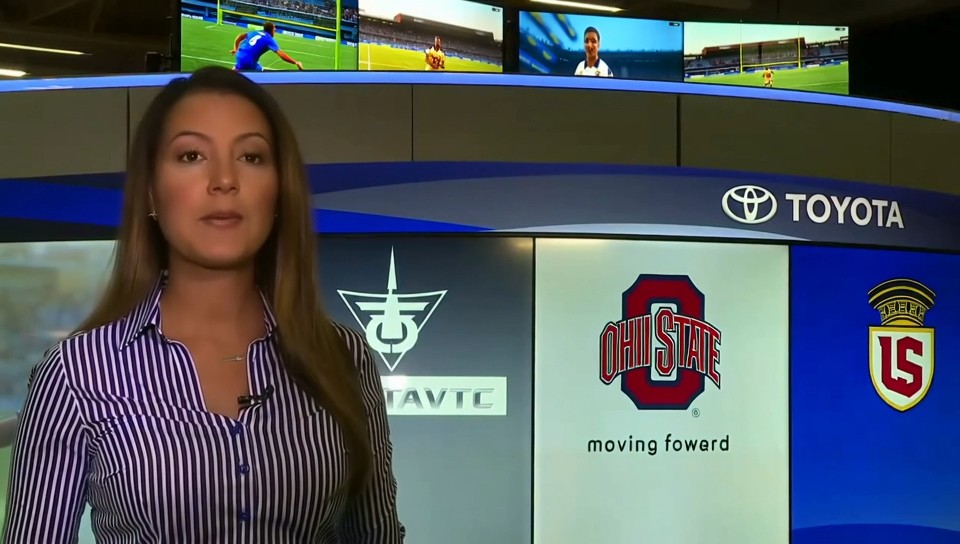} &
        \includegraphics[width=0.14\textwidth]{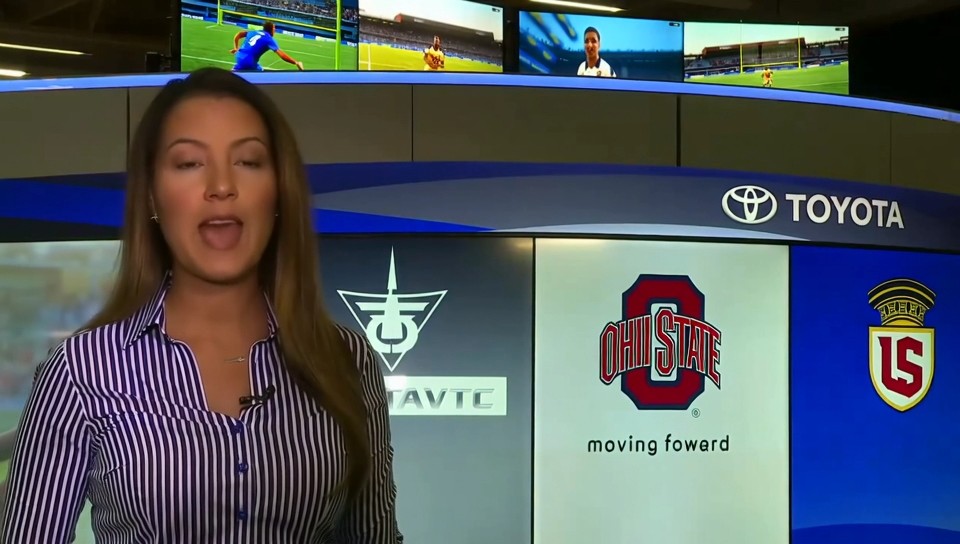} &
        \includegraphics[width=0.14\textwidth]{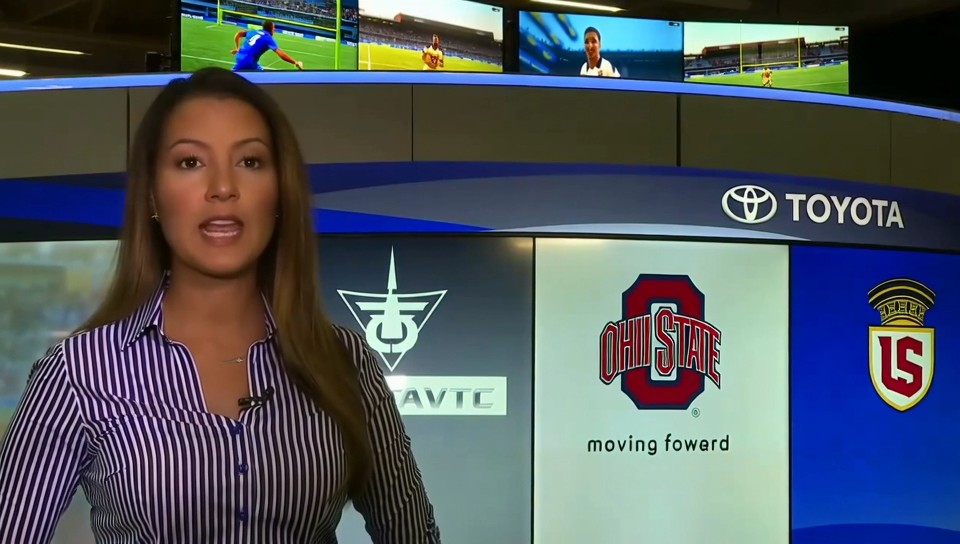} &
        \includegraphics[width=0.14\textwidth]{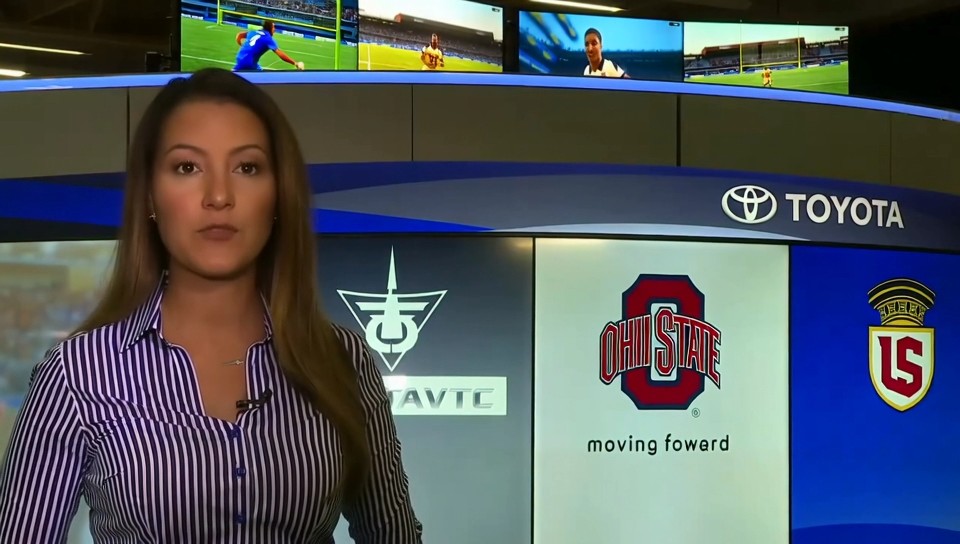} \\
        
        \rotatebox{90}{\tiny \textbf{\textsc{MAGI-1}}} &
        \includegraphics[width=0.14\textwidth]{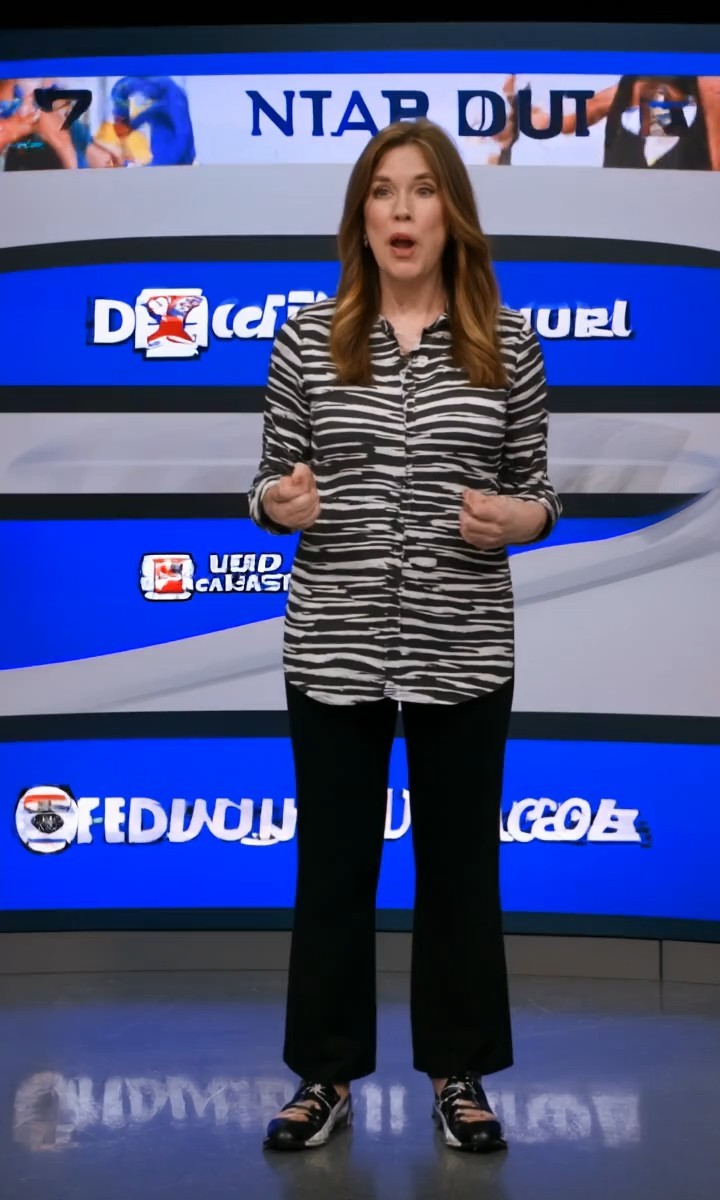} &
        \includegraphics[width=0.14\textwidth]{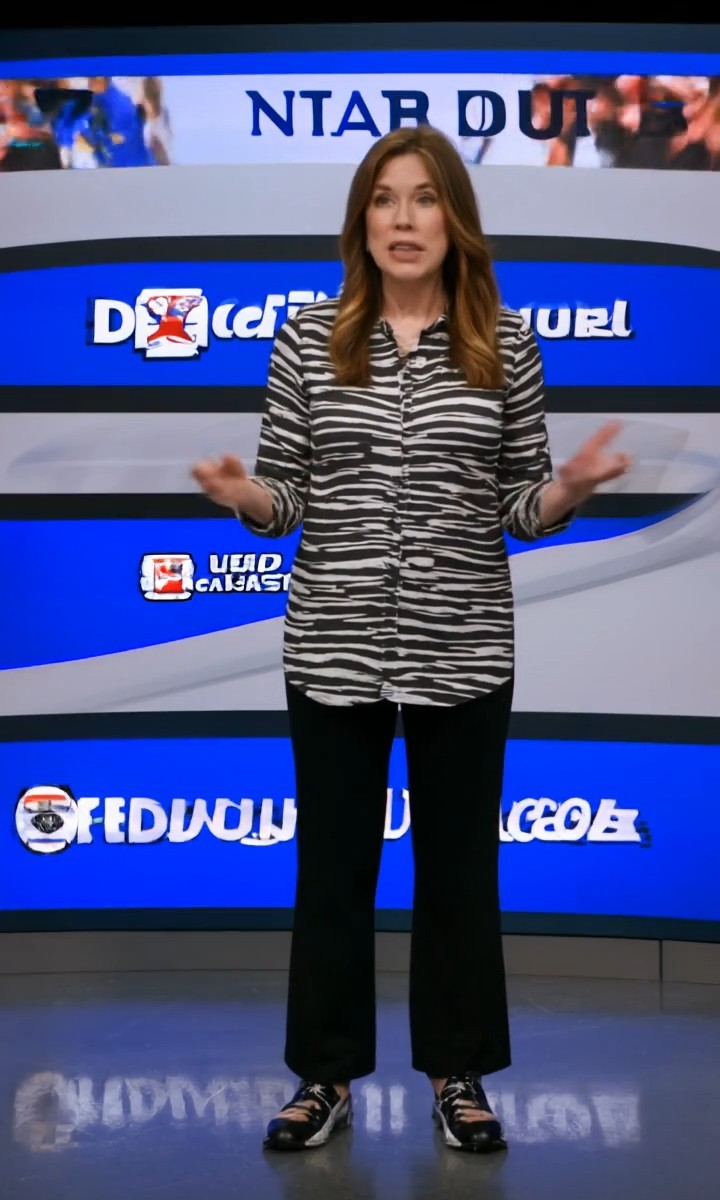} &
        \includegraphics[width=0.14\textwidth]{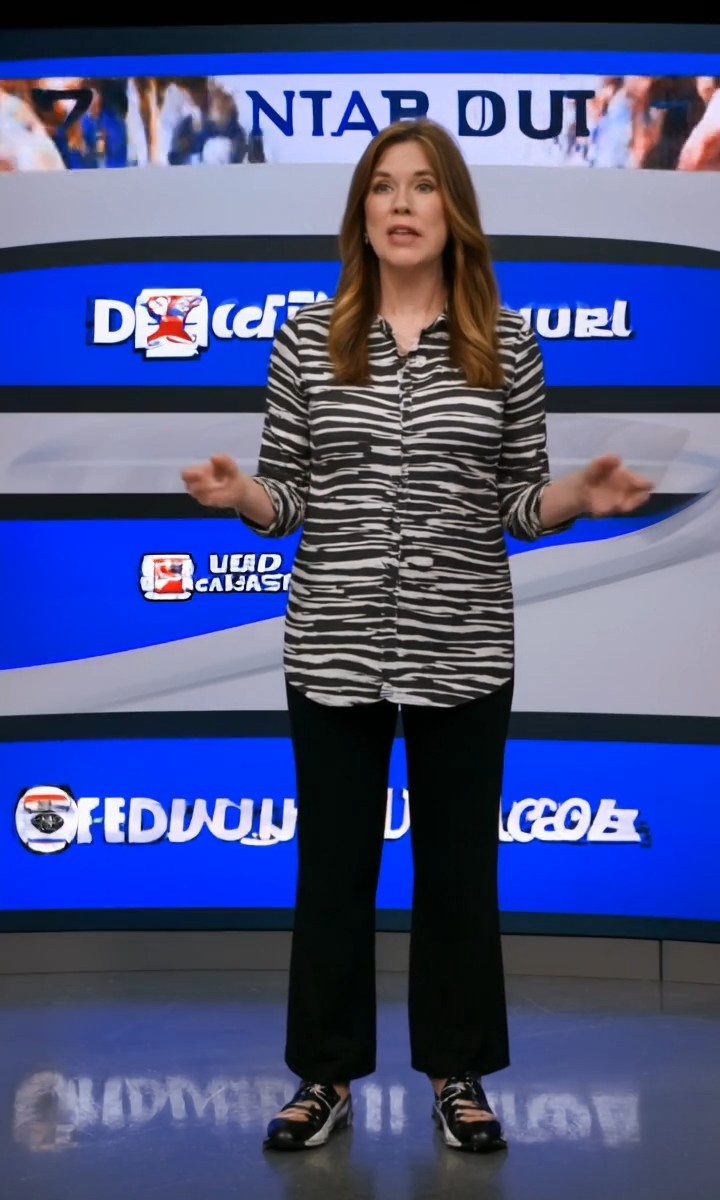} &
        \includegraphics[width=0.14\textwidth]{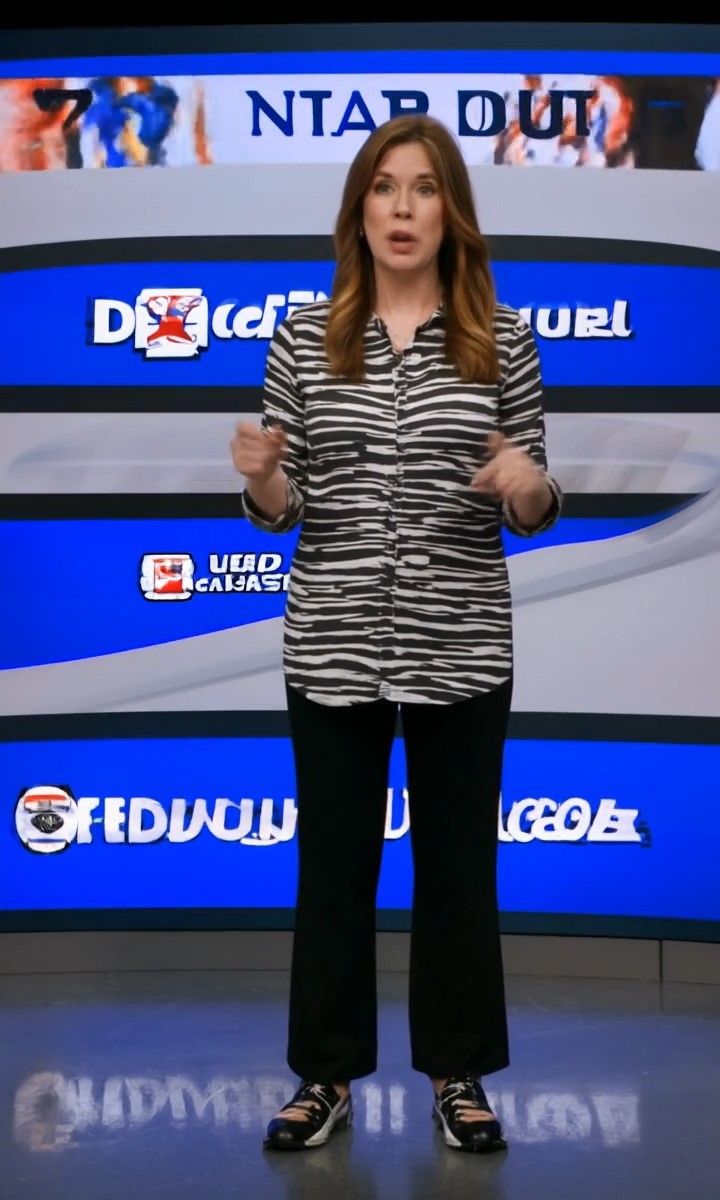} &
        \includegraphics[width=0.14\textwidth]{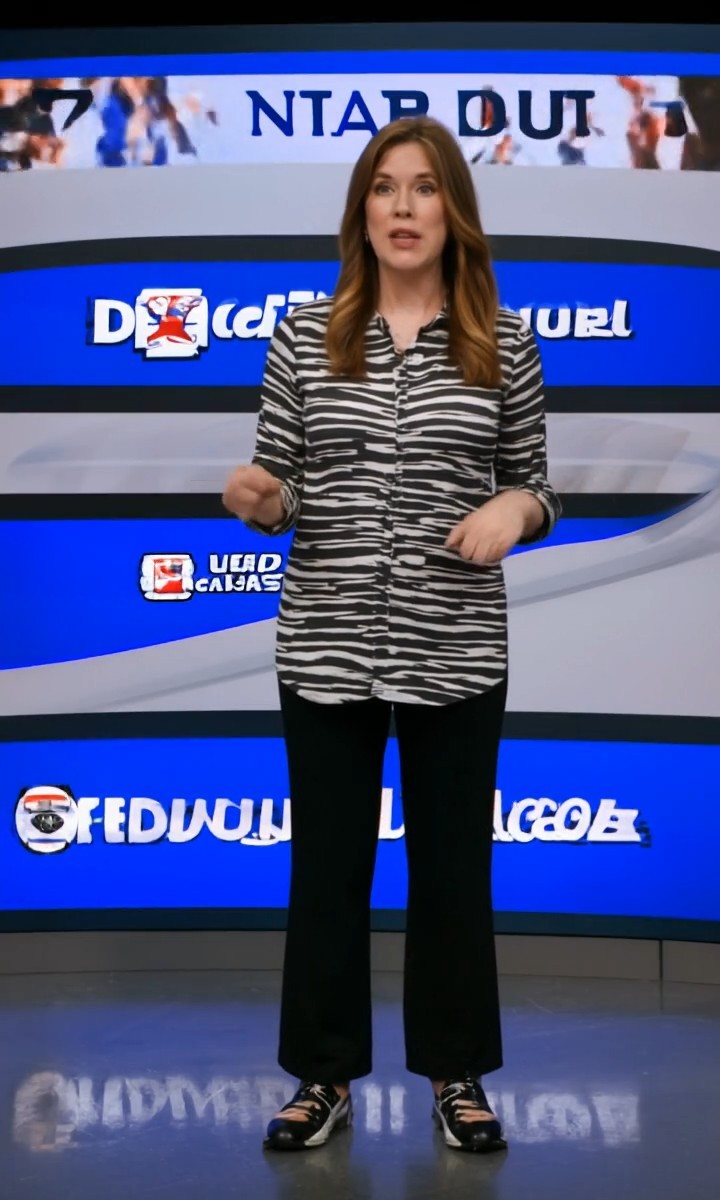} \\
        
        \rotatebox{90}{\tiny \textbf{\textsc{LTX-2.3}}} &
        \includegraphics[width=0.14\textwidth]{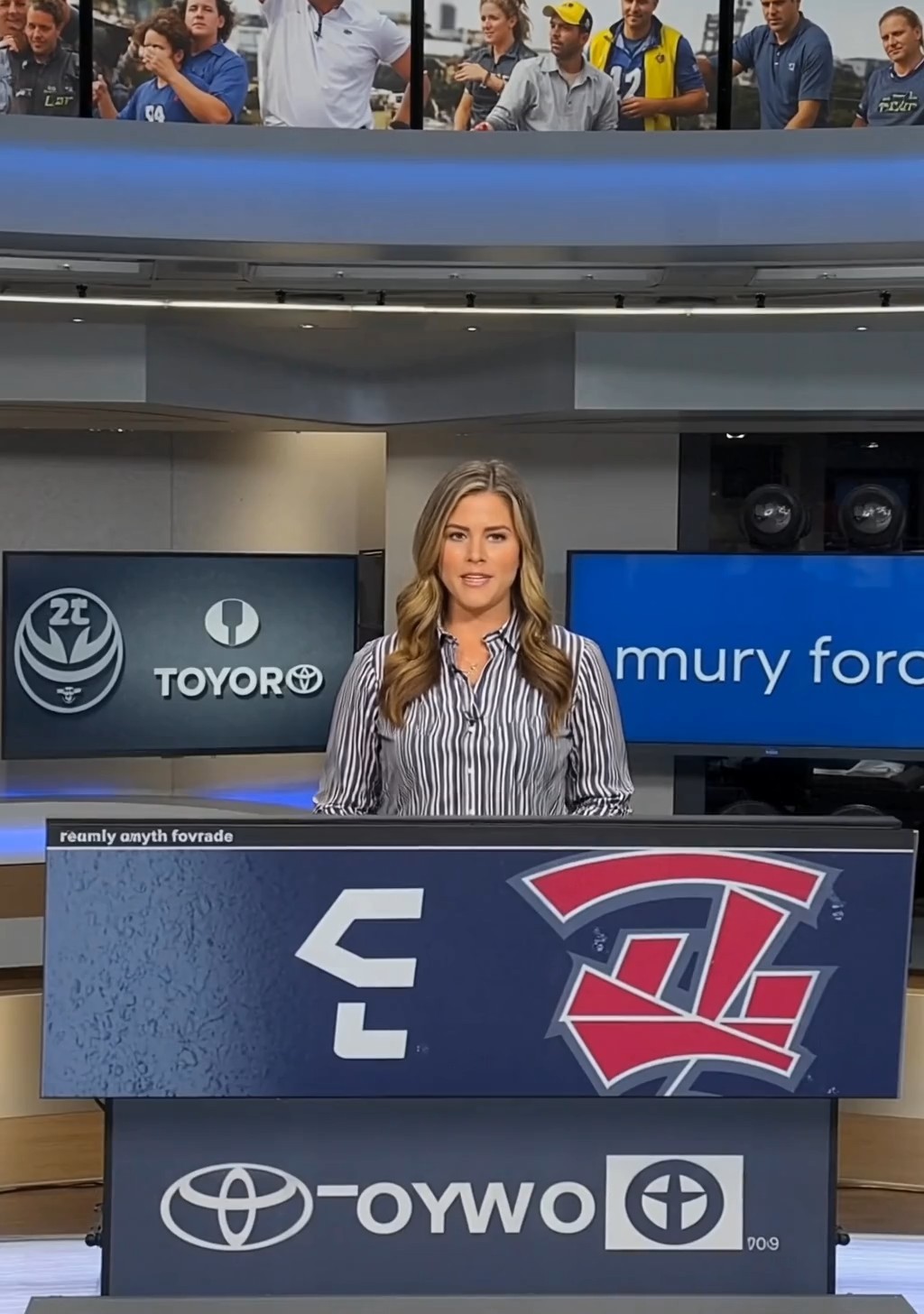} &
        \includegraphics[width=0.14\textwidth]{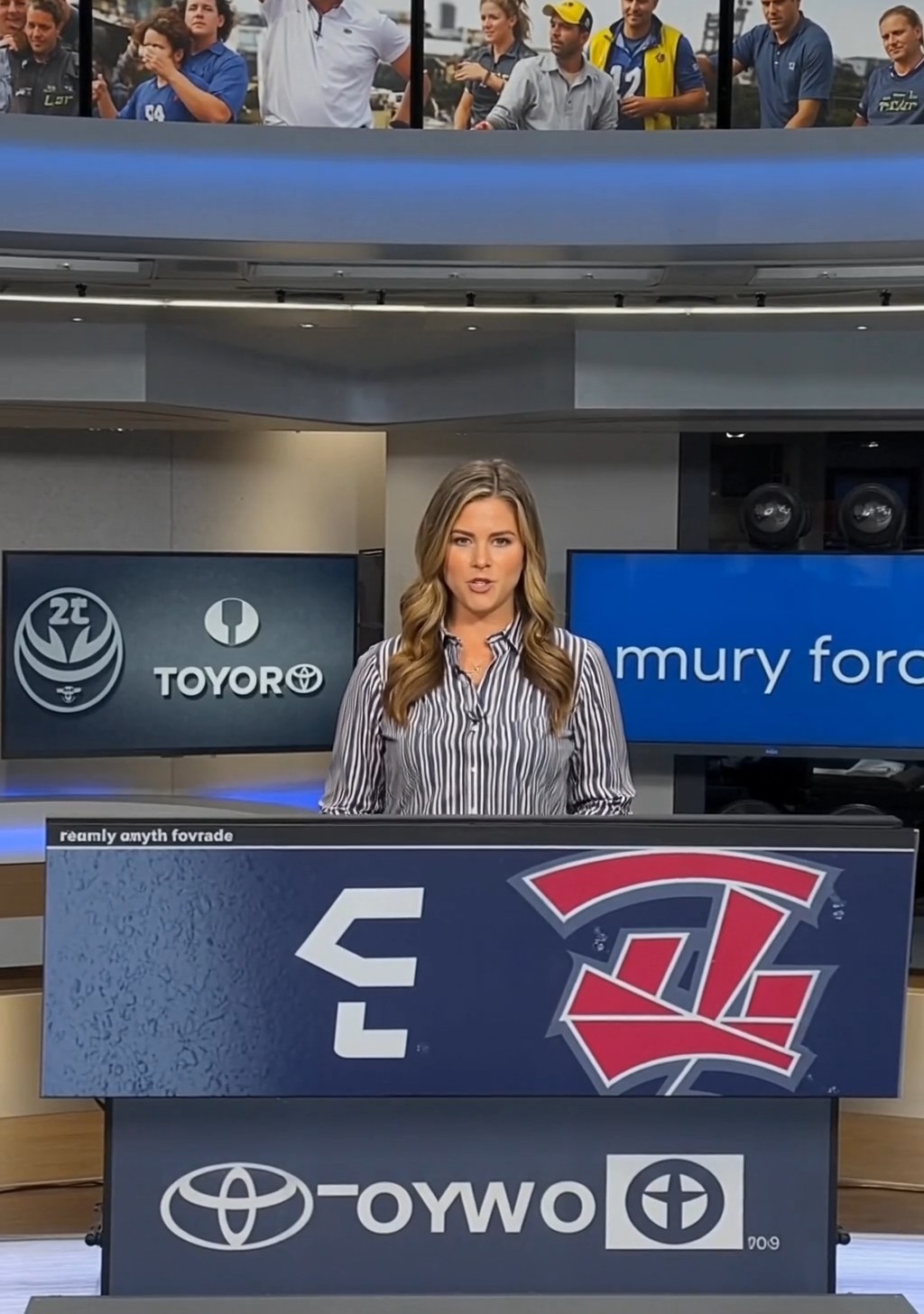} &
        \includegraphics[width=0.14\textwidth]{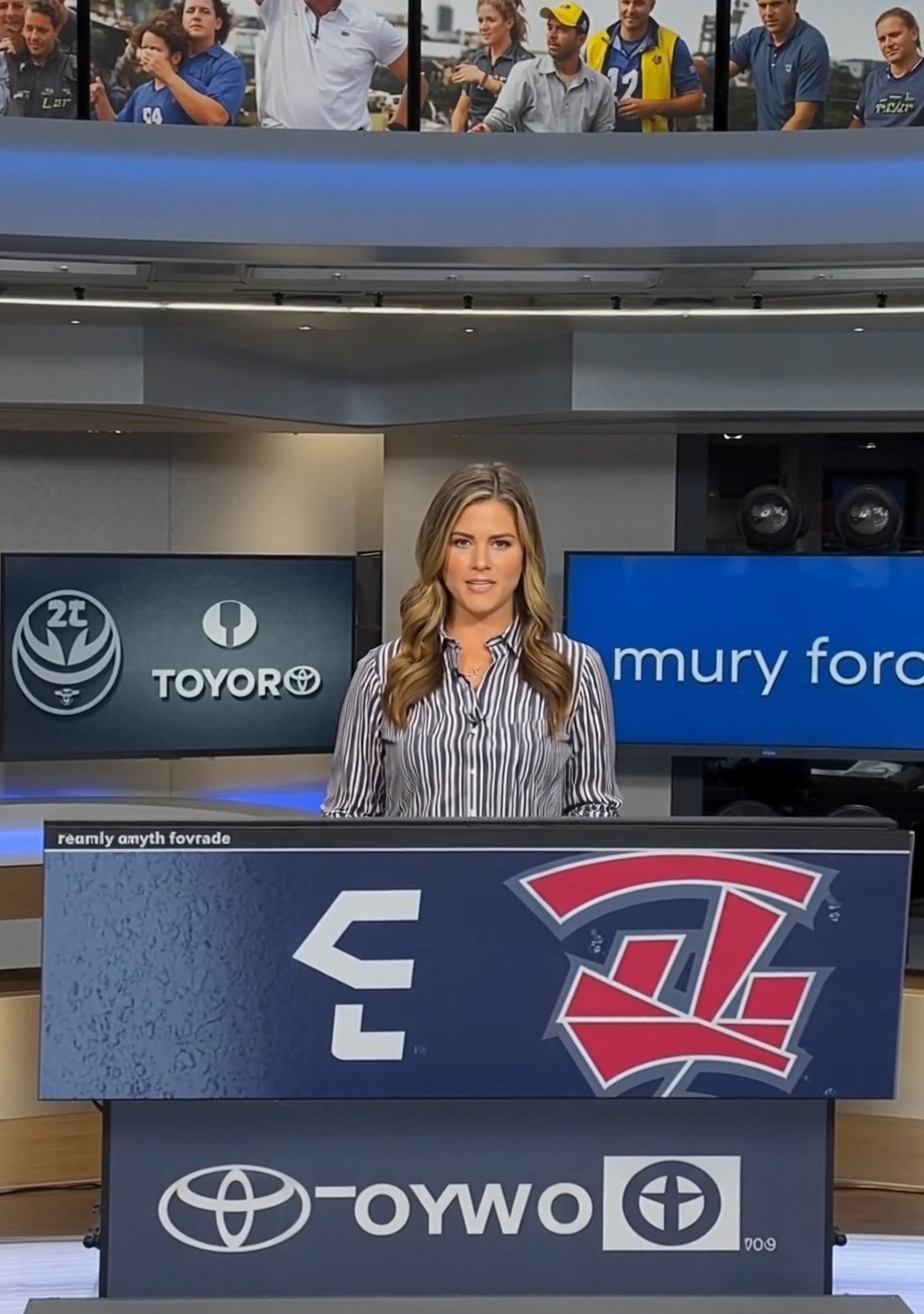} &
        \includegraphics[width=0.14\textwidth]{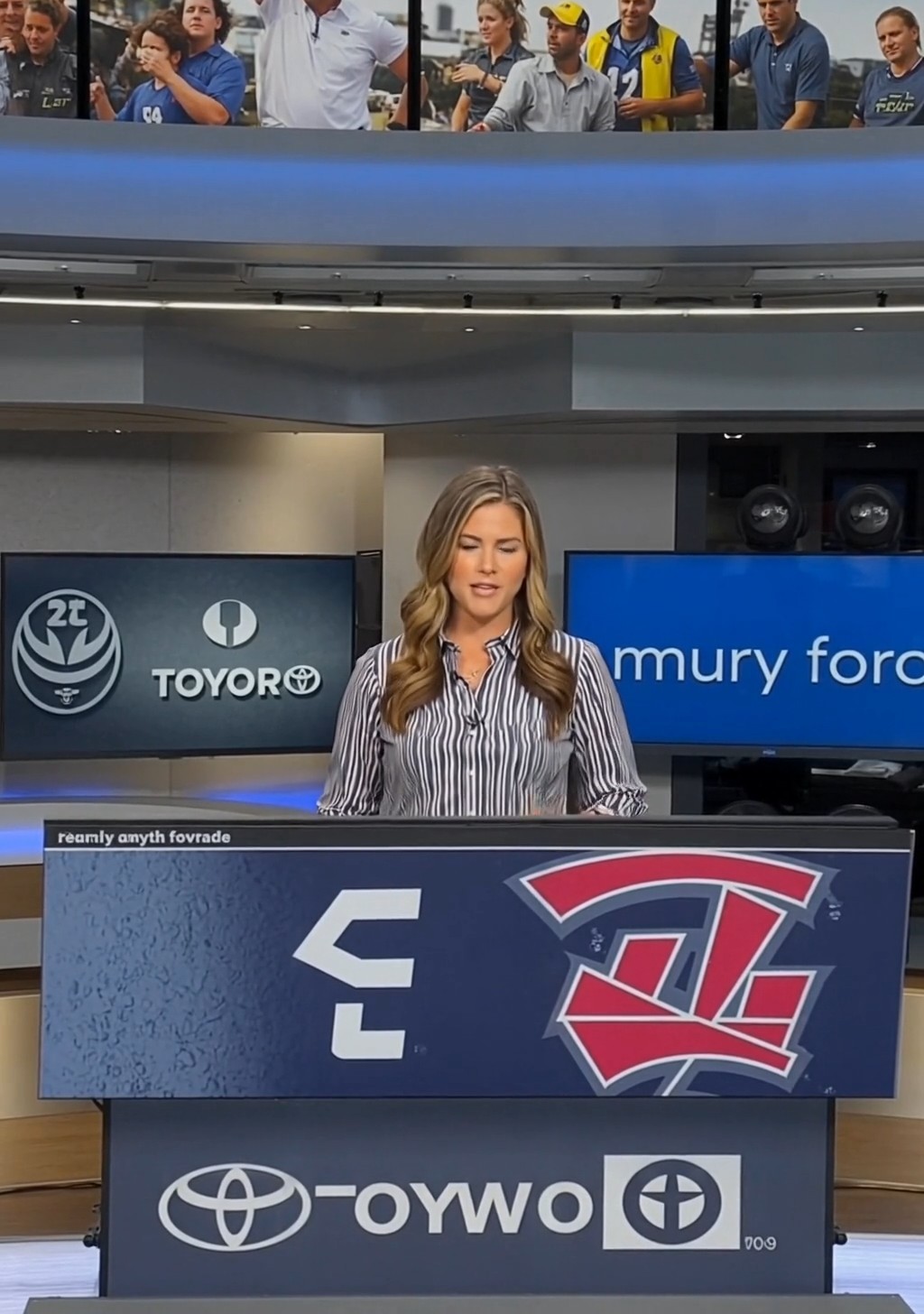} &
        \includegraphics[width=0.14\textwidth]{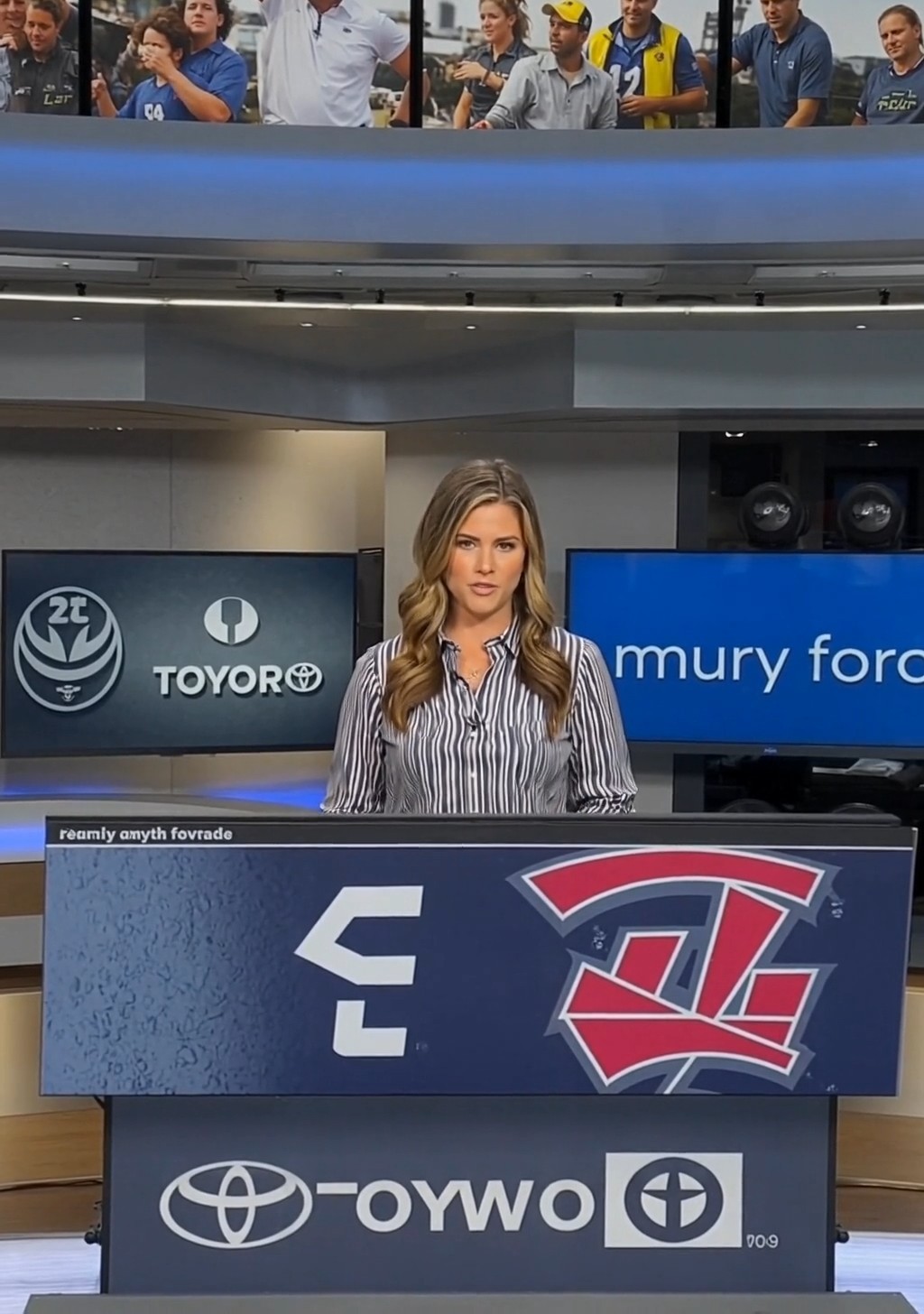} \\
        
        \rotatebox{90}{\tiny \textbf{\textsc{Helios}}} &
        \includegraphics[width=0.14\textwidth]{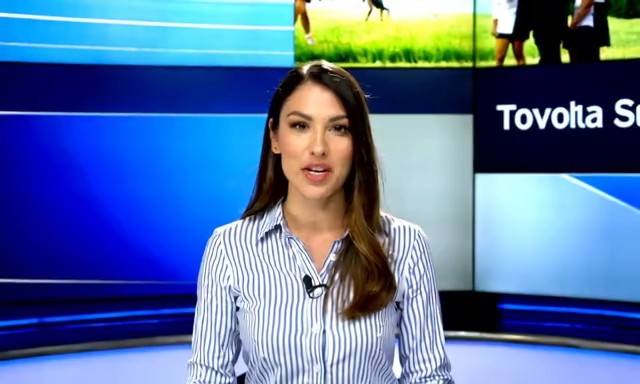} &
        \includegraphics[width=0.14\textwidth]{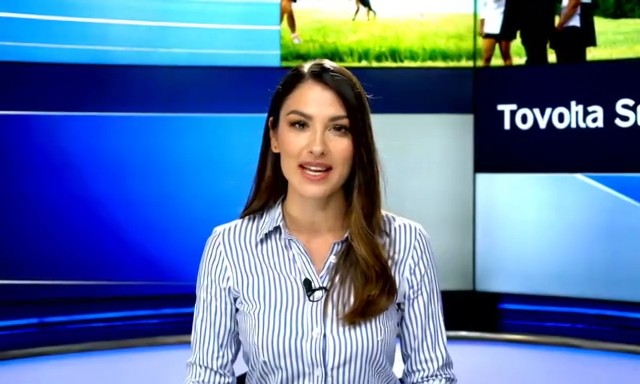} &
        \includegraphics[width=0.14\textwidth]{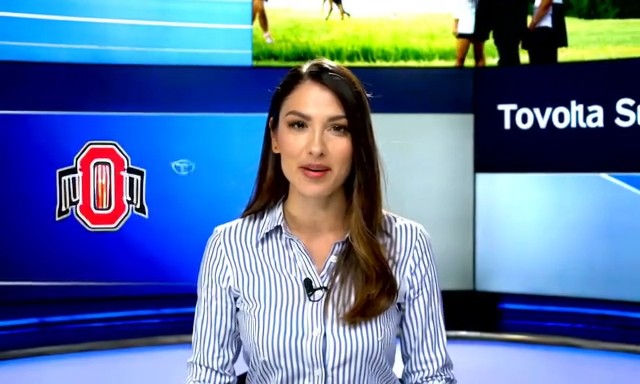} &
        \includegraphics[width=0.14\textwidth]{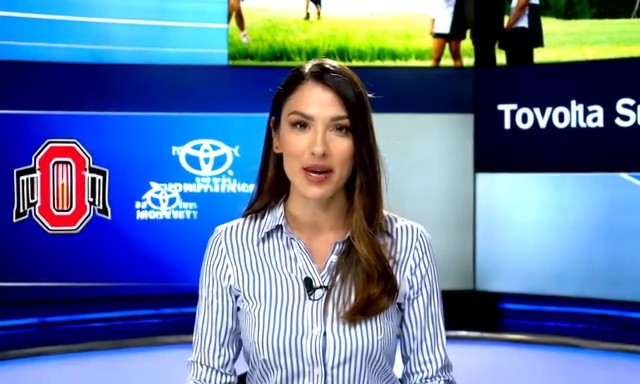} &
        \includegraphics[width=0.14\textwidth]{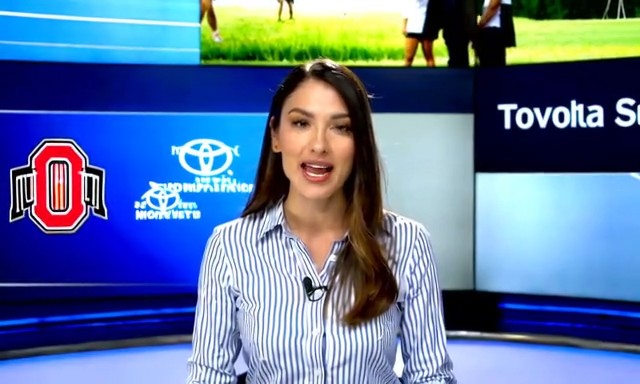} \\
        
        \rotatebox{90}{\tiny \textbf{\textsc{daVinci}}} &
        \includegraphics[width=0.14\textwidth]{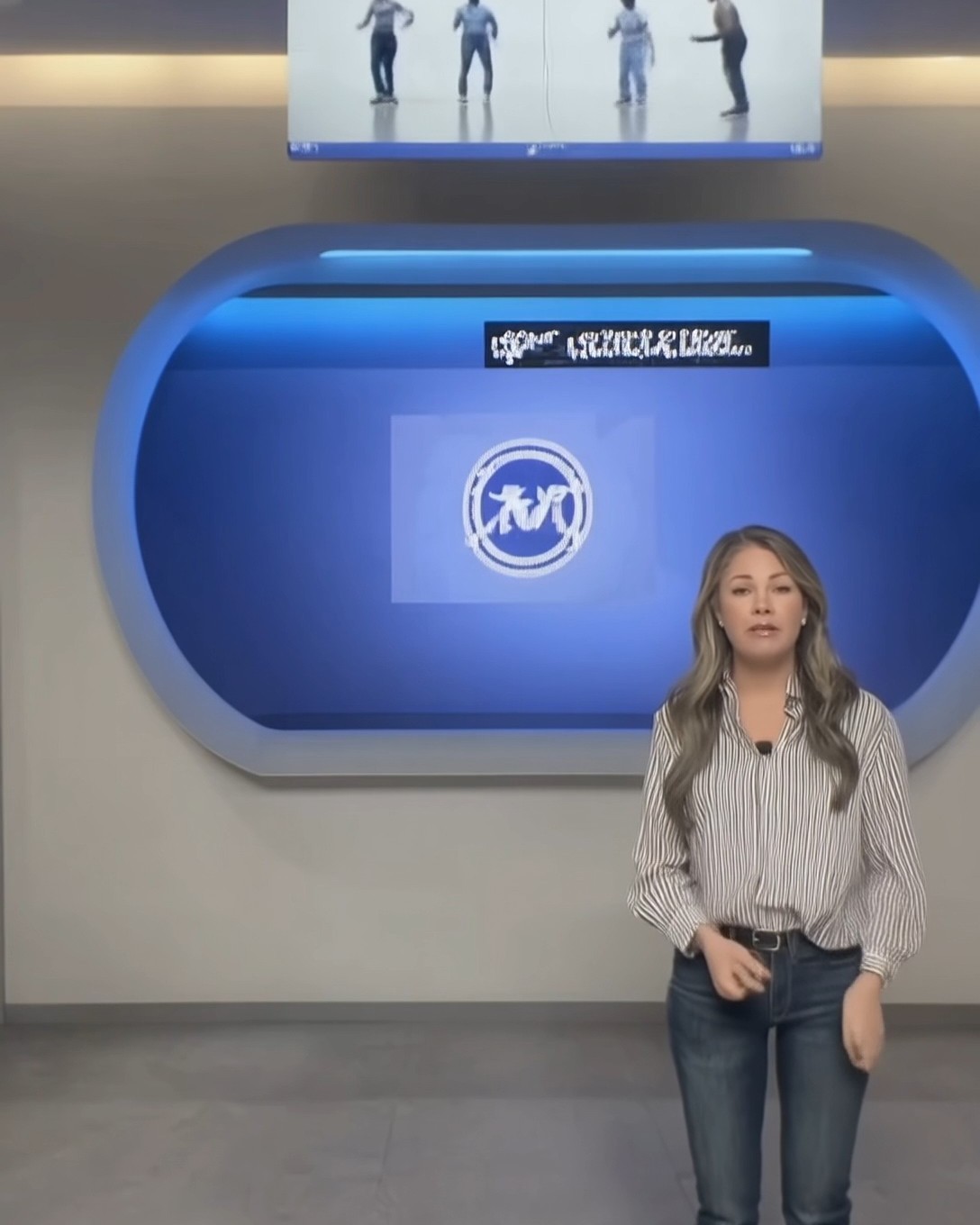} &
        \includegraphics[width=0.14\textwidth]{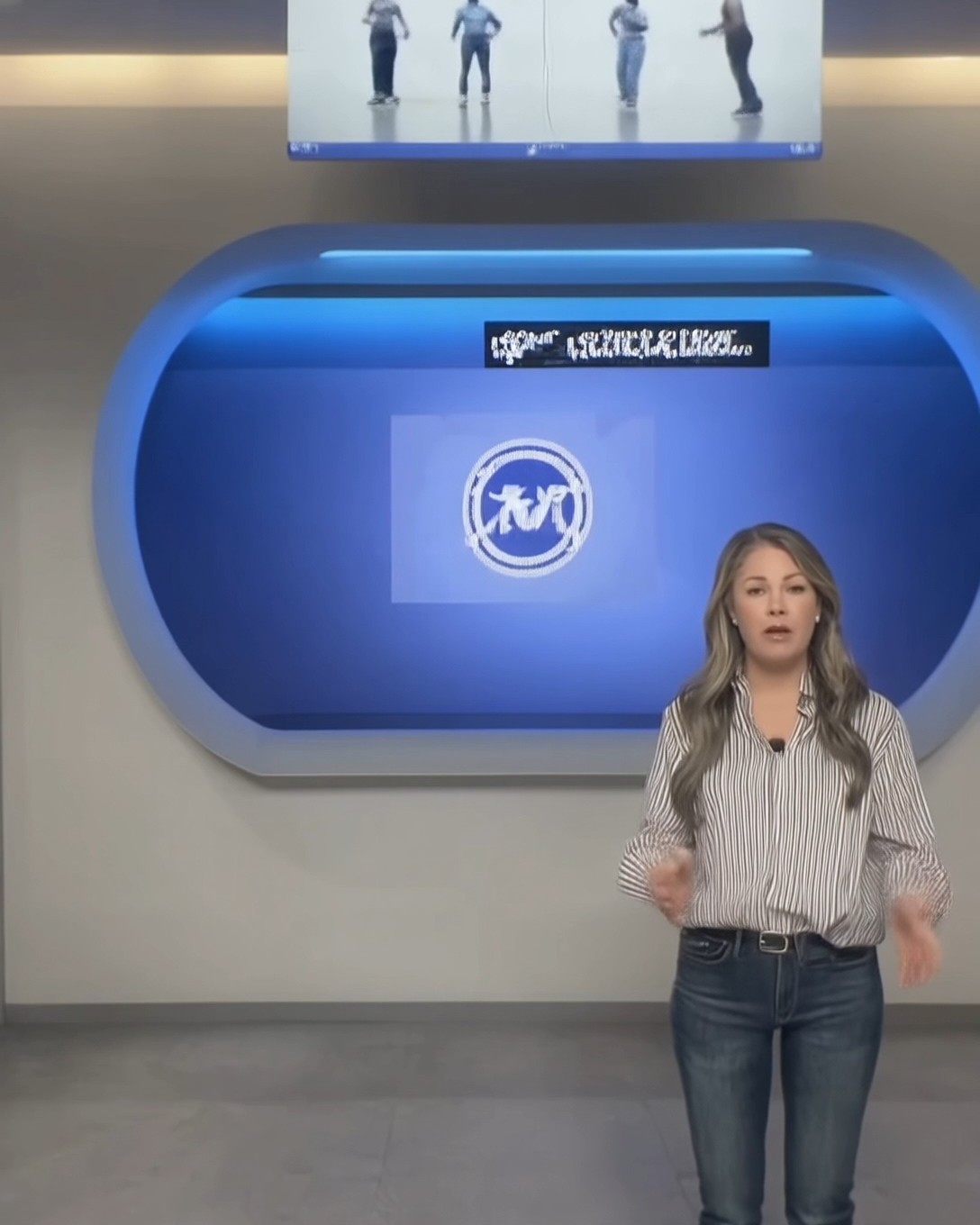} &
        \includegraphics[width=0.14\textwidth]{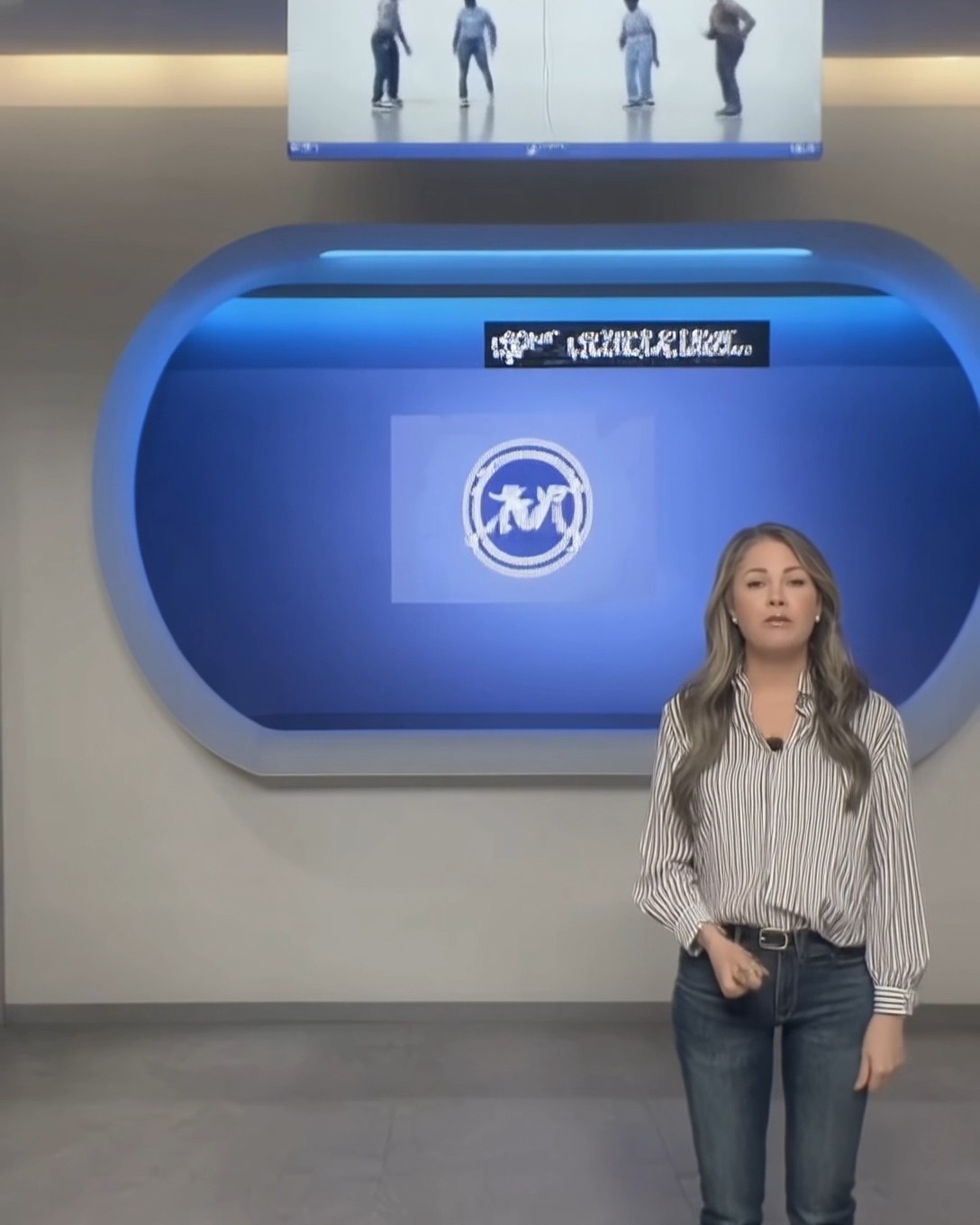} &
        \includegraphics[width=0.14\textwidth]{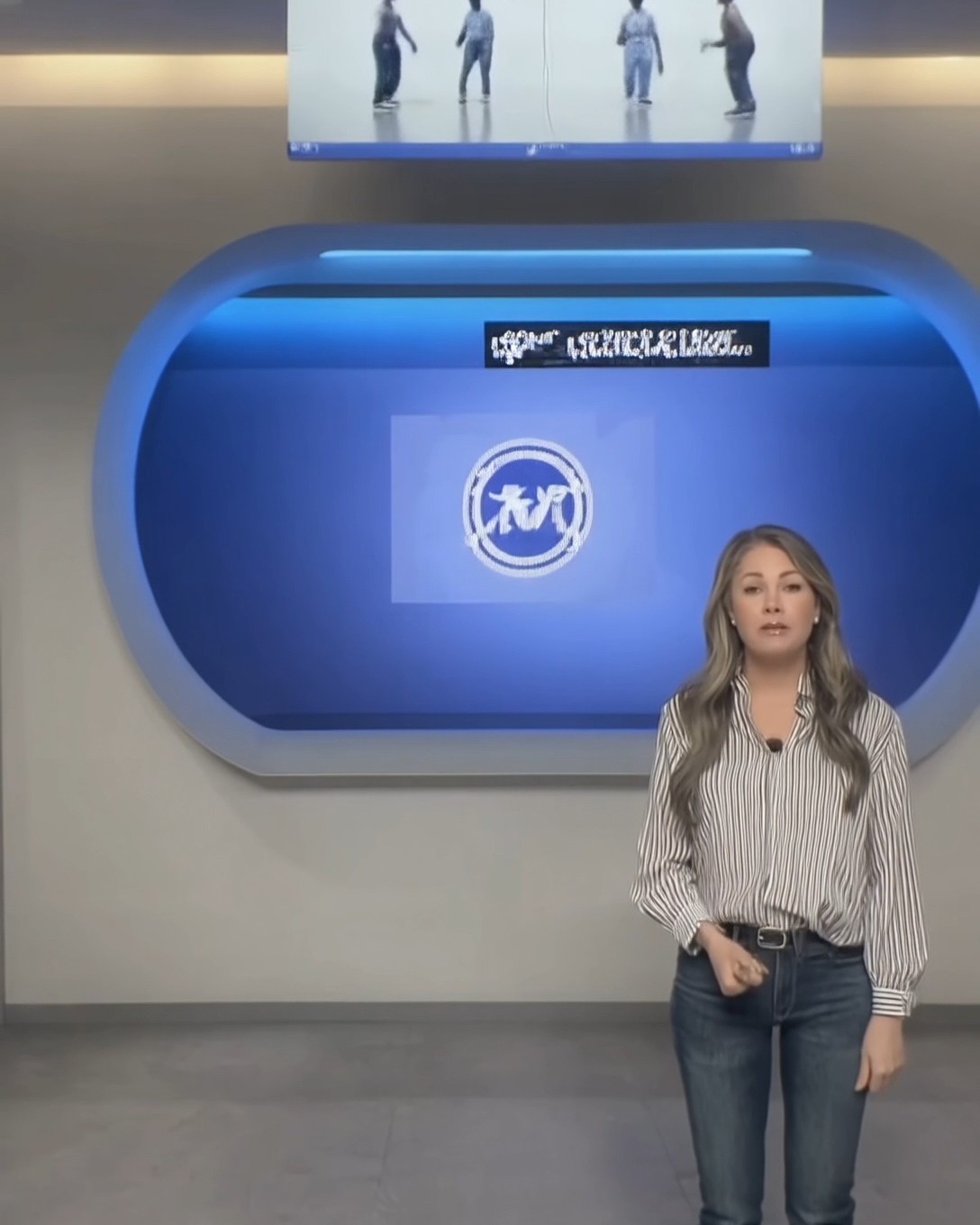} &
        \includegraphics[width=0.14\textwidth]{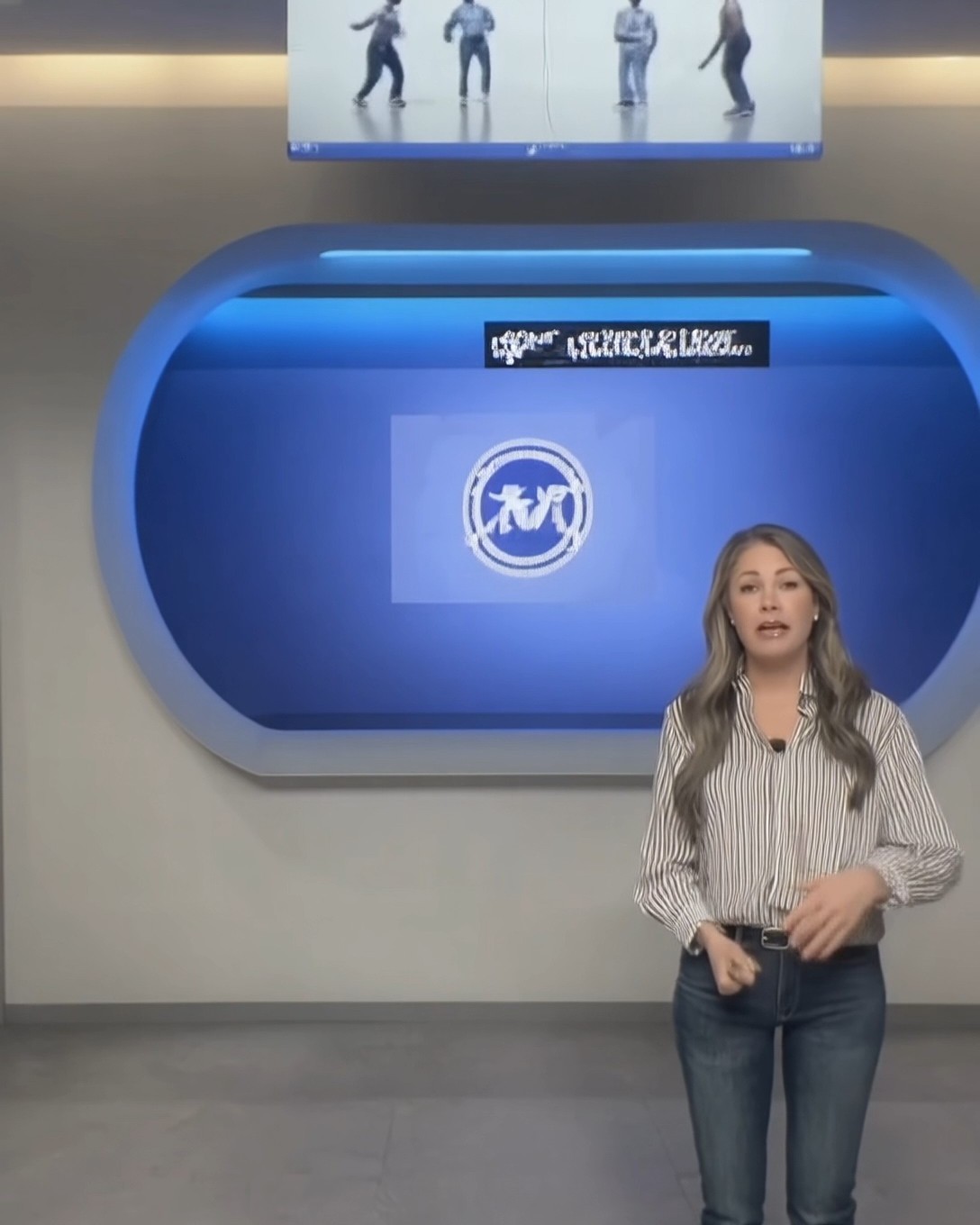} \\
        
        \rotatebox{90}{\tiny \textbf{\textsc{Self-Forcing}}} &
        \includegraphics[width=0.14\textwidth]{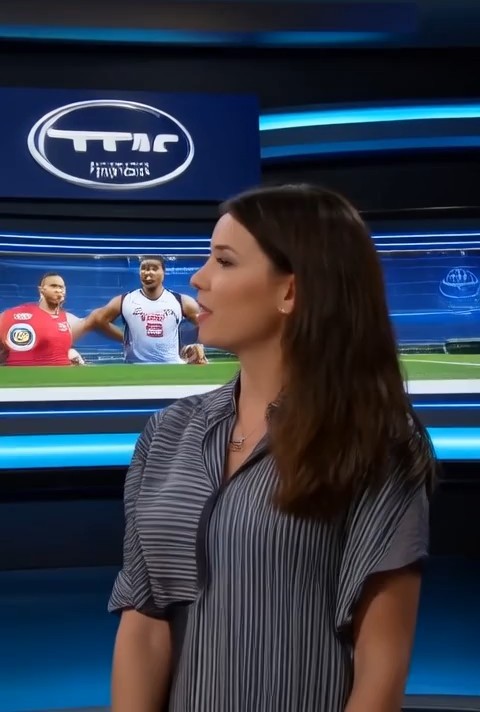} &
        \includegraphics[width=0.14\textwidth]{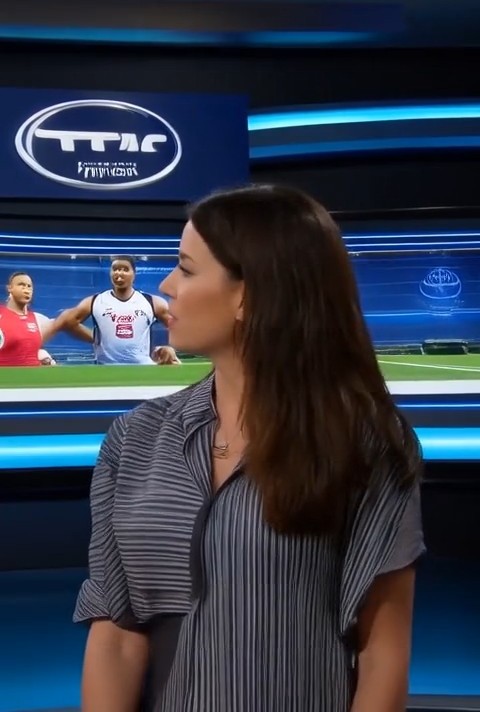} &
        \includegraphics[width=0.14\textwidth]{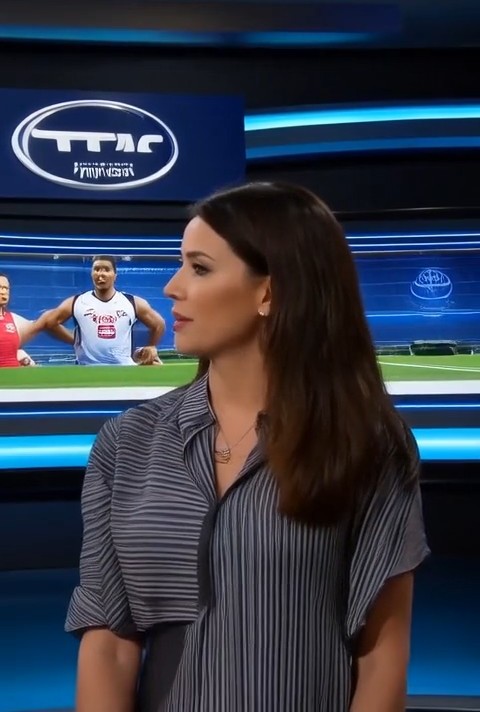} &
        \includegraphics[width=0.14\textwidth]{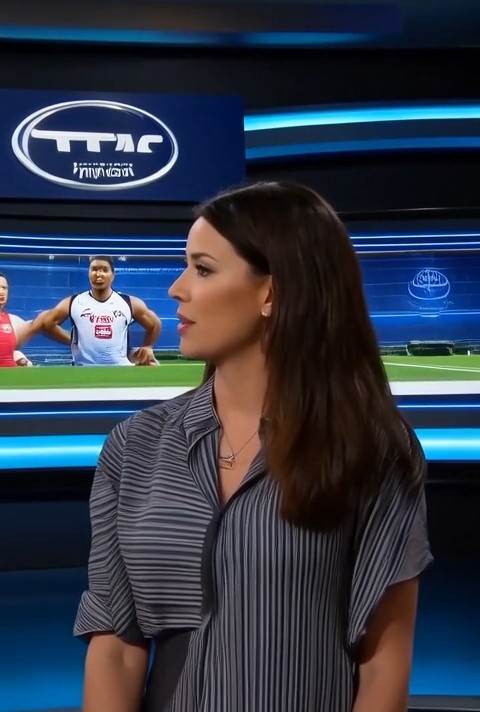} &
        \includegraphics[width=0.14\textwidth]{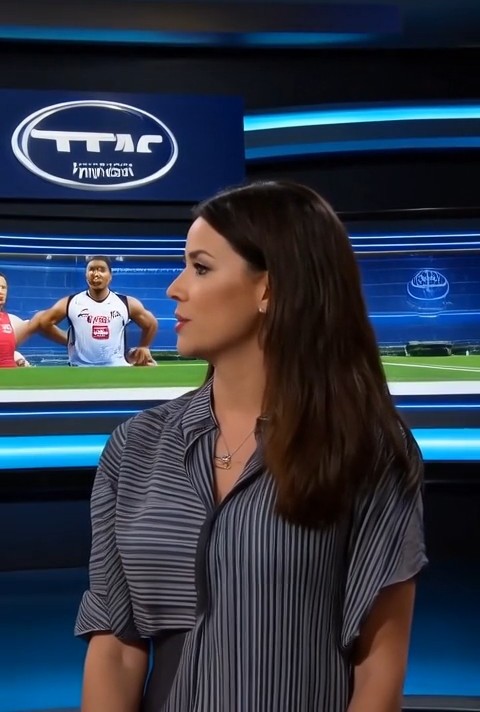} \\
    \end{tabular}
    \caption{Qualitative T2V comparison on FF++ pristine video 240 (sports broadcast). Frame sequences ($t=0$ to $t=4$) comparing the source video (top row) with synthetic outputs from the eight T2V generators. This scenario tests the models' ability to handle complex backgrounds (screens) and dynamic subject motion.}
    \label{fig:qualitative_240_t2v_v2}
\end{figure}

This appendix presents qualitative visual samples drawn from SynthForensics on representative source content. Figures~\ref{fig:qualitative_271_t2v_v2} and~\ref{fig:qualitative_240_t2v_v2} illustrate the per-generator text-to-video outputs on two pristine FaceForensics++ source videos: a news anchor scene and a sports broadcast. Across both scenarios, the eight SynthForensics generators successfully reproduce the principal semantic constraints encoded in the structured caption (e.g., the subject's attire and the studio environment), confirming that the paired-source prompting protocol described in Appendix~\ref{app:benchmark-prompts} elicits semantically aligned outputs from heterogeneous architectures.

\begin{figure}[!p]
    \centering
    \setlength{\tabcolsep}{1pt}
    \renewcommand{\arraystretch}{1.4}
    \begin{tabular}{c ccccc}
        & $t=0$ & $t=1$ & $t=2$ & $t=3$ & $t=4$ \\
        \rotatebox{90}{\tiny \textbf{\textsc{Wan2.1}}} &
        \includegraphics[width=0.16\textwidth]{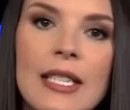} &
        \includegraphics[width=0.16\textwidth]{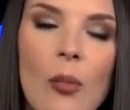} &
        \includegraphics[width=0.16\textwidth]{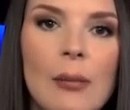} &
        \includegraphics[width=0.16\textwidth]{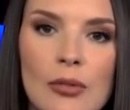} &
        \includegraphics[width=0.16\textwidth]{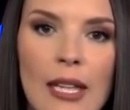} \\
        
        \rotatebox{90}{\tiny \textbf{\textsc{CogVideoX}}} &
        \includegraphics[width=0.16\textwidth]{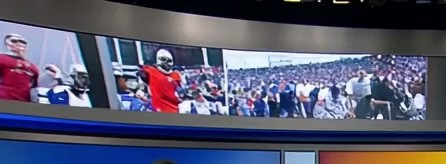} &
        \includegraphics[width=0.16\textwidth]{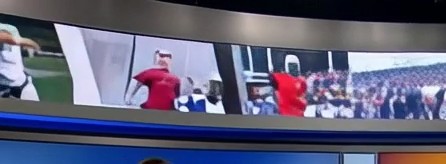} &
        \includegraphics[width=0.16\textwidth]{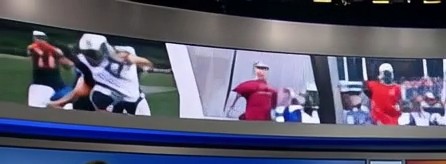} &
        \includegraphics[width=0.16\textwidth]{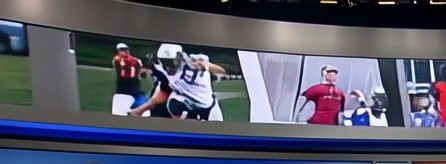} &
        \includegraphics[width=0.16\textwidth]{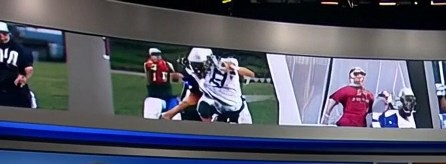} \\
        
        \rotatebox{90}{\tiny \textbf{\textsc{SkyReels-V2}}} &
        \includegraphics[width=0.16\textwidth]{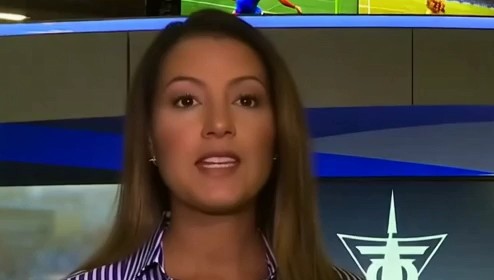} &
        \includegraphics[width=0.16\textwidth]{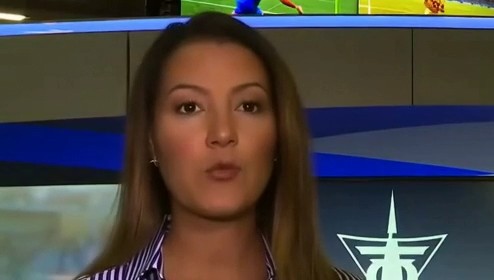} &
        \includegraphics[width=0.16\textwidth]{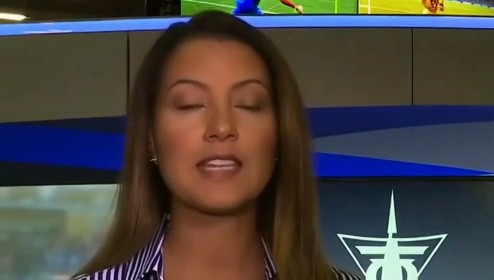} &
        \includegraphics[width=0.16\textwidth]{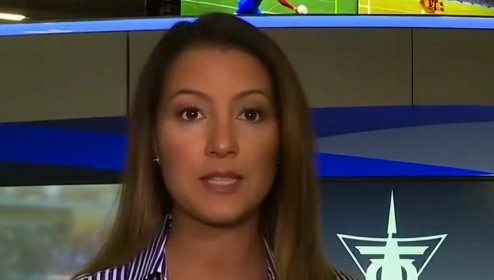} &
        \includegraphics[width=0.16\textwidth]{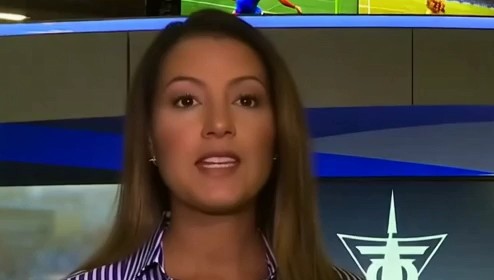} \\
        
        \rotatebox{90}{\tiny \textbf{\textsc{MAGI-1}}} &
        \includegraphics[width=0.16\textwidth]{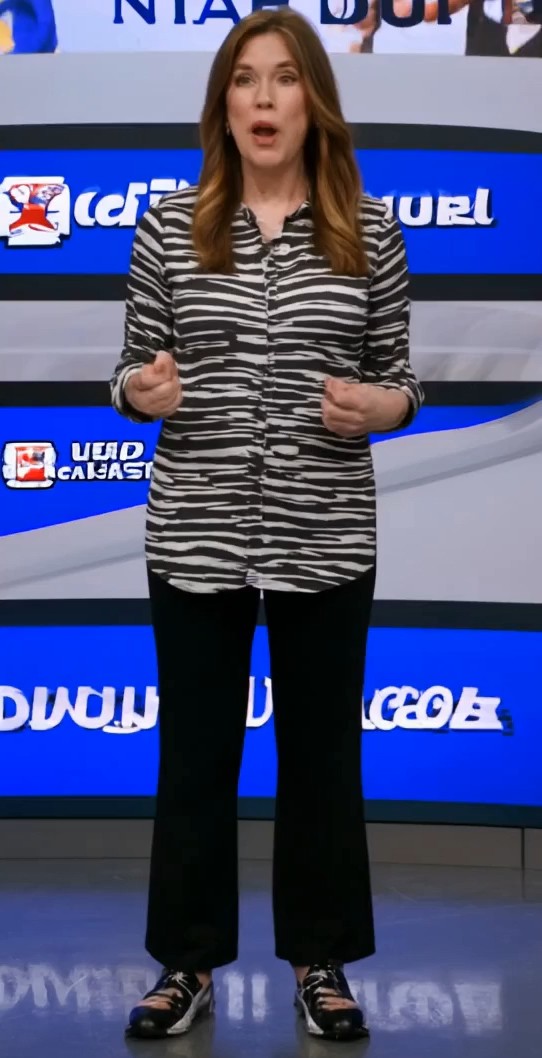} &
        \includegraphics[width=0.16\textwidth]{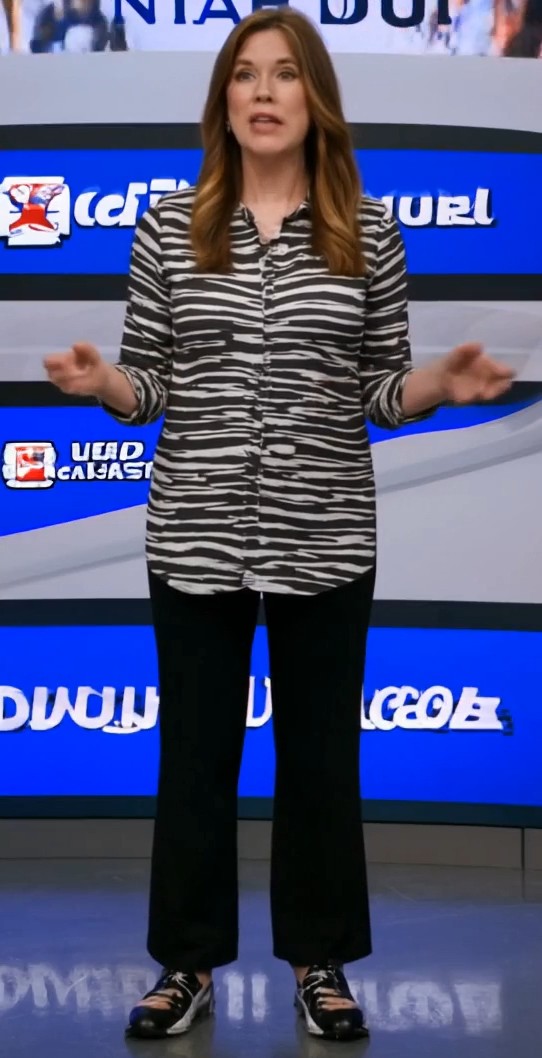} &
        \includegraphics[width=0.16\textwidth]{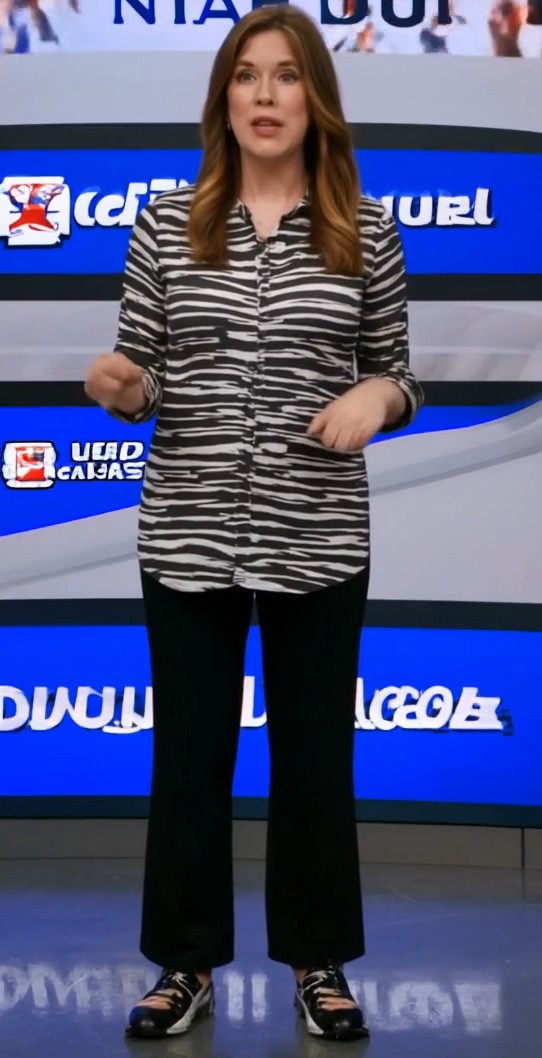} &
        \includegraphics[width=0.16\textwidth]{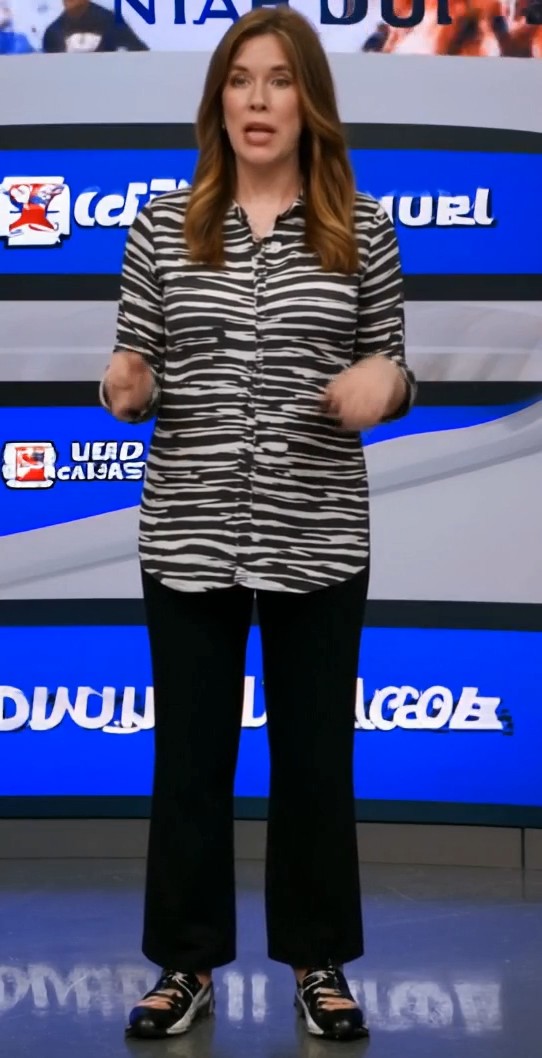} &
        \includegraphics[width=0.16\textwidth]{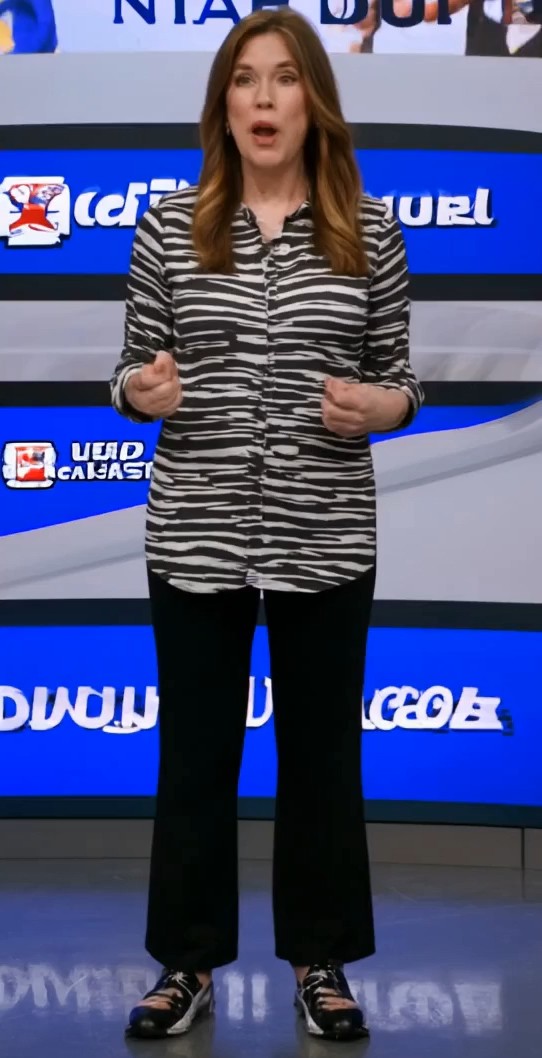} \\
        
        \rotatebox{90}{\tiny \textbf{\textsc{LTX-2.3}}} &
        \includegraphics[width=0.16\textwidth]{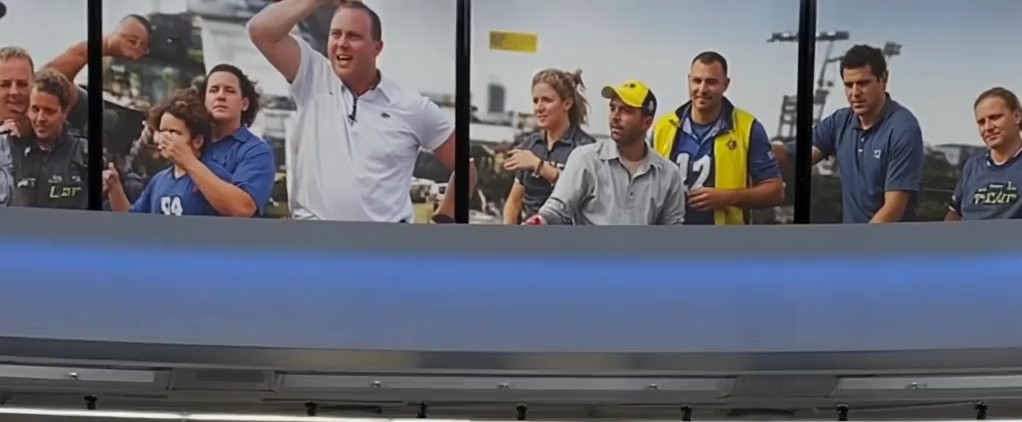} &
        \includegraphics[width=0.16\textwidth]{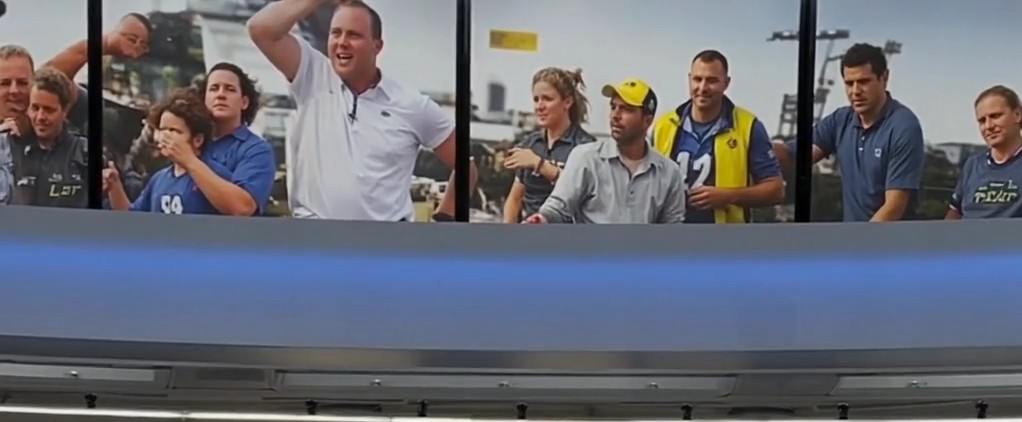} &
        \includegraphics[width=0.16\textwidth]{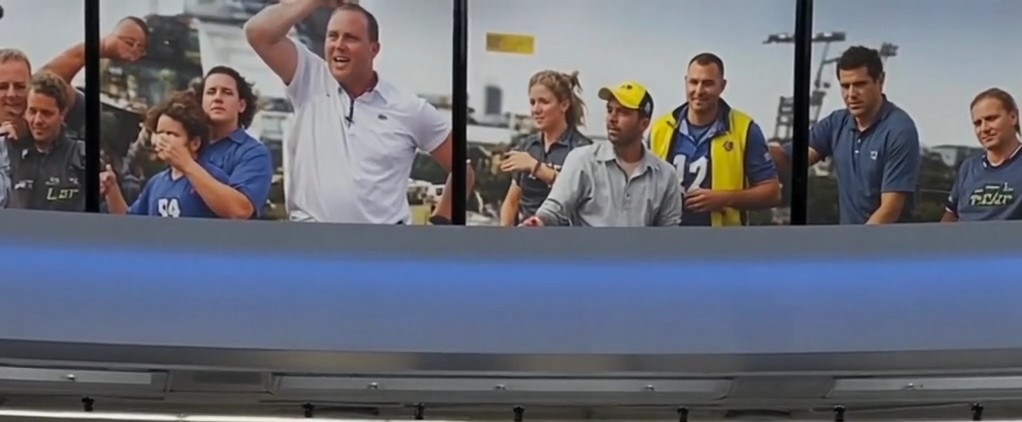} &
        \includegraphics[width=0.16\textwidth]{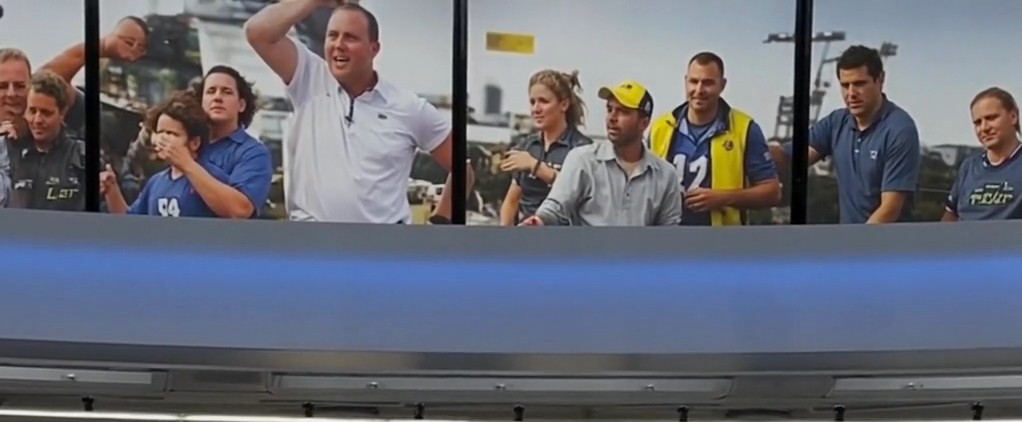} &
        \includegraphics[width=0.16\textwidth]{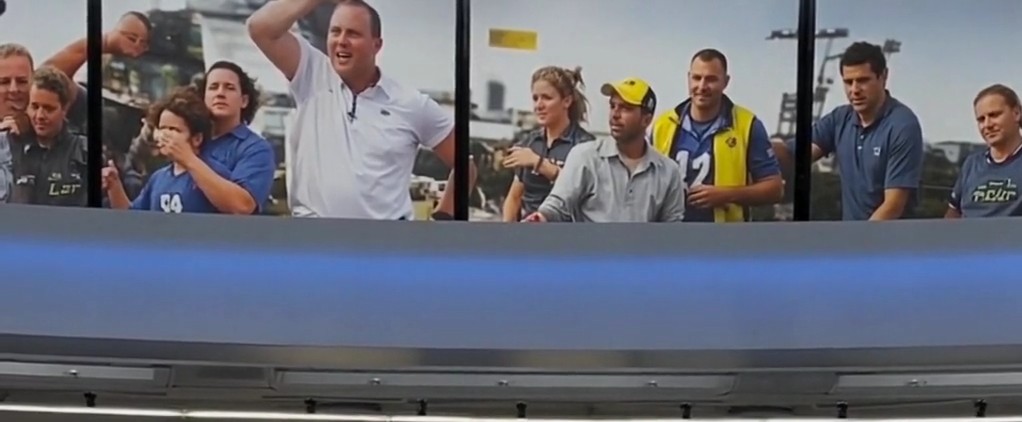} \\
        
        \rotatebox{90}{\tiny \textbf{\textsc{Helios}}} &
        \includegraphics[width=0.16\textwidth]{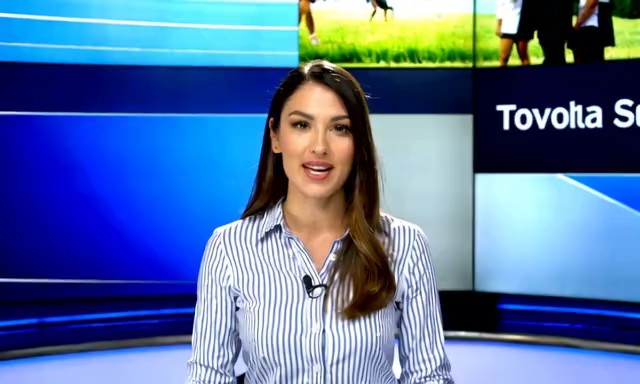} &
        \includegraphics[width=0.16\textwidth]{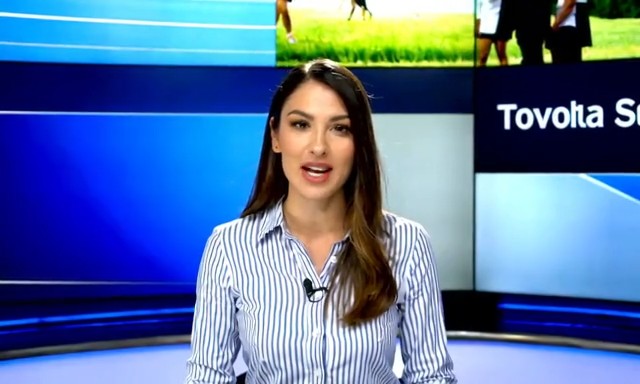} &
        \includegraphics[width=0.16\textwidth]{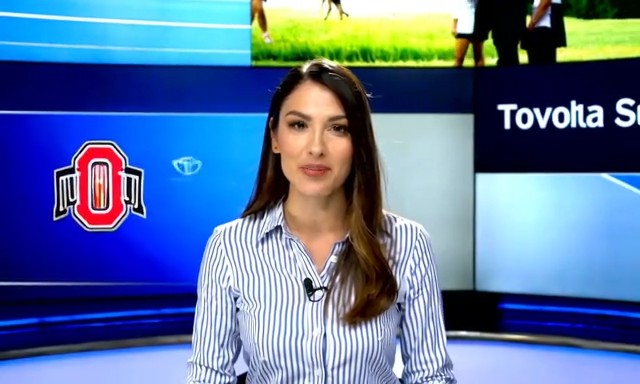} &
        \includegraphics[width=0.16\textwidth]{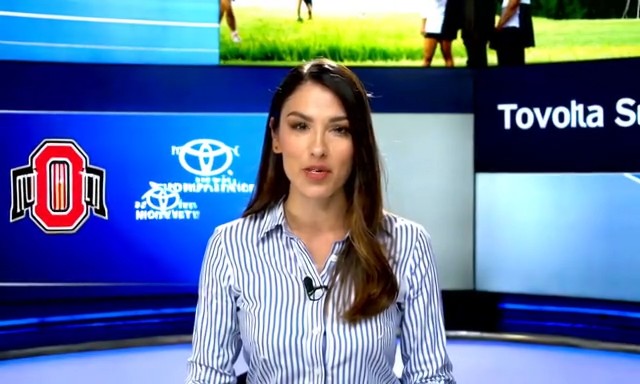} &
        \includegraphics[width=0.16\textwidth]{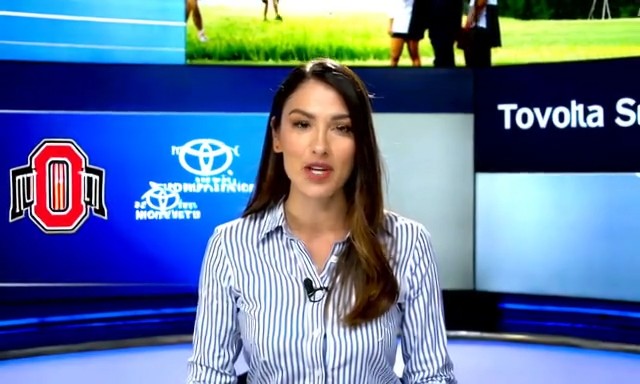} \\
        
        \rotatebox{90}{\tiny \textbf{\textsc{daVinci-MagiHuman}}} &
        \includegraphics[width=0.16\textwidth]{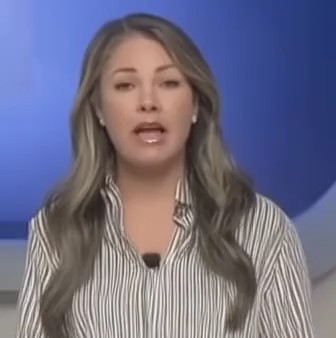} &
        \includegraphics[width=0.16\textwidth]{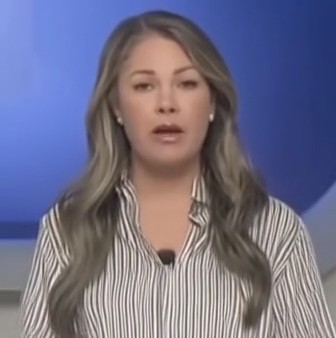} &
        \includegraphics[width=0.16\textwidth]{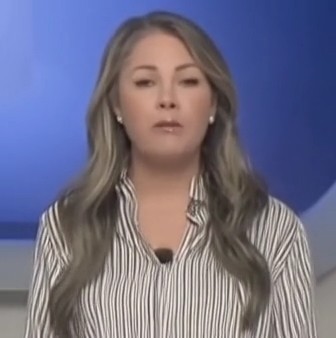} &
        \includegraphics[width=0.16\textwidth]{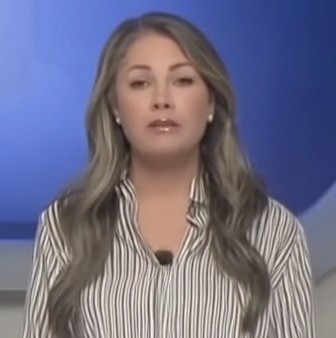} &
        \includegraphics[width=0.16\textwidth]{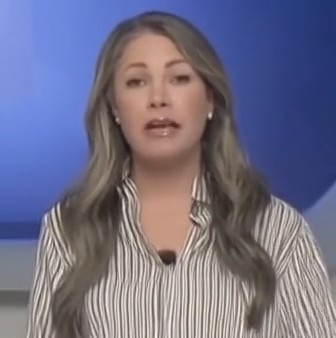} \\
        
        \rotatebox{90}{\tiny \textbf{\textsc{Self-Forcing}}} &
        \includegraphics[width=0.16\textwidth]{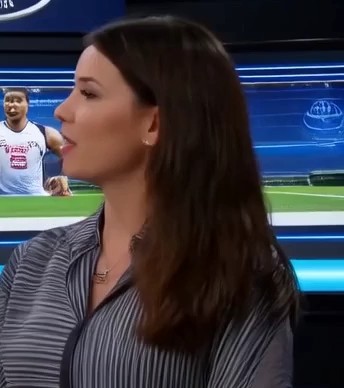} &
        \includegraphics[width=0.16\textwidth]{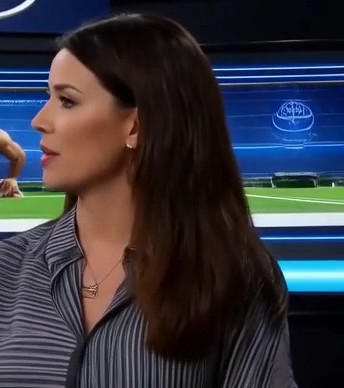} &
        \includegraphics[width=0.16\textwidth]{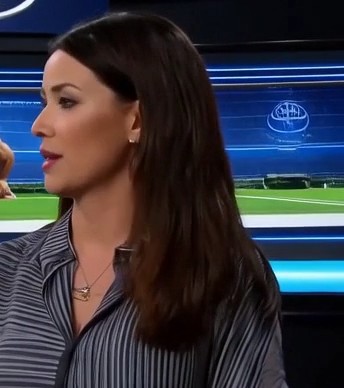} &
        \includegraphics[width=0.16\textwidth]{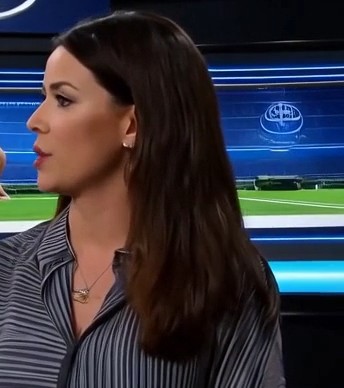} &
        \includegraphics[width=0.16\textwidth]{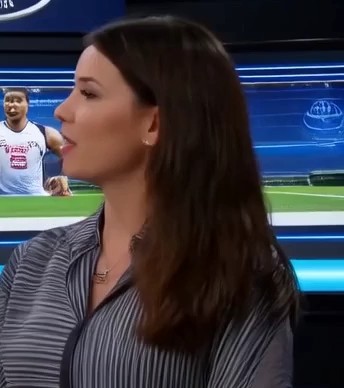} \\
    \end{tabular}
    \caption{Fine-grained generation fidelity analysis on FF++ pristine video 240 (sports broadcast), with magnified per-generator crops of the outputs shown in Figure~\ref{fig:qualitative_240_t2v_v2}. Frame sequences ($t=0$ to $t=4$) highlight architecture-specific strengths.}
    \label{fig:qualitative_details_v2}
\end{figure}

To probe fine-grained generation fidelity, Figure~\ref{fig:qualitative_details_v2} reports magnified per-generator crops extracted from the sports broadcast sequence. The crops surface architecture-specific qualitative strengths: Wan2.1 maintains anatomical integrity during transient facial events, with naturalistic blinking motion at $t=1$ free of iris deformation; CogVideoX preserves exceptional temporal coherence in the animated background screens, rendering the embedded ``video-within-video'' motion without texture boiling; SkyReels-V2 sustains consistent facial-landmark geometry during the blinking event at $t=2$; MAGI-1 produces natural kinematic flow during articulated body motion, with the high-frequency zebra pattern on the presenter's shirt deforming realistically with body geometry; LTX-2.3 reconstructs the multi-person background composition without per-individual identity drift, with the additional subjects maintaining distinct attire and posture across the sequence; Helios renders high-frequency studio graphics (logos, scrolling banner text) and the presenter's striped shirt with crisp pattern continuity; daVinci-MagiHuman delivers portrait-level facial fidelity with layered hair strands and stable striped-garment rendering throughout the sequence; and Self-Forcing exhibits high-fidelity physics simulation in hair dynamics, with strands flowing naturally over the shoulder without temporal blurring.

\begin{figure}[!p]
    \centering
    \setlength{\tabcolsep}{0pt}
    \renewcommand{\arraystretch}{1.2}
    \begin{tabular}{c ccccc}
        & $t=0$ & $t=1$ & $t=2$ & $t=3$ & $t=4$ \\
        \rotatebox{90}{\tiny \textbf{\textsc{Pristine}}} &
        \includegraphics[width=0.195\textwidth]{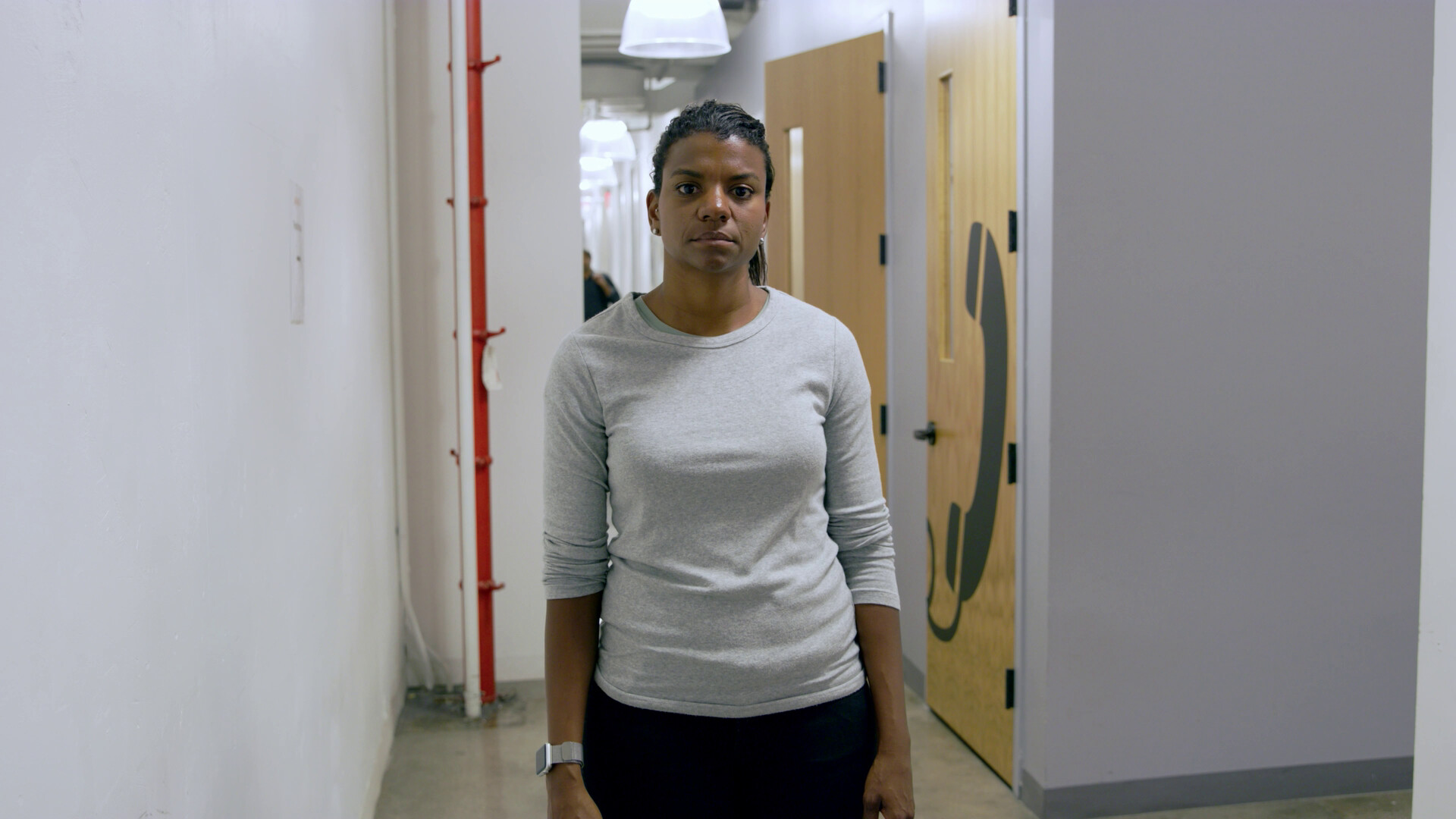} &
        \includegraphics[width=0.195\textwidth]{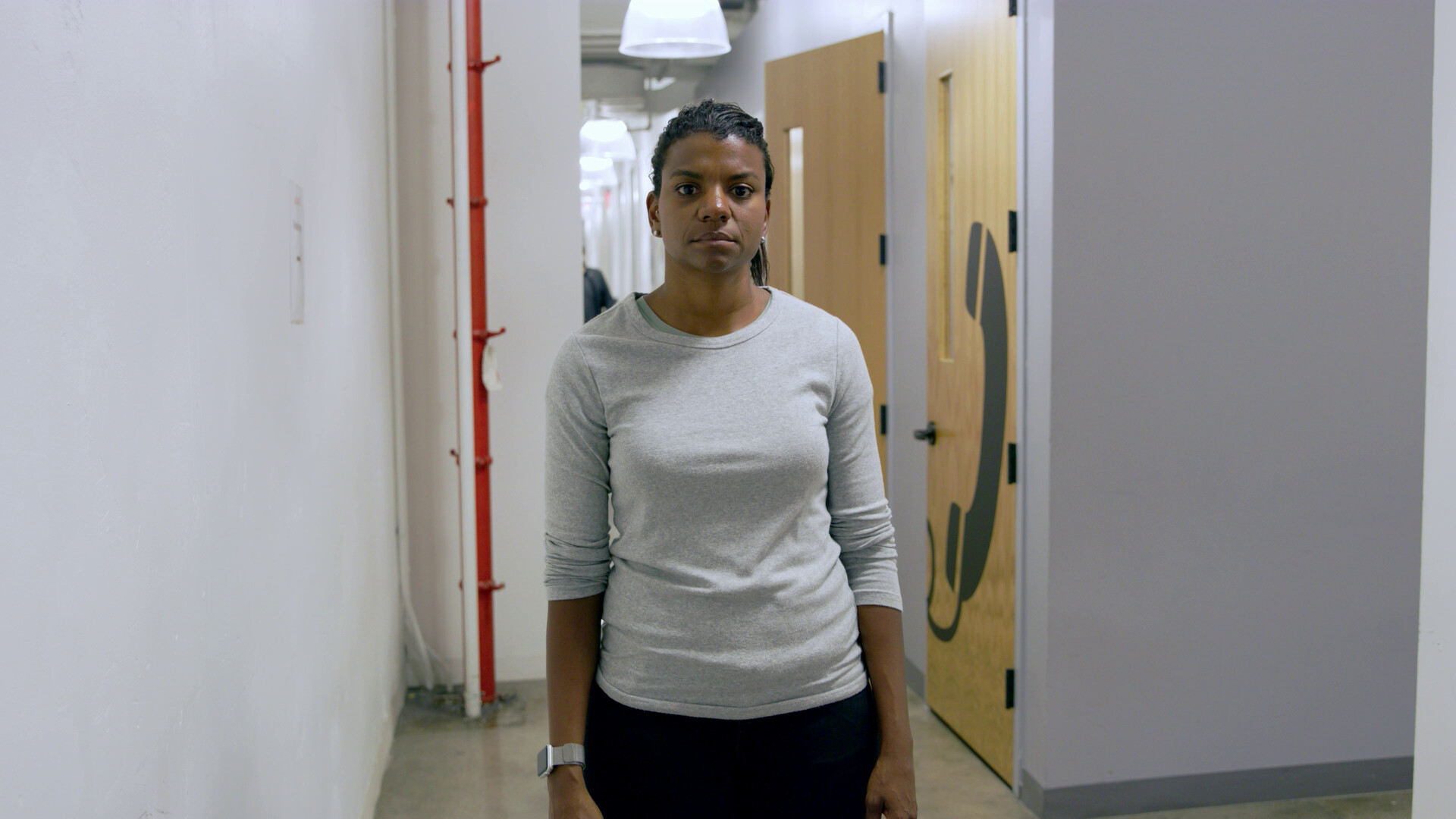} &
        \includegraphics[width=0.195\textwidth]{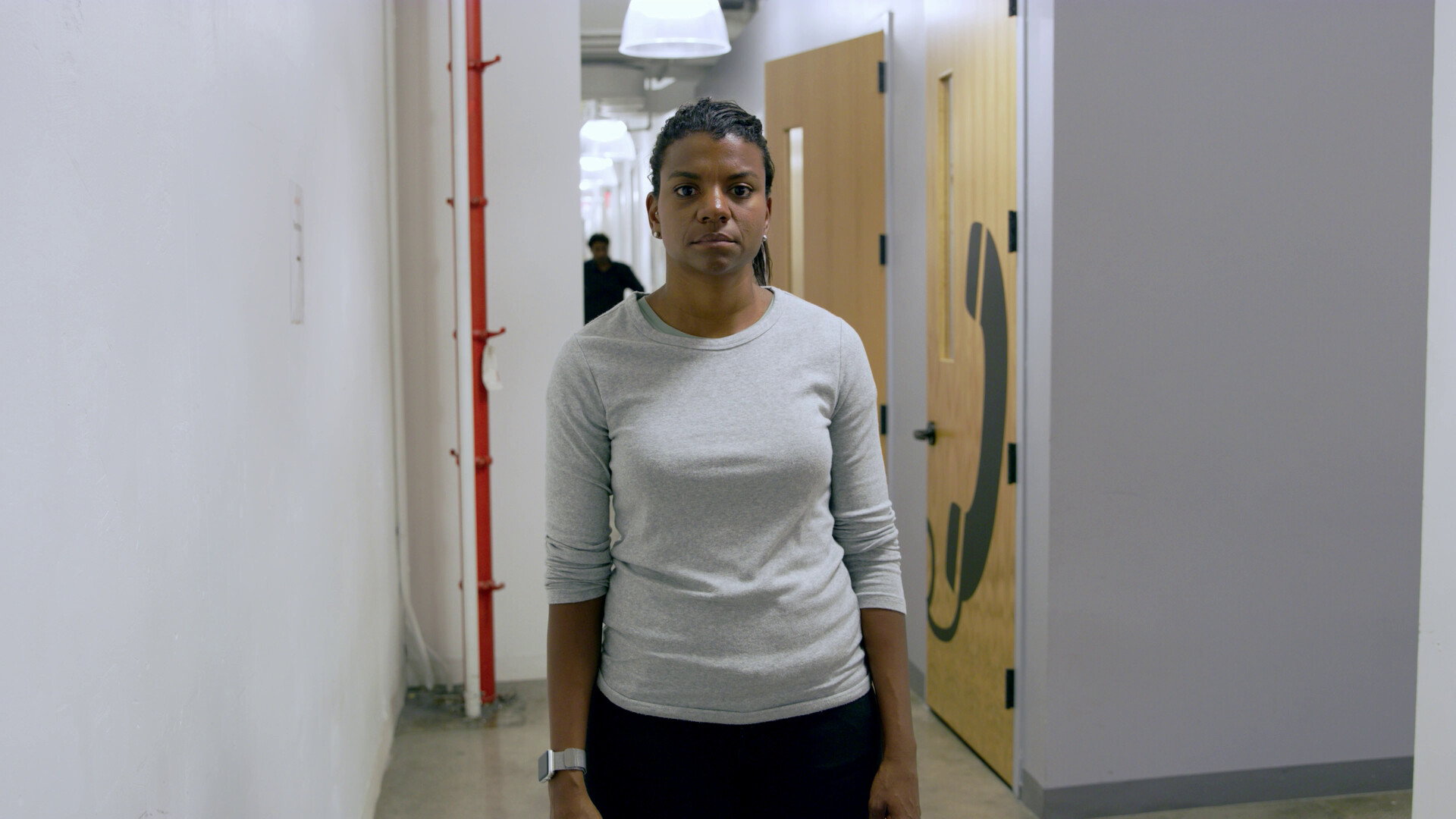} &
        \includegraphics[width=0.195\textwidth]{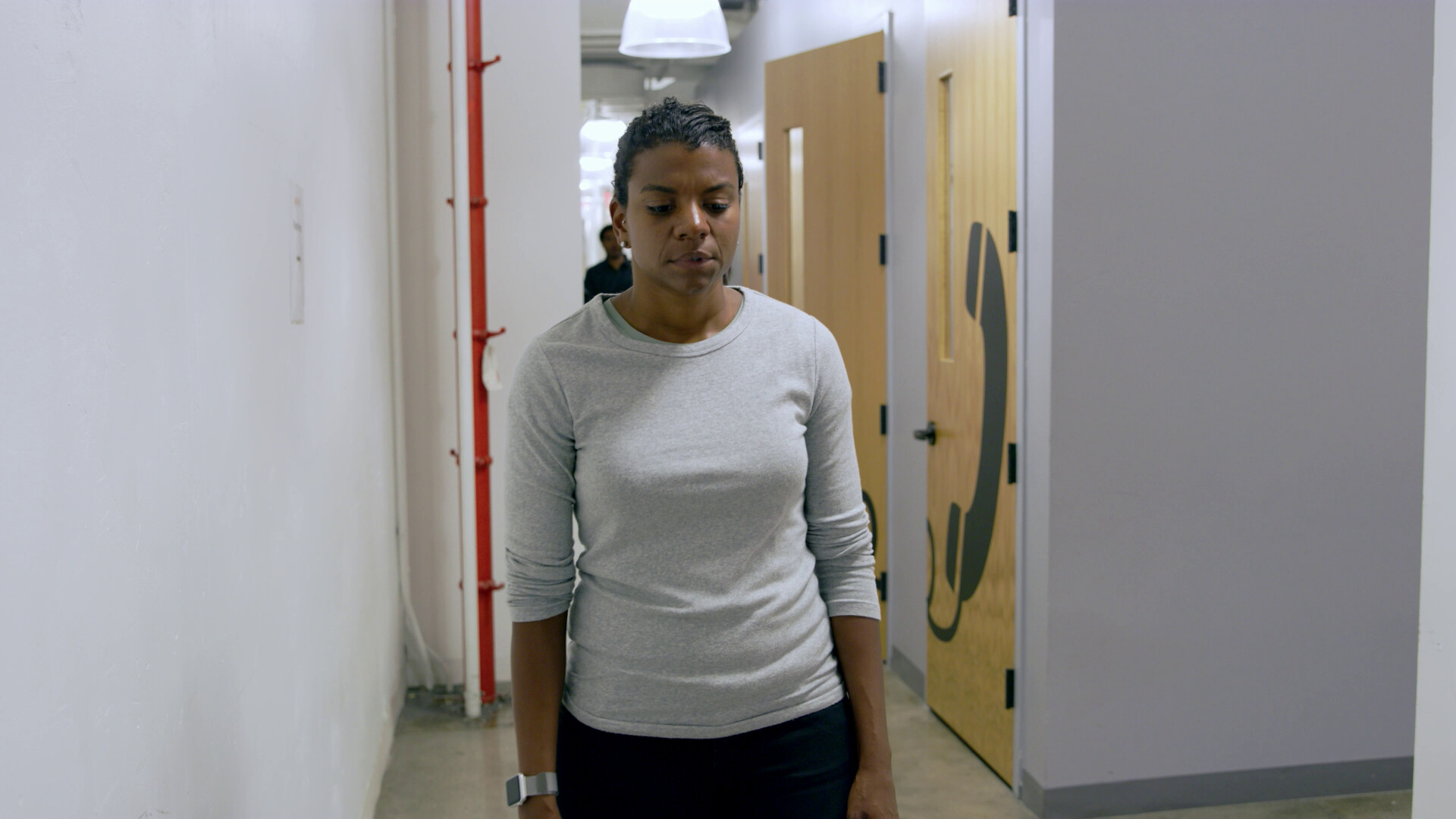} &
        \includegraphics[width=0.195\textwidth]{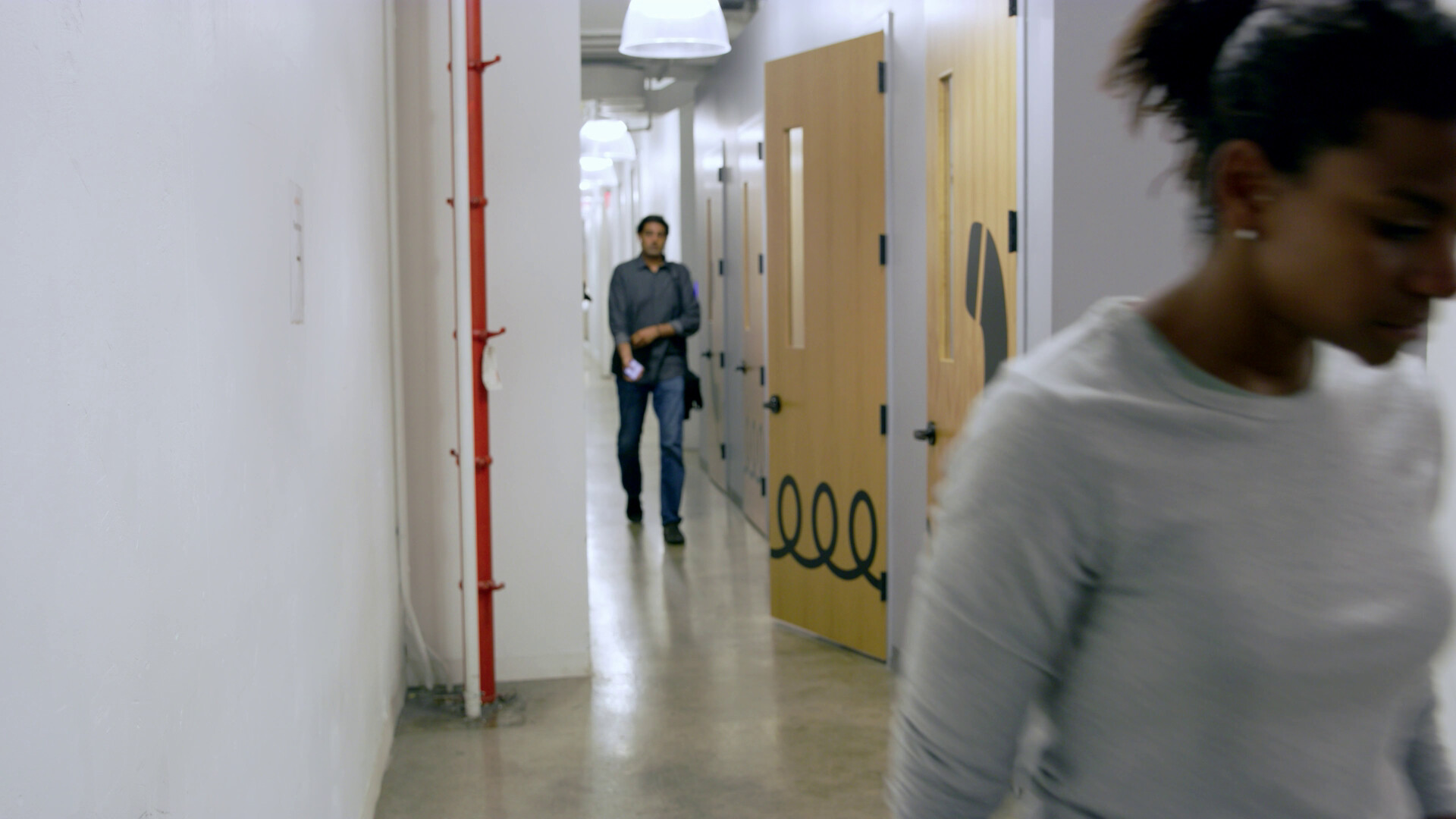} \\

        \rotatebox{90}{\tiny \textbf{\textsc{Wan2.1}}} &
        \includegraphics[width=0.195\textwidth]{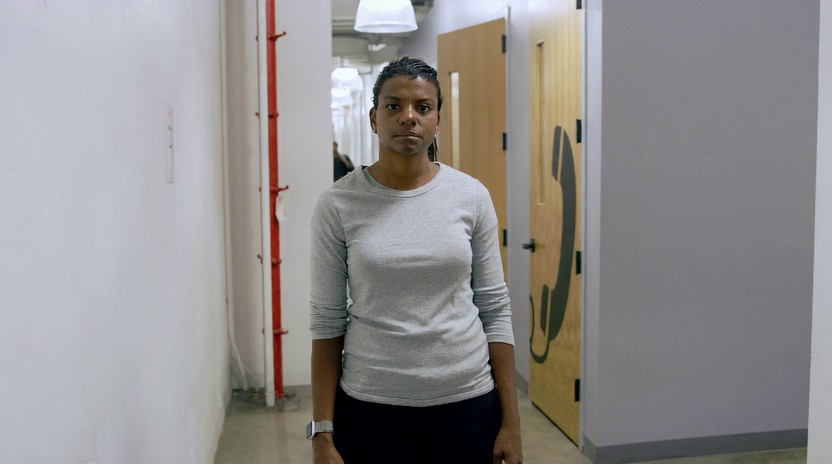} &
        \includegraphics[width=0.195\textwidth]{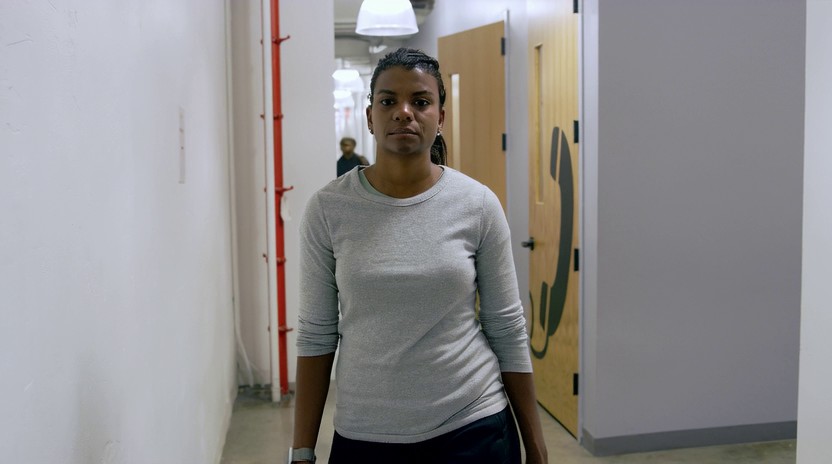} &
        \includegraphics[width=0.195\textwidth]{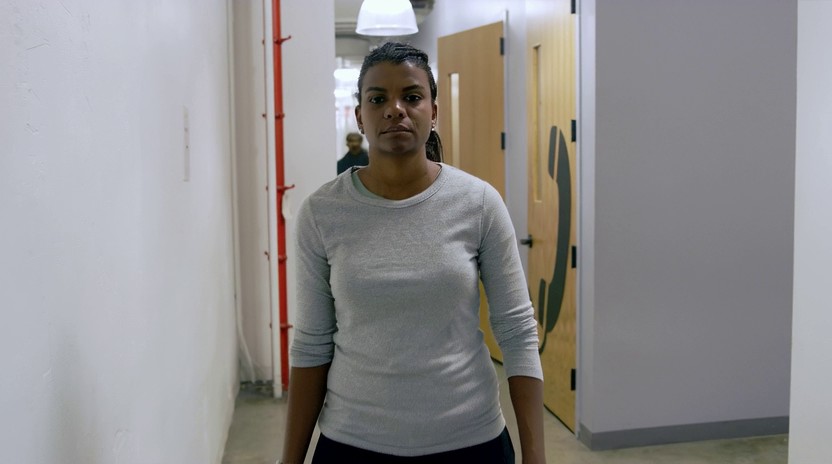} &
        \includegraphics[width=0.195\textwidth]{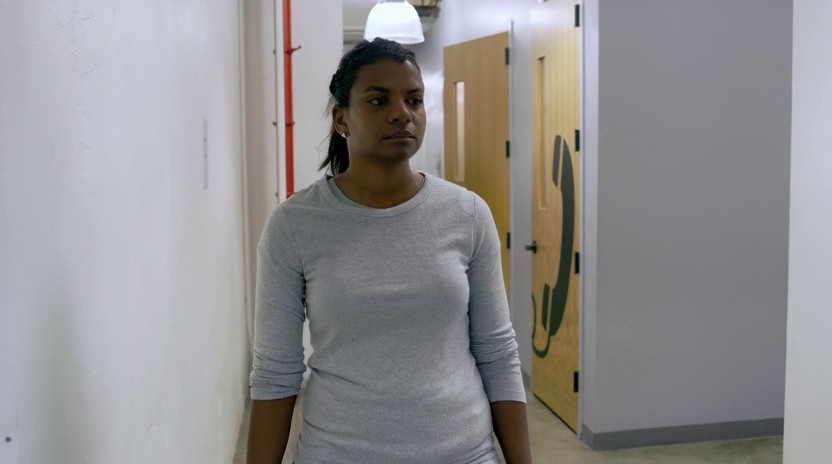} &
        \includegraphics[width=0.195\textwidth]{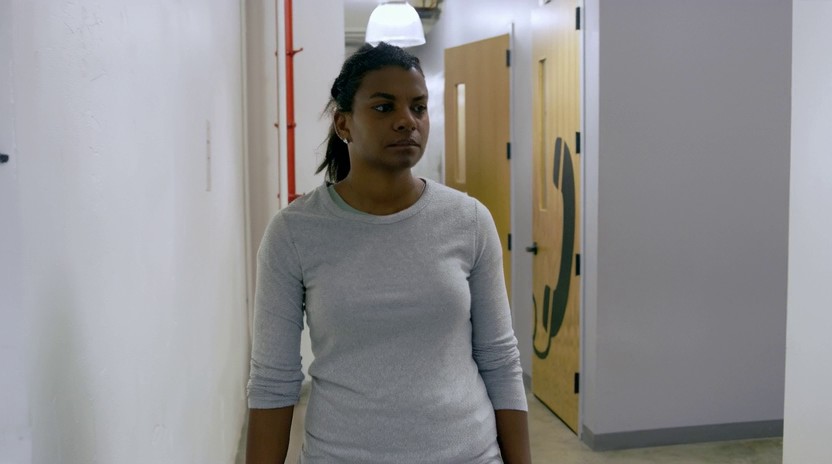} \\

        \rotatebox{90}{\tiny \textbf{\textsc{CogVideoX}}} &
        \includegraphics[width=0.195\textwidth]{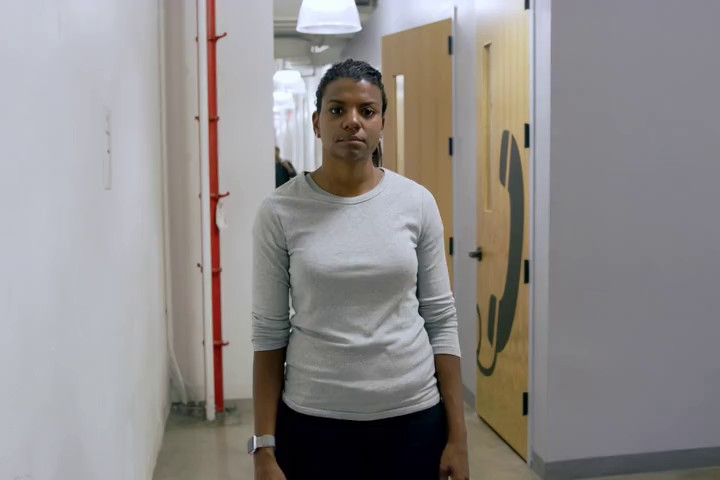} &
        \includegraphics[width=0.195\textwidth]{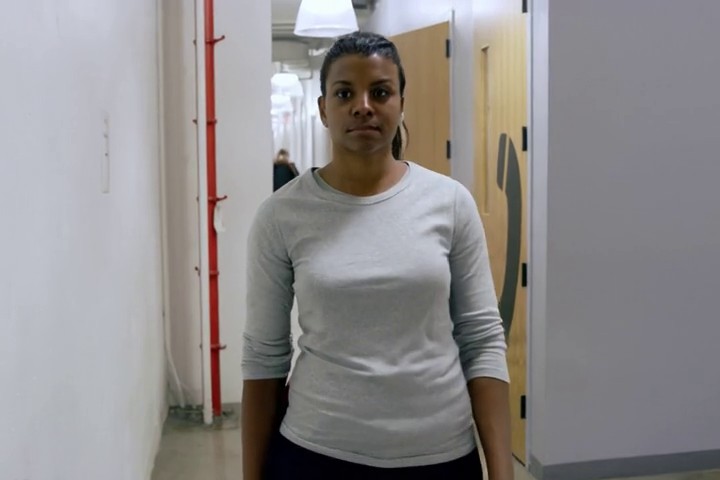} &
        \includegraphics[width=0.195\textwidth]{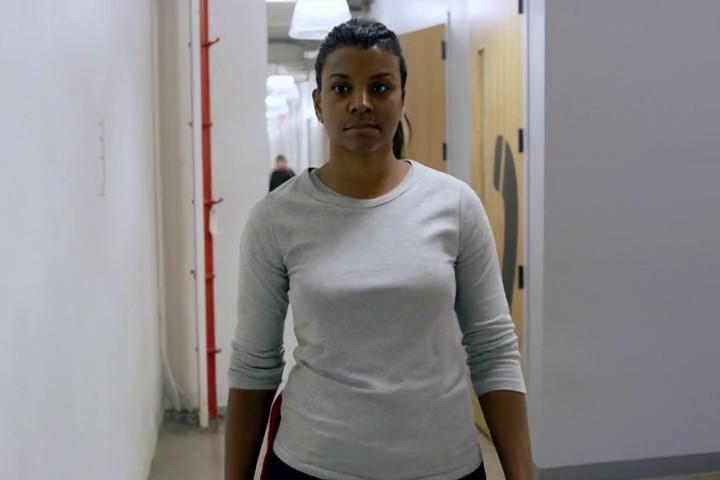} &
        \includegraphics[width=0.195\textwidth]{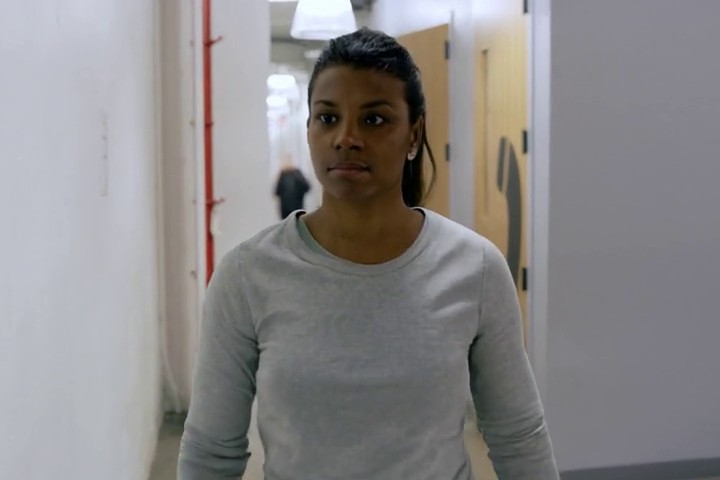} &
        \includegraphics[width=0.195\textwidth]{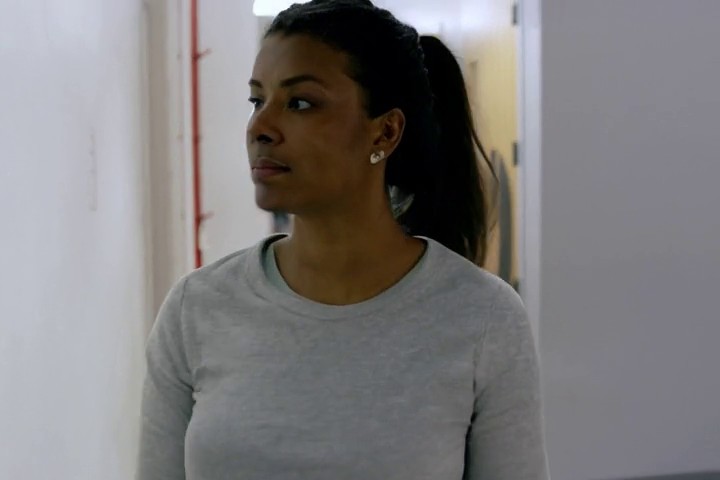} \\

        \rotatebox{90}{\tiny \textbf{\textsc{SkyReels-V2}}} &
        \includegraphics[width=0.195\textwidth]{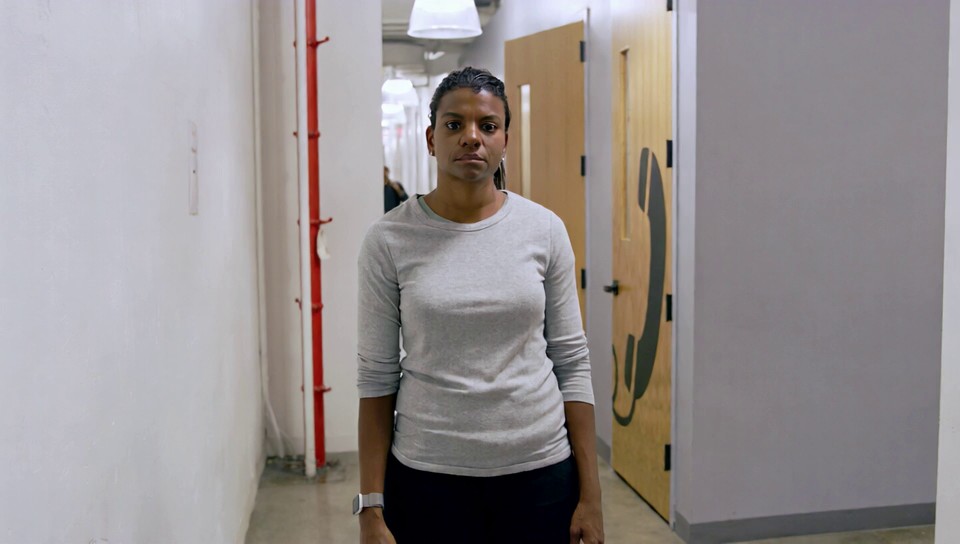} &
        \includegraphics[width=0.195\textwidth]{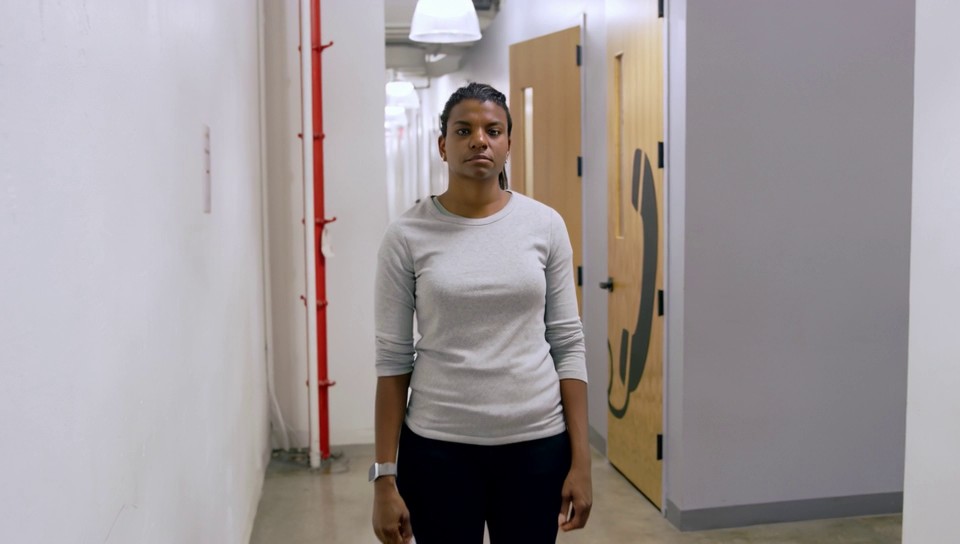} &
        \includegraphics[width=0.195\textwidth]{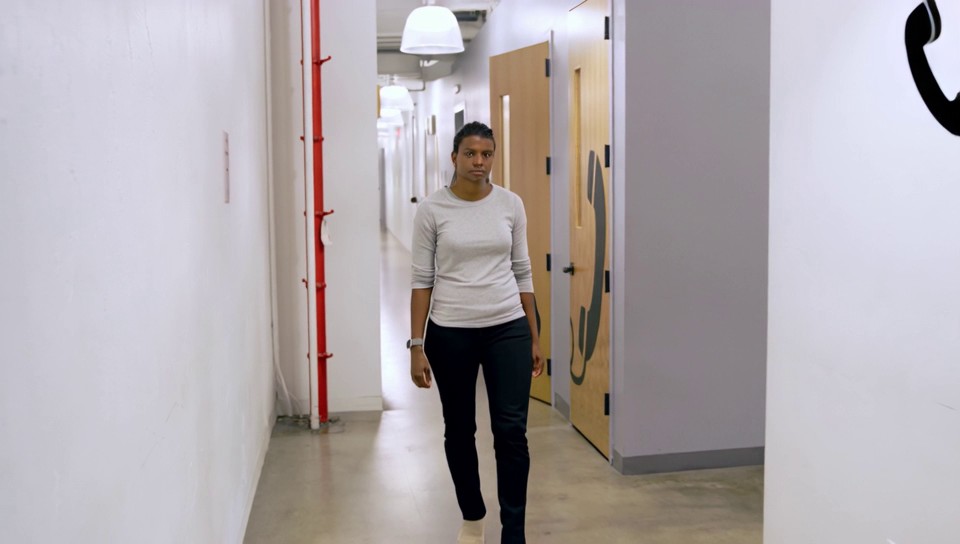} &
        \includegraphics[width=0.195\textwidth]{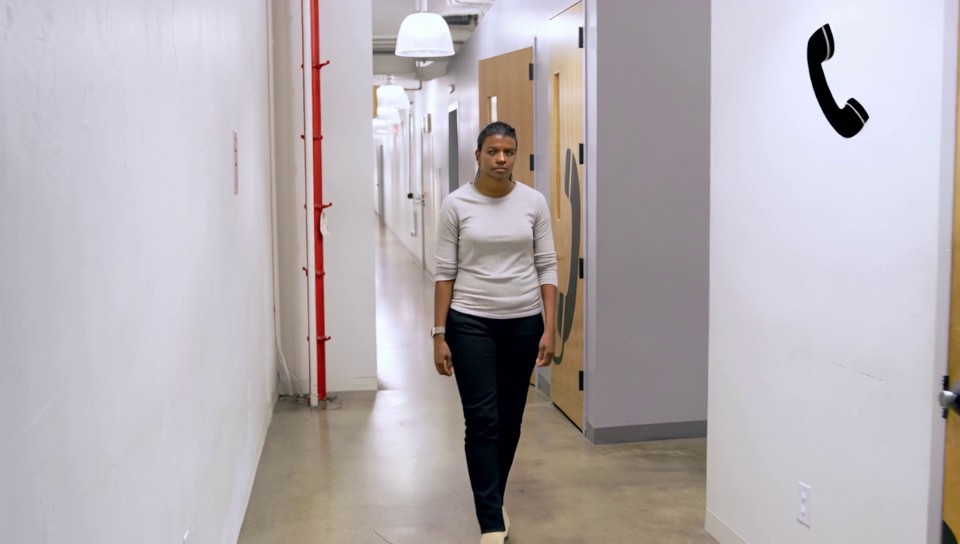} &
        \includegraphics[width=0.195\textwidth]{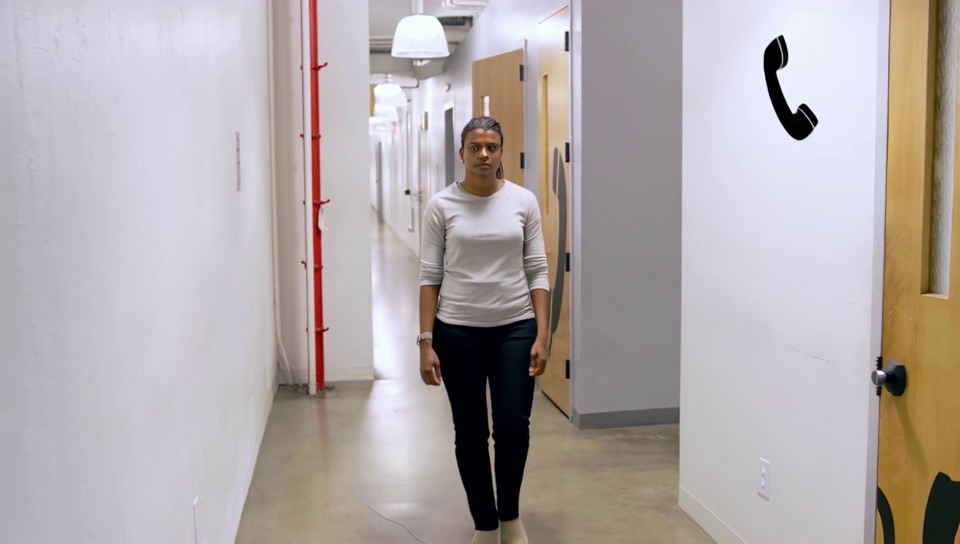} \\
        
        \rotatebox{90}{\tiny \textbf{\textsc{MAGI-1}}} &
        \includegraphics[width=0.195\textwidth]{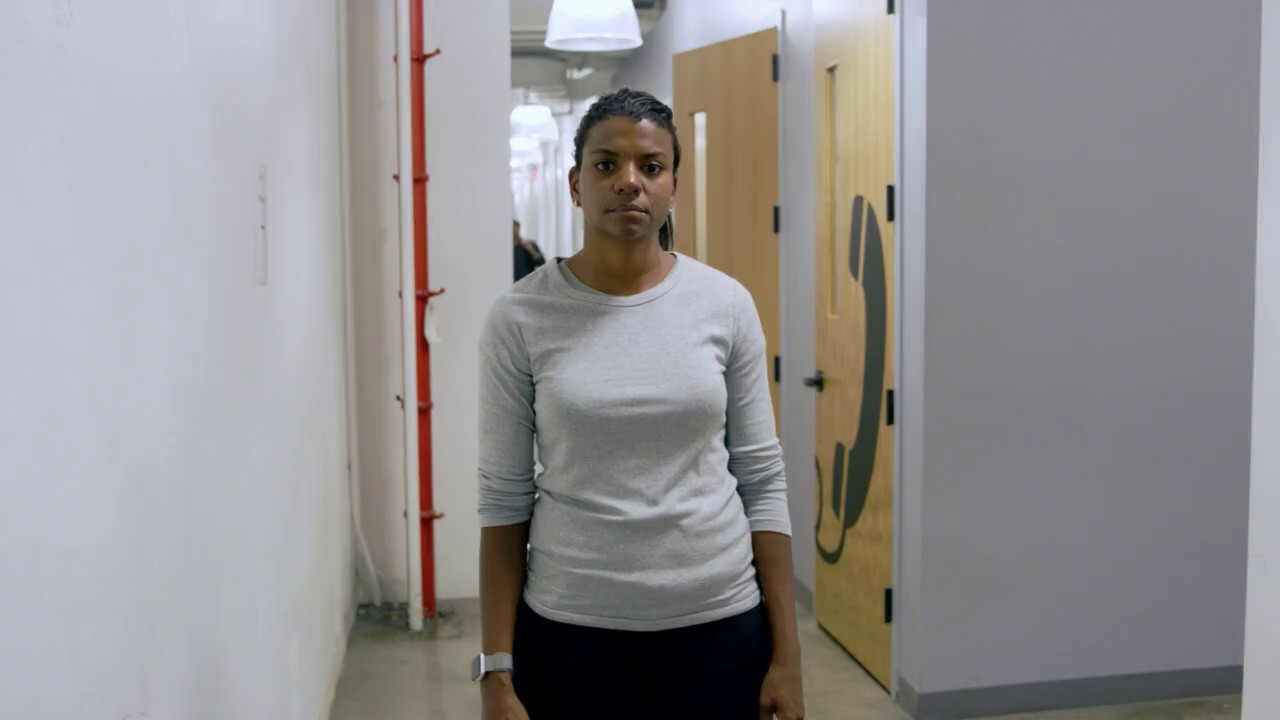} &
        \includegraphics[width=0.195\textwidth]{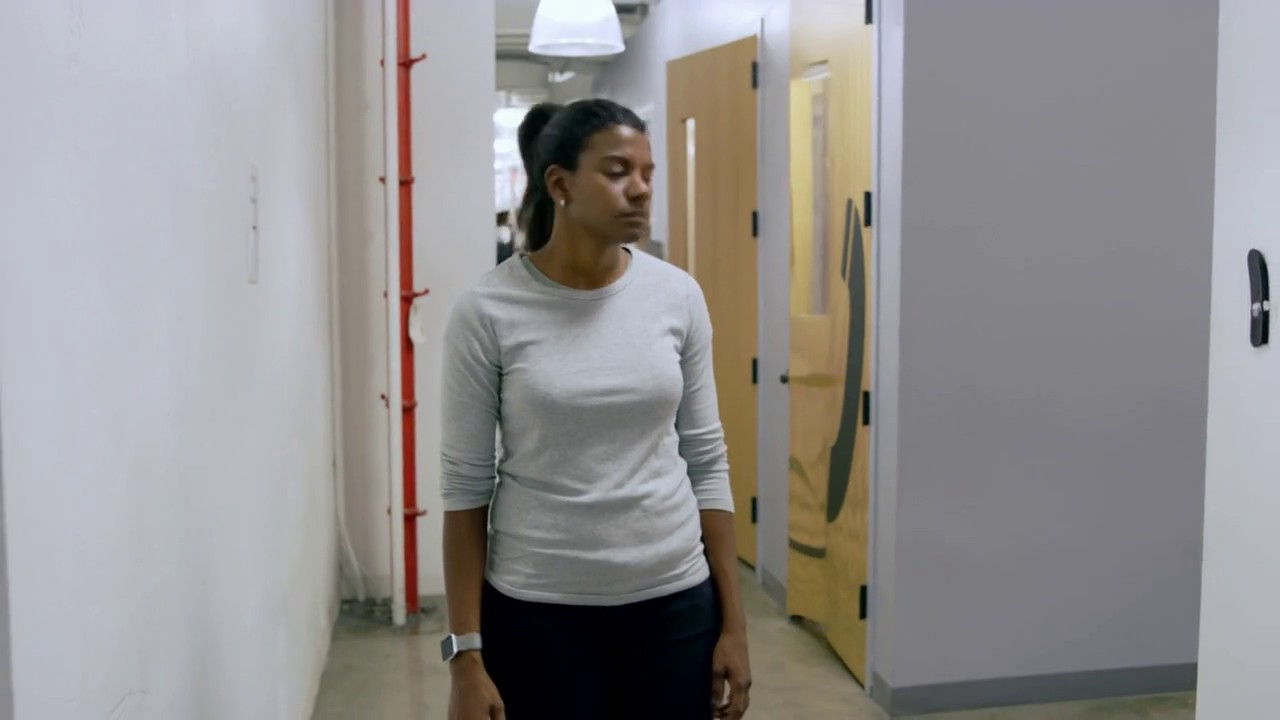} &
        \includegraphics[width=0.195\textwidth]{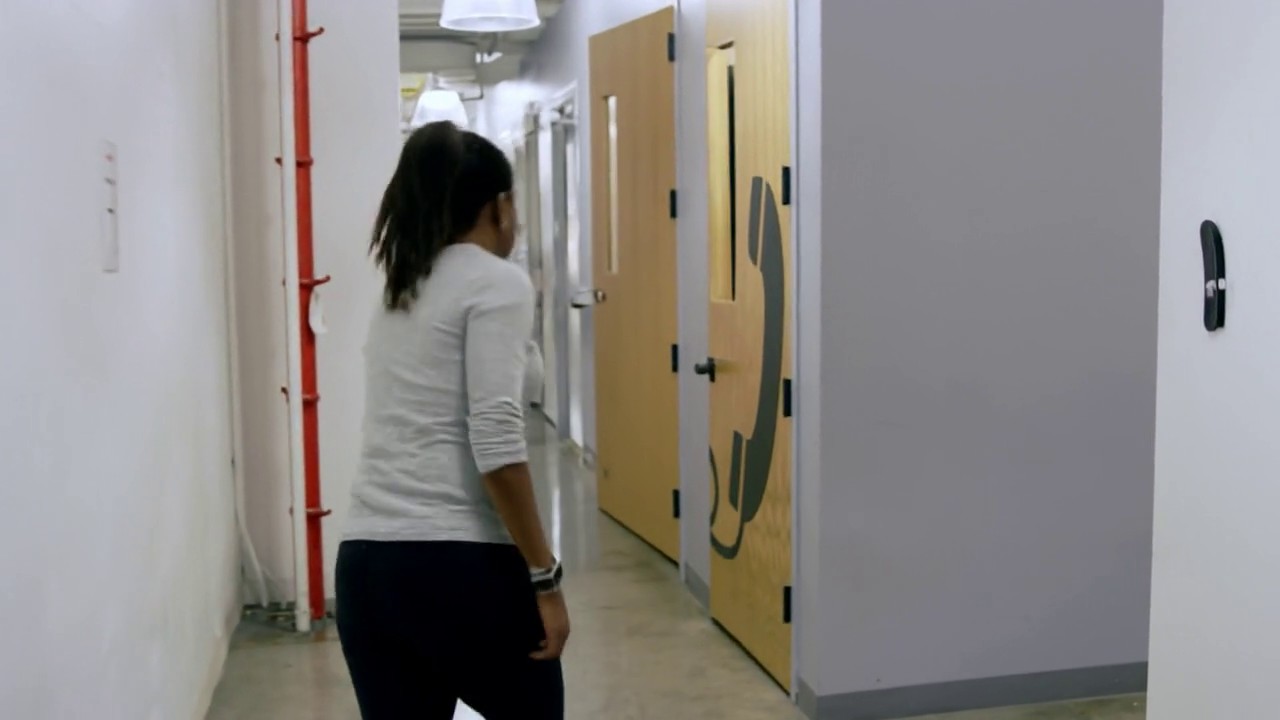} &
        \includegraphics[width=0.195\textwidth]{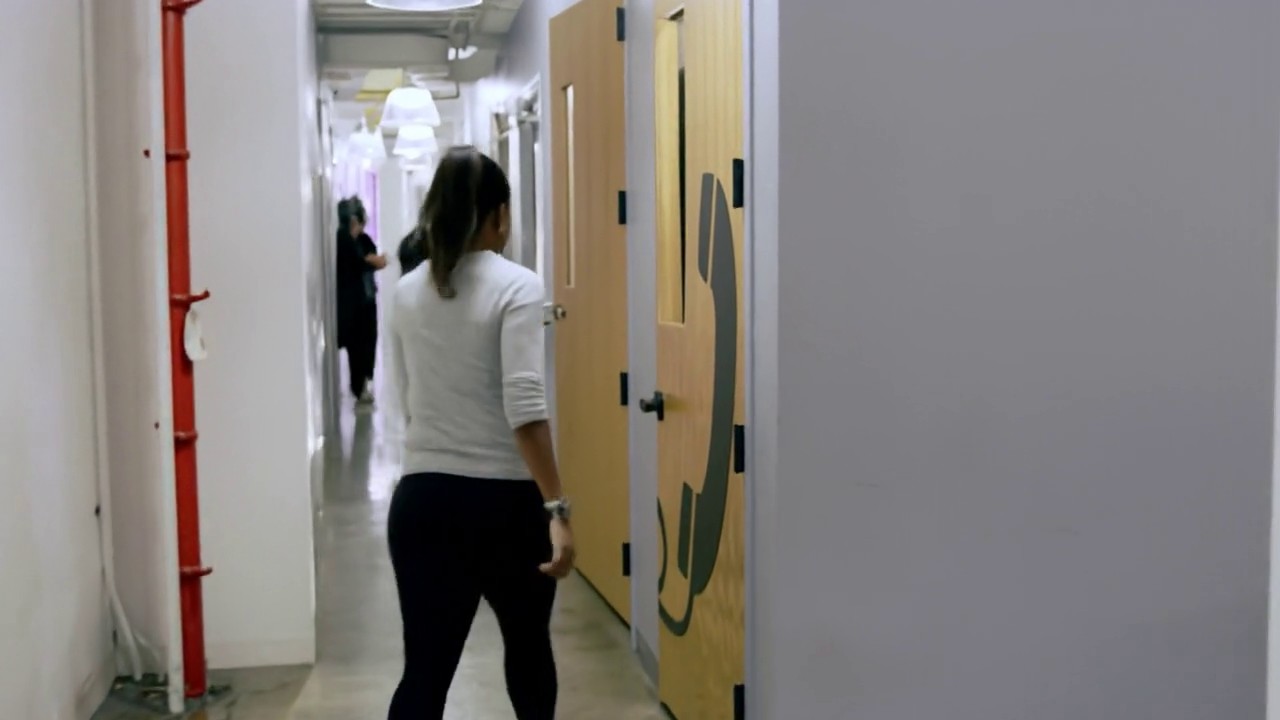} &
        \includegraphics[width=0.195\textwidth]{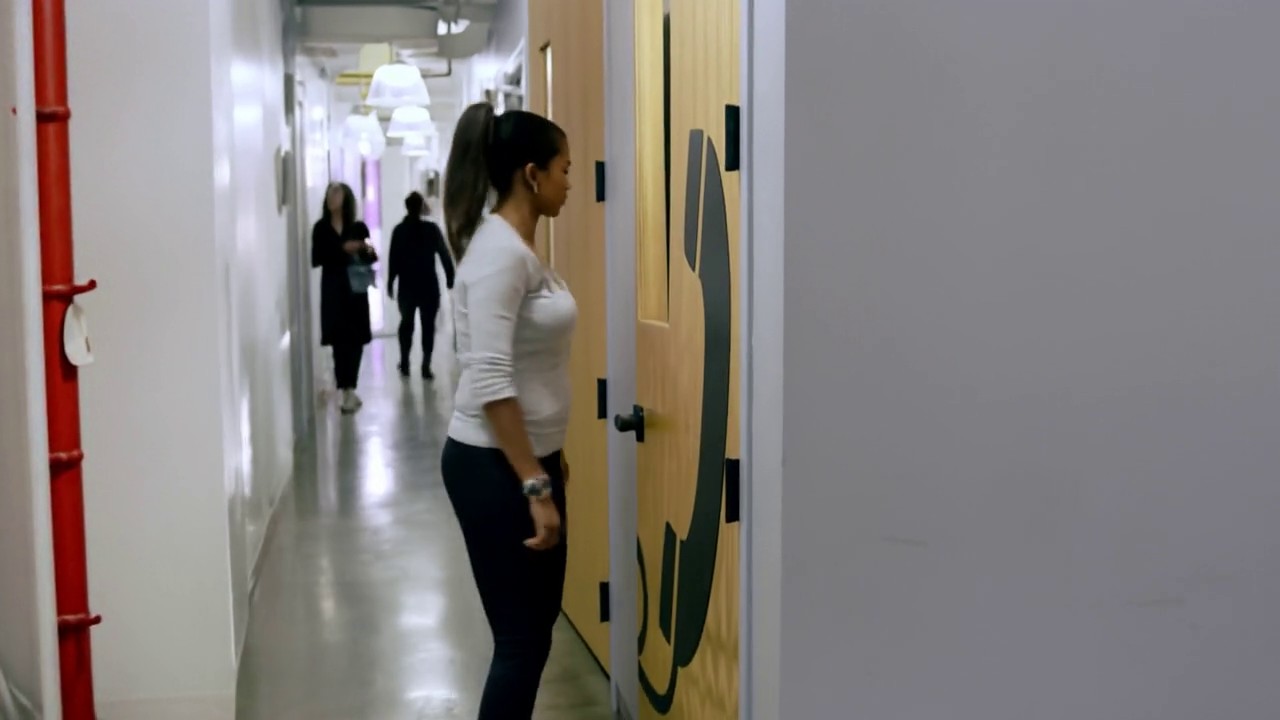} \\
        
        \rotatebox{90}{\tiny \textbf{\textsc{LTX-2.3}}} &
        \includegraphics[width=0.195\textwidth]{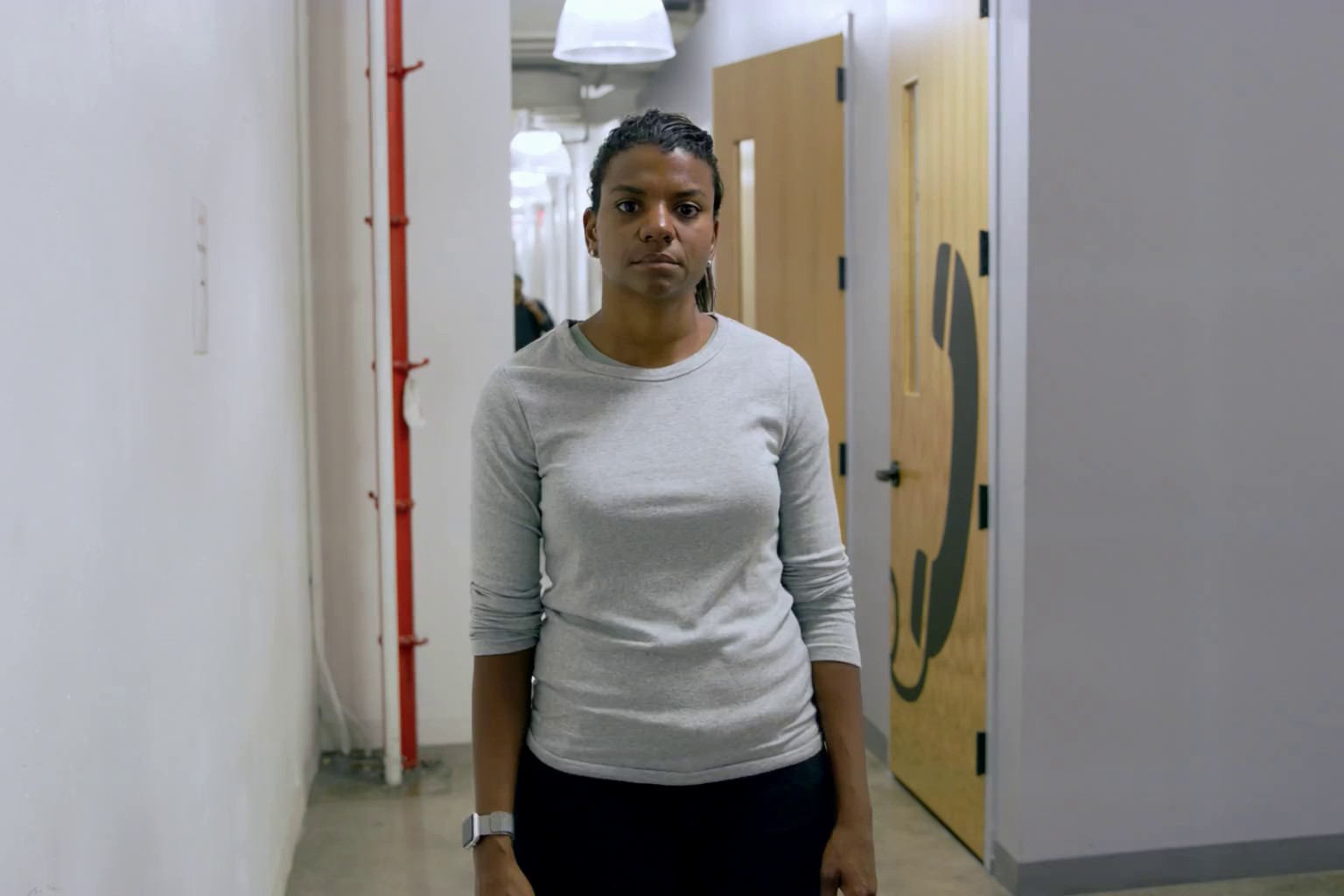} &
        \includegraphics[width=0.195\textwidth]{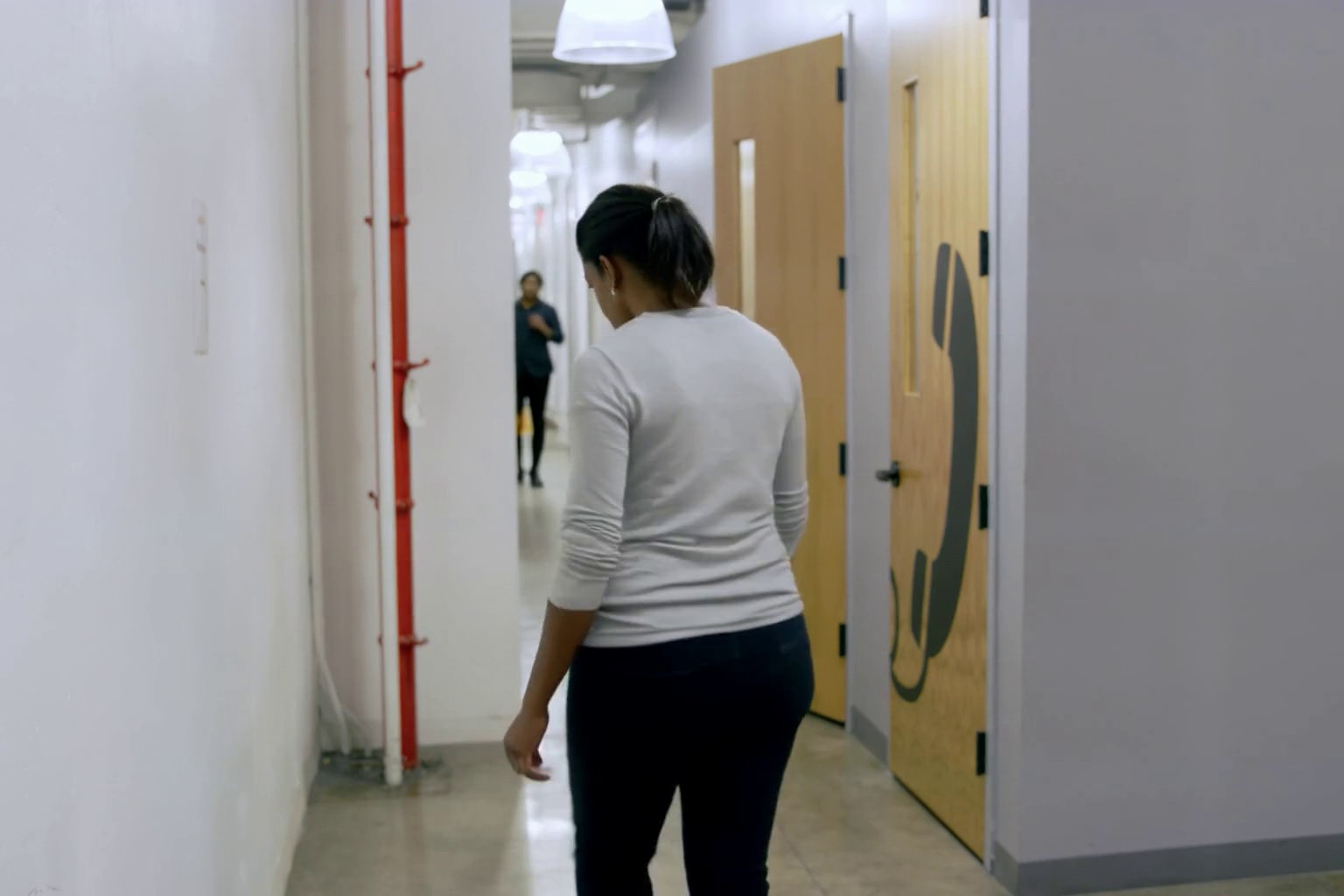} &
        \includegraphics[width=0.195\textwidth]{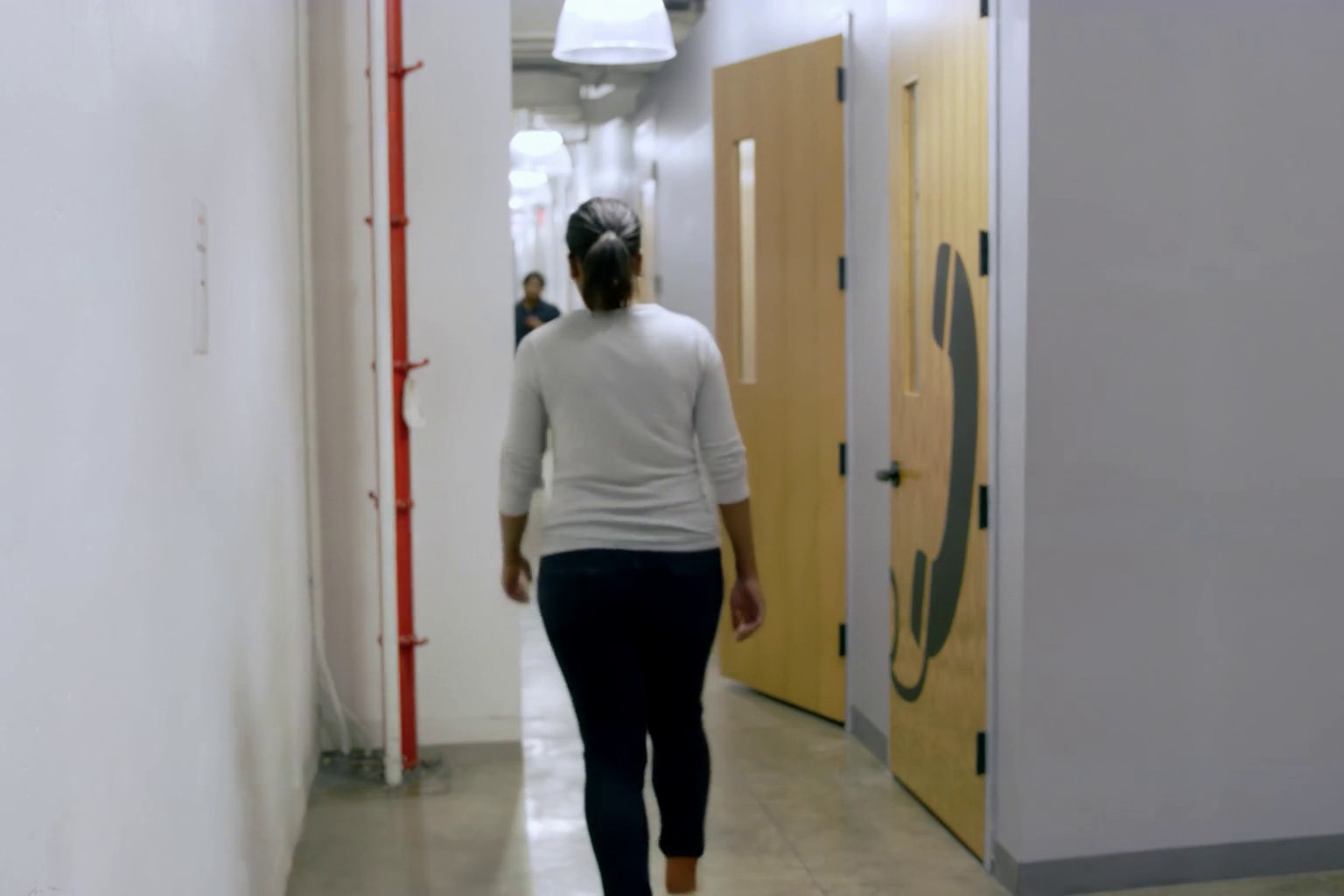} &
        \includegraphics[width=0.195\textwidth]{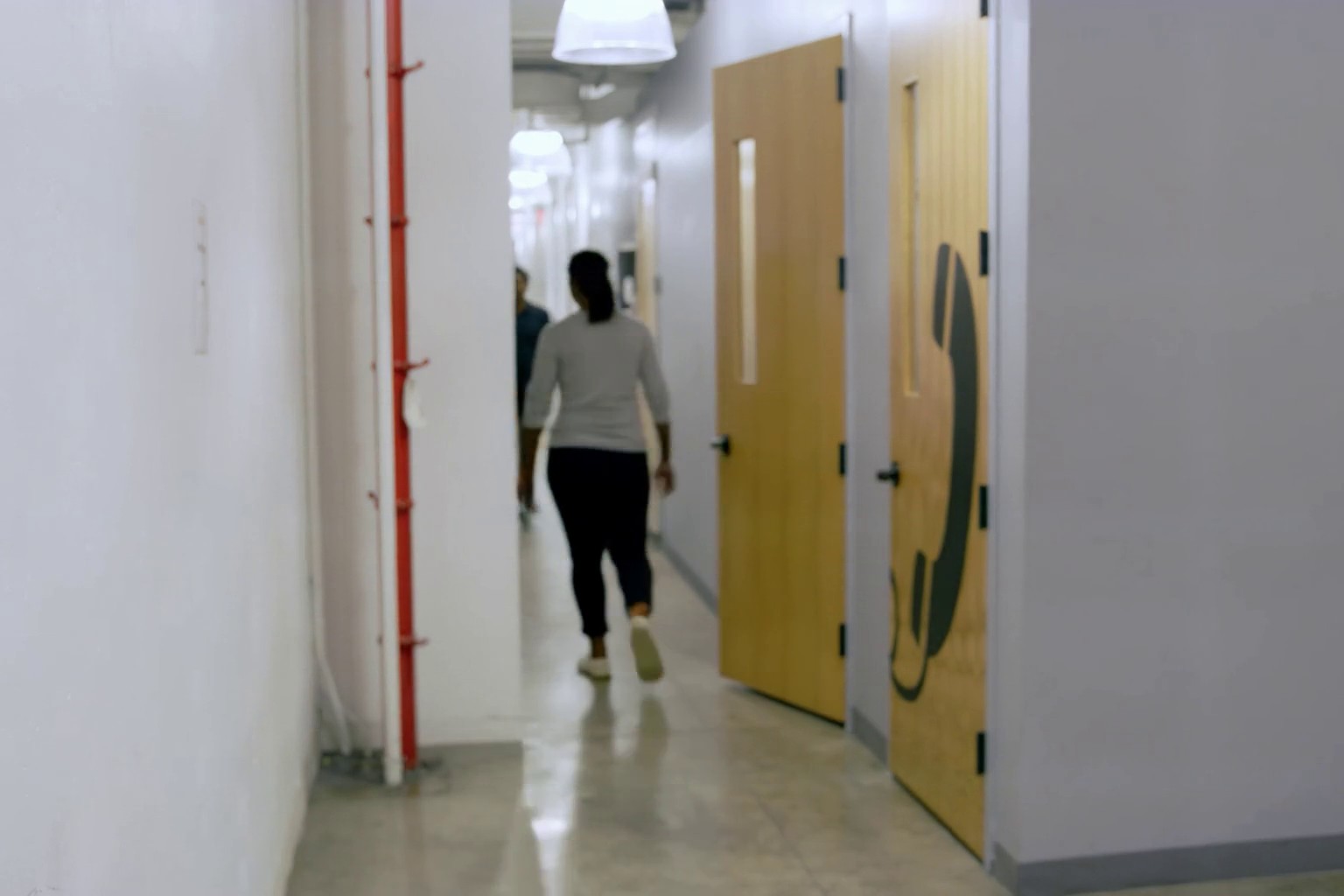} &
        \includegraphics[width=0.195\textwidth]{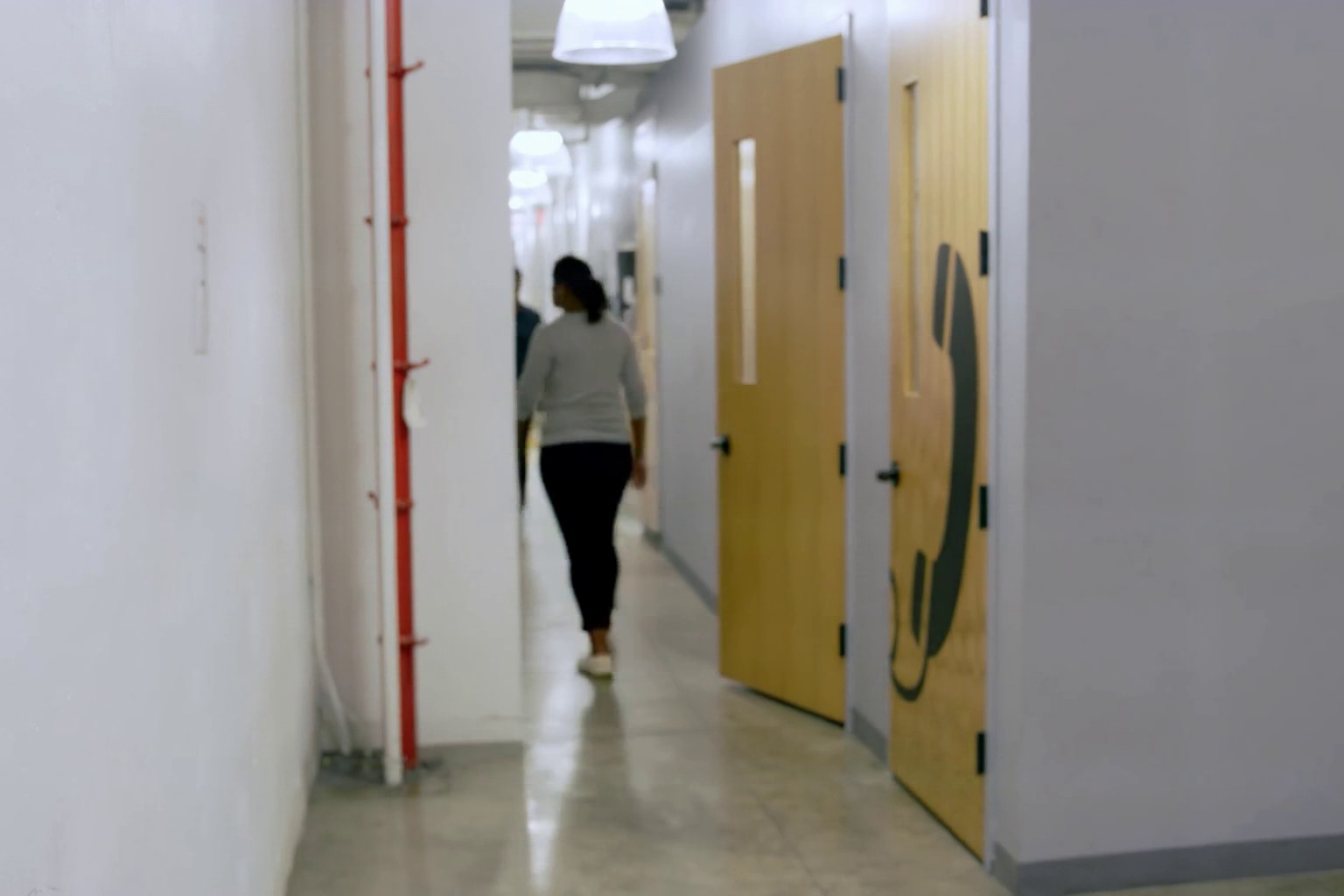} \\
        
        \rotatebox{90}{\tiny \textbf{\textsc{Helios}}} &
        \includegraphics[width=0.195\textwidth]{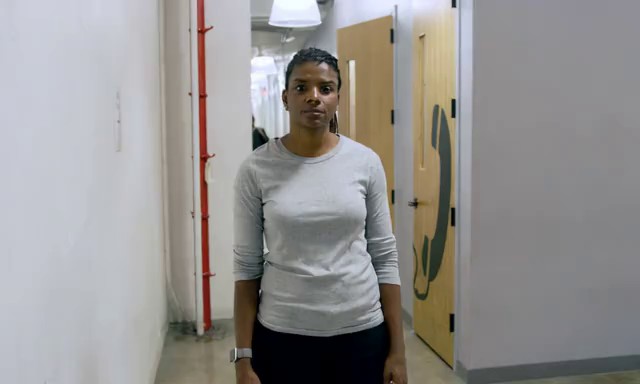} &
        \includegraphics[width=0.195\textwidth]{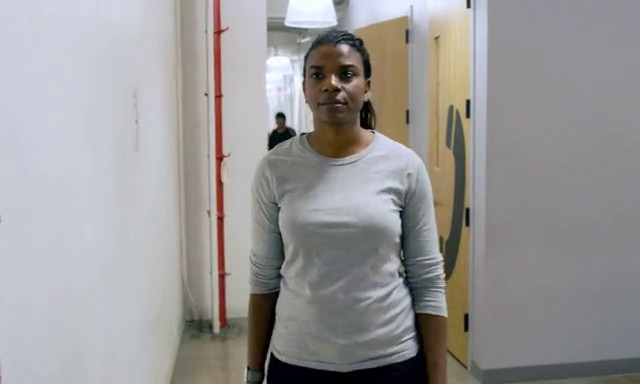} &
        \includegraphics[width=0.195\textwidth]{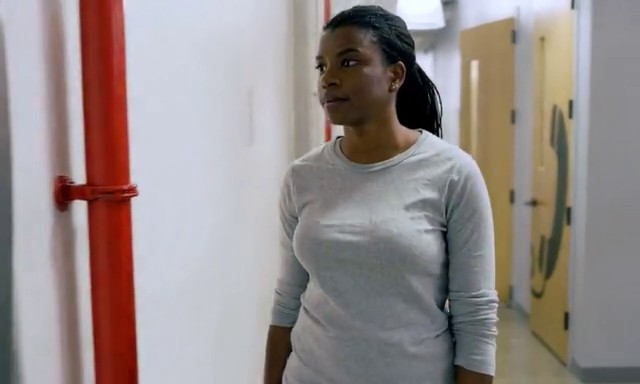} &
        \includegraphics[width=0.195\textwidth]{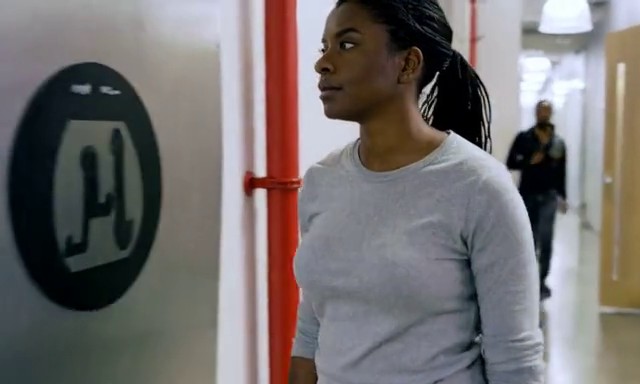} &
        \includegraphics[width=0.195\textwidth]{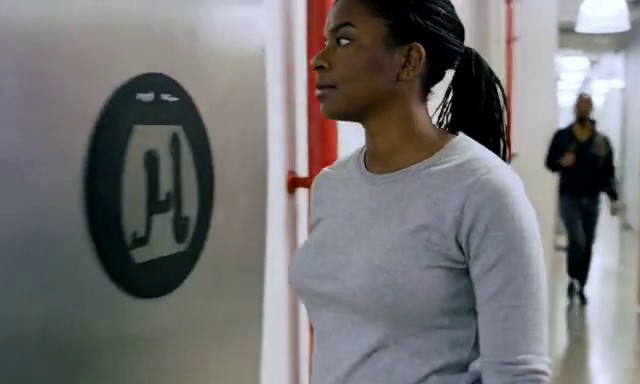} \\
        
        \rotatebox{90}{\tiny \textbf{\textsc{daVinci}}} &
        \includegraphics[width=0.195\textwidth]{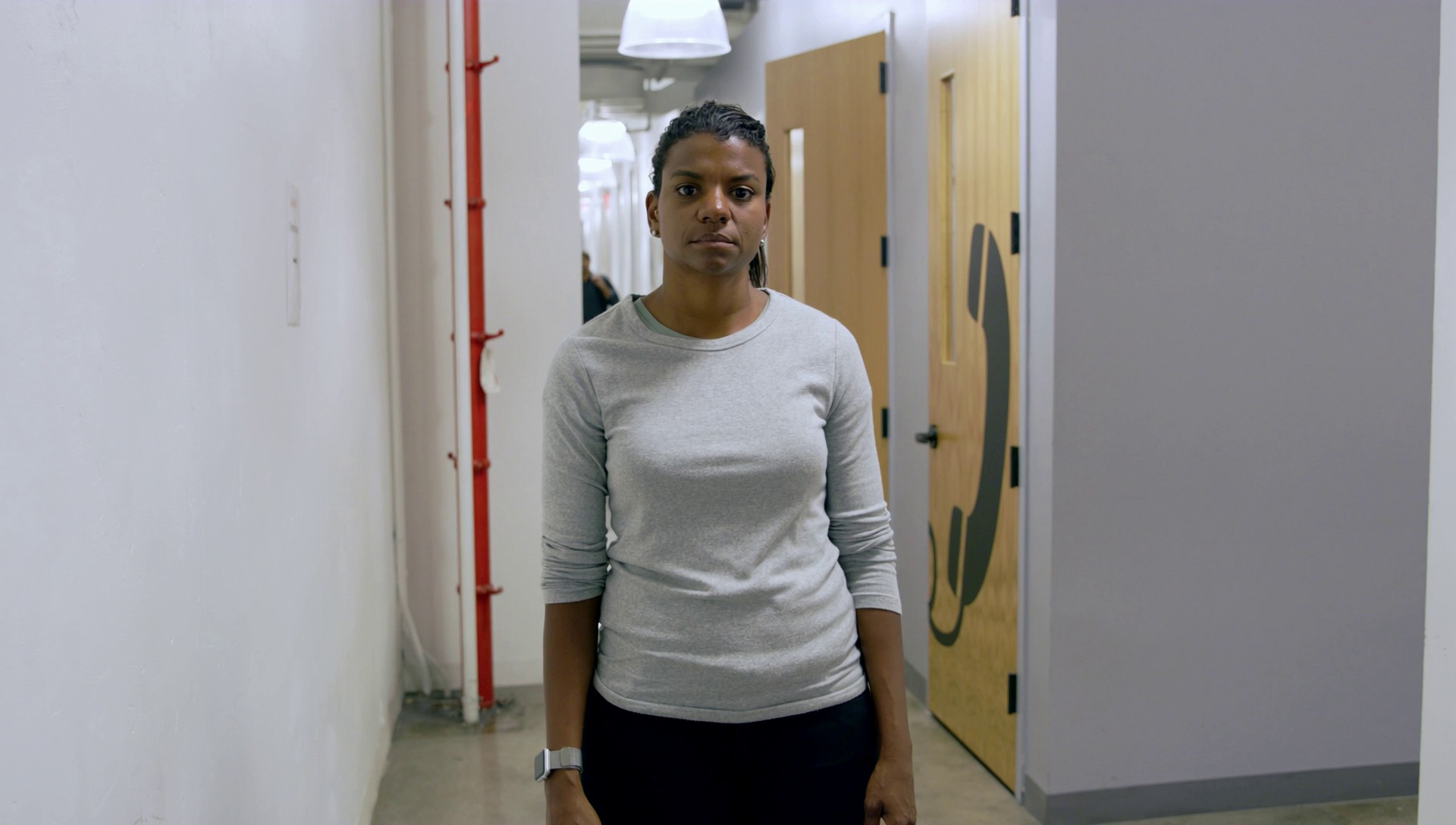} &
        \includegraphics[width=0.195\textwidth]{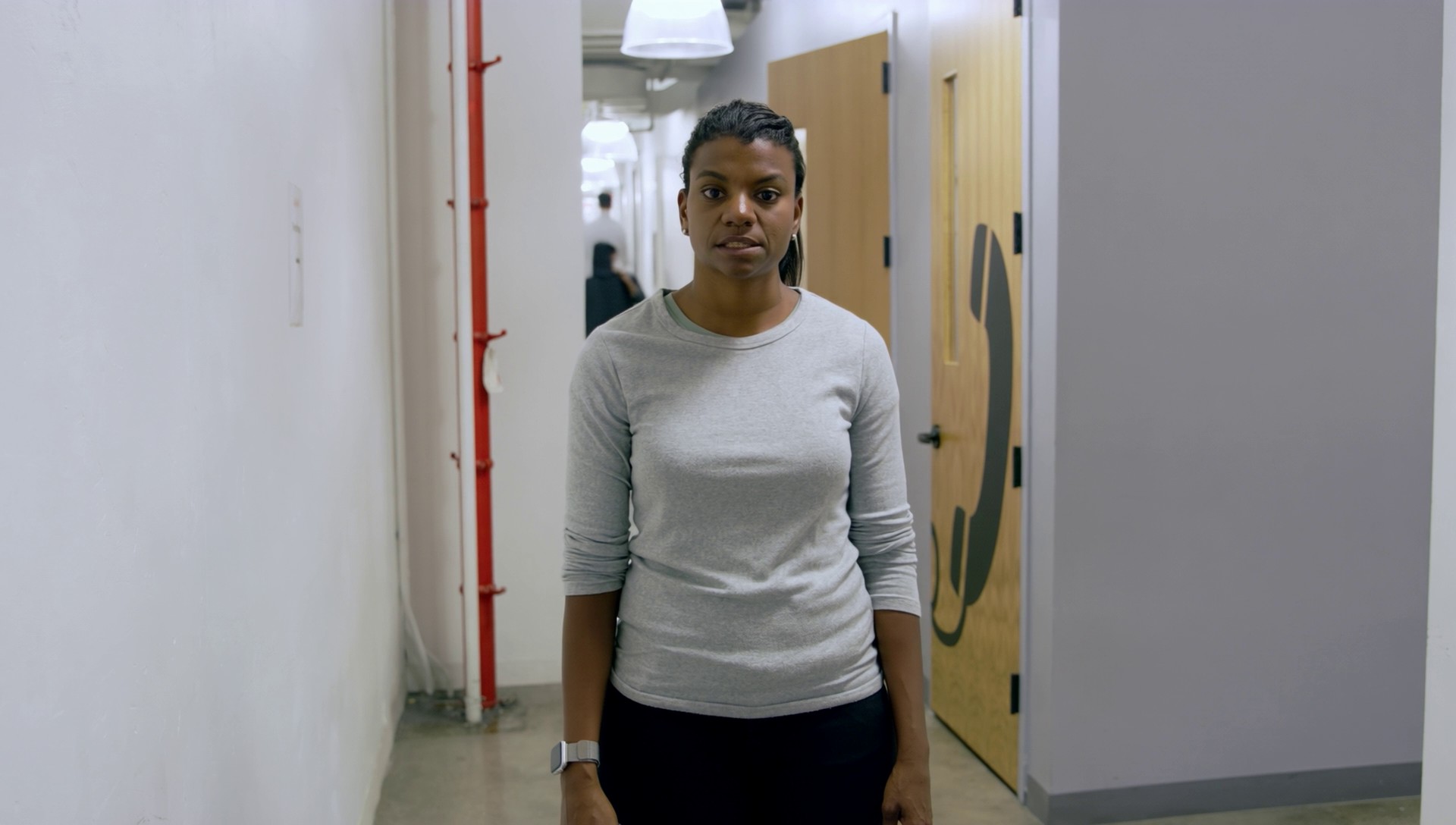} &
        \includegraphics[width=0.195\textwidth]{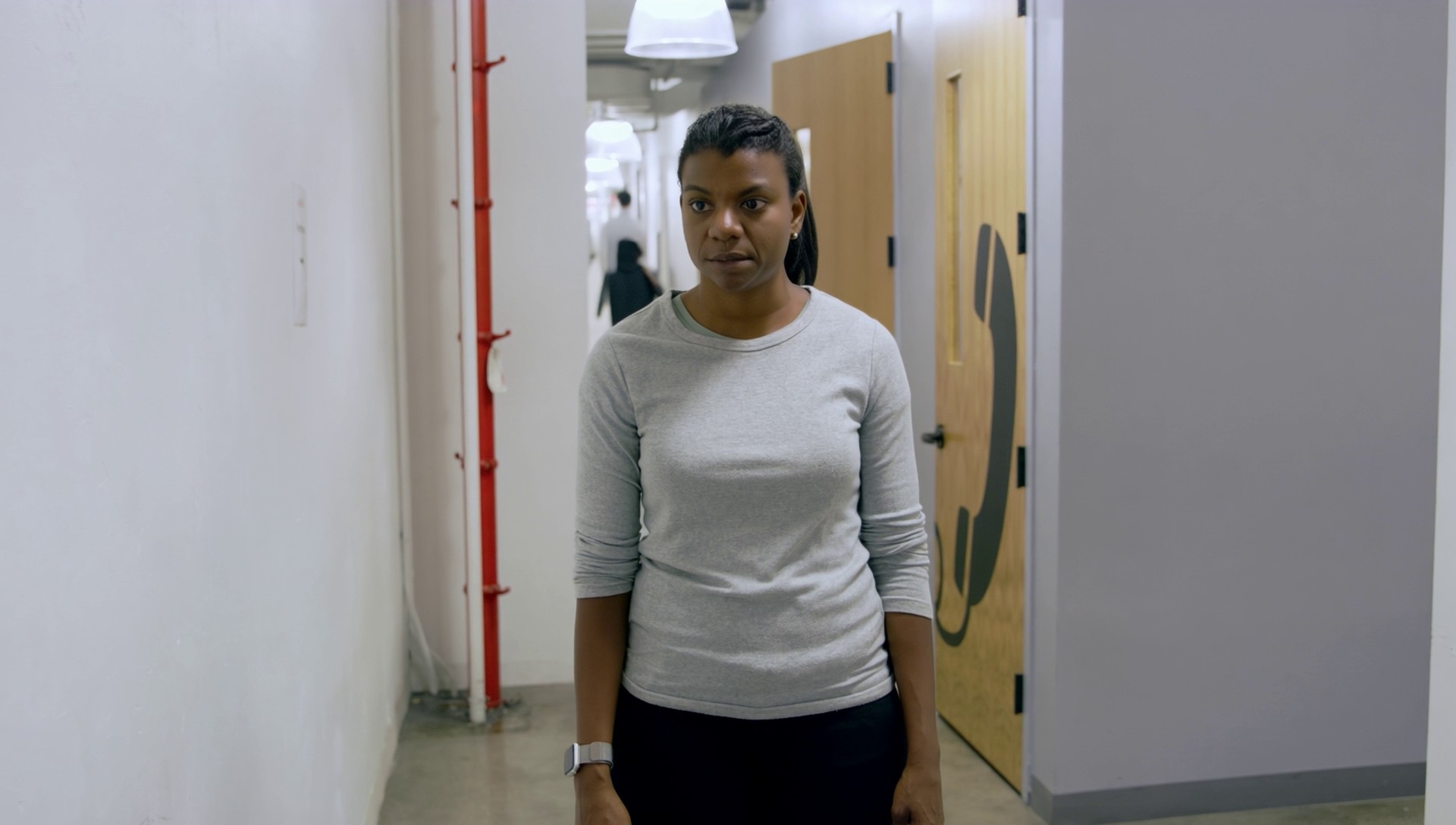} &
        \includegraphics[width=0.195\textwidth]{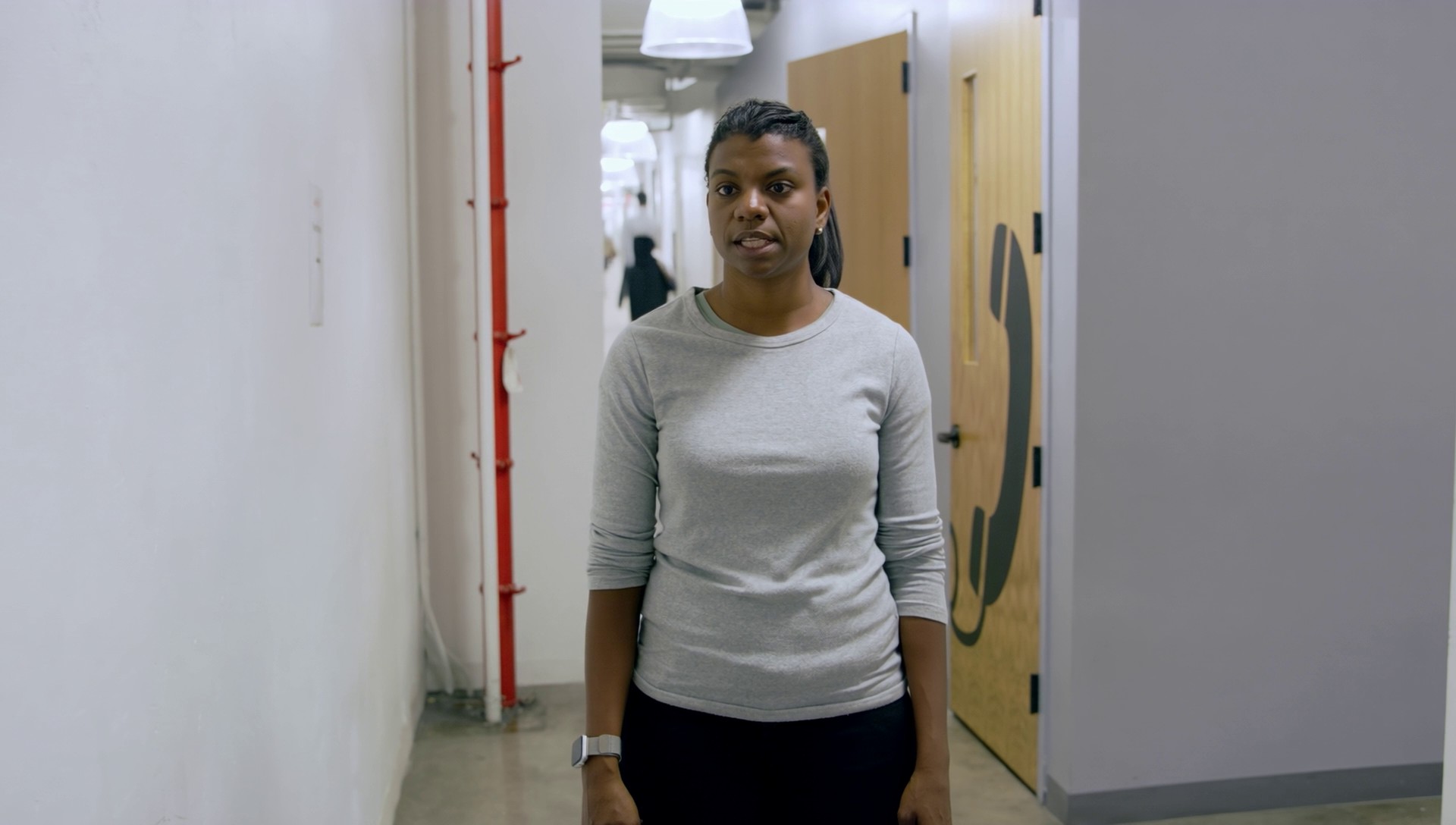} &
        \includegraphics[width=0.195\textwidth]{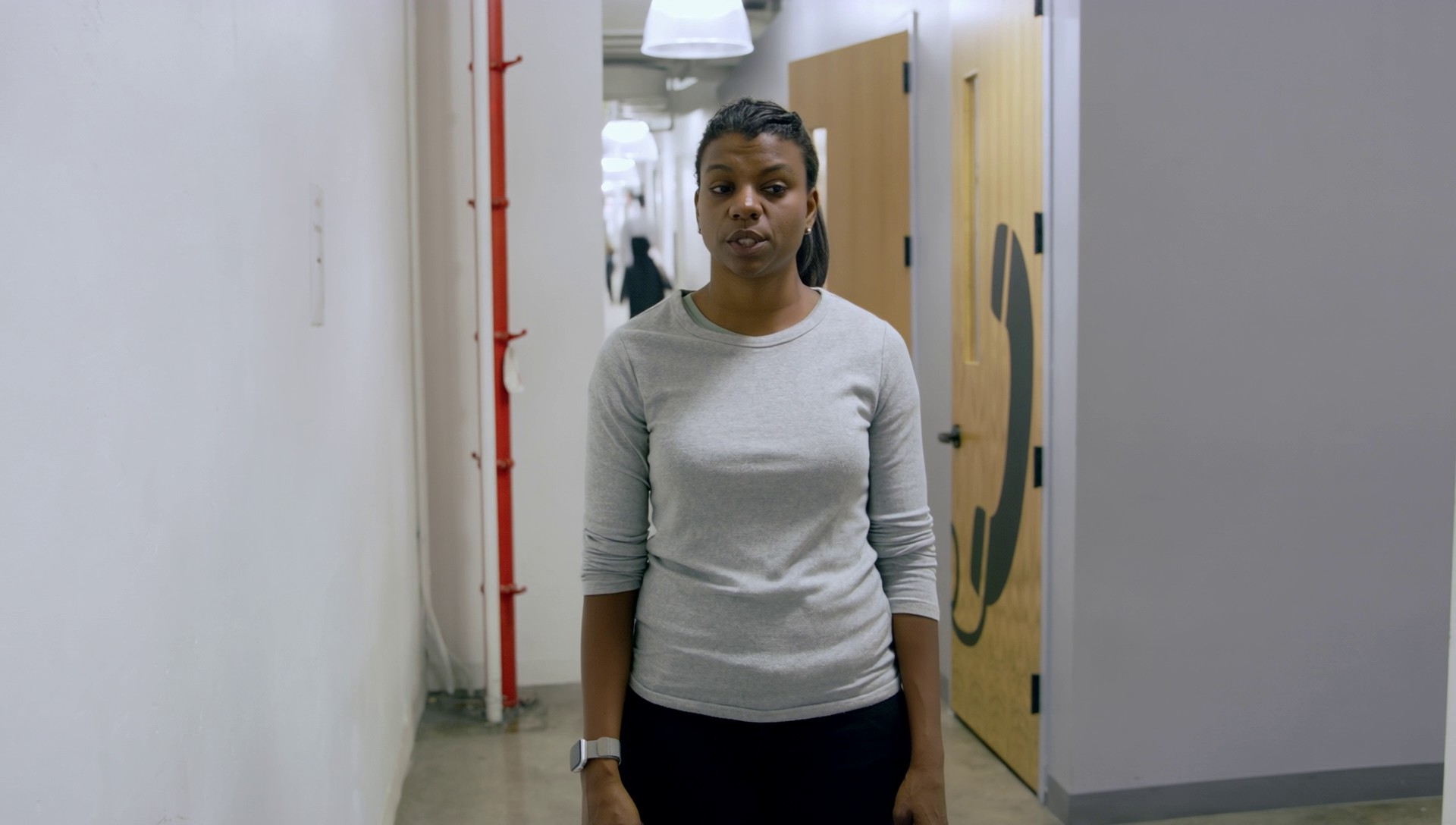} \\
    \end{tabular}
    \caption{Qualitative I2V comparison on DFD pristine video \texttt{26\_\_exit\_phone\_room}. Frame sequences ($t=0$ to $t=4$) comparing the source video (top row) with synthetic outputs from the seven I2V generators. Conditioned on the human-selected reference frame (Figure~\ref{fig:i2v_frame_selection}), all I2V generators preserve subject identity and background composition while extending the scene with temporally coherent motion.}
    \label{fig:qualitative_26_dfd_i2v_v2}
\end{figure}

Figure~\ref{fig:qualitative_26_dfd_i2v_v2} reports the corresponding I2V comparison on DFD pristine video \texttt{26\_\_exit\_phone\_room}, a corridor walking shot with a single subject framed against a service-area background containing wall fixtures, fluorescent ceiling lights, and the eponymous phone-icon door signage. Because every I2V output is anchored to the same human-selected reference frame (Figure~\ref{fig:i2v_frame_selection}), all seven generators preserve the subject's identity, garment, and ambient lighting almost without deviation; the visible variability lies in how each model extends motion beyond the conditioning frame. Wan2.1 stays closest to the pristine walk-toward-camera trajectory; SkyReels-V2 has the subject move away from the camera; LTX-2.3 has the subject walk down the corridor away from the viewer; daVinci-MagiHuman tightens the framing into a medium close-up; CogVideoX, MAGI-1, and Helios reorient the subject sideways towards the door with varying degrees of motion. Across all seven, ambient elements (red fire pipe, wooden doors, phone signage) remain spatially coherent, with no texture boiling or geometry drift over the five sampled timestamps.

\clearpage
\clearpage

\clearpage

\clearpage


\clearpage
\clearpage
\section{Comparative Analysis: Extended Details}
\label{app:comparative}

This appendix complements Section~\ref{sec:comparative}, which characterizes SynthForensics relative to nine publicly available synthetic-video benchmarks through landmark stability, face quality, and a paired-comparison human study, with the preprocessing pipeline and face-positive subset construction (Section~\ref{app:comparative-preprocessing}), the per-benchmark and per-generator breakdowns of the landmark and face-quality metrics (Sections~\ref{app:comparative-landmark} and~\ref{app:comparative-face-quality}), and the full protocol, screening procedure, statistical methodology, per-benchmark results, and inter-rater agreement of the human study (Section~\ref{app:comparative-human-study}).

\subsection{Preprocessing for Face-Level Analyses}
\label{app:comparative-preprocessing}

Starting from the nine benchmarks retained in Section~\ref{sec:comparative}, we downloaded from each benchmark the full synthetic-video corpus released by its authors; the present analysis is restricted to fake content, with real videos used elsewhere as baselines (FF++ and DFD) rather than as benchmark targets. Where a benchmark releases multiple compression versions (e.g., GVD with several H.264 CRF levels), we use the highest-quality (least-compressed) version; for SynthForensics we use the \textit{Raw} version (uncompressed generator output). This ensures the comparison reflects intrinsic generator quality rather than compression artifacts. For most benchmarks the released set matches the declared corpus, with three exceptions. GenVidBench distributes 142{,}968 of its 6{,}353{,}071 declared synthetic videos~\citep{ni2026genvidbench}, with the remaining ones inherited from the upstream VidProM dataset~\citep{wang2024vidprom}. AEGIS releases the 436-video Hard Test split (218 synthetic, 218 real) of its 5{,}199 declared synthetic videos~\citep{li2025aegis}, with most of the remaining content drawn from the upstream TIP-I2V dataset~\citep{wang2025tipi2v}. DeepAction releases 2{,}600 of its 3{,}100 declared synthetic videos~\citep{bohacek2025deepaction}. We use the released subsets in all three cases. Within each downloaded set we then remove byte-identical duplicates via MD5 hashing; the per-benchmark counts so identified are negligible relative to the corresponding totals. Table~\ref{tab:preprocessing} reports declared, available, and analyzed counts side by side.

\begin{table}[!h]
\centering
\caption{Per-benchmark video counts: declared in the original publications, available for download, MD5 duplicates, and face-positive cardinalities under FAN~\citep{bulat2017fan} for landmark stability (Section~\ref{subsec:landmark_stability}) and TOPIQ-NR-Face~\citep{chen2024topiq} for face quality (Section~\ref{subsec:face_quality}).}
\label{tab:preprocessing}
\small
\setlength{\tabcolsep}{4pt}
\begin{tabular}{lrrrrr}
\toprule
 & & & & \multicolumn{2}{c}{Face-positive subset} \\
\cmidrule(lr){5-6}
Benchmark & Declared & Available & Duplicates & Landmark stability & Face quality \\
\midrule
GenVideo            & 1{,}078{,}838 & 1{,}078{,}838 & 68  & 264{,}849         & 245{,}431         \\
AIGVDBench          & 422{,}000    & 422{,}000    & 189  & 175{,}961         & 171{,}961         \\
GenVidBench         & 6{,}353{,}071 & 142{,}968    & 0    & 10{,}350          & 9{,}998           \\
GVD                 & 11{,}618     & 11{,}618     & 15   & 2{,}501           & 2{,}313           \\
DVF                 & 3{,}938      & 3{,}938      & 1    & 1{,}198           & 1{,}129           \\
DeepAction          & 3{,}100      & 2{,}600      & 0    & 1{,}063           & 1{,}004           \\
GVF                 & 964          & 964          & 3    & 863               & 798               \\
LOKI                & 593          & 593          & 3    & 67                & 64                \\
AEGIS               & 5{,}199      & 218          & 0    & 67                & 57                \\
\midrule
SynthForensics      & 20{,}445     & 20{,}445     & 0    & 20{,}356          & 20{,}216          \\
\midrule
FF++ (real)         & 1{,}000      & 1{,}000      & 0    & 1{,}000           & 1{,}000           \\
DFD (real)          & 363          & 363          & 0    & 363               & 363               \\
\bottomrule
\end{tabular}
\end{table}

Each face-level analysis pipeline (Sections~\ref{subsec:landmark_stability} and~\ref{subsec:face_quality}) is executed on the full deduplicated set of videos for every benchmark; the per-benchmark aggregate, however, is computed only over the subset of videos on which the pipeline's face-detection module returns at least one detection across the 32 uniformly sampled frames, as per-video metrics are undefined on videos lacking a detected face. We refer to this set as the \emph{face-positive subset} of the benchmark under that pipeline. The two pipelines therefore induce distinct face-positive subsets, defined by the upstream face detector of FAN~\citep{bulat2017fan} and by the internal face detector of TOPIQ-NR-Face~\citep{chen2024topiq} respectively, with cardinalities reported in the last two columns of Table~\ref{tab:preprocessing}. The paired-comparison human study of Section~\ref{subsec:human_study} samples its stimuli from the intersection of the two subsets, ensuring that every video presented to participants admits a valid face localization under both pipelines.

The face-positive cardinalities in Table~\ref{tab:preprocessing} differ across the eleven entities by more than three orders of magnitude (from $57$ for AEGIS to $\sim 250{,}000$ for GenVideo), and the resulting per-benchmark sampling variance differs accordingly. We report $95\%$ percentile bootstrap confidence intervals~\citep{efron1979bootstrap} alongside every per-benchmark aggregate of the comparative analysis: $\bar{a}$ and $\mathrm{Comp}_\tau$ in Table~\ref{tab:landmark-comparative}, $Q$ in Table~\ref{tab:face-quality-wasserstein}, and the human-study win rates and detection rates of Section~\ref{app:comparative-human-study} (cluster-bootstrap over participants). The cardinality asymmetry is therefore made explicit on every metric of Section~\ref{sec:comparative}.

\subsection{Landmark Stability: Per-Benchmark Breakdown}
\label{app:comparative-landmark}

The use of FAN's per-landmark heatmap peak as a localization-reliability proxy in Section~\ref{subsec:landmark_stability} rests on its training objective. FAN~\citep{bulat2017fan} is trained with an L2 loss against 2D Gaussian target heatmaps of unit peak amplitude centered on each ground-truth landmark. Under this objective, the output heatmap peak at a given landmark approaches the Gaussian peak of $1$ when the image structure locally supports a sharp, unambiguous response, and degrades to a lower-valued, more diffuse peak when that structure is weak or inconsistent.

In addition to $\mathrm{Comp}_\tau$, Table~\ref{tab:landmark-comparative} reports the per-benchmark mean of the average per-landmark peak activation,
\begin{equation}
\label{eq:mean_peak}
\bar{a}(v) = \frac{1}{68 \cdot |\mathcal{F}_v^{\mathrm{det}}|} \sum_{f \in \mathcal{F}_v^{\mathrm{det}}} \sum_{\ell=1}^{68} a_{v,f,\ell},
\end{equation}
which summarizes localization reliability without committing to a particular threshold. The two metrics differ in how they handle isolated weak landmarks: $\mathrm{Comp}_\tau$ takes a worst-case-per-frame view in which a single landmark below $\tau$ disqualifies the frame, whereas $\bar{a}$ aggregates all peaks uniformly and is therefore less sensitive to such cases. The threshold range $\tau \in \{0.3, 0.4, 0.5, 0.6, 0.7\}$ is chosen to span the discriminative window of $\mathrm{Comp}_\tau$: outside this window the metric loses resolution, with all benchmarks (including FF++ and DFD) saturating near $1$ for $\tau \leq 0.2$ and collapsing near $0$ for $\tau \geq 0.8$.

Across both metrics, SynthForensics closely tracks the real-video baselines: $\bar{a}=0.821$ trails the FF++/DFD pair $(0.832, 0.828)$ by less than one percentage point, and $\mathrm{Comp}_\tau$ stays within five percentage points of FF++ at every $\tau$. All nine benchmarks fall well below this regime, with $\bar{a}$ ranging from $0.616$ (AEGIS) to $0.765$ (AIGVDBench), all below DFD's $0.828$. The leading benchmark additionally shifts with $\tau$: AIGVDBench tops the field at low $\tau$ (e.g.\ $0.900$ at $\tau=0.3$), while GenVideo overtakes at high $\tau$ ($0.509$ at $\tau=0.7$), with the gap to SynthForensics widening from approximately $8$ to over $37$ percentage points across the same range.

The $95\%$ percentile bootstrap CIs ($B=1{,}000$ resamples~\citep{efron1979bootstrap}) reported alongside each value make the cardinality asymmetry of Section~\ref{app:comparative-preprocessing} explicit: half-widths sit below $0.005$ for large-cardinality entities (GenVideo, AIGVDBench, SynthForensics, FF++) and reach $0.04$--$0.10$ for LOKI and AEGIS ($n=67$). The qualitative gaps reported above hold under uncertainty: at $\tau=0.7$ the SynthForensics interval $[0.882, 0.889]$ sits more than $37$ percentage points above the highest benchmark interval (GenVideo, $[0.507, 0.511]$), well beyond any CI overlap; on $\bar{a}$, the SynthForensics interval $[0.820, 0.822]$ lies above the highest benchmark interval (AIGVDBench, $[0.764, 0.765]$).

Table~\ref{tab:landmark-sf-breakdown} disaggregates these statistics by SynthForensics generator and modality. All 15 generator-modality pairs sit within the FF++/DFD range, and the T2V-to-I2V transition consistently improves both metrics (no pair regresses).

\clearpage

\begin{table}[t]
\centering
\caption{Per-benchmark landmark statistics: mean per-landmark peak activation $\bar{a}$ and mean landmark completeness $\mathrm{Comp}_\tau$ for $\tau \in \{0.3, 0.4, 0.5, 0.6, 0.7\}$, computed on the face-positive subset of each benchmark (Section~\ref{app:comparative-preprocessing}), with $95\%$ percentile bootstrap confidence intervals ($B = 1{,}000$ resamples over face-positive videos~\citep{efron1979bootstrap}) in brackets. Real-video baselines (FF++, DFD) on separate rows.}
\label{tab:landmark-comparative}
\footnotesize
\setlength{\tabcolsep}{3pt}
\renewcommand{\arraystretch}{1.15}
\resizebox{\linewidth}{!}{%
\begin{tabular}{lcccccc}
\toprule
Benchmark & $\bar{a}$ & $\tau{=}0.3$ & $\tau{=}0.4$ & $\tau{=}0.5$ & $\tau{=}0.6$ & $\tau{=}0.7$ \\
\midrule
GenVideo       & $0.715$\,{\tiny $[0.715, 0.716]$} & $0.819$\,{\tiny $[0.818, 0.820]$} & $0.803$\,{\tiny $[0.802, 0.804]$} & $0.775$\,{\tiny $[0.774, 0.777]$} & $0.712$\,{\tiny $[0.711, 0.714]$} & $0.509$\,{\tiny $[0.507, 0.511]$} \\
AIGVDBench     & $0.765$\,{\tiny $[0.764, 0.765]$} & $0.900$\,{\tiny $[0.899, 0.901]$} & $0.880$\,{\tiny $[0.879, 0.882]$} & $0.845$\,{\tiny $[0.844, 0.847]$} & $0.762$\,{\tiny $[0.760, 0.763]$} & $0.502$\,{\tiny $[0.501, 0.504]$} \\
GenVidBench    & $0.745$\,{\tiny $[0.741, 0.748]$} & $0.871$\,{\tiny $[0.865, 0.876]$} & $0.847$\,{\tiny $[0.840, 0.852]$} & $0.803$\,{\tiny $[0.796, 0.810]$} & $0.705$\,{\tiny $[0.697, 0.713]$} & $0.424$\,{\tiny $[0.416, 0.433]$} \\
GVD            & $0.714$\,{\tiny $[0.705, 0.723]$} & $0.815$\,{\tiny $[0.800, 0.829]$} & $0.795$\,{\tiny $[0.780, 0.810]$} & $0.764$\,{\tiny $[0.748, 0.779]$} & $0.694$\,{\tiny $[0.678, 0.711]$} & $0.477$\,{\tiny $[0.460, 0.494]$} \\
DVF            & $0.635$\,{\tiny $[0.620, 0.649]$} & $0.696$\,{\tiny $[0.671, 0.717]$} & $0.670$\,{\tiny $[0.644, 0.694]$} & $0.632$\,{\tiny $[0.605, 0.656]$} & $0.559$\,{\tiny $[0.531, 0.585]$} & $0.376$\,{\tiny $[0.351, 0.401]$} \\
DeepAction     & $0.686$\,{\tiny $[0.675, 0.699]$} & $0.745$\,{\tiny $[0.725, 0.766]$} & $0.705$\,{\tiny $[0.683, 0.727]$} & $0.646$\,{\tiny $[0.623, 0.670]$} & $0.539$\,{\tiny $[0.513, 0.564]$} & $0.315$\,{\tiny $[0.294, 0.337]$} \\
GVF            & $0.649$\,{\tiny $[0.634, 0.663]$} & $0.706$\,{\tiny $[0.681, 0.731]$} & $0.666$\,{\tiny $[0.639, 0.692]$} & $0.608$\,{\tiny $[0.581, 0.636]$} & $0.486$\,{\tiny $[0.458, 0.512]$} & $0.235$\,{\tiny $[0.213, 0.256]$} \\
LOKI           & $0.722$\,{\tiny $[0.677, 0.763]$} & $0.817$\,{\tiny $[0.736, 0.891]$} & $0.787$\,{\tiny $[0.702, 0.863]$} & $0.728$\,{\tiny $[0.642, 0.808]$} & $0.607$\,{\tiny $[0.517, 0.701]$} & $0.345$\,{\tiny $[0.269, 0.439]$} \\
AEGIS          & $0.616$\,{\tiny $[0.558, 0.677]$} & $0.656$\,{\tiny $[0.564, 0.754]$} & $0.623$\,{\tiny $[0.524, 0.720]$} & $0.564$\,{\tiny $[0.466, 0.669]$} & $0.449$\,{\tiny $[0.346, 0.553]$} & $0.265$\,{\tiny $[0.185, 0.351]$} \\
\midrule
SynthForensics & $0.821$\,{\tiny $[0.820, 0.822]$} & $0.979$\,{\tiny $[0.977, 0.980]$} & $0.977$\,{\tiny $[0.975, 0.978]$} & $0.972$\,{\tiny $[0.970, 0.974]$} & $0.960$\,{\tiny $[0.958, 0.962]$} & $0.886$\,{\tiny $[0.882, 0.889]$} \\
\midrule
FF++ (real)    & $0.832$\,{\tiny $[0.832, 0.833]$} & $1.000$\,{\tiny $[0.999, 1.000]$} & $0.999$\,{\tiny $[0.999, 1.000]$} & $0.998$\,{\tiny $[0.997, 0.999]$} & $0.994$\,{\tiny $[0.991, 0.997]$} & $0.932$\,{\tiny $[0.923, 0.941]$} \\
DFD (real)     & $0.828$\,{\tiny $[0.824, 0.832]$} & $0.981$\,{\tiny $[0.976, 0.985]$} & $0.979$\,{\tiny $[0.973, 0.984]$} & $0.974$\,{\tiny $[0.969, 0.980]$} & $0.963$\,{\tiny $[0.956, 0.969]$} & $0.891$\,{\tiny $[0.875, 0.906]$} \\
\bottomrule
\end{tabular}%
}
\end{table}

\begin{table}[t]
\centering
\caption{Mean per-landmark peak $\bar{a}$ and landmark completeness $\mathrm{Comp}_\tau$ per SynthForensics generator-modality pair, computed on the face-positive subset (Section~\ref{app:comparative-preprocessing}). Self-Forcing has no I2V variant (Section~\ref{sec:sf_bench}).}
\label{tab:landmark-sf-breakdown}
\small
\setlength{\tabcolsep}{6pt}
\begin{tabular}{llcccccc}
\toprule
Modality & Generator & $\bar{a}$ & $\tau{=}0.3$ & $\tau{=}0.4$ & $\tau{=}0.5$ & $\tau{=}0.6$ & $\tau{=}0.7$ \\
\midrule
\multirow{9}{*}{T2V}
  & Wan2.1              & 0.809 & 0.955 & 0.951 & 0.945 & 0.933 & 0.865 \\
  & CogVideoX           & 0.809 & 0.960 & 0.957 & 0.952 & 0.941 & 0.877 \\
  & SkyReels-V2         & 0.808 & 0.960 & 0.956 & 0.949 & 0.935 & 0.852 \\
  & Self-Forcing        & 0.818 & 0.961 & 0.956 & 0.949 & 0.927 & 0.870 \\
  & MAGI-1              & 0.817 & 0.970 & 0.969 & 0.965 & 0.956 & 0.885 \\
  & LTX-2.3               & 0.811 & 0.961 & 0.957 & 0.951 & 0.927 & 0.828 \\
  & Helios              & 0.826 & 0.992 & 0.990 & 0.986 & 0.973 & 0.870 \\
  & daVinci-MagiHuman   & 0.827 & 0.987 & 0.984 & 0.979 & 0.968 & 0.888 \\
  \cmidrule(lr){2-8}
  & \textit{Mean}       & 0.816 & 0.968 & 0.965 & 0.960 & 0.945 & 0.867 \\
\midrule
\multirow{8}{*}{I2V}
  & Wan2.1              & 0.832 & 0.996 & 0.995 & 0.993 & 0.986 & 0.929 \\
  & CogVideoX           & 0.827 & 0.994 & 0.993 & 0.990 & 0.982 & 0.908 \\
  & SkyReels-V2         & 0.829 & 0.991 & 0.990 & 0.986 & 0.978 & 0.928 \\
  & MAGI-1              & 0.828 & 0.991 & 0.990 & 0.987 & 0.980 & 0.916 \\
  & LTX-2.3               & 0.820 & 0.977 & 0.974 & 0.970 & 0.957 & 0.889 \\
  & Helios              & 0.826 & 0.993 & 0.992 & 0.989 & 0.980 & 0.895 \\
  & daVinci-MagiHuman   & 0.830 & 0.994 & 0.993 & 0.990 & 0.980 & 0.888 \\
  \cmidrule(lr){2-8}
  & \textit{Mean}       & 0.827 & 0.991 & 0.990 & 0.986 & 0.978 & 0.908 \\
\bottomrule
\end{tabular}
\end{table}

\clearpage

\subsection{Face Quality: Per-Benchmark Breakdown}
\label{app:comparative-face-quality}

The TOPIQ-NR-Face metric used in Section~\ref{subsec:face_quality} extends TOPIQ~\citep{chen2024topiq}, a top-down image-quality assessment network that propagates high-level semantic features to guide the regression of low-level distortion features through cross-scale attention. The no-reference face variant is fine-tuned on the CGFIQA-40k dataset~\citep{chen2024dslfiqa} described in Section~\ref{subsec:face_quality}. Authenticity-invariance manifests operationally as follows: a clean, sharply defined face scores high regardless of whether it is real or synthetic, while a face with visible perceptual degradations scores low. We use the implementation distributed in the IQA-PyTorch toolkit\footnote{\url{https://github.com/chaofengc/IQA-PyTorch}}.

Table~\ref{tab:face-quality-sf-breakdown} reports the per-generator $Q$ for SynthForensics. The 15 pairs span from $Q=0.361$ (Helios I2V) to $Q=0.564$ (Self-Forcing T2V), with Helios alone falling below the FF++ baseline ($Q=0.442$) in both modes. Unlike landmark stability, where I2V improves consistently over T2V (Table~\ref{tab:landmark-sf-breakdown}), $Q$ is higher in T2V for five of the seven generators with both variants (Wan2.1, CogVideoX, MAGI-1, LTX-2.3, Helios), with I2V leading only on SkyReels-V2 and daVinci-MagiHuman. This inversion has a direct interpretation. The reference frames conditioning the I2V mode are extracted from the $1{,}000$ FF++ source videos (YouTube-sourced, compressed) and the $363$ DFD source videos of Section~\ref{sec:sf_bench}. I2V pipelines are faithful enough to the conditioning that the compression artifacts of the FF++ reference frames propagate into the generated outputs. Consequently, the SynthForensics I2V aggregate ($Q_{\mathrm{I2V}}=0.450$) matches the FF++ baseline ($Q_{\mathrm{FF++}}=0.442$).

\begin{table}[!h]
\centering
\caption{Mean TOPIQ-NR-Face score $Q$ per SynthForensics generator-modality pair, computed on the face-positive subset (Section~\ref{app:comparative-preprocessing}). Self-Forcing has no I2V variant (Section~\ref{sec:sf_bench}).}
\label{tab:face-quality-sf-breakdown}
\small
\setlength{\tabcolsep}{8pt}
\begin{tabular}{llc}
\toprule
Modality & Generator & $Q$ \\
\midrule
\multirow{9}{*}{T2V}
  & Wan2.1              & 0.514 \\
  & CogVideoX           & 0.472 \\
  & SkyReels-V2         & 0.461 \\
  & Self-Forcing        & 0.564 \\
  & MAGI-1              & 0.500 \\
  & LTX-2.3               & 0.531 \\
  & Helios              & 0.383 \\
  & daVinci-MagiHuman   & 0.495 \\
  \cmidrule(lr){2-3}
  & \textit{Mean}       & 0.490 \\
\midrule
\multirow{8}{*}{I2V}
  & Wan2.1              & 0.465 \\
  & CogVideoX           & 0.401 \\
  & SkyReels-V2         & 0.478 \\
  & MAGI-1              & 0.449 \\
  & LTX-2.3               & 0.485 \\
  & Helios              & 0.361 \\
  & daVinci-MagiHuman   & 0.508 \\
  \cmidrule(lr){2-3}
  & \textit{Mean}       & 0.450 \\
\bottomrule
\end{tabular}
\end{table}

The mean-aggregate framing of $Q$ across benchmarks warrants one caveat. The two real-video baselines are themselves $0.159$ apart in mean ($Q_{\mathrm{FF++}}=0.442$ vs $Q_{\mathrm{DFD}}=0.601$), reflecting the YouTube-compressed origin of FF++ versus the lab-recorded origin of DFD. To complement the mean comparison we compute the 1D Wasserstein distance~\citep{peyre2019computational} between per-video $Q$ distributions on the face-positive subset of every entity in the pool (Table~\ref{tab:face-quality-wasserstein}). The two real baselines are $W(\mathrm{FF++}, \mathrm{DFD}) = 0.158$ apart, confirming that they occupy distinct regions of the quality space. SynthForensics is the closest entity to FF++ in the entire pool ($W(\mathrm{SF}, \mathrm{FF++}) = 0.037$) and closer to FF++ than any of the nine existing synthetic-video benchmarks: the next-closest is GenVidBench at $W = 0.051$, with the remaining eight ranging from $W = 0.070$ to $0.133$. SynthForensics therefore sits within the FF++ tail of the real-video range, consistent with the YouTube-domain origin of $1{,}000$ of the $1{,}363$ source videos and with the I2V-conditioning analysis above.

The $95\%$ percentile bootstrap CIs ($B=1{,}000$ resamples~\citep{efron1979bootstrap}) reported alongside each $Q$ in Table~\ref{tab:face-quality-wasserstein} make the cardinality asymmetry visible: half-widths sit below $0.005$ for large-cardinality entities (SynthForensics, GenVideo, AIGVDBench) and reach $\pm 0.045$ (LOKI) and $\pm 0.066$ (AEGIS) at the small end. Two implications for the mean-based comparison follow. The SynthForensics CI $[0.469, 0.473]$ lies cleanly above FF++ $[0.435, 0.450]$ and well below DFD $[0.590, 0.612]$, placing SynthForensics unambiguously between the two real-video baselines in mean. LOKI's CI $[0.458, 0.549]$ at $n=64$ overlaps SynthForensics instead, so a mean-level distinction cannot be confirmed for that entity; the Wasserstein comparison is more informative in this regime, with LOKI at $W(\mathrm{LOKI}, \mathrm{FF\text{++}}) = 0.094$, more than twice as far from FF++ as SynthForensics ($W = 0.037$).

\begin{table}[!h]
\centering
\caption{Per-benchmark mean face quality $Q$ (with $95\%$ percentile bootstrap CIs, $B=1{,}000$ resamples~\citep{efron1979bootstrap}) and 1D Wasserstein distances $W(\cdot, \mathrm{FF\text{++}})$, $W(\cdot, \mathrm{DFD})$ between per-video $Q$ distributions, computed on the face-positive subset of each entity (Section~\ref{app:comparative-preprocessing}). Lower $W$ is closer in distribution. Real-video baselines (FF++, DFD) on separate rows.}
\label{tab:face-quality-wasserstein}
\small
\setlength{\tabcolsep}{6pt}
\renewcommand{\arraystretch}{1.15}
\begin{tabular}{lccc}
\toprule
Benchmark & $Q$ & $W(\cdot, \mathrm{FF\text{++}})$ & $W(\cdot, \mathrm{DFD})$ \\
\midrule
GenVideo       & $0.369$\,{\tiny $[0.368, 0.370]$} & 0.077 & 0.232 \\
AIGVDBench     & $0.371$\,{\tiny $[0.371, 0.372]$} & 0.071 & 0.230 \\
GenVidBench    & $0.401$\,{\tiny $[0.397, 0.404]$} & 0.051 & 0.200 \\
GVD            & $0.456$\,{\tiny $[0.448, 0.465]$} & 0.080 & 0.150 \\
DVF            & $0.311$\,{\tiny $[0.301, 0.322]$} & 0.131 & 0.290 \\
DeepAction     & $0.401$\,{\tiny $[0.390, 0.413]$} & 0.070 & 0.202 \\
GVF            & $0.333$\,{\tiny $[0.319, 0.346]$} & 0.124 & 0.270 \\
LOKI           & $0.505$\,{\tiny $[0.458, 0.549]$} & 0.094 & 0.101 \\
AEGIS          & $0.381$\,{\tiny $[0.319, 0.450]$} & 0.133 & 0.235 \\
\midrule
SynthForensics & $0.471$\,{\tiny $[0.469, 0.473]$} & 0.037 & 0.130 \\
\midrule
FF++ (real)    & $0.442$\,{\tiny $[0.435, 0.450]$} & ---   & 0.158 \\
DFD (real)     & $0.601$\,{\tiny $[0.590, 0.612]$} & 0.158 & ---   \\
\bottomrule
\end{tabular}
\end{table}

\subsection{Human Perceptual Study: Full Protocol}
\label{app:comparative-human-study}

This appendix complements the brief description of the human study in Section~\ref{subsec:human_study} with the full protocol, screening procedure, statistical methodology, per-benchmark results, and inter-rater agreement analysis.

\subsubsection{Setup, recruitment, demographics, and survey UI}
\label{app:human-study-setup}

The study was conducted as a remote, web-based survey, available in English (default) and Italian. Participants were recruited via direct invitation from a convenience sample external to the research team; participation was voluntary and uncompensated. Recruitment yielded $159$ unique sessions.

Each session presented $35$ pair-comparison trials drawn from a fixed pool of $110$ video pairs (sampled with random seed $42$ for reproducibility): $100$ pairs couple a SynthForensics video with a video from one of the nine retained benchmarks of Section~\ref{sec:comparative} (approximately $11$ per benchmark), and $10$ pairs couple a SynthForensics video with a pristine video from FF++ or DFD ($5$ from each). The $110$ SynthForensics videos in the pool are themselves evenly distributed across the eight generators and the T2V/I2V modalities. All videos in the pool are sampled from the intersection of the two face-positive subsets of Section~\ref{app:comparative-preprocessing}, so that every video presented to participants admits a valid face localization under both face-detection pipelines.

Before starting the rating session, each participant acknowledged the survey's GDPR-compliant privacy policy and gave informed consent to participate, then completed a short anonymous demographic questionnaire recording age range (four brackets: $18$--$25$, $26$--$35$, $36$--$50$, $50+$), AI or computer vision expertise (yes/no), and prior familiarity with deepfakes (three options: familiar, heard of the term, not familiar). Gender was deliberately not collected, applying GDPR data-minimization principles as it was not required for the research objectives.

Table~\ref{tab:human-demographics} reports the demographic breakdown for the $159$ initially recruited and the $118$ retained participants (filtering procedure described in Section~\ref{app:human-study-filtering-stats}). The kept-vs-initial deltas are within $2$ percentage points across all dimensions, indicating that screening did not bias the demographic composition.

The protocol conforms to ITU-T P.910 (2023 edition)~\citep{itu_p910_2023}: the Pair Comparison method (\S8.4); $159$ unique participants exceed the $35$-subject floor for uncontrolled-environment studies (\S10.1); the $\sim$30-minute session duration sits well below the $1.5$-hour fatigue ceiling (\S11.1); and per-pair recordings include subject id, both stimulus paths, response, and time stamps (\S12.8). The three-option layout (A/B/Equal) used for Q1 and Q2 corresponds to the ``paired comparison with no-preference option'' variant explicitly recognized in \S8.4.1 and adopted by T2V-HE~\citep{zhang2024t2vhe}; the four-option layout (A/B/Both/Neither) used for Q3 is a non-standard extension motivated in Section~\ref{subsec:human_study} as a deliberate departure for the absolute-property nature of Q3. Our deviations from \S10.3 (gender balance, addressed above) and \S12.4 (post-screening of subjects, see Section~\ref{app:human-study-filtering-stats}) are deliberate.

\begin{figure}[!htbp]
\centering
\includegraphics[width=0.75\linewidth]{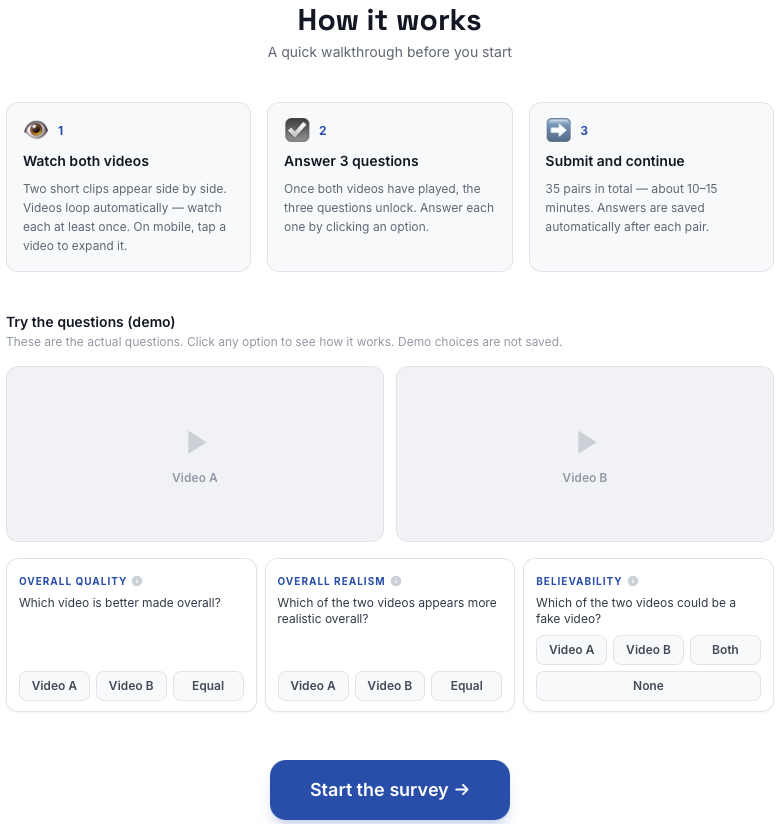}
\caption{Pre-experiment introduction page describing the study purpose, the three questions, and the meaning of each response option.}
\label{fig:survey-ui-intro}
\end{figure}

\begin{figure}[!htbp]
\centering
\includegraphics[width=0.75\linewidth]{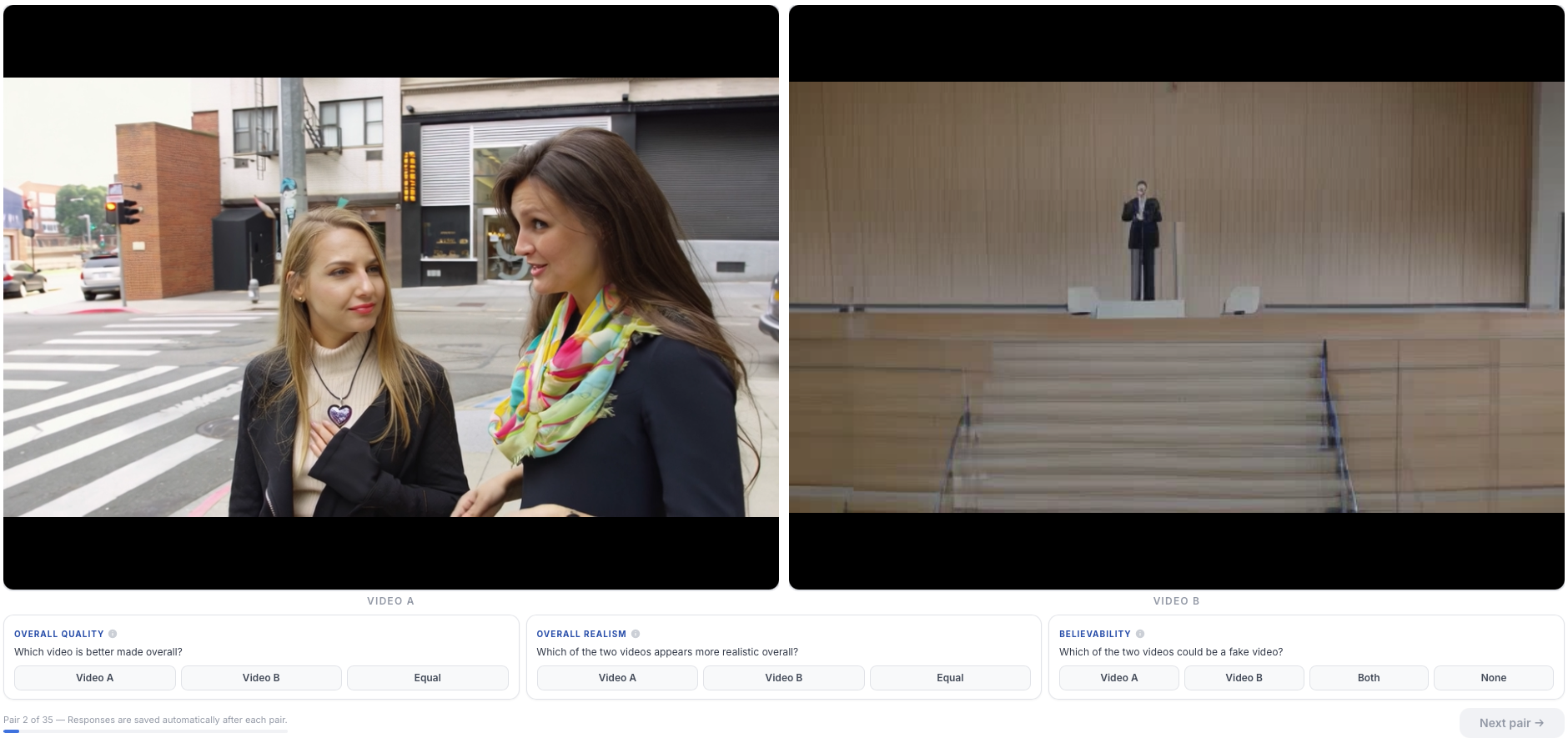}
\caption{Pair-comparison trial page with the two videos side by side and the three questions Q1 (Better Quality), Q2 (More Realistic), and Q3 (Flagged as Fake) displayed simultaneously.}
\label{fig:survey-ui-qs}
\end{figure}

Each session begins with the pre-experiment introduction page (Figure~\ref{fig:survey-ui-intro}), which describes the study purpose, the three questions, and the meaning of every response option; in T2V-HE's terminology this corresponds to instruction-based annotator training~\citep{zhang2024t2vhe}. Each pair-comparison trial is then presented on a single page (Figure~\ref{fig:survey-ui-qs}) with the two videos side by side and the three questions displayed simultaneously. The standard three-option layout (A/B/Equal) is used for Q1 and Q2, while Q3 employs the four-option layout (A/B/Both/Neither) motivated in Section~\ref{subsec:human_study}. To prevent rushed responses, the response buttons remain disabled until both videos have played in full, plus an additional 5-second buffer.

\begin{table}[!h]
\centering
\caption{Demographic breakdown of survey participants. ``Initial'' denotes the $159$ recruited participants; ``Kept'' denotes the $118$ retained after the filtering procedure of Section~\ref{app:human-study-filtering-stats}.}
\label{tab:human-demographics}
\small
\setlength{\tabcolsep}{8pt}
\begin{tabular}{llrrrr}
\toprule
Dimension & Value & \multicolumn{2}{c}{Initial} & \multicolumn{2}{c}{Kept} \\
\cmidrule(lr){3-4}\cmidrule(lr){5-6}
 &  & $n$ & $\%$ & $n$ & $\%$ \\
\midrule
\multirow{4}{*}{Age range}
  & 18--25 & 56 & 35.2 & 40 & 33.9 \\
  & 26--35 & 70 & 44.0 & 55 & 46.6 \\
  & 36--50 & 30 & 18.9 & 21 & 17.8 \\
  & 50+    & 3  & 1.9  & 2  & 1.7  \\
\midrule
\multirow{2}{*}{AI/CV expertise}
  & Yes & 81 & 50.9 & 57 & 48.3 \\
  & No  & 78 & 49.1 & 61 & 51.7 \\
\midrule
\multirow{3}{*}{Deepfake familiarity}
  & Familiar           & 115 & 72.3 & 87 & 73.7 \\
  & Heard of the term  & 32  & 20.1 & 24 & 20.3 \\
  & Not familiar       & 12  & 7.5  & 7  & 5.9  \\
\bottomrule
\end{tabular}
\end{table}

\clearpage

\clearpage

\subsubsection{Filtering and statistical methodology}
\label{app:human-study-filtering-stats}

Following T2V-HE~\citep{zhang2024t2vhe}, we apply a minimal post-hoc filtering procedure restricted to objective behavioral checks, in deliberate contrast to the consensus-based screening of ITU-T P.910~\citep{itu_p910_2023} \S12.4, for the reasons described below. The procedure has two stages. Stage~1 enforces a completion criterion: a session is retained only if the participant completed at least $28$ of the $35$ trials ($\geq 80\%$ completion rate), to exclude trivially partial sessions. Stage~2 applies a disengagement check based on the participant's behavioral variance across Q1 and Q2 trials: for each session, we count how often each response category (A, B, Equal) was chosen, and reject the session if a single category accounts for $\geq 95\%$ of the responses, since this indicates that the participant clicked nearly the same button on almost every trial, a pattern incompatible with attentive evaluation. Of the $159$ recruited participants, Stage~1 rejects $39$ sessions and Stage~2 rejects an additional $2$, leaving $118$ retained sessions. Across these $118$ sessions, the partial completions allowed under the $\geq 80\%$ threshold yield $4{,}121$ ratings per question rather than the maximum $35 \times 118 = 4{,}130$, a shortfall of $9$ ratings. The screening logs are released alongside the data.

ITU-T P.910~\citep{itu_p910_2023} \S12.4 lists four appropriate post-screening methods: clause 13.6 of the same Recommendation, Annex~A (Pearson linear correlation against the group consensus, with iterative worst-first rejection; a discard threshold of $r_1 < 0.75$ is recommended for ACR and ACR-HR tests of entertainment video), Annex~1 \S A1-2.3 of ITU-R BT.500~\citep{itu_bt500_2023}, and post-experiment questionnaires or interviews. We deliberately depart from all three, and from Annex~A in particular, because consensus-based screening assumes the existence of an objective ground truth, typical of ACR-style video-quality experiments where compressed processed sequences are rated against a known uncompressed reference, against which an individual subject's deviation can be classified as error. Our task is instead a paired-comparison preference judgment for which no such ground truth exists: a participant whose response pattern diverges from the group consensus may simply hold a minority but valid preference, and rejecting such participants would systematically bias the estimated preferences toward majority opinions. The same precedent is set in T2V-HE~\citep{zhang2024t2vhe}, which evaluates open- and closed-source text-to-video models without consensus-based filtering and reports Krippendorff's $\alpha$ as a quality measure of the resulting annotations rather than as a screening criterion. We follow this precedent.

Per-benchmark win rates~\citep{david1988method} are computed as the empirical proportion of pairs in which the SynthForensics video is preferred over the paired entity, separately for Q1 and Q2 and for each of the eleven entities (the nine retained benchmarks of Section~\ref{sec:comparative}, FF++, and DFD). $95\%$ confidence intervals are obtained by participant-clustered bootstrap with $B = 1{,}000$ resamples and a fixed random seed (\texttt{42}): in each resample, participant identifiers are drawn with replacement from the $118$ retained sessions and the three response proportions ($w_{\mathrm{SF}}, w_{\mathrm{Equal}}, w_{\mathrm{Other}}$) are recomputed on the resampled data; the $2.5$th and $97.5$th percentiles of the bootstrap distribution define the CI bounds. Clustering by participant accounts for the within-participant correlation induced by each participant rating multiple pairs~\citep{efron1979bootstrap}. For Q3, the four-option response is decomposed into two binary judgments, ``is video A a fake?'' and ``is video B a fake?'', according to the mapping $\mathrm{A} \mapsto (1, 0)$, $\mathrm{B} \mapsto (0, 1)$, $\mathrm{Both} \mapsto (1, 1)$, $\mathrm{Neither} \mapsto (0, 0)$. The per-entity human-detection rate is the resulting proportion of fake-flagged judgments on videos belonging to that entity, with CIs obtained by the same participant-clustered bootstrap procedure. We do not fit a global Bradley--Terry~\citep{bradley1952rank} or Rao--Kupper~\citep{rao1967ties} model across the eleven entities: the survey design is a star pattern in which every pair couples SynthForensics with one other entity, and pairwise comparisons among non-SynthForensics entities are never collected; estimating non-SynthForensics rankings from such a design would rely on transitivity assumptions that the data cannot directly support, while per-benchmark win rates honestly report what was measured.

Inter-rater agreement is reported as Krippendorff's $\alpha$~\citep{krippendorff2011computing}, computed on the raw response categories of each question (three categories for Q1 and Q2; four categories for Q3). $\alpha$ is preferred over Fleiss' $\kappa$~\citep{fleiss1971kappa} in the present setting because it natively accommodates missing data and unequal numbers of raters per unit, both of which apply here: the $110$ stimulus pairs are rated by partially overlapping subsets of the $118$ retained participants, with approximately $37$ ratings per pair on average. Following T2V-HE~\citep{zhang2024t2vhe}, we report $\alpha$ as a quality measure of the annotations rather than as a screening criterion.

\newpage

\subsubsection{Detailed per-benchmark results}
\label{app:human-study-results}

Tables~\ref{tab:human-q12} and~\ref{tab:human-q3} report the per-benchmark point estimates and $95\%$ cluster-bootstrap confidence intervals that underlie the \emph{vs Benchmarks} and \emph{vs Real} aggregates of Figure~\ref{fig:human_study}.

Against the nine benchmarks of Section~\ref{sec:comparative}, SynthForensics is preferred in every row of both Q1 and Q2: $9$ of $9$ benchmarks on overall quality, $9$ of $9$ on realism. The margin is decisive on the older generator pools, with Q1 win rates of $89.9\%$ on GenVideo, $85.9\%$ on GVF, and $81.7\%$ on GenVidBench, and remains substantial even on the most recent and competitive pools, with Q1 above $50\%$ on LOKI, AEGIS, and DeepAction and Q2 between $61\%$ and $65\%$ on the same three. Against the FF++ and DFD pristine baselines the win rate naturally falls below $50\%$, yet a substantial $33$--$37\%$ of responses are \emph{Equal}: participants are frequently unable to express a preference between SynthForensics and unaltered real footage.

Table~\ref{tab:human-q3} amplifies this picture. SynthForensics videos are flagged as fake in a narrow $31.4$--$44.8\%$ band across all eleven comparison contexts, essentially constant whether SynthForensics is paired with one of the nine benchmarks or with pristine real footage. The fraction of fake-flags on the paired entity, in contrast, spans more than an order of magnitude: it climbs to $92.3\%$ on GenVideo, $91.8\%$ on GenVidBench, and $89.5\%$ on GVF (older pools), sits at $68.3\%$ on DeepAction and $72.7\%$ on LOKI, and reaches just $13.1\%$ for FF++ (real footage). The flat band on SynthForensics, contrasted with the order-of-magnitude spread on the paired entity, indicates that the perceived fakeness of SynthForensics videos is essentially independent of the comparison context.

Tables~\ref{tab:human-by-generator} and~\ref{tab:human-by-modality} re-aggregate the same benchmark responses by SynthForensics generator and by SynthForensics modality, pooling across the nine benchmarks. SkyReels-V2 leads on every metric, with both Q1 and Q2 win rates above $90\%$ and the lowest Q3 SF-flag rate (fraction of SynthForensics videos in the pair flagged as fake) in the pool ($26.4\%$); LTX-2.3 follows close behind. The remaining four large-scale generators (Helios, MAGI-1, Wan2.1, daVinci-MagiHuman) form a tight cluster with Q1 and Q2 win rates in the $70$--$80\%$ band and Q3 SF-flag rates between $30\%$ and $40\%$. The two least competitive entries are the smallest model (CogVideoX) and the few-step distilled Self-Forcing, the latter being the only generator on which SynthForensics is preferred in less than $50\%$ of pairs and is more often flagged as fake than the paired benchmark. Across modalities (Table~\ref{tab:human-by-modality}), image-to-video outputs are perceived as marginally more realistic and less identifiable as fake than text-to-video outputs (Q2 win rate $78.5\%$ vs $74.8\%$; Q3 SF-flag $33.6\%$ vs $40.8\%$), consistent with the I2V mode inheriting structural realism from the conditioning frame, while text-to-video shows a small edge on Q1 ($72.5\%$ vs $70.4\%$).

\clearpage

\begin{table}[!t]
\centering
\caption{Per-benchmark Q1 (Better Quality) and Q2 (More Realistic) win rates with $95\%$ cluster-bootstrap confidence intervals over participants. $n$ is the number of ratings. $w_{\mathrm{SF}}$ is the proportion of pairs in which SynthForensics is preferred; $w_{\mathrm{Equal}}$ and $w_{\mathrm{Other}}$ are the proportions of \emph{Equal} responses and of preferences for the paired entity respectively. All values in percent.}
\label{tab:human-q12}
\small
\setlength{\tabcolsep}{10pt}
\renewcommand{\arraystretch}{1.3}
\begin{tabular}{llrlll}
\toprule
Benchmark & Q & $n$ & $w_{\mathrm{SF}}$ [CI] & $w_{\mathrm{Equal}}$ [CI] & $w_{\mathrm{Other}}$ [CI] \\
\midrule
\multirow{2}{*}{GenVideo}    & Q1 & 388 & 89.9 [86.7, 93.1] & 4.9 [2.6, 7.6]   & 5.2 [3.0, 7.6]   \\
                             & Q2 & 388 & 90.5 [87.0, 93.9] & 4.9 [2.4, 7.6]   & 4.6 [2.4, 7.2]   \\
\multirow{2}{*}{AIGVDBench}  & Q1 & 389 & 67.6 [62.3, 72.5] & 13.4 [9.8, 16.9] & 19.0 [14.8, 23.6] \\
                             & Q2 & 389 & 74.6 [70.0, 78.8] & 12.6 [9.3, 16.1] & 12.9 [9.7, 16.4]  \\
\multirow{2}{*}{GenVidBench} & Q1 & 389 & 81.7 [77.2, 85.9] & 9.8 [6.8, 13.0]  & 8.5 [5.6, 11.9]  \\
                             & Q2 & 389 & 84.6 [80.5, 88.1] & 9.8 [6.8, 13.1]  & 5.7 [3.5, 8.1]   \\
\multirow{2}{*}{GVD}         & Q1 & 393 & 79.1 [74.8, 83.1] & 7.4 [5.2, 10.0]  & 13.5 [10.3, 16.8] \\
                             & Q2 & 393 & 84.0 [79.6, 88.0] & 5.3 [3.2, 8.1]   & 10.7 [7.6, 13.9]  \\
\multirow{2}{*}{DVF}         & Q1 & 387 & 74.2 [69.0, 78.6] & 9.6 [6.1, 13.5]  & 16.3 [12.8, 20.0] \\
                             & Q2 & 387 & 78.6 [73.9, 82.9] & 8.8 [5.9, 12.1]  & 12.7 [9.6, 15.7]  \\
\multirow{2}{*}{DeepAction}  & Q1 & 397 & 54.2 [49.4, 59.1] & 14.6 [10.5, 18.9] & 31.2 [26.5, 36.5] \\
                             & Q2 & 397 & 61.5 [56.7, 66.3] & 14.1 [10.5, 17.7] & 24.4 [20.2, 28.9] \\
\multirow{2}{*}{GVF}         & Q1 & 418 & 85.9 [81.6, 89.9] & 7.7 [5.0, 10.9]  & 6.5 [3.7, 9.3]   \\
                             & Q2 & 418 & 86.8 [83.1, 90.4] & 7.4 [4.8, 10.5]  & 5.7 [3.2, 8.6]   \\
\multirow{2}{*}{LOKI}        & Q1 & 384 & 53.9 [48.3, 59.6] & 13.0 [9.7, 16.8] & 33.1 [27.7, 38.1] \\
                             & Q2 & 384 & 62.0 [57.1, 66.9] & 12.8 [9.6, 16.2] & 25.3 [21.0, 29.3] \\
\multirow{2}{*}{AEGIS}       & Q1 & 387 & 55.8 [50.1, 61.0] & 12.7 [8.6, 17.0] & 31.5 [26.7, 37.0] \\
                             & Q2 & 387 & 65.4 [59.9, 70.7] & 15.0 [11.2, 18.7] & 19.6 [15.4, 24.1] \\
\midrule
\multirow{2}{*}{FF++}        & Q1 & 298 & 31.2 [25.3, 36.7] & 32.9 [26.4, 39.3] & 35.9 [29.5, 42.9] \\
                             & Q2 & 298 & 20.8 [16.0, 26.0] & 37.2 [30.8, 43.5] & 41.9 [35.4, 48.5] \\
\multirow{2}{*}{DFD}         & Q1 & 291 & 23.0 [17.6, 29.1] & 35.7 [29.1, 42.7] & 41.2 [34.3, 48.5] \\
                             & Q2 & 291 & 26.8 [21.6, 32.7] & 36.8 [30.4, 43.2] & 36.4 [30.2, 42.4] \\
\bottomrule
\end{tabular}
\end{table}

\begin{table}[!t]
\centering
\caption{Per-benchmark Q3 (Flagged as Fake) detection rates with $95\%$ cluster-bootstrap confidence intervals over participants. $n$ is the number of ratings. ``SF flagged'' is the fraction of SynthForensics videos in the pair flagged as fake; ``Other flagged'' is the same fraction for the paired entity (a benchmark or a real-video baseline). All values in percent.}
\label{tab:human-q3}
\small
\setlength{\tabcolsep}{14pt}
\renewcommand{\arraystretch}{1.3}
\begin{tabular}{lrll}
\toprule
Benchmark & $n$ & SF flagged [CI] & Other flagged [CI] \\
\midrule
GenVideo    & 388 & 31.4 [26.0, 37.1] & 92.3 [87.4, 96.1] \\
AIGVDBench  & 389 & 34.4 [29.0, 39.8] & 88.9 [84.7, 92.5] \\
GenVidBench & 389 & 33.2 [27.9, 38.9] & 91.8 [87.0, 95.4] \\
GVD         & 393 & 37.9 [32.1, 44.0] & 87.5 [83.5, 90.9] \\
DVF         & 387 & 39.0 [33.1, 45.0] & 86.0 [82.4, 90.1] \\
DeepAction  & 397 & 44.8 [38.5, 50.9] & 68.3 [63.3, 73.0] \\
GVF         & 418 & 39.2 [33.7, 45.5] & 89.5 [85.0, 93.5] \\
LOKI        & 384 & 43.5 [36.8, 50.5] & 72.7 [67.8, 77.3] \\
AEGIS       & 387 & 33.3 [27.5, 39.1] & 80.1 [75.7, 84.4] \\
\midrule
FF++        & 298 & 38.9 [32.0, 45.8] & 13.1 [8.7, 17.6]  \\
DFD         & 291 & 41.6 [35.7, 47.4] & 31.6 [25.3, 38.4] \\
\bottomrule
\end{tabular}
\end{table}

\clearpage

\begin{table}[!t]
\centering
\caption{Per-SynthForensics-generator breakdown of the human-study results, pooled across the nine benchmarks. Win rates (Q1, Q2) and Q3 SF-flag rates with $95\%$ cluster-bootstrap CIs over participants. $n$ is the number of ratings; Self-Forcing's lower count reflects that it has no I2V variant. Rows ordered by Q1 win rate descending. All values in percent.}
\label{tab:human-by-generator}
\small
\setlength{\tabcolsep}{8pt}
\renewcommand{\arraystretch}{1.3}
\begin{tabular}{lrlll}
\toprule
Generator & $n$ & Q1 SF win [CI] & Q2 SF win [CI] & Q3 SF flagged [CI] \\
\midrule
SkyReels-V2       & 459 & 92.4 [89.6, 94.9] & 91.5 [89.0, 94.1] & 26.4 [21.1, 31.8] \\
LTX-2.3             & 468 & 78.2 [73.4, 82.3] & 83.5 [79.6, 87.0] & 29.7 [24.6, 35.1] \\
Helios        & 465 & 73.5 [69.0, 77.5] & 79.4 [75.0, 83.5] & 37.0 [30.8, 43.3] \\
MAGI-1            & 557 & 73.4 [69.2, 77.3] & 77.4 [73.4, 80.9] & 39.7 [33.9, 45.4] \\
Wan2.1            & 416 & 70.0 [64.9, 74.9] & 74.8 [70.2, 79.7] & 32.2 [26.7, 37.8] \\
daVinci-MagiHuman & 424 & 69.1 [64.4, 73.6] & 78.5 [74.6, 82.2] & 30.7 [25.4, 36.1] \\
CogVideoX         & 492 & 56.9 [51.6, 61.5] & 65.7 [61.1, 70.3] & 52.4 [46.2, 58.8] \\
Self-Forcing      & 251 & 47.8 [41.1, 55.0] & 49.4 [42.2, 57.0] & 59.0 [50.2, 66.5] \\
\bottomrule
\end{tabular}
\end{table}

\begin{table}[!t]
\centering
\caption{Per-SynthForensics-modality breakdown of the human-study results, pooled across the nine benchmarks. $n$ is the number of ratings. All values in percent, with $95\%$ cluster-bootstrap CIs over participants.}
\label{tab:human-by-modality}
\small
\setlength{\tabcolsep}{12pt}
\renewcommand{\arraystretch}{1.3}
\begin{tabular}{lrlll}
\toprule
Modality & $n$ & Q1 SF win [CI] & Q2 SF win [CI] & Q3 SF flagged [CI] \\
\midrule
T2V & 1878 & 72.5 [69.4, 75.8] & 74.8 [71.8, 77.7] & 40.8 [36.2, 45.2] \\
I2V & 1654 & 70.4 [67.0, 73.7] & 78.5 [75.6, 81.4] & 33.6 [29.0, 38.4] \\
\bottomrule
\end{tabular}
\end{table}

\clearpage

\titlespacing*{\subsubsection}{0pt}{2pt}{2pt}

\subsubsection{Inter-rater agreement and sensitivity analyses}

\titlespacing*{\subsubsection}{0pt}{1.5ex}{0.8ex}  

\label{app:human-study-agreement}
Inter-rater agreement is reported as Krippendorff's $\alpha$~\citep{krippendorff2011computing} on the raw response categories of each question, computed on the $118$ retained sessions. The overall values are $\alpha_{\mathrm{Q1}} = 0.260$, $\alpha_{\mathrm{Q2}} = 0.247$, and $\alpha_{\mathrm{Q3}} = 0.152$. As a reference, T2V-HE~\citep{zhang2024t2vhe} (Tables~2--3) reports post-training LRA agreement in the $0.107$--$0.339$ band and AMT crowdworker agreement in the $0.177$--$0.451$ band across their video-quality metrics; our values for Q1 and Q2 fall in the upper half of the post-training-LRA band, while Q3 sits in the lower portion of the same band. We treat these values as a quality measure of the annotations rather than a screening threshold, consistent with the rationale of Section~\ref{app:human-study-filtering-stats}.

Stratifying $\alpha$ by stimulus type (Table~\ref{tab:human-alpha}) exposes a gap between the two strata. On the SF-vs-benchmark pairs the agreement remains within the T2V-HE band ($0.242$, $0.195$, $0.112$ for Q1, Q2, Q3), whereas on the SF-vs-real pairs (FF++ and DFD pristine videos) it collapses toward zero ($0.050$, $0.022$, $0.066$). This is consistent with SynthForensics videos approaching real footage in visual realism: when SynthForensics is paired with one of the nine benchmarks, participants converge on a clear preference (high agreement); when paired with pristine real video, they cannot reliably tell the two apart, and their responses concentrate on the indecision option rather than spreading uniformly across categories. The Equal rate on Q1 and Q2 rises from $5$--$15\%$ on SF-vs-benchmark pairs to $33$--$37\%$ on SF-vs-real (Table~\ref{tab:human-q12}), and a chi-square test of the uniform-choice null on the combined Q2-vs-real distribution ($n=589$, the focal cell at $\alpha_{Q2}=0.022$) yields $\chi^2 \approx 24.6$ ($\mathrm{df}=2$, $p < 10^{-5}$), so the indecision pattern is statistically distinguishable from uniformly random responses. Agreement values across demographic strata (age range, AI/CV background, deepfake familiarity) remain in a narrow band around the overall values for the well-populated cells; the small cells (knows-deepfakes ``no'' with $n_{\mathrm{sess}} = 7$ and age ``50+'' with $n_{\mathrm{sess}} = 2$) yield numerically unstable estimates that we report for completeness only.

We complement the inter-rater agreement analysis with two robustness checks on the headline numbers reported in Section~\ref{subsec:human_study} and Tables~\ref{tab:human-q12}--\ref{tab:human-q3}. The first is a demographic-subgroup sensitivity (Table~\ref{tab:human-sensitivity-subgroup}): the SF-vs-benchmark pool of $\sim$$3{,}500$ ratings per question is re-aggregated within each demographic stratum, with $95\%$ cluster-bootstrap confidence intervals over participants~\citep{efron1979bootstrap}. SynthForensics is preferred over the benchmark pool in every well-populated stratum: Q1 win rates remain in the $59$--$76\%$ band across the three informative age brackets, in $68$--$76\%$ across AI/CV background, and in $71$--$75\%$ across the two informative familiarity levels, with Q2 win rates uniformly higher and Q3 SF-flag rates uniformly within $26$--$41\%$. The two cells with $n_{\mathrm{sess}} \leq 7$ (knows-deepfakes ``not familiar'', age ``50+'') are reported for completeness only and not interpreted, consistently with the corresponding caveat in the agreement analysis.

To probe whether the headline numbers depend on the choice of not applying consensus-based post-screening (Section~\ref{app:human-study-filtering-stats}), we run a fixed-fraction sensitivity check inspired by the correlation step of ITU-T P.910 Annex~A.1~\citep{itu_p910_2023}. The procedure operates on the $120$-session post-Stage-1 set of Section~\ref{app:human-study-filtering-stats}, that is, the participants that passed the completion filter regardless of the Stage-2 disengagement check. We encode Q1 and Q2 responses as $A = +1$, $B = -1$, $\mathrm{Equal} = 0$, compute the consensus mean per (stimulus pair, question) across the $120$ sessions, and compute each session's Pearson correlation $r_1$ between its response vector and the matched consensus vector. We then sort sessions by $r_1$ and drop the bottom $25\%$ ($30$ sessions; the $90$ most consensus-aligned remain). The $25\%$ fraction is chosen as a conventional quartile-based trim that removes a substantial tail of consensus-divergent sessions while keeping the retained $90$ sessions well above the $35$-subject floor recommended by ITU-T P.910 \S10.1 for uncontrolled-environment studies~\citep{itu_p910_2023}. On this trimmed subset, the headline numbers shift only marginally and in the direction of stronger preference for SynthForensics: Q1 win rate $71.5 \to 72.6$, Q2 win rate $76.5 \to 77.6$, Q3 SF-flag rate $37.5 \to 37.9$ (all changes within $1.2$ percentage points). The conclusions of Section~\ref{subsec:human_study} therefore do not depend on the absence of consensus-based participant trimming.

As concluding notes, two considerations bear on the headline numbers. First, the recruited sample is not representative of a general consumer population: as shown in Table~\ref{tab:human-demographics}, most participants are aged 18--35 ($80\%$) and report AI/CV familiarity ($48\%$ self-reported expertise, $74\%$ deepfake-familiar). The technical/AI-literate composition acts as an upper-bound stress test on perceived realism, since familiarity with synthetic media should plausibly aid identification of fakes; the headline rates from this sample should therefore be conservative relative to a less-expert population. Second, Q3 carries the lowest per-rater agreement of the three questions ($\alpha_{Q3}=0.152$ vs.\ $0.260$ for Q1 and $0.247$ for Q2), consistent with the harder absolute-property nature of the question (``is this a fake?'') versus the relative A/B judgments of Q1 and Q2. Both considerations are addressed empirically by the analyses above: the headline rates remain stable across demographic strata (Table~\ref{tab:human-sensitivity-subgroup}, Q3 SF-flag in $26$--$41\%$ band) and shift by less than $1$pp under the ITU-T P.910 Annex~A.1 sensitivity check~\citep{itu_p910_2023} ($37.5 \to 37.9$ on Q3); the headline numbers therefore survive both the demographic skew and the per-rater Q3 noise.

\begin{table}[H]
\centering
\caption{Krippendorff's $\alpha$ on the raw response categories of Q1, Q2, Q3, computed overall and stratified by stimulus type and demographic dimensions. $n_{\mathrm{sess}}$ is the number of retained sessions in each stratum.}
\label{tab:human-alpha}
\small
\setlength{\tabcolsep}{10pt}
\renewcommand{\arraystretch}{1.2}
\begin{tabular}{llrrrr}
\toprule
Stratum & Value & $n_{\mathrm{sess}}$ & $\alpha_{\mathrm{Q1}}$ & $\alpha_{\mathrm{Q2}}$ & $\alpha_{\mathrm{Q3}}$ \\
\midrule
Overall          & all retained sessions    & 118 & 0.260 & 0.247 & 0.152 \\
\midrule
\multirow{2}{*}{Stimulus type}
                 & vs benchmark             & 118 & 0.242 & 0.195 & 0.112 \\
                 & vs real (FF++/DFD)       & 118 & 0.050 & 0.022 & 0.066 \\
\midrule
\multirow{4}{*}{Age range}
                 & 18--25                   &  40 & 0.298 & 0.256 & 0.138 \\
                 & 26--35                   &  55 & 0.267 & 0.253 & 0.161 \\
                 & 36--50                   &  21 & 0.197 & 0.209 & 0.093 \\
                 & 50+                      &   2 & 0.222 & 0.510 & 0.746 \\
\midrule
\multirow{2}{*}{AI/CV background}
                 & yes                      &  57 & 0.293 & 0.260 & 0.180 \\
                 & no                       &  61 & 0.233 & 0.239 & 0.129 \\
\midrule
\multirow{3}{*}{Deepfake familiarity}
                 & familiar                 &  87 & 0.271 & 0.255 & 0.166 \\
                 & heard of the term        &  24 & 0.281 & 0.261 & 0.170 \\
                 & not familiar             &   7 & $-0.031$ & 0.166 & $-0.030$ \\
\bottomrule
\end{tabular}
\end{table}

\begin{table}[H]
\centering
\caption{Demographic-subgroup sensitivity for the headline SF-vs-benchmark pool. Q1, Q2 are SynthForensics win rates; Q3 is the SF-flagged-as-fake rate. $95\%$ cluster-bootstrap CIs over participants; $n_{\mathrm{sess}}$ is the number of retained sessions in each stratum. All values in percent.}
\label{tab:human-sensitivity-subgroup}
\small
\setlength{\tabcolsep}{4pt}
\renewcommand{\arraystretch}{1.2}
\resizebox{\linewidth}{!}{%
\begin{tabular}{llrlll}
\toprule
Stratum & Value & $n_{\mathrm{sess}}$ & Q1 SF win [CI] & Q2 SF win [CI] & Q3 SF flagged [CI] \\
\midrule
Baseline & all retained sessions  & 118 & 71.5 [68.7, 74.4] & 76.5 [74.1, 78.9] & 37.5 [33.4, 41.6] \\
\midrule
\multirow{4}{*}{Age range}
         & 18--25                 &  40 & 75.7 [70.8, 80.0] & 79.8 [74.6, 84.4] & 33.0 [25.0, 41.7] \\
         & 26--35                 &  55 & 73.5 [69.5, 77.3] & 77.3 [74.1, 80.4] & 40.1 [33.9, 46.9] \\
         & 36--50                 &  21 & 59.2 [52.7, 65.1] & 68.0 [63.7, 72.5] & 40.5 [31.4, 49.4] \\
         & 50+                    &   2 & 61.7 [56.7, 66.7] & 78.3 [73.3, 83.3] & 23.3 [23.3, 23.3] \\
\midrule
\multirow{2}{*}{AI/CV background}
         & yes                    &  57 & 75.0 [71.2, 79.0] & 79.3 [75.5, 83.1] & 37.1 [29.8, 44.4] \\
         & no                     &  61 & 68.2 [64.2, 72.4] & 73.9 [70.7, 77.2] & 37.8 [32.5, 43.7] \\
\midrule
\multirow{3}{*}{Deepfake familiarity}
         & familiar               &  87 & 71.3 [68.3, 74.4] & 75.8 [73.0, 78.5] & 40.6 [35.6, 46.0] \\
         & heard of the term      &  24 & 75.0 [68.9, 80.6] & 79.9 [73.8, 85.7] & 26.4 [19.7, 33.5] \\
         & not familiar           &   7 & 61.4 [44.3, 78.6] & 73.3 [64.3, 83.8] & 36.2 [19.5, 51.9] \\
\bottomrule
\end{tabular}%
}
\end{table}

\clearpage
\section{Detection Experiments: Extended Details}
\label{app:detection}

This appendix complements Section~\ref{sec:detection} with: per-detector preprocessing, score aggregation, evaluation metric, and computing hardware (Appendix~\ref{app:detection-setup}); per-generator and compression breakdowns of the zero-shot results (Appendix~\ref{app:detection-zeroshot}); per-generator results and backward compatibility on the Legacy Benchmark Sets for the fine-tuning protocol (Appendix~\ref{app:detection-finetuning}); and the same breakdowns for the training-from-scratch protocol (Appendix~\ref{app:detection-retraining}).

\subsection{Detailed Setup}
\label{app:detection-setup}

To ensure a comprehensive and reproducible assessment of detector robustness, we follow each detector's official preprocessing pipeline, including temporal frame sampling, face detection, alignment, spatial cropping, and resolution normalization. By default, we apply the exact extraction and transformation routines prescribed by the respective authors of CFM~\citep{luo2023beyond}, LAA-Net~\citep{nguyen2024laa}, GenD~\citep{yermakov2026deepfake}, GenConViT~\citep{deressa2025genconvit}, DFD-FCG~\citep{han2025towards}, FakeSTormer~\citep{nguyen2025vulnerability}, MM-Det~\citep{song2024learning}, D3~\citep{zheng2025d3}, and NSG-VD~\citep{zhang2025physics}. The remaining six (RECCE~\citep{cao2022end}, ProDet~\citep{cheng2024can}, UCF~\citep{yan2023ucf}, Effort~\citep{yan2024orthogonal}, AltFreezing~\citep{wang2023altfreezing}, and FTCN~\citep{zheng2021exploring}) are integrated via the DeepfakeBench framework~\citep{yan2023deepfakebench}\footnote{\url{https://github.com/SCLBD/DeepfakeBench}}, adopting the framework's standardized preprocessing and sampling protocols to maintain a consistent evaluation baseline. Full per-detector preprocessing schemes and temporal sampling configurations are released with our source code.

We standardize the output into a single anomaly score $S_V \in [0,1]$ for each video, representing the aggregate probability that the content is synthetic. The calculation of this score inherently depends on the detector's architectural category. In the case of frame-level models, each sampled frame is analyzed independently to generate an individual probability score $s_i$. The final video-level score is then derived by computing the arithmetic mean of the scores across all $K$ sampled frames:
\begin{equation}
    S_V = \frac{1}{K} \sum_{i=1}^{K} s_i
\end{equation}
In contrast, video-level models process the input sequence as a coherent spatiotemporal volume. These architectures directly yield the global score $S_V$ in a single inference pass, without requiring any post-processing temporal aggregation.

We quantify detection performance using the video-level Area Under the Receiver Operating Characteristic Curve (AUC). This metric provides a threshold-independent assessment, evaluating the global ranking quality of the detector by measuring how well the model separates the positive (synthetic) and negative (real) distributions. The AUC is formally defined as the integral of the True Positive Rate (TPR) with respect to the False Positive Rate (FPR):
\begin{equation}
    \text{AUC} = \int_{0}^{1} \text{TPR}(x) \, dx
\end{equation}
where $x$ denotes the False Positive Rate. An AUC of $50\%$ implies performance equivalent to random guessing, while $100\%$ indicates perfect separability between the real and synthetic distributions.

\subsubsection{Computing Resources}
\label{app:detection-hardware}

We conduct experimental evaluations, training procedures, and video generation workloads (Appendix~\ref{app:benchmark-hyperparams}) on a combination of on-site infrastructure and cloud computing resources. The primary computational tasks are distributed across two dedicated on-site servers. The first server features a dual-processor configuration with two Intel(R) Xeon(R) Gold 5416S CPUs, supported by 256 GB of DDR4 ECC RAM and two NVIDIA A100 (80GB) GPUs. The second server is equipped with two AMD EPYC 9135 16-Core Processors, 512 GB of DDR5 ECC RAM, one NVIDIA A100 (80GB) GPU, and two NVIDIA RTX PRO 6000 Blackwell Server Edition (96 GB) GPUs. Additionally, we use an Amazon Web Services (AWS) instance provisioned with an NVIDIA H100 (80GB) GPU to support the large-scale video generation phase.

\newpage

\subsection{Zero-Shot Evaluation}
\label{app:detection-zeroshot}

This section reports the full breakdowns of the zero-shot results of Section~\ref{subsec:zero_shot}: per-detector AUC by generator and modality on each Primary Evaluation Set (Section~\ref{app:detection-zeroshot-pergenerator}), and the compression-robustness ablation (Section~\ref{app:detection-zeroshot-compression}).

\subsubsection{Per-Generator and Per-Modality Breakdown}
\label{app:detection-zeroshot-pergenerator}

Tables~\ref{tab:zeroshot-pergenerator-sf-ffpp}--\ref{tab:zeroshot-pergenerator-sf-cdf} report the per-detector zero-shot AUC at \textit{Raw} compression on each Primary Evaluation Set, broken down by SynthForensics generator and modality (T2V vs I2V). Generator short codes: Cog = CogVideoX, MAGI = MAGI-1-Distilled, Wan = Wan2.1, SF = Self-Forcing, Sky = SkyReels-V2, Hel = Helios-Distilled, LTX = LTX-2.3, dV = daVinci-MagiHuman-Distilled.

The per-generator breakdowns reveal a sharp separation in detection difficulty. Among the eight T2V generators, Helios-Distilled and LTX-2.3 are consistently the hardest: average detector AUC drops to $53.9\%$ and $50.5\%$ on SF-FF++ T2V, near random chance, with similar values on SF-DFD and SF-CDF. SkyReels-V2 and daVinci-MagiHuman are at the opposite end (both around $79\%$ average AUC on SF-FF++ T2V); these are also the two generators rated most realistic in the human study (Section~\ref{subsec:human_study}), suggesting the detection-difficulty axis tracks low-level generative artifacts rather than human-perceived realism.

Across modalities, T2V and I2V perform comparably on SF-FF++ ($68.2\%$ vs $68.8\%$ average detector AUC), with I2V becoming progressively harder under stronger distribution shifts: $69.8\%$ vs $67.8\%$ on SF-DFD and $64.9\%$ vs $60.8\%$ on SF-CDF. We attribute this trend to the I2V conditioning frame anchoring synthetic outputs to a real face, leaving subtler artifacts than the fully-novel identities of T2V outputs.

Among detectors, DFD-FCG, GenConViT, GenD, and RECCE form the strongest tier (overall mean above $77\%$), all face-based methods built on CLIP/ViT backbones. FTCN ($52.8\%$ average) and the three purpose-built synthetic-video methods (MM-Det $48.5\%$, D3 $53.3\%$, NSG-VD $60.6\%$) cluster near random chance across all three Primary Sets, confirming that synthetic-video generic detection does not transfer to people-centric content of SynthForensics's quality bar.

\clearpage

\begin{table}[t]
\centering
\caption{Per-detector zero-shot AUC (\%) on SF--FF++ (\textit{Raw}), broken down by generator and modality. \textsuperscript{*}Frame-level detector.}
\label{tab:zeroshot-pergenerator-sf-ffpp}
\footnotesize
\setlength{\tabcolsep}{3pt}
\renewcommand{\arraystretch}{1.05}
\resizebox{\linewidth}{!}{%
\begin{tabular}{l|ccccccccc|cccccccc|c}
\toprule
 & \multicolumn{9}{c|}{\textbf{T2V}} & \multicolumn{8}{c|}{\textbf{I2V}} & \textbf{Overall} \\
\cmidrule(lr){2-10} \cmidrule(lr){11-18}
\textbf{Detector} & Cog & MAGI & Wan & SF & Sky & Hel & LTX & dV & \textit{Mean} & Cog & MAGI & Wan & Sky & Hel & LTX & dV & \textit{Mean} & \textit{Mean} \\
\midrule
CFM\textsuperscript{*} & 71.9 & 80.1 & 73.0 & 76.3 & 85.7 & 49.1 & 45.0 & 87.3 & 71.1 & 85.5 & 84.1 & 68.0 & 78.8 & 58.5 & 69.5 & 90.2 & 76.4 & 73.5 \\
RECCE\textsuperscript{*} & 75.3 & 81.3 & 80.5 & 83.2 & 91.1 & 57.0 & 59.3 & 96.7 & 78.1 & 86.3 & 85.9 & 80.1 & 86.3 & 59.5 & 76.8 & 93.2 & 81.1 & 79.5 \\
ProDet\textsuperscript{*} & 72.2 & 83.1 & 78.3 & 75.5 & 90.5 & 52.8 & 54.1 & 90.5 & 74.6 & 78.7 & 87.1 & 71.9 & 83.8 & 47.4 & 73.8 & 90.6 & 76.2 & 75.3 \\
UCF\textsuperscript{*} & 73.7 & 74.0 & 79.1 & 77.1 & 88.8 & 42.5 & 49.2 & 95.8 & 72.5 & 83.8 & 81.7 & 76.9 & 83.4 & 60.3 & 71.5 & 93.4 & 78.7 & 75.4 \\
Effort\textsuperscript{*} & 83.7 & 80.0 & 71.9 & 70.9 & 85.1 & 27.4 & 50.2 & 91.8 & 70.1 & 71.6 & 77.7 & 72.5 & 77.6 & 42.4 & 65.3 & 86.8 & 70.5 & 70.3 \\
LAA-Net\textsuperscript{*} & 71.9 & 72.8 & 73.7 & 57.5 & 82.7 & 71.3 & 43.2 & 89.5 & 70.3 & 68.6 & 72.7 & 58.3 & 78.2 & 52.7 & 55.6 & 92.9 & 68.4 & 69.4 \\
GenD\textsuperscript{*} & 86.7 & 86.2 & 91.8 & 81.4 & 94.6 & 67.6 & 50.9 & 97.7 & 82.1 & 87.5 & 84.7 & 86.4 & 90.1 & 59.2 & 74.3 & 96.2 & 82.6 & 82.3 \\
\addlinespace
AltFreezing & 60.3 & 55.9 & 56.1 & 50.6 & 59.8 & 69.2 & 31.6 & 66.8 & 56.3 & 59.9 & 46.7 & 52.6 & 63.0 & 68.2 & 39.1 & 62.0 & 55.9 & 56.1 \\
FTCN & 43.6 & 44.5 & 49.6 & 54.7 & 60.6 & 40.1 & 33.6 & 54.3 & 47.6 & 44.9 & 45.0 & 36.3 & 44.8 & 34.4 & 38.5 & 46.4 & 41.5 & 44.8 \\
GenConViT & 79.7 & 83.8 & 91.2 & 88.4 & 94.1 & 42.4 & 65.6 & 97.5 & 80.3 & 93.5 & 96.4 & 90.0 & 95.0 & 53.9 & 84.4 & 98.4 & 87.4 & 83.6 \\
DFD-FCG & 87.1 & 94.4 & 94.2 & 95.3 & 95.2 & 64.4 & 66.6 & 97.0 & 86.8 & 96.5 & 95.6 & 87.7 & 91.1 & 55.7 & 81.7 & 95.9 & 86.3 & 86.6 \\
FakeSTormer & 75.3 & 79.8 & 79.0 & 74.8 & 88.5 & 71.4 & 43.9 & 86.6 & 74.9 & 71.8 & 72.6 & 63.5 & 73.3 & 56.5 & 65.1 & 86.8 & 69.9 & 72.6 \\
\midrule
MM-Det & 47.4 & 48.4 & 48.2 & 38.9 & 50.7 & 50.1 & 46.4 & 46.9 & 47.1 & 48.5 & 51.4 & 52.9 & 51.9 & 50.8 & 52.9 & 52.2 & 51.5 & 49.2 \\
NSG-VD & 49.7 & 56.9 & 63.9 & 50.4 & 72.6 & 65.4 & 61.6 & 51.3 & 59.0 & 55.4 & 54.5 & 53.4 & 59.3 & 65.1 & 60.0 & 64.9 & 58.9 & 59.0 \\
D3 & 47.9 & 54.4 & 49.6 & 70.6 & 47.9 & 38.4 & 56.4 & 46.7 & 51.5 & 49.4 & 53.2 & 50.3 & 45.8 & 35.8 & 44.6 & 41.8 & 45.9 & 48.9 \\
\bottomrule
\end{tabular}%
}
\end{table}

\begin{table}[t]
\centering
\caption{Per-detector zero-shot AUC (\%) on SF--DFD (\textit{Raw}), broken down by generator and modality. \textsuperscript{*}Frame-level detector.}
\label{tab:zeroshot-pergenerator-sf-dfd}
\footnotesize
\setlength{\tabcolsep}{3pt}
\renewcommand{\arraystretch}{1.05}
\resizebox{\linewidth}{!}{%
\begin{tabular}{l|ccccccccc|cccccccc|c}
\toprule
 & \multicolumn{9}{c|}{\textbf{T2V}} & \multicolumn{8}{c|}{\textbf{I2V}} & \textbf{Overall} \\
\cmidrule(lr){2-10} \cmidrule(lr){11-18}
\textbf{Detector} & Cog & MAGI & Wan & SF & Sky & Hel & LTX & dV & \textit{Mean} & Cog & MAGI & Wan & Sky & Hel & LTX & dV & \textit{Mean} & \textit{Mean} \\
\midrule
CFM\textsuperscript{*} & 62.3 & 81.6 & 74.6 & 85.1 & 84.2 & 64.3 & 67.2 & 88.9 & 76.0 & 59.2 & 71.4 & 71.6 & 72.4 & 57.5 & 67.1 & 78.7 & 68.3 & 72.4 \\
RECCE\textsuperscript{*} & 68.8 & 83.0 & 81.2 & 85.7 & 88.6 & 72.1 & 76.0 & 93.6 & 81.1 & 67.4 & 78.7 & 79.5 & 80.8 & 70.4 & 70.1 & 83.5 & 75.8 & 78.6 \\
ProDet\textsuperscript{*} & 59.2 & 72.1 & 74.1 & 71.6 & 81.8 & 48.3 & 62.0 & 81.4 & 68.8 & 53.6 & 68.9 & 71.3 & 78.2 & 42.1 & 67.6 & 83.9 & 66.5 & 67.7 \\
UCF\textsuperscript{*} & 67.0 & 76.5 & 78.0 & 84.8 & 85.6 & 64.0 & 66.1 & 91.7 & 76.7 & 65.8 & 68.9 & 78.5 & 77.0 & 67.1 & 66.9 & 77.0 & 71.6 & 74.3 \\
Effort\textsuperscript{*} & 71.5 & 64.9 & 69.5 & 65.0 & 76.0 & 37.5 & 61.8 & 71.7 & 64.7 & 61.4 & 61.8 & 80.6 & 79.2 & 45.3 & 57.0 & 64.7 & 64.3 & 64.5 \\
LAA-Net\textsuperscript{*} & 50.7 & 57.0 & 60.5 & 44.9 & 66.1 & 49.9 & 38.5 & 76.7 & 55.5 & 52.8 & 54.6 & 82.7 & 75.6 & 41.4 & 52.6 & 82.1 & 63.1 & 59.1 \\
GenD\textsuperscript{*} & 83.8 & 80.4 & 92.5 & 85.6 & 92.7 & 77.2 & 65.4 & 93.1 & 83.8 & 80.4 & 77.7 & 91.6 & 89.3 & 73.7 & 68.6 & 82.2 & 80.5 & 82.3 \\
\addlinespace
AltFreezing & 70.4 & 71.5 & 75.7 & 77.1 & 74.4 & 81.9 & 60.7 & 78.3 & 73.8 & 59.1 & 62.5 & 71.9 & 72.2 & 79.2 & 68.1 & 71.9 & 69.3 & 71.7 \\
FTCN & 57.2 & 55.8 & 58.1 & 70.8 & 71.9 & 48.8 & 42.0 & 69.4 & 59.2 & 65.9 & 61.3 & 60.2 & 68.4 & 47.8 & 52.6 & 67.4 & 60.5 & 59.8 \\
GenConViT & 92.1 & 96.4 & 94.4 & 95.9 & 98.9 & 75.3 & 90.2 & 98.3 & 92.7 & 93.1 & 97.0 & 81.3 & 91.2 & 85.7 & 86.7 & 83.6 & 88.4 & 90.7 \\
DFD-FCG & 82.9 & 87.2 & 88.8 & 91.0 & 92.2 & 72.3 & 75.2 & 91.9 & 85.2 & 85.3 & 82.5 & 81.6 & 86.1 & 65.9 & 71.7 & 84.8 & 79.7 & 82.6 \\
FakeSTormer & 74.3 & 78.9 & 78.6 & 74.9 & 85.3 & 71.0 & 67.7 & 77.0 & 76.0 & 70.1 & 72.8 & 67.5 & 69.5 & 64.5 & 68.6 & 71.7 & 69.2 & 72.8 \\
\midrule
MM-Det & 48.5 & 62.1 & 47.0 & 32.8 & 61.9 & 58.8 & 54.7 & 57.4 & 52.9 & 48.4 & 56.8 & 57.0 & 60.3 & 59.8 & 56.5 & 65.5 & 57.8 & 55.2 \\
NSG-VD & 49.3 & 50.4 & 52.4 & 48.8 & 56.7 & 41.8 & 50.9 & 61.3 & 51.4 & 51.9 & 50.0 & 53.4 & 49.8 & 56.5 & 50.1 & 58.8 & 52.9 & 52.1 \\
D3 & 46.8 & 48.5 & 46.0 & 61.4 & 46.1 & 42.4 & 49.7 & 55.4 & 49.5 & 61.7 & 51.8 & 47.7 & 44.6 & 40.2 & 46.9 & 51.9 & 49.2 & 49.4 \\
\bottomrule
\end{tabular}%
}
\end{table}

\begin{table}[t]
\centering
\caption{Per-detector zero-shot AUC (\%) on SF--CDF (\textit{Raw}), broken down by generator and modality. \textsuperscript{*}Frame-level detector.}
\label{tab:zeroshot-pergenerator-sf-cdf}
\footnotesize
\setlength{\tabcolsep}{3pt}
\renewcommand{\arraystretch}{1.05}
\resizebox{\linewidth}{!}{%
\begin{tabular}{l|ccccccccc|cccccccc|c}
\toprule
 & \multicolumn{9}{c|}{\textbf{T2V}} & \multicolumn{8}{c|}{\textbf{I2V}} & \textbf{Overall} \\
\cmidrule(lr){2-10} \cmidrule(lr){11-18}
\textbf{Detector} & Cog & MAGI & Wan & SF & Sky & Hel & LTX & dV & \textit{Mean} & Cog & MAGI & Wan & Sky & Hel & LTX & dV & \textit{Mean} & \textit{Mean} \\
\midrule
CFM\textsuperscript{*} & 60.2 & 76.4 & 68.4 & 74.8 & 85.4 & 52.3 & 50.9 & 84.6 & 69.1 & 66.4 & 69.9 & 56.3 & 65.4 & 44.7 & 56.9 & 77.8 & 62.5 & 66.0 \\
RECCE\textsuperscript{*} & 64.3 & 78.5 & 77.9 & 78.2 & 91.8 & 58.7 & 60.4 & 95.4 & 75.6 & 70.9 & 74.2 & 70.3 & 75.3 & 51.6 & 62.3 & 84.2 & 69.8 & 72.9 \\
ProDet\textsuperscript{*} & 64.2 & 78.9 & 77.0 & 75.6 & 87.6 & 48.3 & 59.1 & 88.2 & 72.4 & 62.0 & 74.4 & 61.7 & 75.7 & 37.2 & 66.1 & 85.2 & 66.0 & 69.4 \\
UCF\textsuperscript{*} & 71.3 & 78.6 & 81.3 & 81.4 & 91.5 & 59.2 & 57.9 & 96.0 & 77.1 & 75.6 & 74.9 & 73.7 & 78.8 & 61.3 & 65.0 & 84.4 & 73.4 & 75.4 \\
Effort\textsuperscript{*} & 67.0 & 62.5 & 65.8 & 54.6 & 76.8 & 25.5 & 47.6 & 76.5 & 59.6 & 61.9 & 62.3 & 69.1 & 72.6 & 34.9 & 53.6 & 72.5 & 61.0 & 60.2 \\
LAA-Net\textsuperscript{*} & 44.5 & 49.5 & 50.4 & 32.6 & 64.3 & 43.2 & 21.8 & 74.7 & 47.6 & 41.5 & 42.9 & 51.2 & 58.2 & 25.9 & 33.9 & 77.7 & 47.3 & 47.5 \\
GenD\textsuperscript{*} & 68.6 & 69.3 & 80.9 & 65.9 & 85.6 & 59.2 & 38.8 & 89.3 & 69.7 & 73.0 & 64.8 & 76.6 & 80.2 & 46.5 & 51.8 & 81.8 & 67.8 & 68.8 \\
\addlinespace
AltFreezing & 52.8 & 48.7 & 55.3 & 51.3 & 56.5 & 63.8 & 27.6 & 60.3 & 52.0 & 46.2 & 39.9 & 46.1 & 52.0 & 61.0 & 39.5 & 50.7 & 47.9 & 50.1 \\
FTCN & 47.1 & 53.7 & 56.0 & 66.6 & 69.5 & 50.6 & 36.4 & 59.7 & 54.9 & 55.9 & 54.1 & 49.5 & 58.3 & 41.5 & 49.0 & 57.6 & 52.2 & 53.7 \\
GenConViT & 57.9 & 76.6 & 74.1 & 74.8 & 89.2 & 26.8 & 52.5 & 91.7 & 68.0 & 66.7 & 79.8 & 58.1 & 75.0 & 31.7 & 62.3 & 68.8 & 63.2 & 65.7 \\
DFD-FCG & 73.8 & 85.0 & 87.6 & 89.5 & 92.2 & 54.2 & 58.7 & 89.2 & 78.8 & 88.3 & 82.8 & 75.4 & 82.3 & 43.0 & 66.3 & 87.5 & 75.1 & 77.0 \\
FakeSTormer & 69.1 & 74.1 & 74.2 & 67.6 & 84.9 & 63.6 & 48.9 & 76.5 & 69.9 & 61.0 & 60.6 & 52.9 & 59.1 & 47.9 & 53.4 & 70.6 & 57.9 & 64.3 \\
\midrule
MM-Det & 38.6 & 47.1 & 37.2 & 25.1 & 46.6 & 40.6 & 40.9 & 39.5 & 39.4 & 38.9 & 41.2 & 44.0 & 45.2 & 43.2 & 43.1 & 47.8 & 43.4 & 41.3 \\
NSG-VD & 59.6 & 81.8 & 76.8 & 66.6 & 85.7 & 69.3 & 82.7 & 81.4 & 75.5 & 50.4 & 57.7 & 56.4 & 61.8 & 66.7 & 78.4 & 83.3 & 65.0 & 70.6 \\
D3 & 60.2 & 64.4 & 59.5 & 79.8 & 59.6 & 55.3 & 65.3 & 65.5 & 63.7 & 64.4 & 63.8 & 60.7 & 60.7 & 50.8 & 59.8 & 57.2 & 59.6 & 61.8 \\
\bottomrule
\end{tabular}%
}
\end{table}

\clearpage

\subsubsection{Compression Robustness Across All Primary Sets}
\label{app:detection-zeroshot-compression}

Table~\ref{tab:zeroshot-compression} reports per-detector zero-shot AUC across the four compression versions (\textit{Raw}, \textit{Canonical}, \textit{CRF23}, \textit{CRF40}) on each Primary Evaluation Set, aggregated over the SynthForensics generators (8 T2V + 7 I2V).

Three patterns stand out. First, \textit{Raw} and \textit{Canonical} are essentially equivalent on every Primary Set ($\leq 1$pp average difference), confirming that the H.264 CRF=0 re-encoding of the \textit{Canonical} version preserves the detector-relevant signal of the unprocessed generator output and neutralizes format-specific confounds without introducing detection cues of its own.

Second, aggressive compression flips the ranking of detectors. The top face-based detectors at \textit{Raw} (GenD, RECCE, ProDet, UCF, CFM, GenConViT, DFD-FCG) collapse by $25$--$40$ percentage points moving from \textit{Raw} to \textit{CRF40} on SF-FF++, while the three purpose-built synthetic-video methods (MM-Det, NSG-VD, D3), which are already near random chance at \textit{Raw}, show no drop or marginally rise ($+0.7$ to $+6.7$pp). The strongest detectors thus rely on fine-grained generative artifacts that are progressively destroyed by H.264 quantization; the weakest detectors latch onto coarser cues that survive (or are introduced by) the compression itself.

Third, the magnitude of the compression collapse depends sharply on the Primary Set. The average AUC across the 15 detectors drops by $23$pp on SF-FF++ ($68.4\% \to 45.4\%$), only $8$pp on SF-DFD ($68.9\% \to 61.1\%$), and $13$pp on SF-CDF ($63.0\% \to 50.5\%$). SF-DFD's robustness reflects the fact that its DFD reals (lab recordings) and the matched compression pipeline align the noise statistics of real and synthetic videos at heavy CRF, leaving the detectors with cleaner residual cues than on the more heterogeneous YouTube-sourced SF-FF++.

\begin{table}[!h]
\centering
\caption{Per-detector zero-shot AUC (\%) across the four compression versions (\textit{Raw} (R), \textit{Canonical} (C), \textit{CRF23} (23), \textit{CRF40} (40)) on each Primary Evaluation Set, aggregated over the SynthForensics generators (8 T2V + 7 I2V). \textsuperscript{*}Frame-level detector.}
\label{tab:zeroshot-compression}
\footnotesize
\setlength{\tabcolsep}{4pt}
\renewcommand{\arraystretch}{1.05}
\resizebox{\linewidth}{!}{%
\begin{tabular}{l|cccc|cccc|cccc}
\toprule
 & \multicolumn{4}{c|}{\textbf{SF-FF++}} & \multicolumn{4}{c|}{\textbf{SF-DFD}} & \multicolumn{4}{c}{\textbf{SF-CDF}} \\
\cmidrule(lr){2-5} \cmidrule(lr){6-9} \cmidrule(lr){10-13}
\textbf{Detector} & R & C & 23 & 40 & R & C & 23 & 40 & R & C & 23 & 40 \\
\midrule
CFM\textsuperscript{*} & 73.5 & 71.6 & 66.3 & 37.5 & 72.4 & 72.5 & 65.6 & 64.1 & 66.0 & 64.7 & 55.7 & 40.0 \\
RECCE\textsuperscript{*} & 79.5 & 78.7 & 71.3 & 40.5 & 78.6 & 77.6 & 71.2 & 72.9 & 72.9 & 71.9 & 60.2 & 47.1 \\
ProDet\textsuperscript{*} & 75.3 & 74.8 & 68.0 & 36.9 & 67.7 & 67.2 & 57.0 & 64.8 & 69.4 & 67.9 & 65.7 & 50.7 \\
UCF\textsuperscript{*} & 75.4 & 74.5 & 65.1 & 37.0 & 74.3 & 72.9 & 66.3 & 65.6 & 75.4 & 73.4 & 67.9 & 50.0 \\
Effort\textsuperscript{*} & 70.3 & 69.8 & 63.6 & 46.7 & 64.5 & 63.9 & 55.8 & 51.9 & 60.2 & 60.2 & 59.0 & 46.6 \\
LAA-Net\textsuperscript{*} & 69.4 & 71.6 & 58.6 & 44.0 & 59.1 & 56.8 & 48.6 & 49.8 & 47.5 & 45.1 & 57.9 & 52.1 \\
GenD\textsuperscript{*} & 82.3 & 81.2 & 72.3 & 42.4 & 82.3 & 81.9 & 72.0 & 64.5 & 68.8 & 68.5 & 67.9 & 48.4 \\
\addlinespace
AltFreezing & 56.1 & 55.8 & 53.3 & 40.4 & 71.7 & 71.1 & 69.3 & 58.0 & 50.1 & 49.4 & 47.2 & 35.6 \\
FTCN & 44.8 & 43.9 & 40.5 & 34.5 & 59.8 & 60.7 & 54.1 & 38.6 & 53.7 & 51.8 & 49.6 & 50.4 \\
GenConViT & 83.6 & 83.7 & 76.4 & 50.6 & 90.7 & 89.6 & 85.9 & 65.9 & 65.7 & 61.9 & 62.3 & 50.1 \\
DFD-FCG & 86.6 & 86.6 & 80.6 & 54.5 & 82.6 & 81.8 & 78.7 & 73.7 & 77.0 & 76.1 & 73.6 & 56.7 \\
FakeSTormer & 72.6 & 72.7 & 65.5 & 46.0 & 72.8 & 71.5 & 69.4 & 74.4 & 64.3 & 62.9 & 61.0 & 53.0 \\
\midrule
MM-Det & 49.2 & 48.8 & 49.4 & 49.9 & 55.2 & 54.1 & 53.1 & 54.6 & 41.3 & 40.7 & 40.3 & 45.2 \\
NSG-VD & 59.0 & 60.6 & 57.0 & 64.5 & 52.1 & 52.4 & 56.6 & 67.9 & 70.6 & 70.0 & 69.6 & 71.8 \\
D3 & 48.9 & 51.0 & 52.2 & 55.5 & 49.4 & 48.6 & 51.7 & 49.9 & 61.8 & 62.8 & 59.2 & 59.4 \\
\bottomrule
\end{tabular}%
}
\end{table}

\clearpage

\titlespacing*{\subsection}{0pt}{2pt}{2pt}

\subsection{Fine-Tuning}

\titlespacing*{\subsection}{0pt}{1.5ex}{0.8ex}  

\label{app:detection-finetuning}

This section expands Section~\ref{subsec:fine_tuning} with the rationale for the trained-protocol detector pool (Section~\ref{app:detection-trained-pool}), the per-detector fine-tuned AUC by generator and modality on each Primary Evaluation Set (Section~\ref{app:detection-finetuning-pergenerator}), and backward compatibility on the Legacy Benchmark Sets (Section~\ref{app:detection-finetuning-legacy}).

Training follows a conservative protocol designed to preserve pre-trained representations while adapting to synthetic artifacts: each detector uses its native optimizer and batch size, learning rate $10^{-5}$, weight decay $10^{-4}$, standard augmentations, a maximum of 25 epochs, and early stopping (patience 5) on validation AUC.

\subsubsection{Detector Pool for Trained-Protocol Evaluation}
\label{app:detection-trained-pool}

The fine-tuning and training-from-scratch pools are the subset of the 15-detector zero-shot pool of Section~\ref{subsec:zero_shot} for which a supervised trained-protocol is well-defined. The remaining six detectors are excluded by construction, not for compute or coverage reasons.

CFM~\citep{luo2023beyond} is omitted because the original authors do not release training code, precluding faithful re-implementation of its critical-forgery-mining pipeline.

LAA-Net~\citep{nguyen2024laa} and FakeSTormer~\citep{nguyen2025vulnerability} train on real videos only and dynamically synthesize pseudo-fakes via Blended-Image and Self-Blended-Video generation; their multi-task auxiliary heads (vulnerable-pixel heatmap and self-consistency mask for LAA-Net; temporal blending-boundary derivatives and per-frame spatial vulnerabilities for FakeSTormer) are supervised by ground-truth signals computed from the blending mask, which is undefined on fully-synthetic videos that contain no blending boundary. The auxiliary supervisions therefore become undefined and the multi-task framework reduces to a binary classifier, defeating the architectural premise that the auxiliary heads disentangle subtle artifact-prone regions; this rationale applies equally to fine-tuning and to training from scratch.

D3~\citep{zheng2025d3} is training-free by construction, consisting of a frozen pre-trained encoder and a closed-form second-order temporal statistic with no trainable parameters; the authors explicitly state that the method operates solely during inference and requires no training datasets, so neither fine-tuning nor training from scratch is defined. NSG-VD~\citep{zhang2025physics} is a non-parametric Maximum Mean Discrepancy test between the test video and a real-video reference set, computed over a physics-derived statistic on top of a frozen pre-trained diffusion score function; the only trainable parameters are the deep-kernel hyperparameters of the MMD test, optimized through kernel-learning rather than supervised classifier learning. Adapting NSG-VD to the supervised fine-tuning and training-from-scratch protocols would require a fundamentally different optimization paradigm, falling outside the scope of a like-for-like comparison with the face-based classifier pool. MM-Det~\citep{song2024learning} is designed for diffusion-generated content rather than face manipulation, with the Large Multimodal Model, CLIP visual encoder, and VQ-VAE all frozen by construction and only the projection layer and Spatio-Temporal branch trainable; its training set (DVF) consists of eight diffusion methods with no face-swap content, and the authors have explicitly stated that face-swap benchmarks fall outside the method's design scope, so training from scratch on synthetic-only data preserves the design but does not provide a meaningful comparison against face-based detectors.

The disjointness analysis of Section~\ref{subsec:re_training} additionally restricts the pool to face-based detectors that can in principle learn either the synthetic or the manipulation feature space; the three synthetic-specific methods (D3, NSG-VD, MM-Det) would enter the analysis as null cases since they are training-free or synthetic-specific by construction, tautologically confirming rather than testing the claim.

\subsubsection{Per-Generator and Per-Modality Breakdown}
\label{app:detection-finetuning-pergenerator}

Tables~\ref{tab:finetuning-pergenerator-sf-ffpp}--\ref{tab:finetuning-pergenerator-sf-cdf} report the per-detector fine-tuned AUC at \textit{Raw} compression on each Primary Evaluation Set, broken down by SynthForensics generator and modality (T2V vs I2V). Generator short codes follow Section~\ref{app:detection-zeroshot-pergenerator}.

After fine-tuning on SynthForensics-Train, almost all detectors reach $96$--$99\%$ AUC on SF-FF++, closing the zero-shot gap of Section~\ref{app:detection-zeroshot-pergenerator} on every generator including the ones that were hardest at zero-shot (Helios-Distilled, LTX-2.3). Generalization to SF-DFD ($87$--$96\%$) and SF-CDF ($91$--$97\%$) remains strong, indicating that the artifacts learned from FF++-paired SynthForensics videos transfer to lab-recorded (DFD) and celebrity (CDF) real-content shifts. Two detectors deviate: GenD plateaus at $80\%$ (the LayerNorm-only adaptation has limited capacity to absorb the new distribution), and DFD-FCG drops on SF-DFD ($79.1\%$) and SF-CDF ($82.8\%$) despite reaching $92.8\%$ on SF-FF++, suggesting a more FF-specific adaptation.

\begin{table}[H]
\centering
\caption{Per-detector fine-tuned AUC (\%) on SF--FF++ (\textit{Raw}), broken down by generator and modality. \textsuperscript{*}Frame-level detector.}
\label{tab:finetuning-pergenerator-sf-ffpp}
\footnotesize
\setlength{\tabcolsep}{3pt}
\renewcommand{\arraystretch}{1.05}
\resizebox{\linewidth}{!}{%
\begin{tabular}{l|ccccccccc|cccccccc|c}
\toprule
 & \multicolumn{9}{c|}{\textbf{T2V}} & \multicolumn{8}{c|}{\textbf{I2V}} & \textbf{Overall} \\
\cmidrule(lr){2-10} \cmidrule(lr){11-18}
\textbf{Detector} & Cog & MAGI & Wan & SF & Sky & Hel & LTX & dV & \textit{Mean} & Cog & MAGI & Wan & Sky & Hel & LTX & dV & \textit{Mean} & \textit{Mean} \\
\midrule
RECCE\textsuperscript{*} & 99.5 & 99.5 & 99.9 & 99.8 & 99.9 & 99.9 & 99.0 & 99.8 & 99.7 & 96.0 & 95.8 & 96.0 & 98.0 & 95.4 & 95.1 & 97.1 & 96.2 & 98.0 \\
ProDet\textsuperscript{*} & 96.7 & 98.7 & 99.2 & 99.5 & 99.6 & 98.1 & 98.2 & 99.2 & 98.7 & 89.9 & 94.8 & 92.2 & 96.3 & 90.2 & 92.3 & 97.1 & 93.3 & 96.1 \\
UCF\textsuperscript{*} & 98.8 & 99.4 & 100.0 & 99.7 & 99.9 & 99.7 & 99.2 & 100.0 & 99.6 & 94.0 & 93.1 & 93.5 & 95.3 & 92.0 & 91.8 & 95.2 & 93.6 & 96.8 \\
Effort\textsuperscript{*} & 100.0 & 99.7 & 100.0 & 99.9 & 99.8 & 99.7 & 98.9 & 100.0 & 99.7 & 92.6 & 91.9 & 91.5 & 95.6 & 88.0 & 92.4 & 96.1 & 92.6 & 96.4 \\
GenD\textsuperscript{*} & 83.1 & 84.6 & 88.1 & 78.2 & 91.2 & 65.1 & 46.1 & 97.7 & 79.3 & 88.1 & 84.6 & 86.3 & 89.4 & 60.4 & 72.7 & 93.3 & 82.1 & 80.6 \\
\addlinespace
AltFreezing & 98.4 & 99.8 & 99.5 & 99.6 & 99.7 & 99.2 & 99.1 & 99.0 & 99.3 & 98.4 & 98.9 & 99.5 & 99.7 & 98.9 & 98.7 & 98.3 & 98.9 & 99.1 \\
FTCN & 97.2 & 99.7 & 99.1 & 98.8 & 99.2 & 99.1 & 98.1 & 97.8 & 98.6 & 97.6 & 98.3 & 98.9 & 99.3 & 97.6 & 97.7 & 97.2 & 98.1 & 98.4 \\
GenConViT & 98.5 & 98.8 & 100.0 & 99.3 & 99.9 & 99.8 & 97.8 & 99.8 & 99.2 & 93.1 & 97.2 & 97.5 & 98.9 & 90.9 & 93.1 & 98.9 & 95.7 & 97.6 \\
DFD-FCG & 96.9 & 98.2 & 98.5 & 99.1 & 99.0 & 84.8 & 92.9 & 98.4 & 96.0 & 95.2 & 97.9 & 90.4 & 93.0 & 61.3 & 90.6 & 95.9 & 89.2 & 92.8 \\
\bottomrule
\end{tabular}%
}
\end{table}

\begin{table}[H]
\centering
\caption{Per-detector fine-tuned AUC (\%) on SF--DFD (\textit{Raw}), broken down by generator and modality. \textsuperscript{*}Frame-level detector.}
\label{tab:finetuning-pergenerator-sf-dfd}
\footnotesize
\setlength{\tabcolsep}{3pt}
\renewcommand{\arraystretch}{1.05}
\resizebox{\linewidth}{!}{%
\begin{tabular}{l|ccccccccc|cccccccc|c}
\toprule
 & \multicolumn{9}{c|}{\textbf{T2V}} & \multicolumn{8}{c|}{\textbf{I2V}} & \textbf{Overall} \\
\cmidrule(lr){2-10} \cmidrule(lr){11-18}
\textbf{Detector} & Cog & MAGI & Wan & SF & Sky & Hel & LTX & dV & \textit{Mean} & Cog & MAGI & Wan & Sky & Hel & LTX & dV & \textit{Mean} & \textit{Mean} \\
\midrule
RECCE\textsuperscript{*} & 95.4 & 97.4 & 97.7 & 94.7 & 99.3 & 97.4 & 96.3 & 99.1 & 97.2 & 84.1 & 86.0 & 92.9 & 97.4 & 77.8 & 83.3 & 97.1 & 88.4 & 93.0 \\
ProDet\textsuperscript{*} & 87.2 & 91.4 & 97.5 & 95.9 & 99.1 & 90.8 & 91.7 & 98.5 & 94.0 & 64.5 & 74.5 & 88.9 & 93.3 & 59.4 & 75.3 & 97.1 & 79.0 & 87.0 \\
UCF\textsuperscript{*} & 94.4 & 96.1 & 98.0 & 94.0 & 99.3 & 96.2 & 94.7 & 99.0 & 96.5 & 83.3 & 83.2 & 89.7 & 94.4 & 78.7 & 80.9 & 91.8 & 86.0 & 91.6 \\
Effort\textsuperscript{*} & 95.0 & 93.8 & 97.6 & 90.9 & 97.3 & 96.2 & 84.0 & 90.7 & 93.2 & 91.7 & 88.2 & 95.0 & 98.0 & 81.6 & 81.1 & 95.4 & 90.2 & 91.8 \\
GenD\textsuperscript{*} & 76.3 & 79.5 & 88.7 & 85.0 & 89.2 & 78.6 & 62.7 & 92.3 & 81.5 & 80.9 & 77.2 & 88.8 & 86.6 & 74.6 & 65.8 & 78.5 & 78.9 & 80.3 \\
\addlinespace
AltFreezing & 95.5 & 96.5 & 96.3 & 95.6 & 97.7 & 93.3 & 90.4 & 95.7 & 95.1 & 95.9 & 95.1 & 94.2 & 96.7 & 93.0 & 95.5 & 96.9 & 95.3 & 95.2 \\
FTCN & 92.3 & 95.8 & 95.3 & 95.4 & 97.6 & 93.8 & 90.6 & 92.8 & 94.2 & 95.7 & 94.1 & 91.9 & 96.0 & 93.2 & 93.6 & 94.2 & 94.1 & 94.2 \\
GenConViT & 95.9 & 96.8 & 99.1 & 96.2 & 99.7 & 97.5 & 94.0 & 98.7 & 97.2 & 93.7 & 94.9 & 96.9 & 99.3 & 91.7 & 89.3 & 98.5 & 94.9 & 96.1 \\
DFD-FCG & 81.0 & 83.8 & 85.9 & 87.6 & 88.1 & 66.2 & 76.2 & 87.1 & 82.0 & 83.5 & 78.0 & 78.4 & 82.6 & 47.5 & 78.0 & 83.4 & 75.9 & 79.1 \\
\bottomrule
\end{tabular}%
}
\end{table}

\begin{table}[H]
\centering
\caption{Per-detector fine-tuned AUC (\%) on SF--CDF (\textit{Raw}), broken down by generator and modality. \textsuperscript{*}Frame-level detector.}
\label{tab:finetuning-pergenerator-sf-cdf}
\footnotesize
\setlength{\tabcolsep}{3pt}
\renewcommand{\arraystretch}{1.05}
\resizebox{\linewidth}{!}{%
\begin{tabular}{l|ccccccccc|cccccccc|c}
\toprule
 & \multicolumn{9}{c|}{\textbf{T2V}} & \multicolumn{8}{c|}{\textbf{I2V}} & \textbf{Overall} \\
\cmidrule(lr){2-10} \cmidrule(lr){11-18}
\textbf{Detector} & Cog & MAGI & Wan & SF & Sky & Hel & LTX & dV & \textit{Mean} & Cog & MAGI & Wan & Sky & Hel & LTX & dV & \textit{Mean} & \textit{Mean} \\
\midrule
RECCE\textsuperscript{*} & 98.6 & 99.0 & 99.2 & 97.7 & 99.8 & 99.2 & 98.7 & 99.8 & 99.0 & 92.5 & 93.1 & 95.5 & 98.7 & 89.4 & 92.0 & 98.0 & 94.2 & 96.8 \\
ProDet\textsuperscript{*} & 93.6 & 95.9 & 98.4 & 98.4 & 99.5 & 95.7 & 96.0 & 99.4 & 97.1 & 72.9 & 83.6 & 88.4 & 95.9 & 72.4 & 80.7 & 96.8 & 84.4 & 91.2 \\
UCF\textsuperscript{*} & 97.8 & 98.2 & 99.4 & 97.6 & 99.8 & 98.8 & 96.8 & 99.6 & 98.5 & 90.3 & 90.4 & 92.8 & 96.2 & 88.8 & 89.3 & 93.6 & 91.6 & 95.3 \\
Effort\textsuperscript{*} & 95.7 & 97.0 & 98.8 & 97.0 & 98.6 & 97.3 & 92.7 & 96.8 & 96.7 & 87.2 & 85.5 & 88.1 & 93.3 & 80.6 & 85.5 & 92.4 & 87.5 & 92.4 \\
GenD\textsuperscript{*} & 65.5 & 69.6 & 78.7 & 64.8 & 82.2 & 59.2 & 36.9 & 86.7 & 67.9 & 72.7 & 64.2 & 76.2 & 78.6 & 48.5 & 49.5 & 75.1 & 66.4 & 67.2 \\
\addlinespace
AltFreezing & 93.4 & 96.3 & 95.3 & 95.5 & 96.6 & 91.3 & 92.0 & 93.4 & 94.2 & 92.1 & 92.3 & 93.2 & 94.2 & 90.8 & 94.1 & 92.9 & 92.8 & 93.5 \\
FTCN & 92.9 & 97.3 & 97.5 & 96.9 & 98.7 & 95.9 & 95.2 & 94.3 & 96.1 & 94.8 & 94.8 & 94.8 & 96.9 & 94.0 & 94.9 & 92.7 & 94.7 & 95.4 \\
GenConViT & 95.3 & 96.7 & 98.2 & 95.8 & 99.7 & 96.5 & 90.6 & 98.3 & 96.4 & 85.0 & 91.6 & 94.6 & 97.6 & 82.3 & 82.8 & 96.9 & 90.1 & 93.5 \\
DFD-FCG & 86.0 & 91.9 & 93.9 & 96.1 & 95.8 & 63.8 & 80.7 & 91.8 & 87.5 & 88.6 & 87.9 & 79.1 & 84.6 & 28.8 & 83.7 & 89.8 & 77.5 & 82.8 \\
\bottomrule
\end{tabular}%
}
\end{table}

\clearpage

\subsubsection{Backward Compatibility on Legacy Benchmark Sets}
\label{app:detection-finetuning-legacy}

Table~\ref{tab:finetuning-legacy} reports the fine-tuned AUC on the three Legacy Benchmark Sets (FF++, CDF, DFD) plus their mean, measuring how much of the original manipulation-detection capability is retained after fine-tuning on SynthForensics.

The pool splits sharply. DFD-FCG and GenD retain their Legacy capability nearly intact (Mean Legacy $94.2\%$ and $92.7\%$), with GenD's frozen-backbone LayerNorm-only adaptation paying virtually no backward cost. The remaining face-based frame-level detectors (RECCE, ProDet, UCF, Effort) drop into the $76$--$81\%$ range; the loss is concentrated on CDF ($63$--$74\%$) and DFD ($74$--$88\%$) rather than FF++ ($84$--$92\%$), reflecting the train/test alignment with FF++-derived synthetic content. The two video-level detectors with the largest SF-FF++ gains, AltFreezing and FTCN, exhibit the most pronounced backward collapse (Mean Legacy $57.9\%$ and $51.4\%$), with FTCN at or below random chance on CDF and DFD.

\begin{table}[!h]
\centering
\caption{Backward compatibility: per-detector fine-tuned AUC (\%) on the Legacy Benchmark Sets. \textsuperscript{*}Frame-level detector.}
\label{tab:finetuning-legacy}
\small
\setlength{\tabcolsep}{8pt}
\renewcommand{\arraystretch}{1.15}
\begin{tabular}{l|cccc}
\toprule
\textbf{Detector} & \textbf{FF++} & \textbf{CDF} & \textbf{DFD} & \textbf{Mean} \\
\midrule
RECCE\textsuperscript{*} & 88.8 & 64.8 & 74.3 & 76.8 \\
ProDet\textsuperscript{*} & 84.2 & 74.3 & 80.1 & 79.5 \\
UCF\textsuperscript{*} & 92.5 & 63.7 & 77.0 & 77.7 \\
Effort\textsuperscript{*} & 92.4 & 63.9 & 88.0 & 81.4 \\
GenD\textsuperscript{*} & 98.2 & 93.2 & 86.6 & 92.7 \\
\addlinespace
AltFreezing & 65.5 & 57.0 & 51.1 & 57.9 \\
FTCN & 57.1 & 50.3 & 46.8 & 51.4 \\
GenConViT & 86.4 & 65.9 & 90.8 & 81.0 \\
DFD-FCG & 99.4 & 92.1 & 91.1 & 94.2 \\
\bottomrule
\end{tabular}
\end{table}

\subsection{Training from Scratch}
\label{app:detection-retraining}

This section expands Section~\ref{subsec:re_training} with the per-detector training-from-scratch AUC by generator and modality on each Primary Evaluation Set (Section~\ref{app:detection-retraining-pergenerator}), and backward compatibility on the Legacy Benchmark Sets (Section~\ref{app:detection-retraining-legacy}). The detector pool follows the same selection criteria detailed in Section~\ref{app:detection-trained-pool} for the fine-tuning protocol.

Training follows each detector's native optimizer, learning rate, schedule, and batch size; the full network is trained end-to-end, with backbones (e.g., CLIP, ResNet) initialized from the standard pre-trained checkpoints adopted by the original detector implementations. We apply weight decay $10^{-4}$, standard augmentations, a maximum of 50 epochs, and early stopping (patience 10) on validation AUC.

\subsubsection{Per-Generator and Per-Modality Breakdown}
\label{app:detection-retraining-pergenerator}

Tables~\ref{tab:retraining-pergenerator-sf-ffpp}--\ref{tab:retraining-pergenerator-sf-cdf} report the per-detector training-from-scratch AUC at \textit{Raw} compression on each Primary Evaluation Set, broken down by SynthForensics generator and modality (T2V vs I2V). Generator short codes follow Section~\ref{app:detection-zeroshot-pergenerator}. The In-Domain pool (CogVideoX, MAGI-1-Distilled, Wan2.1, LTX-2.3) and the Out-of-Domain pool (Self-Forcing, SkyReels-V2, Helios-Distilled, daVinci-MagiHuman) are defined in Section~\ref{subsec:re_training}.

After training from scratch, seven of the nine detectors saturate on the In-Domain T2V pool (above $99\%$ on SF-FF++) and remain within $1$--$3$pp on Out-of-Domain T2V. The bulk of the Out-of-Domain generalization cost concentrates on a single generator-modality pair: Helios-I2V drops frame-level detectors (RECCE, UCF, Effort) and DFD-FCG from $\geq 90\%$ on the rest of the I2V pool to $63$--$77\%$ on SF-FF++, and to $32$--$64\%$ on the more challenging SF-DFD and SF-CDF, while AltFreezing and FTCN remain above $96\%$ throughout, suggesting the Helios-I2V artifact survives spatial-only inductive bias but not the explicit temporal modeling of these two detectors. Generalization to SF-DFD ($81$--$96\%$ overall for the seven strong detectors) and SF-CDF ($83$--$95\%$) confirms that the synthetic artifacts learned at training transfer to lab-recorded (DFD) and celebrity (CDF) real-content shifts, mirroring the fine-tuning behavior. ProDet ($79.8\%$ overall on SF-FF++) and GenD ($56.2\%$) remain the two outliers: ProDet collapses on daVinci-T2V ($35$--$48\%$ across the three Primary Sets) and underperforms uniformly on I2V, while GenD's failure is symmetric across In- and Out-of-Domain, with no recovery relative to its zero-shot performance.

\begin{table}[H]
\centering
\caption{Per-detector training-from-scratch AUC (\%) on SF--FF++ (\textit{Raw}), broken down by generator and modality. \textsuperscript{*}Frame-level detector.}
\label{tab:retraining-pergenerator-sf-ffpp}
\footnotesize
\setlength{\tabcolsep}{3pt}
\renewcommand{\arraystretch}{1.05}
\resizebox{\linewidth}{!}{%
\begin{tabular}{l|ccccccccc|cccccccc|c}
\toprule
 & \multicolumn{9}{c|}{\textbf{T2V}} & \multicolumn{8}{c|}{\textbf{I2V}} & \textbf{Overall} \\
\cmidrule(lr){2-10} \cmidrule(lr){11-18}
\textbf{Detector} & Cog & MAGI & Wan & SF & Sky & Hel & LTX & dV & \textit{Mean} & Cog & MAGI & Wan & Sky & Hel & LTX & dV & \textit{Mean} & \textit{Mean} \\
\midrule
RECCE\textsuperscript{*} & 99.2 & 99.3 & 99.6 & 99.9 & 99.7 & 94.2 & 99.4 & 99.7 & 98.9 & 95.0 & 95.6 & 95.1 & 95.2 & 77.3 & 94.5 & 96.5 & 92.7 & 96.0 \\
ProDet\textsuperscript{*} & 94.0 & 93.3 & 94.3 & 96.3 & 91.5 & 97.1 & 87.7 & 72.9 & 90.9 & 66.3 & 67.6 & 62.2 & 72.2 & 70.4 & 65.9 & 66.0 & 67.2 & 79.8 \\
UCF\textsuperscript{*} & 98.4 & 99.2 & 99.6 & 99.8 & 99.5 & 95.2 & 99.5 & 99.5 & 98.9 & 94.3 & 94.9 & 95.0 & 95.0 & 76.2 & 93.7 & 95.1 & 92.0 & 95.7 \\
Effort\textsuperscript{*} & 99.7 & 99.8 & 100.0 & 99.9 & 100.0 & 93.2 & 99.4 & 100.0 & 98.7 & 95.4 & 95.5 & 96.2 & 97.3 & 75.2 & 96.7 & 97.0 & 93.3 & 95.8 \\
GenD\textsuperscript{*} & 61.0 & 66.6 & 62.4 & 66.8 & 50.4 & 62.9 & 55.8 & 59.5 & 60.7 & 49.2 & 54.2 & 51.2 & 47.2 & 49.8 & 54.1 & 51.7 & 51.0 & 56.2 \\
\addlinespace
AltFreezing & 99.1 & 99.7 & 99.4 & 99.8 & 99.6 & 98.0 & 99.5 & 98.9 & 99.2 & 98.8 & 99.2 & 99.2 & 99.4 & 97.9 & 99.0 & 97.9 & 98.8 & 99.0 \\
FTCN & 96.7 & 99.7 & 98.4 & 98.7 & 98.4 & 98.7 & 98.5 & 97.6 & 98.3 & 97.1 & 98.2 & 98.6 & 98.8 & 96.4 & 97.9 & 95.6 & 97.5 & 98.0 \\
GenConViT & 93.8 & 96.8 & 97.5 & 96.0 & 98.0 & 68.6 & 95.6 & 99.3 & 93.2 & 91.5 & 93.7 & 93.0 & 94.1 & 64.8 & 93.0 & 92.4 & 88.9 & 91.2 \\
DFD-FCG & 99.0 & 99.1 & 99.4 & 99.7 & 98.9 & 87.5 & 97.7 & 97.7 & 97.4 & 91.8 & 93.8 & 91.5 & 90.5 & 62.9 & 91.9 & 94.1 & 88.1 & 93.0 \\
\bottomrule
\end{tabular}%
}
\end{table}

\begin{table}[H]
\centering
\caption{Per-detector training-from-scratch AUC (\%) on SF--DFD (\textit{Raw}), broken down by generator and modality. \textsuperscript{*}Frame-level detector.}
\label{tab:retraining-pergenerator-sf-dfd}
\footnotesize
\setlength{\tabcolsep}{3pt}
\renewcommand{\arraystretch}{1.05}
\resizebox{\linewidth}{!}{%
\begin{tabular}{l|ccccccccc|cccccccc|c}
\toprule
 & \multicolumn{9}{c|}{\textbf{T2V}} & \multicolumn{8}{c|}{\textbf{I2V}} & \textbf{Overall} \\
\cmidrule(lr){2-10} \cmidrule(lr){11-18}
\textbf{Detector} & Cog & MAGI & Wan & SF & Sky & Hel & LTX & dV & \textit{Mean} & Cog & MAGI & Wan & Sky & Hel & LTX & dV & \textit{Mean} & \textit{Mean} \\
\midrule
RECCE\textsuperscript{*} & 95.4 & 97.2 & 97.6 & 95.2 & 98.7 & 78.9 & 96.1 & 99.0 & 94.8 & 80.9 & 85.2 & 89.4 & 92.5 & 47.3 & 81.5 & 92.8 & 81.4 & 88.5 \\
ProDet\textsuperscript{*} & 71.8 & 75.2 & 75.1 & 77.4 & 73.1 & 78.6 & 61.5 & 35.6 & 68.5 & 53.5 & 56.7 & 57.1 & 62.0 & 52.1 & 53.6 & 56.3 & 55.9 & 62.6 \\
UCF\textsuperscript{*} & 92.0 & 93.9 & 96.0 & 92.4 & 98.2 & 75.3 & 92.4 & 96.0 & 92.0 & 75.9 & 77.8 & 83.2 & 87.8 & 39.9 & 78.5 & 87.2 & 75.8 & 84.4 \\
Effort\textsuperscript{*} & 96.0 & 94.3 & 98.9 & 90.3 & 99.4 & 79.3 & 91.5 & 98.9 & 93.6 & 91.7 & 88.7 & 98.6 & 99.6 & 51.1 & 84.9 & 99.5 & 87.8 & 90.8 \\
GenD\textsuperscript{*} & 41.4 & 50.4 & 50.4 & 53.7 & 43.2 & 60.9 & 36.5 & 44.5 & 47.6 & 49.5 & 51.1 & 49.7 & 48.6 & 50.8 & 46.5 & 50.9 & 49.6 & 48.5 \\
\addlinespace
AltFreezing & 97.3 & 97.4 & 98.2 & 96.5 & 99.0 & 93.1 & 96.0 & 98.2 & 97.0 & 96.7 & 96.8 & 94.3 & 98.3 & 90.0 & 97.3 & 97.5 & 95.9 & 96.4 \\
FTCN & 91.9 & 95.2 & 92.7 & 93.7 & 93.9 & 90.0 & 89.7 & 93.1 & 92.5 & 92.2 & 90.9 & 88.8 & 93.4 & 87.8 & 91.2 & 92.1 & 90.9 & 91.8 \\
GenConViT & 84.2 & 92.9 & 91.1 & 83.2 & 94.7 & 56.0 & 89.5 & 96.9 & 86.1 & 72.6 & 91.0 & 75.9 & 88.5 & 47.3 & 86.8 & 99.2 & 80.2 & 83.3 \\
DFD-FCG & 89.8 & 89.8 & 92.3 & 89.5 & 90.8 & 61.5 & 82.3 & 87.8 & 85.5 & 73.1 & 84.0 & 80.1 & 91.5 & 33.0 & 80.4 & 93.0 & 76.4 & 81.3 \\
\bottomrule
\end{tabular}%
}
\end{table}

\begin{table}[H]
\centering
\caption{Per-detector training-from-scratch AUC (\%) on SF--CDF (\textit{Raw}), broken down by generator and modality. \textsuperscript{*}Frame-level detector.}
\label{tab:retraining-pergenerator-sf-cdf}
\footnotesize
\setlength{\tabcolsep}{3pt}
\renewcommand{\arraystretch}{1.05}
\resizebox{\linewidth}{!}{%
\begin{tabular}{l|ccccccccc|cccccccc|c}
\toprule
 & \multicolumn{9}{c|}{\textbf{T2V}} & \multicolumn{8}{c|}{\textbf{I2V}} & \textbf{Overall} \\
\cmidrule(lr){2-10} \cmidrule(lr){11-18}
\textbf{Detector} & Cog & MAGI & Wan & SF & Sky & Hel & LTX & dV & \textit{Mean} & Cog & MAGI & Wan & Sky & Hel & LTX & dV & \textit{Mean} & \textit{Mean} \\
\midrule
RECCE\textsuperscript{*} & 98.0 & 98.4 & 98.7 & 98.4 & 99.2 & 86.1 & 97.8 & 99.0 & 97.0 & 92.4 & 93.9 & 94.6 & 96.1 & 59.8 & 93.2 & 96.4 & 89.5 & 93.5 \\
ProDet\textsuperscript{*} & 86.4 & 89.6 & 89.2 & 89.6 & 84.5 & 89.8 & 78.5 & 47.8 & 81.9 & 57.9 & 62.4 & 57.3 & 67.1 & 59.5 & 57.7 & 54.4 & 59.5 & 71.5 \\
UCF\textsuperscript{*} & 97.2 & 98.0 & 98.9 & 98.1 & 99.1 & 88.8 & 97.7 & 98.6 & 97.0 & 91.9 & 92.2 & 94.3 & 95.3 & 63.7 & 93.0 & 94.8 & 89.3 & 93.4 \\
Effort\textsuperscript{*} & 98.2 & 97.7 & 99.4 & 96.7 & 99.7 & 81.2 & 95.9 & 99.5 & 96.0 & 90.5 & 90.3 & 94.2 & 95.4 & 56.5 & 92.2 & 94.7 & 87.7 & 92.1 \\
GenD\textsuperscript{*} & 62.8 & 68.5 & 67.5 & 70.0 & 60.7 & 70.2 & 56.9 & 63.7 & 65.0 & 64.1 & 66.6 & 64.6 & 61.9 & 65.8 & 64.2 & 65.0 & 64.6 & 64.8 \\
\addlinespace
AltFreezing & 95.2 & 97.1 & 97.5 & 96.2 & 98.2 & 87.9 & 96.4 & 95.3 & 95.5 & 94.4 & 95.6 & 94.5 & 96.0 & 87.6 & 95.7 & 92.6 & 93.8 & 94.7 \\
FTCN & 94.8 & 97.8 & 96.9 & 96.4 & 97.1 & 93.7 & 93.7 & 93.6 & 95.5 & 93.8 & 94.2 & 93.6 & 95.9 & 89.7 & 94.0 & 92.1 & 93.3 & 94.5 \\
GenConViT & 91.7 & 96.5 & 95.3 & 90.1 & 97.9 & 49.2 & 92.7 & 98.6 & 89.0 & 81.0 & 96.3 & 86.4 & 95.0 & 45.8 & 91.5 & 98.0 & 84.8 & 87.1 \\
DFD-FCG & 95.0 & 95.1 & 95.2 & 96.5 & 93.7 & 66.5 & 90.3 & 91.5 & 90.5 & 79.0 & 83.6 & 77.8 & 83.7 & 31.6 & 80.6 & 88.4 & 75.0 & 83.2 \\
\bottomrule
\end{tabular}%
}
\end{table}

\clearpage

\subsubsection{Backward Compatibility on Legacy Benchmark Sets}
\label{app:detection-retraining-legacy}

Table~\ref{tab:retraining-legacy} reports the training-from-scratch AUC on the three Legacy Benchmark Sets (FF++, CDF, DFD) plus their mean, measuring how much manipulation-detection capability emerges after training from scratch on synthetic-only data.

The pool collapses uniformly. Eight of the nine detectors land in the $50$--$65\%$ Mean Legacy range, with GenD sliding below $55\%$ on every Legacy set and FTCN at random chance on CDF and DFD. DFD-FCG is the sole exception ($83.4\%$ Mean Legacy, $80.6$ on FF++, $79.4$ on CDF, and $90.1$ on DFD). Compared to the fine-tuning regime, where seven of nine detectors retained $76$--$94\%$ Mean Legacy AUC by inheriting their pre-trained representations, training from scratch on SynthForensics does not transfer to the Legacy Benchmark Sets: the features that discriminate fully-synthetic videos are largely disjoint from those that discriminate traditional face-swap and face-reenactment manipulations.

\begin{table}[!h]
\centering
\caption{Backward compatibility: per-detector training-from-scratch AUC (\%) on the Legacy Benchmark Sets. \textsuperscript{*}Frame-level detector.}
\label{tab:retraining-legacy}
\small
\setlength{\tabcolsep}{8pt}
\renewcommand{\arraystretch}{1.15}
\begin{tabular}{l|cccc}
\toprule
\textbf{Detector} & \textbf{FF++} & \textbf{CDF} & \textbf{DFD} & \textbf{Mean} \\
\midrule
RECCE\textsuperscript{*} & 67.1 & 60.6 & 68.1 & 65.2 \\
ProDet\textsuperscript{*} & 54.8 & 52.4 & 56.5 & 54.6 \\
UCF\textsuperscript{*} & 64.3 & 55.7 & 62.5 & 60.8 \\
Effort\textsuperscript{*} & 69.5 & 54.8 & 64.6 & 63.0 \\
GenD\textsuperscript{*} & 51.2 & 45.4 & 53.8 & 50.1 \\
\addlinespace
AltFreezing & 67.1 & 58.8 & 63.4 & 63.1 \\
FTCN & 60.7 & 50.2 & 50.3 & 53.8 \\
GenConViT & 54.5 & 52.9 & 73.1 & 60.2 \\
DFD-FCG & 80.6 & 79.4 & 90.1 & 83.4 \\
\bottomrule
\end{tabular}
\end{table}

\clearpage

\clearpage
\section{Limitations and Broader Impact}
\label{app:limitations}

\subsection{Limitations}
\label{app:limitations-limitations}

The construction of SynthForensics balances quality, reproducibility, and threat-realism through several deliberate design choices. We document below the resulting scope limitations, each presented with its motivation and, where applicable, the position of SynthForensics relative to comparable synthetic-video benchmarks.

\textbf{Dataset size.} SynthForensics releases $20{,}445$ unique synthetic videos ($81{,}780$ video files across four compression versions), larger than most existing synthetic-video benchmarks but smaller than the largest broad-scope ones (e.g., GenVideo at $\sim 1.1$M, GenVidBench at $\sim 6.4$M declared). The size trade-off is a direct consequence of the construction protocol: the paired-source design ($1{,}363$ FF++/DFD reals), the two-stage human-in-the-loop validation pipeline, the per-generator hyperparameter tuning of eight state-of-the-art open-source generators across both T2V and I2V modalities, and the four compression versions per video together cap the corpus at the volume that can be manually curated to floor-level visual quality. Absolute size, however, is not the design objective. SynthForensics targets the highest visual fidelity attainable from current open-source generators, not the largest raw video count: human raters flag the competing benchmark videos as fake at $\sim 84\%$ against $\sim 38\%$ for SynthForensics in our paired-comparison study (Section~\ref{subsec:human_study}). A larger benchmark whose synthetic content is already trivially identifiable as fake does not exercise modern detectors at the operating point of forensic concern.

\textbf{Generator pool.} SynthForensics is restricted to open-source generators with publicly released weights, excluding the closed-source state of the art (Sora~2~\citep{sora2}, Veo~3.1~\citep{veo3}, Kling~3.0~\citep{kling2}, SeedDance~2.0~\citep{seeddance2}). The exclusion is deliberate: closed-source systems do not release weights, which precludes reproducible offline generation and makes published checkpoints non-archivable; their visible watermarks act as trivial detection artifacts that bypass the structural cues of interest; and their content guardrails restrict the people-centric generation that defines our threat model. The exclusion also defines SynthForensics's threat scope as the consumer-grade open-source attacker, where weights are publicly available, generation can be reproduced offline, and the benchmark can be archived as a stable artifact. The complementary threat scenario in which a sophisticated attacker bypasses proprietary content guardrails (jailbreaks, trial-access abuse, leaked checkpoints) lies outside this scope: such outputs are not reproducible in a stable benchmark form, since jailbreak prompts are patched over time and the underlying weights are unavailable for offline generation. Free-tier or trial access to the proprietary models is also feature-restricted by vendors (shorter durations, lower resolution, stronger content filters, more visible watermarks), so it does not provide a representative proxy for the paid state of the art. Within the open-source pool, three candidates (CausVid, OpenSora~2.0, Step-Video-T2V) were further excluded for failing a $25\%$ rejection-rate threshold on manual visual inspection (Appendix~\ref{app:benchmark-sources}), set a priori from our near-photorealistic threat model rather than tuned to the measured rejection distribution. This filtering trades absolute generator coverage for floor-level visual quality, by design biasing the corpus toward the hard regime where outputs are not trivially identifiable as fake; we argue this is the correct selection bias for a forensic benchmark, since easily detectable content does not stress modern detectors at the operating point of practical concern (see also \emph{Realism vs detection difficulty} below).

\textbf{Source-content domain.} The $1{,}363$ source videos are drawn exclusively from FaceForensics++~\citep{rossler2019faceforensics++} and the Deep Fake Detection dataset~\citep{dufour2019google}, which constrains the subject diversity of SynthForensics to the YouTube and lab-recording domains of the upstream corpora. Positive prompt construction (Appendix~\ref{app:benchmark-prompts}) is performed in English only, mirroring the dominant training distribution of the eight selected generators; negative prompts retain each generator's author-recommended defaults in their native language (English or Chinese, depending on the generator) and are extended by us with English-language artifact-suppression keywords (Appendix~\ref{app:benchmark-prompts-negative}). Both choices preserve compatibility with the established forensic real-video baselines and avoid prompt-distribution mismatch with the generators, but they bound the cultural and linguistic diversity of the generated content. Conversely, this same constraint produces a paired structure where each synthetic video has a content-aligned real counterpart from the same source, and I2V outputs additionally share the depicted subject identity inherited from the conditioning frame (Appendix~\ref{app:benchmark-i2v-frames}); this property may support research beyond deepfake detection alone.

\textbf{Compression coverage.} The four compression versions of SynthForensics are produced with H.264 at CRF=$0$/$23$/$40$, matching the codec and CRF range adopted by the established forensic baselines FF++~\citep{rossler2019faceforensics++} and DFD~\citep{dufour2019google} to preserve cross-benchmark comparability; this also places SynthForensics among the synthetic-video benchmarks with the most extensive compression coverage in the comparable pool, since most existing synthetic-video benchmarks release a single compression configuration or none. H.265/HEVC, AV1, and VP9 (employed by some social platforms) and the multi-pass re-encoding pipelines that arise when content is forwarded across platforms are not represented; extending the coverage to these codecs and to multi-pass re-encoding sequences is a natural future direction for cross-platform distribution scenarios.

\textbf{Static benchmark.} SynthForensics is a static snapshot of the open-source synthetic-video state of the art at submission time. Generators evolve rapidly: new architectures and capabilities will appear within months. We plan to release rolling updates of the benchmark as new state-of-the-art open-source models pass the same selection and validation procedure described in Section~\ref{sec:sf_bench}, treating SynthForensics as a versioned resource rather than a fixed snapshot.

\textbf{People-centric scope.} By construction, SynthForensics targets people-centric synthetic content, which is the threat surface most relevant to identity fraud, non-consensual content, and impersonation at scale. This scope excludes synthetic-content categories such as landscape video, animation, and game footage, which are addressed by other benchmarks (Section~\ref{subsec:benchmarks}). Detection results on SynthForensics should therefore be interpreted as evidence about people-centric synthetic-video forensics rather than synthetic-video forensics in general.

\textbf{Realism vs detection difficulty.} The two-stage validation pipeline of Section~\ref{sec:sf_bench} systematically rejects outputs with visible anatomical, temporal, or rendering artifacts, biasing SynthForensics toward perceptually challenging videos. This selection bias has two simultaneous consequences: it captures a realistic threat (the people-centric attack surface is precisely the regime where synthetic videos cannot be flagged by an unaided observer, Section~\ref{subsec:human_study}), and it constructs a hard benchmark for detectors that rely on perceptually identifiable cues. The two readings describe the same property of the dataset rather than a hidden confound, and we argue this is the correct selection bias for the threat model: a benchmark composed exclusively of videos already identifiable as fake by humans would not stress modern detectors at the operating point of practical concern. The corollary is that absolute AUC values measured on SynthForensics are not directly comparable to those measured on benchmarks dominated by lower-quality outputs; the meaningful comparison is the relative drop from manipulation-based benchmarks (FF++, DFD) to people-centric synthetic content under the same evaluation protocol.

\textbf{Detector evaluation completeness.} Section~\ref{sec:detection} reports zero-shot results for $15$ detectors; Sections~\ref{subsec:fine_tuning} and~\ref{subsec:re_training} report fine-tuning and training-from-scratch results for $9$ of those. The remaining $6$ are excluded by construction rather than for compute: CFM (training code unavailable from the authors); LAA-Net and FakeSTormer (multi-task auxiliary heads supervised by blending-mask-derived ground truths that are undefined for fully-synthetic videos); D3 (training-free by design, no trainable parameters); NSG-VD (non-parametric Maximum Mean Discrepancy test, optimization through kernel-learning rather than supervised classifier learning); MM-Det (designed for diffusion-generated content, with most parameters frozen by construction). Per-detector justifications are detailed in Appendix~\ref{app:detection-trained-pool}.

\subsection{Broader Impact}
\label{app:limitations-impact}

\textbf{Positive societal impacts.} SynthForensics is intended as a defensive resource for the forensic-detection research community. The dataset enables (i) development and benchmarking of deepfake detectors against the latest generation of open-source synthetic-video models, (ii) benchmarking of commercial and governmental deepfake-detection tools against modern people-centric synthetic content, beyond the legacy manipulation-based corpora typically used in vendor evaluations, (iii) generation of empirical evidence on the perceptual realism attainable by current open-source generators, which can inform policy-making on synthetic media, and (iv) reproducible, archivable evaluation supported by full generation metadata (Appendix~\ref{app:benchmark-stats}). Downstream applications include counter-disinformation pipelines, identity-fraud detection systems, and forensic tooling for non-consensual synthetic content.

\textbf{Negative societal impacts.} We consider three vectors of harm associated with releasing a high-fidelity people-centric synthetic-video corpus. First, harms from intended use: detectors trained or evaluated on SynthForensics are imperfect, and a deployed detector that flags real footage as synthetic could harm legitimate speakers (journalists, witnesses, or individuals contesting their own footage); detector outputs should be treated as one signal among several in any decision pipeline. Second, harms from incorrect outputs: T2V outputs synthesize novel identities, but I2V outputs are conditioned on reference frames extracted from FF++ and DFD source videos and therefore depict the visual identity of real individuals, animated according to a synthetic continuation that those individuals did not actually perform; although the underlying source videos are released for research use and the synthetic continuations were validated for ethical compliance (Appendix~\ref{app:benchmark-validation}), out-of-context redistribution of individual I2V videos could be misread as genuine footage. Third, harms from misuse: an adversary could in principle use SynthForensics as training data for detection-evasive generators, or as a corpus to seed disinformation, fake profiles, or non-consensual synthetic content. The mitigation strategies of Appendix~\ref{app:limitations-mitigation} address these risks.

\textbf{Asymmetry of offensive value.} The eight generators used to construct SynthForensics are already publicly available with their weights, prompts, and inference recipes; an attacker seeking to produce people-centric synthetic videos can do so directly, without our dataset and at comparable fidelity. SynthForensics adds substantive value to the defending party (detector training, evaluation, robustness analysis) and only marginal value to the offending party (a curated corpus of outputs already attainable from publicly-released generators). The release therefore does not meaningfully increase the absolute offensive capacity of the threat ecosystem; it equips the defenders with a benchmark calibrated to the threat already present in that ecosystem.

\subsection{Release Policy and Safeguards}
\label{app:limitations-mitigation}

\textbf{Distribution license.} SynthForensics is released under the Creative Commons Attribution-NonCommercial 4.0 International (CC BY-NC 4.0) license, which permits research, educational, and any non-commercial use with attribution, and prohibits commercial use. This choice is consistent with the upstream license of the FF++ and DFD source videos, which are themselves released for non-commercial research purposes only (Appendix~\ref{app:benchmark-sources}); SynthForensics does not relax the upstream constraint.

\textbf{Access control.} The dataset is distributed through a gated repository on Hugging Face. Before downloading, users must explicitly agree to the dataset terms of use (research-only, attribution required, no commercial use, and acknowledgement of the upstream FF++ and DFD terms of use). The gating is automatic, requires no manual approval after acceptance, and records the agreement, providing a lightweight audit trail without imposing access friction on the legitimate research community.

\textbf{Data usage agreement.} By accepting the terms, users acknowledge that (i) they will use SynthForensics only for non-commercial research and educational purposes; (ii) they will cite the SynthForensics paper alongside the upstream FF++ and DFD references in any publication or derivative artifact; (iii) they will not redistribute the dataset or substantial portions of it outside the original Hugging Face repository; and (iv) they will report observed misuse to the maintainers through the contact channel released alongside the dataset.

\textbf{Ethical review.} The dataset construction pipeline incorporates a two-stage human-in-the-loop ethical-compliance review. Structured prompts are validated before generation through manual annotator review and LLM-based safety screening across seven sensitive thematic categories (violence and warfare, weapons, vulnerable individuals, political figures, geopolitical references, national symbols, identifiable real persons; Appendix~\ref{app:benchmark-validation}); each generated video is then inspected against a rubric that includes ethical compliance as one of the rejection criteria (Appendix~\ref{app:benchmark-video-validation}), with flagged outputs regenerated under modified prompts until they pass.

\textbf{Monitoring and revocation.} Hugging Face provides automatic tooling for download statistics, access logs, and terms-acceptance records, which we use to identify anomalous access patterns. We reserve the right to revoke individual or institutional access in cases of documented misuse, and we maintain a contact channel for the community to report observed abuse of the released artifact.

\end{document}